\documentclass[letterpaper, 10 pt, compsoc]{ieeeconf}  

\usepackage{times}
\usepackage{epsfig}
\usepackage{graphicx}
\usepackage{amsmath}
\usepackage{amssymb}
\usepackage{cite}
\usepackage{amsmath,amssymb,amsfonts}
\usepackage{textcomp}
\usepackage{subfigure}
\usepackage[linesnumbered,boxed,ruled,commentsnumbered]{algorithm2e}
\usepackage{algpseudocode}  
\usepackage{amsmath}  
\usepackage{multirow}
\usepackage{color}
\UseRawInputEncoding

\newtheorem{lemma}{Lemma}

\newtheorem{prooof}{Proof} 
\usepackage{caption}
\usepackage{url}

\IEEEoverridecommandlockouts                              

\title{\LARGE \bf
TriVoC: Efficient Voting-based Consensus Maximization for Robust Point Cloud Registration with Extreme Outlier Ratios}

\author{Lei Sun$^{1,2,*}$, Lu Deng$^{1}$
\thanks{*Corresponding author. This work was not supported by any organization.}
\thanks{$^{1}$The authors are with School of Mechanical and Power Engineering, East China University of Science and Technology, Shanghai 200237, China;  
{\tt\small $\{$leisunjames,ludeng2021$\}$@126.com}}%
\thanks{$^{2}$Lei Sun is also with Shanghai FOCS Instrument Analysis Co., Ltd., Shanghai 201802, China.}
}

\begin{document}

\maketitle
\thispagestyle{empty}
\pagestyle{empty}

\begin{abstract}

Correspondence-based point cloud registration is a cornerstone in robotics perception and computer vision, which seeks to estimate the best rigid transformation aligning two point clouds from the putative correspondences. However, due to the limited robustness of 3D keypoint matching approaches, outliers, probably in large numbers, are prone to exist among the correspondences, which makes robust registration methods imperative. Unfortunately, existing robust methods have their own limitations (e.g. high computational cost or limited robustness) when facing high or extreme outlier ratios, probably unsuitable for practical use. In this paper, we present a novel, fast, deterministic and guaranteed robust solver, named TriVoC (Triple-layered Voting with Consensus maximization), for the robust registration problem. We decompose the selecting of the minimal 3-point sets into 3 consecutive layers, and in each layer we design an efficient voting and correspondence sorting framework on the basis of the pairwise equal-length constraint. In this manner, the 3-point sets can be selected independently from the reduced correspondence sets according to the sorted sequence, which can significantly lower the computational cost and meanwhile provide a strong guarantee to achieve the largest consensus set (as the final inlier set) as long as a probabilistic termination condition is fulfilled. Varied experiments show that our solver TriVoC is robust against up to 99\% outliers, highly accurate, time-efficient even with extreme outlier ratios, and also practical for real-world applications, showing performance superior to other state-of-the-art competitors.

\end{abstract}


\section{Introduction}

3D point cloud registration is a crucial building block in robotics, and 3D computer vision. It aims to align two point clouds by finding the best rigid transformation (including rotation and translation) between them. It has been broadly applied in scene reconstruction and mapping~\cite{henry2012rgb,choi2015robust,zhang2015visual}, object recognition and localization~\cite{drost2010model,zeng2017multi,wong2017segicp,marion2018label}, SLAM~\cite{zhang2014loam}, medical imaging~\cite{audette2000algorithmic}, archaeology~\cite{chase2012geospatial}, etc.

Building correspondences between point clouds using 3D keypoints has been an increasingly popular way for registration. Compared to the ICP~\cite{besl1992method} method, it does not rely on the initial guess and is convenient and cheap in practice. However, recent 3D keypoint matching methods have relatively low accuracy (less accurate than 2D keypoint matching like SIFT~\cite{lowe2004distinctive} or SURF~\cite{bay2006surf}) due to low texture, partiality, repetitive patterns, etc, so they are liable to generate spurious matches (\textit{outliers}) among the correspondences. Moreover, as discussed in~\cite{bustos2017guaranteed}, correspondences with more than 95\% outliers are fairly common in reality. This necessitates robust estimation methods tolerant to high or even extreme outliers.

\begin{figure}[t]
\centering

\setlength\tabcolsep{3pt}
\addtolength{\tabcolsep}{0pt}
\begin{tabular}{cc}

\footnotesize{(a) Correspondences with 99\% outliers}
&
\footnotesize{(b) Registration by TriVoC}
\\

\begin{minipage}[t]{0.49\linewidth}
\centering
\includegraphics[width=1\linewidth]{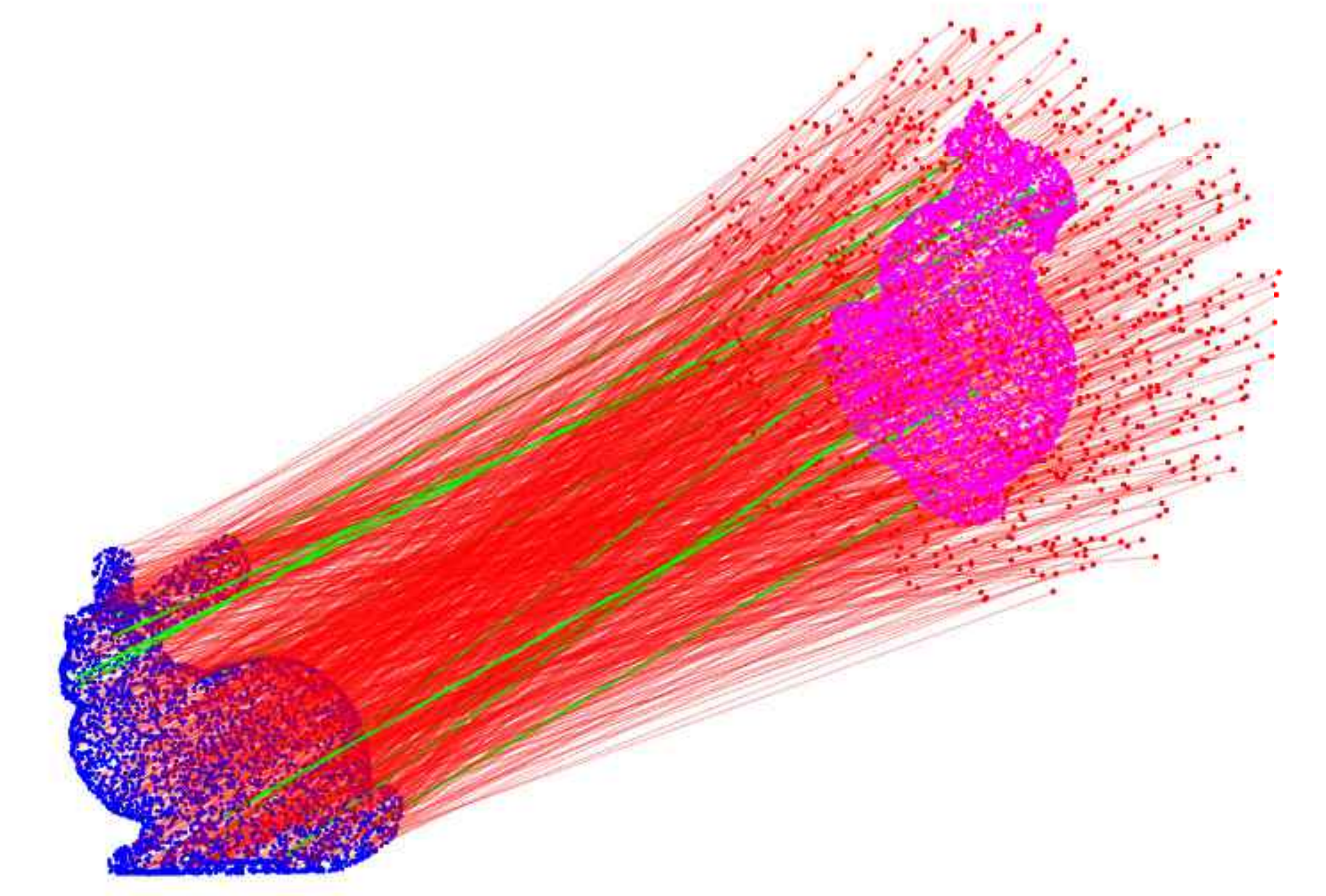}
\end{minipage}
&
\begin{minipage}[t]{0.36\linewidth}
\centering
\includegraphics[width=1\linewidth]{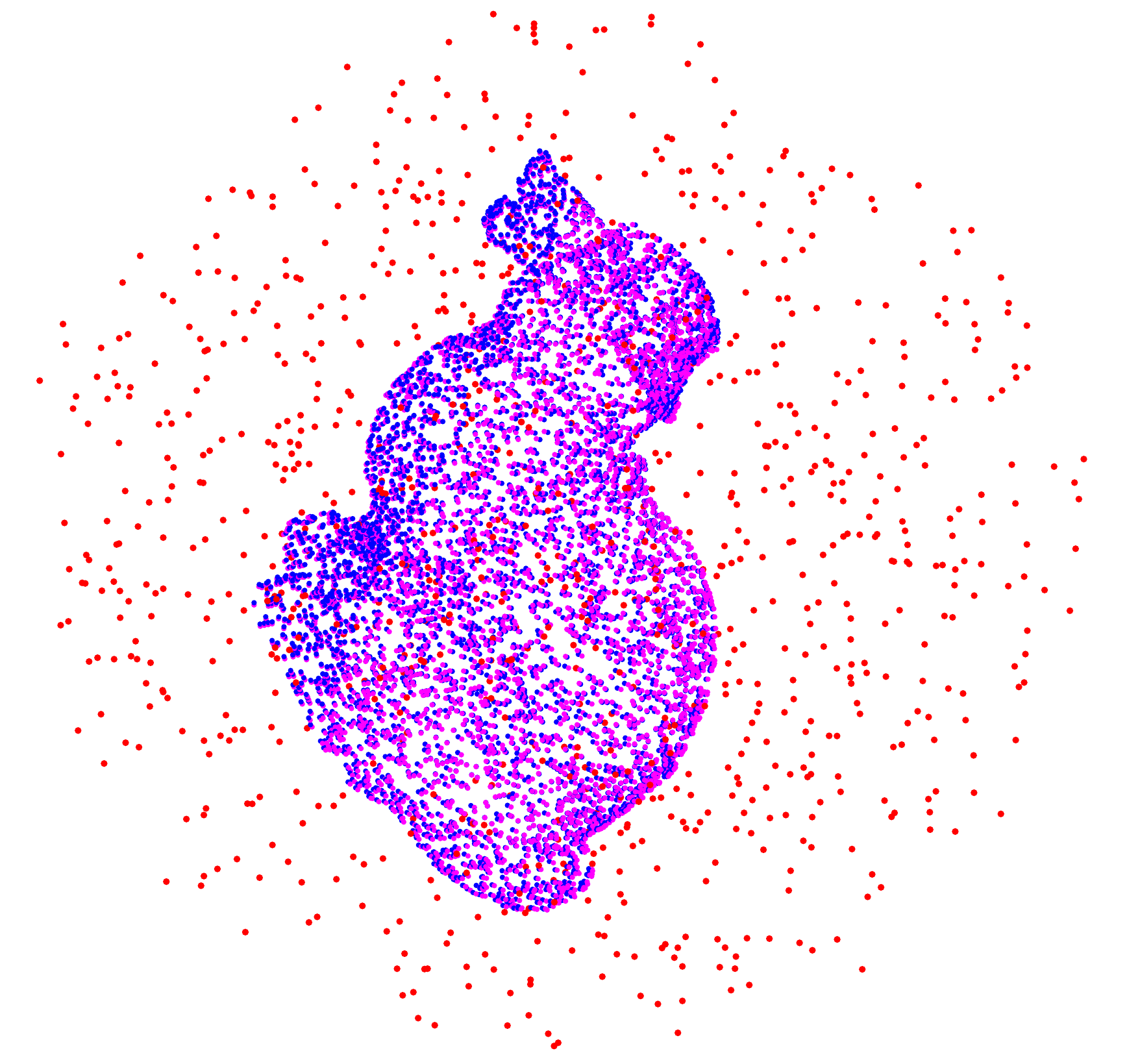}
\end{minipage}

\end{tabular}

\caption{Point cloud registration using our TriVoC with 1000 correspondences and 99\% outliers. Outliers are in \textcolor[rgb]{1,0,0}{red} while inliers are in \textcolor[rgb]{0,0.7,0}{green}.}
\label{show-emp}
\vspace{-1mm}
\end{figure}

Nonetheless, many existing robust solvers have their own drawbacks in practical use. As a famous robust estimator, RANSAC~\cite{fischler1981random} maximizes the consensus with random sampling and model fitting, but its runtime grows exponentially with the outlier ratio, thus infeasible to deal with high-outlier problems. Branch-and-Bound (BnB)~\cite{parra2014fast,horst2013global} is another consensus maximization robust method which can solve the problem globally optimally, but it scales poorly with the problem size also owing to exponential time cost. Non-minimal robust solvers including FGR~\cite{zhou2016fast}, GNC~\cite{yang2020graduated} and ADAPT~\cite{tzoumas2019outlier} run fast with SVD~\cite{arun1987least}, but they have limited robustness and would become brittle with outlier ratios exceeding 90\%. GORE~\cite{bustos2017guaranteed} is a guaranteed outlier removal solver but may also suffer from high computational cost due to its potential use of BnB. The certifiably optimal solver TEASER~\cite{yang2019polynomial,yang2020teaser} can also be slow when parallelism programming is not used for getting maximal cliques. Therefore, a fast and highly robust registration solver is imperative.

In this paper, we render a novel registration method named TriVoC (Triple-layered Voting with Consensus maximization), which is deterministic, efficient and robust against extreme outliers.

\textbf{Contributions.} We reformulate robust point cloud registration into a consensus maximization problem over 3-point sets. We decompose the process of selecting the 3-point sets into 3 embedded layers where only one single point is selected each time in each layer. During this triple-layered selecting process, we then introduce the pairwise equal-length constraint to seek inlier candidates for each correspondence selected, and meanwhile based on it, we design a novel time-efficient framework of voting and correspondence sorting. The strategies applied above lead to the proposed robust solver TriVoC, which is deterministic, has a strong guarantee of correctness, tolerates as many as 99\% outliers, and most often runs faster than other state-of-the-art solvers in multiple experiments and application problems over realistic datasets.

\section{Related Work}

We briefly review some typical robust point cloud registration solvers with correspondences by categories.

\textbf{Consensus Maximization Methods.} RANSAC~\cite{fischler1981random} and BnB~\cite{parra2014fast,horst2013global} are two well-known consensus maximization methods, where the former adopts a hypothesize-and-test paradigm with random minimal subsets and the latter conducts searching in the parameter space (e.g. $SO(3)$ or $SE(3)$). However, both of them suffer from the worst-case exponential computational cost (the former with the outlier ratio while the latter with the correspondence number), so neither of them is ideal for practical use. Our TriVoC is also a consensus maximizer. Slightly similar to RANSAC, TriVoC also needs to select minimal 3-point sets, but much differently, it selects points from the reduced and sorted subsets of the full correspondence set, so it is deterministic and fast even with high outlier ratios.

\textbf{M-estimation Methods.} M-estimation adopts robust cost functions to realize robust estimation by decreasing the effect of outliers. Local M-estimation solvers (e.g.~\cite{agarwal2013robust,kummerle2011g,sunderhauf2012towards}) requires the initial guess, so they could easily converge to local minima if the initialization is not good enough. But the need of initial guess is circumvented by Graduated Non-Convexity (GNC). FGR~\cite{zhou2016fast} is the first GNC-based solver, and then GNC is extended to more robotics problems by~\cite{yang2020graduated}. The major downside of GNC is that its solvers generally only tolerates about 80--90\% outliers, not robust enough for the realistic cases with many outliers.

\textbf{Invariant-based Methods.} Invariants have been employed to solve the registration problem. The equal-length~\cite{michel2017global,zach2015dynamic,quan2020compatibility} (called rigidity or scale-invariant) constraint is a common invariant for roughly differentiating inliers from outliers, which also underlies the framework of our TriVoC. Moreover, invariants~\cite{9552513} on rotation and translation have also been explored and employed to robust estimation.

\section{Our Methodology}

\subsection{Problem Formulation}

With two 3D point sets: $\mathcal{P}=\{\boldsymbol{p}_i\}_{i=1}^{N}$ and $\mathcal{Q}=\{\boldsymbol{q}_i\}_{i=1}^{N}$ ($\boldsymbol{p}_i,\boldsymbol{q}_i\in\mathbb{R}^3$ and $i\in\boldsymbol{N}=[1,2,\dots,N]^{\top}$) where $(\boldsymbol{p}_i,\boldsymbol{q}_i)$ is a putative correspondence (also abbreviated as correspondence $i$), if 
correspondence $i$ is a true inlier, then we can write:
\begin{equation}\label{prob-form1}
\boldsymbol{q}_i=\boldsymbol{R}\boldsymbol{p}_i+\boldsymbol{t}+\boldsymbol{\epsilon}_i,
\end{equation}
where rotation $\boldsymbol{R}\in SO(3)$ and translation $\boldsymbol{t}\in \mathbb{R}^{3}$ jointly constitute the rigid transformation and $\boldsymbol{\epsilon}_i\in\mathbb{R}^3$ denotes the noise measurement. The goal of robust point cloud registration is to estimate the best rigid transformation aligning set $\mathcal{P}$ and $\mathcal{Q}$ even though there exist outliers (correspondences that do not satisfy relation~\eqref{prob-form1}) in the correspondence set $\boldsymbol{N}$.

Hence, the robust registration problem can be formulated as a \textit{consensus maximization} problem such that
\begin{equation}\label{CM}
\begin{gathered}
\underset{\boldsymbol{I}\subset \boldsymbol{N}}{\max}\, |\boldsymbol{I}|, \\
s.t. \,\|\boldsymbol{R}^{\star}\boldsymbol{p}_i+\boldsymbol{t}^{\star}-\boldsymbol{q}_i\| \leq \gamma,\,(\forall i\in\boldsymbol{I})
\end{gathered}
\end{equation}
where $\boldsymbol{I}$ is the consensus set of rigid transformation ($\boldsymbol{R}^{\star}$, $\boldsymbol{t}^{\star}$) and $\gamma$ is the inlier threshold (if we assume the noise to be isotropic Gaussian, $\gamma=5\sim 6\sigma$ where $\sigma$ is the noise standard deviation). In this paper, we intentionally blur the difference between \textit{vector} and \textit{set} to facilitate our method presentation; for example, vector $\boldsymbol{I}$ also denotes a correspondence set whose elements are exactly the same as the entries of $\boldsymbol{I}$.

\subsection{Equal-length Constraint and Consistency Matrix}

Pairwise equal-length~\cite{michel2017global,zach2015dynamic,quan2020compatibility} is a common constraint in 3D registration, which indicates that the length between two points is fixed after rigid transformation and can be derived:

\begin{lemma}[Equal-length Constraint]
Given one pair of point correspondences: $(\boldsymbol{p}_i,\boldsymbol{q}_i)$ and $(\boldsymbol{p}_j,\boldsymbol{q}_j)$, we can have the inequality condition such that
\begin{equation}\label{rigidity}
\left|\left\|\boldsymbol{q}_i-\boldsymbol{q}_j\right\|-\left\|\boldsymbol{p}_i-\boldsymbol{p}_j\right\|\right| \leq 2\gamma,
\end{equation}
as long as correspondence $i$ and $j$ are both inliers.
\end{lemma}
\begin{prooof} 
This can be derived according to the triangular inequality and the norm-invariant property of $\boldsymbol{R}$ such that
\begin{equation}\label{proof-of-rigidity}
\begin{gathered}
\left|{\|\boldsymbol{q}_i-\boldsymbol{q}_j\|}-{\|\boldsymbol{p}_i-\boldsymbol{p}_j\|}\right| \\
=\left|{\left\|\boldsymbol{R} (\boldsymbol{p}_i-\boldsymbol{p}_j+\boldsymbol{R}^{\top}\boldsymbol{\epsilon}_i-\boldsymbol{R}^{\top}\boldsymbol{\epsilon}_j)\right\|}-{\|\boldsymbol{p}_i-\boldsymbol{p}_j\|}\right| \\
\leq \left\| (\boldsymbol{p}_i-\boldsymbol{p}_j)-({\boldsymbol{p}_i-\boldsymbol{p}_j})+\boldsymbol{R}^{\top}\boldsymbol{\epsilon}_i-\boldsymbol{R}^{\top}\boldsymbol{\epsilon}_j \right\| \\ 
= {2\|(\boldsymbol{\epsilon}_i-\boldsymbol{\epsilon}_j)\|}\leq 2\gamma.
\end{gathered}
\end{equation}
\end{prooof} 

This constraint can be used as a prerequisite for any pair of correspondences to be true inliers. Hence, our first step is to conduct test~\eqref{rigidity} over all the correspondence pairs from full set $\boldsymbol{N}$, based on which we can build a binary \textit{consistency matrix} $\boldsymbol{M}$ as rendered in Algorithm~\ref{algo-consistency} where $\boldsymbol{M}_{(i,j)}$ denotes the entry in the $i_{th}$ row and $j_{th}$ column of $\boldsymbol{M}$.
\begin{algorithm}[h]
\caption{\textit{buildConsistencyMatrix} (subroutine)}
\label{algo-consistency}
\SetKwInOut{Input}{\textbf{Input}}
\Input{correspondences $\{(\mathbf{p}_i,\mathbf{q}_i)\}_{i=1}^N$\;} 
Set a $N\times N$ matrix $\boldsymbol{M}$ with all entries equal to $0$\;
\For{$i=1:N$}{ 
\For{$j=(i+1):N$}{
\If{correspondence pair $(i,j)$ satisfies~\eqref{rigidity}}{
$\boldsymbol{M}_{(i,j)}\leftarrow 1$, $\boldsymbol{M}_{(j,i)}\leftarrow 1$\;
}
}
}
\Return consistency matrix $\boldsymbol{M}$\;
\end{algorithm}
 
Despite having the time complexity of $O(N^2)$, this process is generally not time-consuming in practice since it merely consists of the computation of subtracting and norms as well as the checking of boolean conditions.

\begin{figure}[t]
\centering
\setlength\tabcolsep{1pt}
\addtolength{\tabcolsep}{0pt}
\begin{tabular}{cc}

\footnotesize{(a) Correspondences}
&
\footnotesize{(b) First layer}

\\

\begin{minipage}[t]{0.46\linewidth}
\centering
\includegraphics[width=1\linewidth]{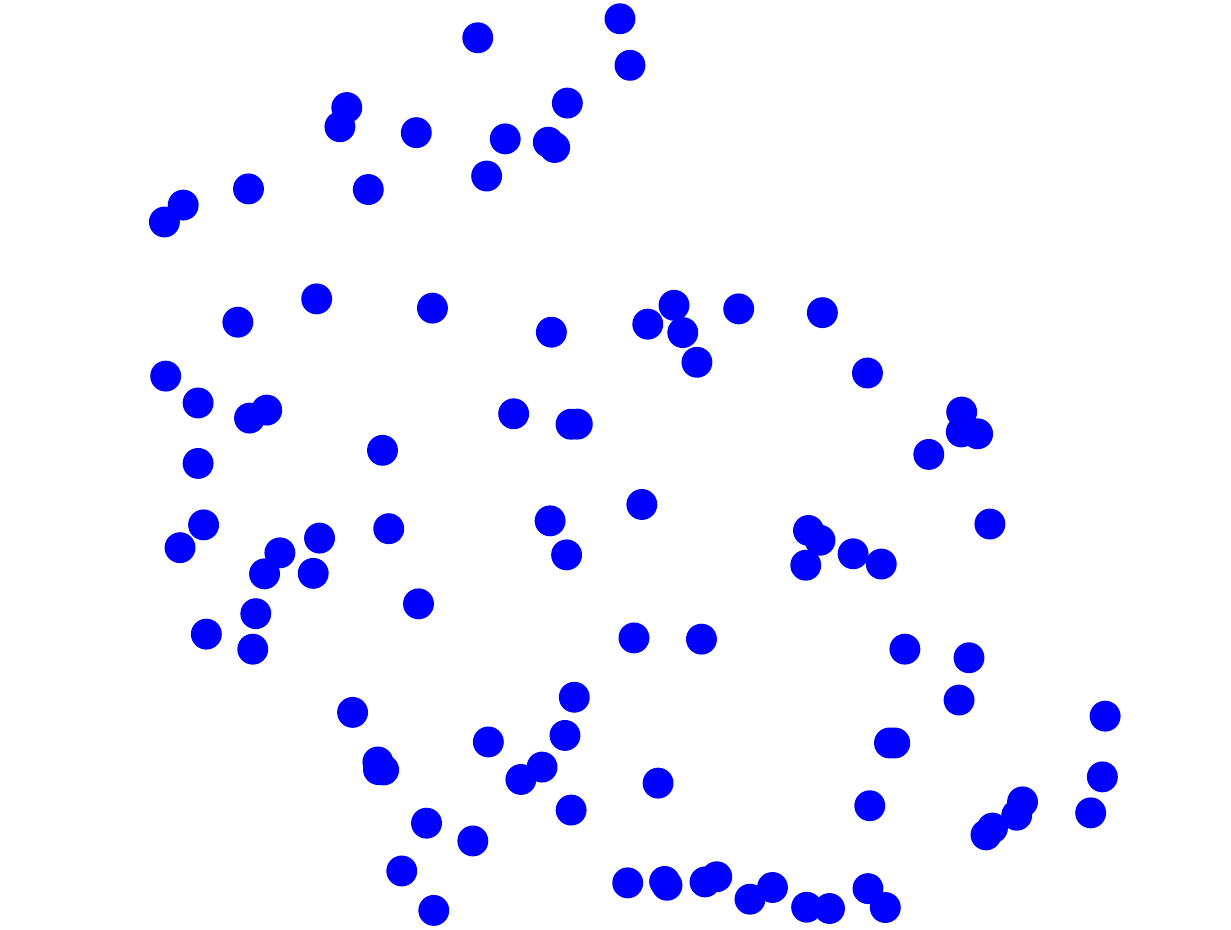}
\end{minipage}

&

\begin{minipage}[t]{0.46\linewidth}
\centering
\includegraphics[width=1\linewidth]{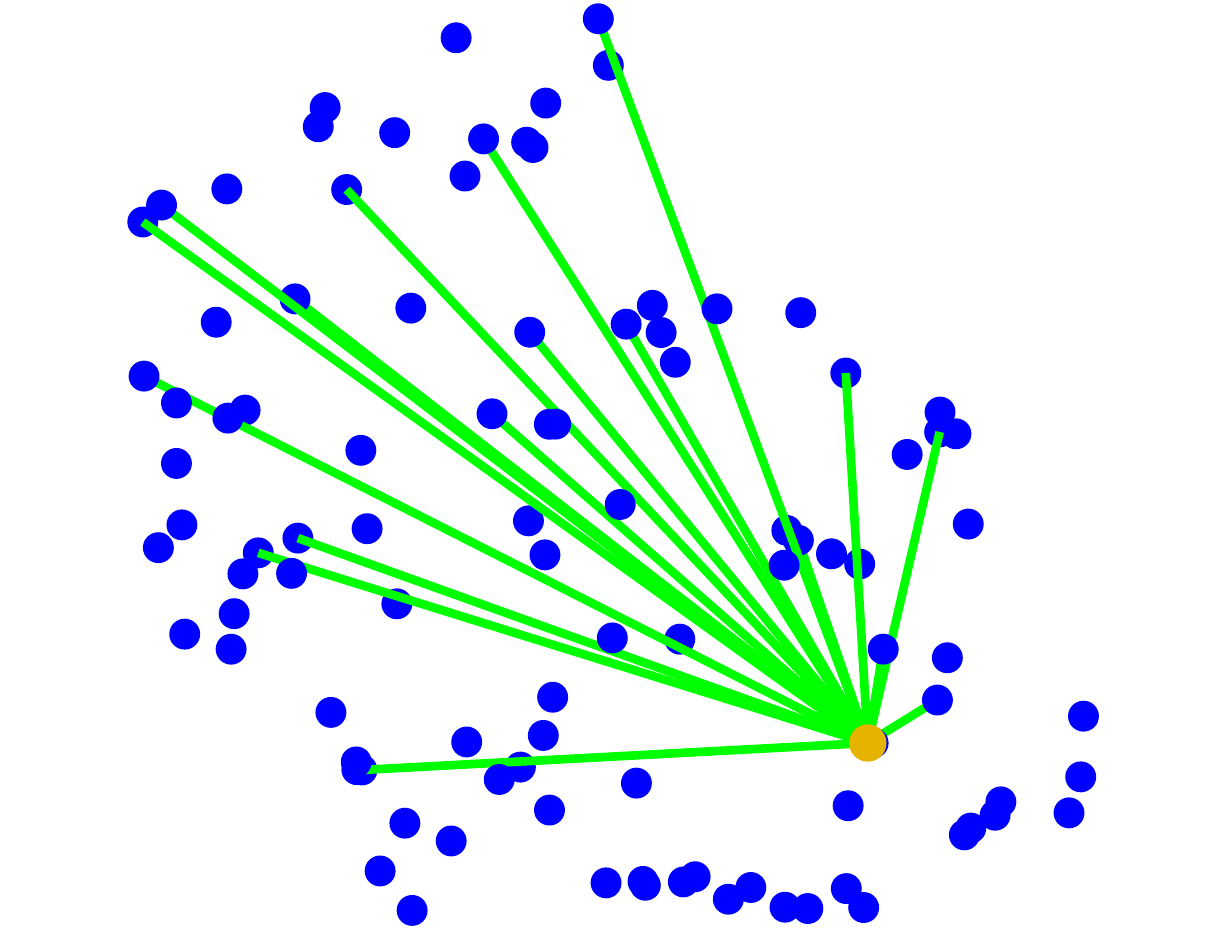}
\end{minipage}

\\
\footnotesize{(c) Second layer}
&
\footnotesize{(d) Third layer}

\\

\begin{minipage}[t]{0.46\linewidth}
\centering
\includegraphics[width=1\linewidth]{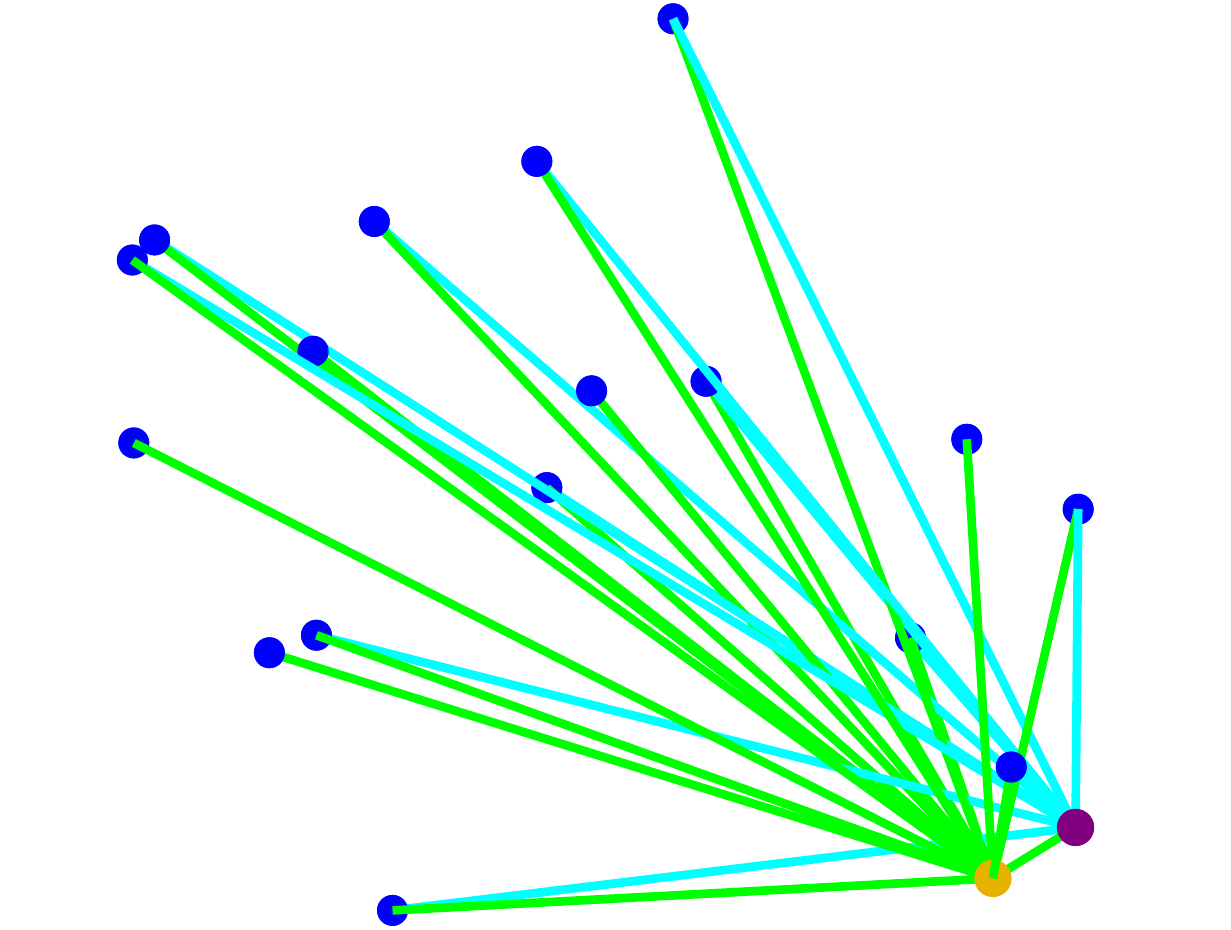}
\end{minipage}

&

\begin{minipage}[t]{0.46\linewidth}
\centering
\includegraphics[width=1\linewidth]{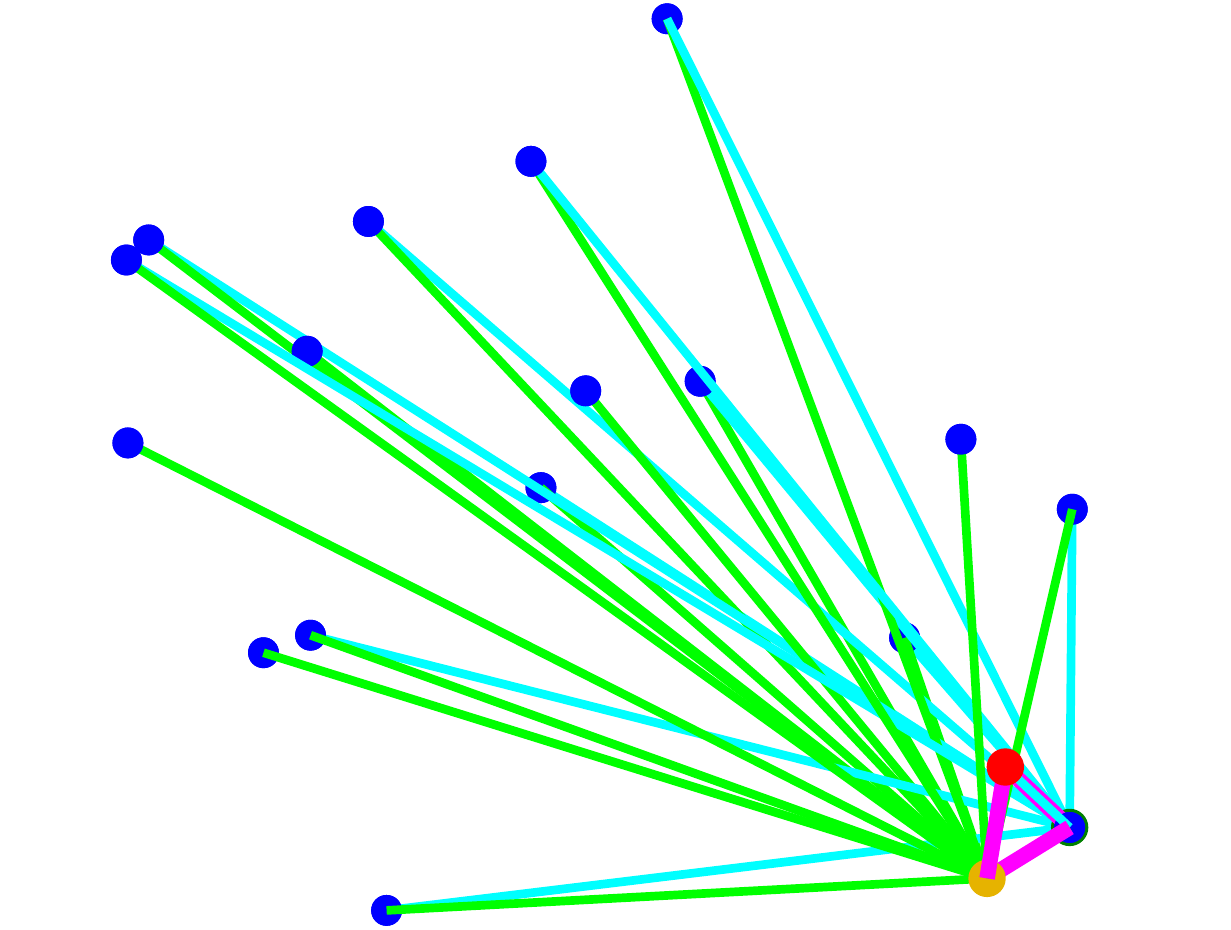}
\end{minipage}

\end{tabular}
\vspace{-1mm}
\caption{Illustration of the main procedures of our robust solver TriVoC. The intuitive descriptions are given throughout Section~\ref{sec-3-point} and \ref{sec-main}.}
\label{demo-concrete}
\vspace{0mm}
\end{figure}

\subsection{3-point Model and Consensus Maximization}\label{sec-3-point}

In registration, 3 points are required to solve the rigid transformation minimally~\cite{horn1987closed}. Our goal is to obtain at least one pure-inlier 3-point set and then further seek the full inlier set with it. Hence, problem~\eqref{CM} can be rewritten as:
\begin{equation}\label{CM-minimal}
\begin{gathered}
\underset{\boldsymbol{C}\subset \boldsymbol{N}}{\max}, \, |\boldsymbol{C}|, \\
s.t. \,\|\boldsymbol{R}^{\circ}\boldsymbol{p}_i+\boldsymbol{t}^{\circ}-\boldsymbol{q}_i\| \leq \gamma,\,(\forall i\in\boldsymbol{C})
\end{gathered}
\end{equation}
where $(\boldsymbol{R}^{\circ},\boldsymbol{t}^{\circ})$ denotes a minimal transformation model computed from a 3-point set and $\boldsymbol{C}$ is its consensus set.

Rather than sampling the 3 points completely randomly as in RANSAC~\cite{fischler1981random} which requires high time cost with high outlier ratios, we prefer to first sort all the putative correspondences according to their probability to be inliers. Voting with the equal-length constraint~\cite{li2021practical} is an efficient way for correspondence sorting. Specifically, if correspondence pair $i$ and $j$ satisfies condition~\eqref{rigidity}, both $i$ and $j$ will get 1 vote. After all correspondence pairs have engaged in voting, we can sort the correspondences according to their respective votes obtained in this process, since the more votes a correspondence can get, the more likely it is to be an inlier.

\begin{figure}[b]
\centering

\setlength\tabcolsep{0.6pt}
\addtolength{\tabcolsep}{0pt}

\begin{tabular}{ccc}

\footnotesize{(a) $N=100$, 95\%}&\footnotesize{(b) $N=500$, 98\%}&\footnotesize{(c) $N=1000$, 99\%}
\\

\begin{minipage}[t]{0.32\linewidth}
\centering
\includegraphics[width=1\linewidth]{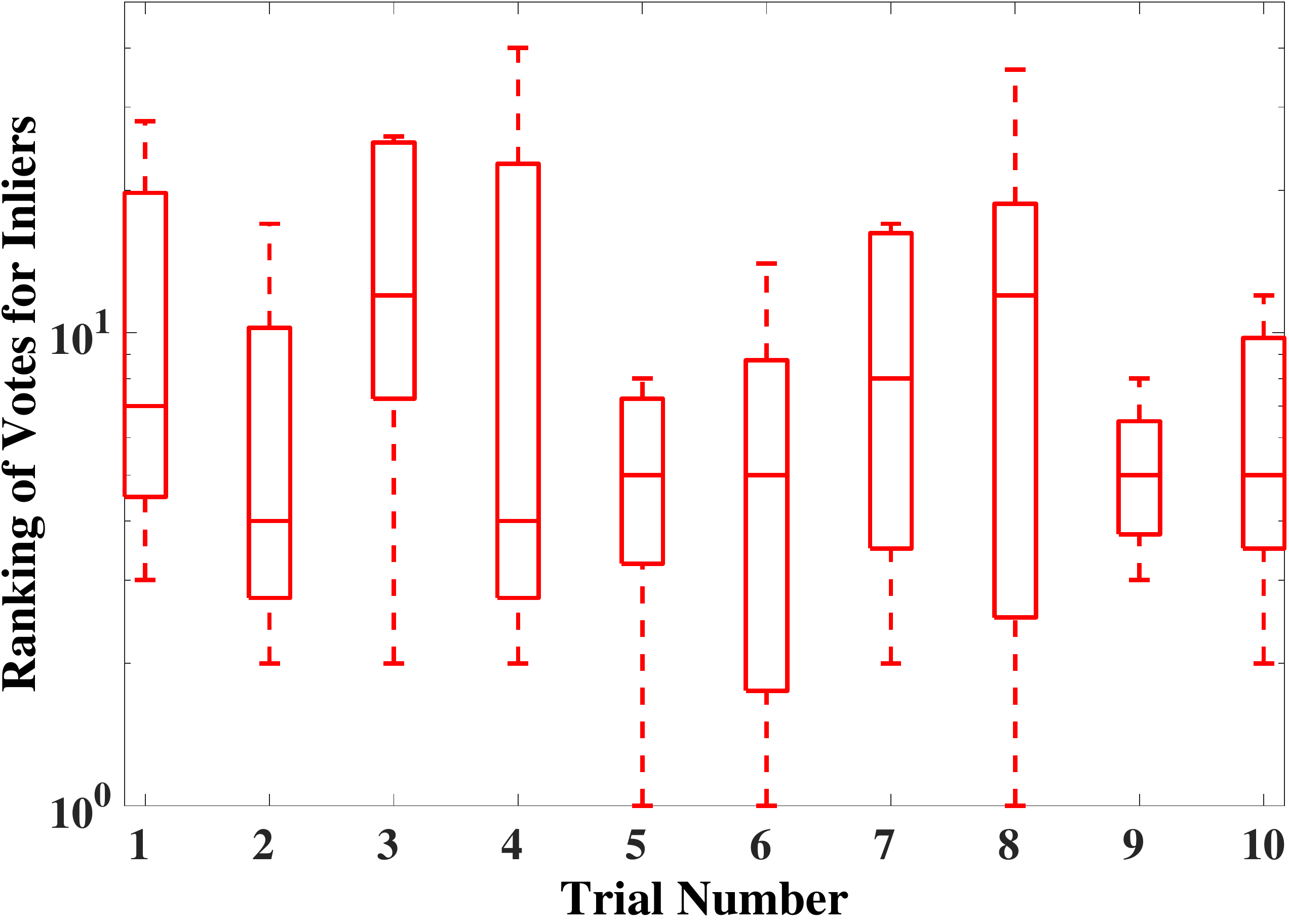}
\end{minipage}
&
\begin{minipage}[t]{0.32\linewidth}
\centering
\includegraphics[width=1\linewidth]{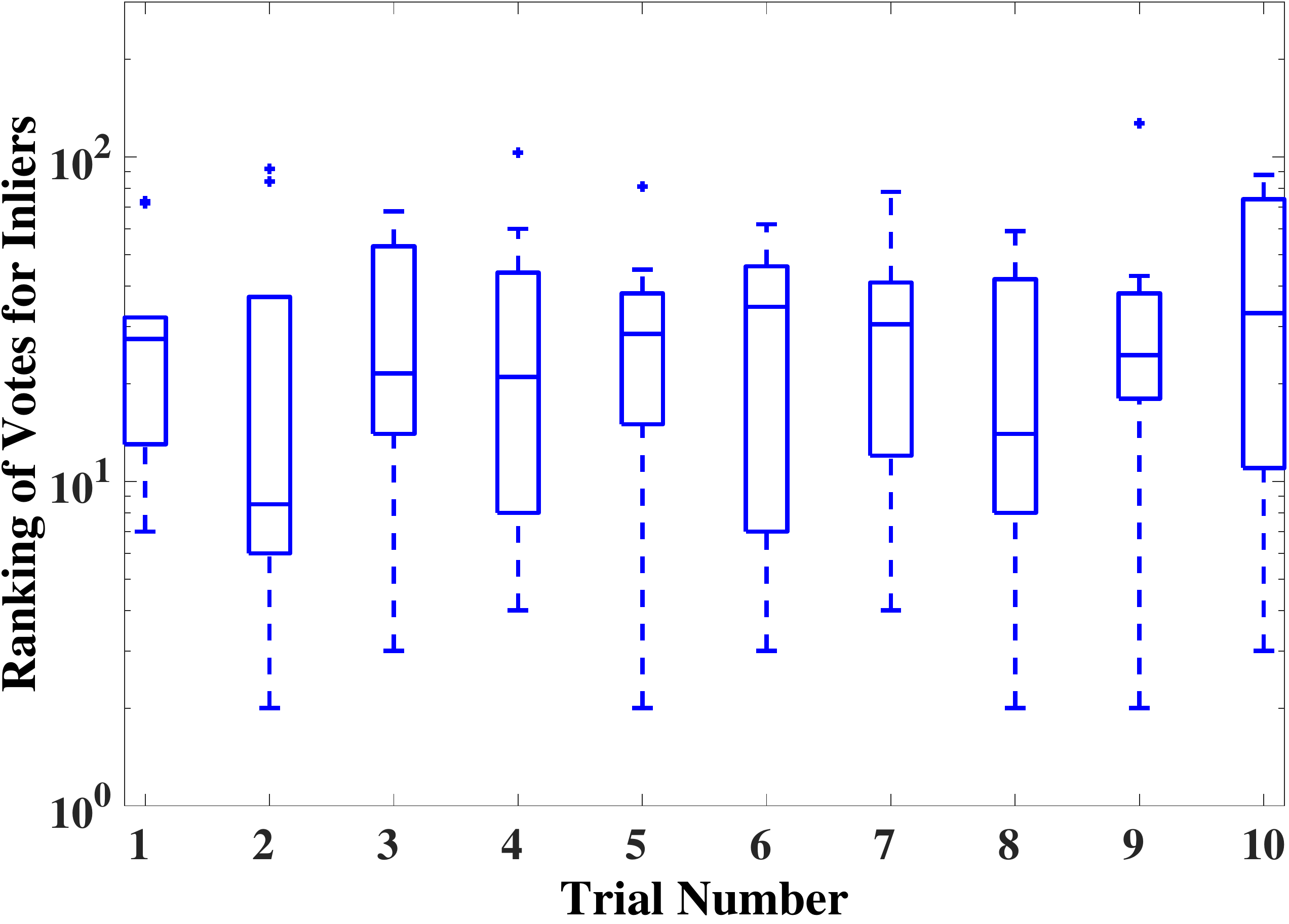}
\end{minipage}
&
\begin{minipage}[t]{0.32\linewidth}
\centering
\includegraphics[width=1\linewidth]{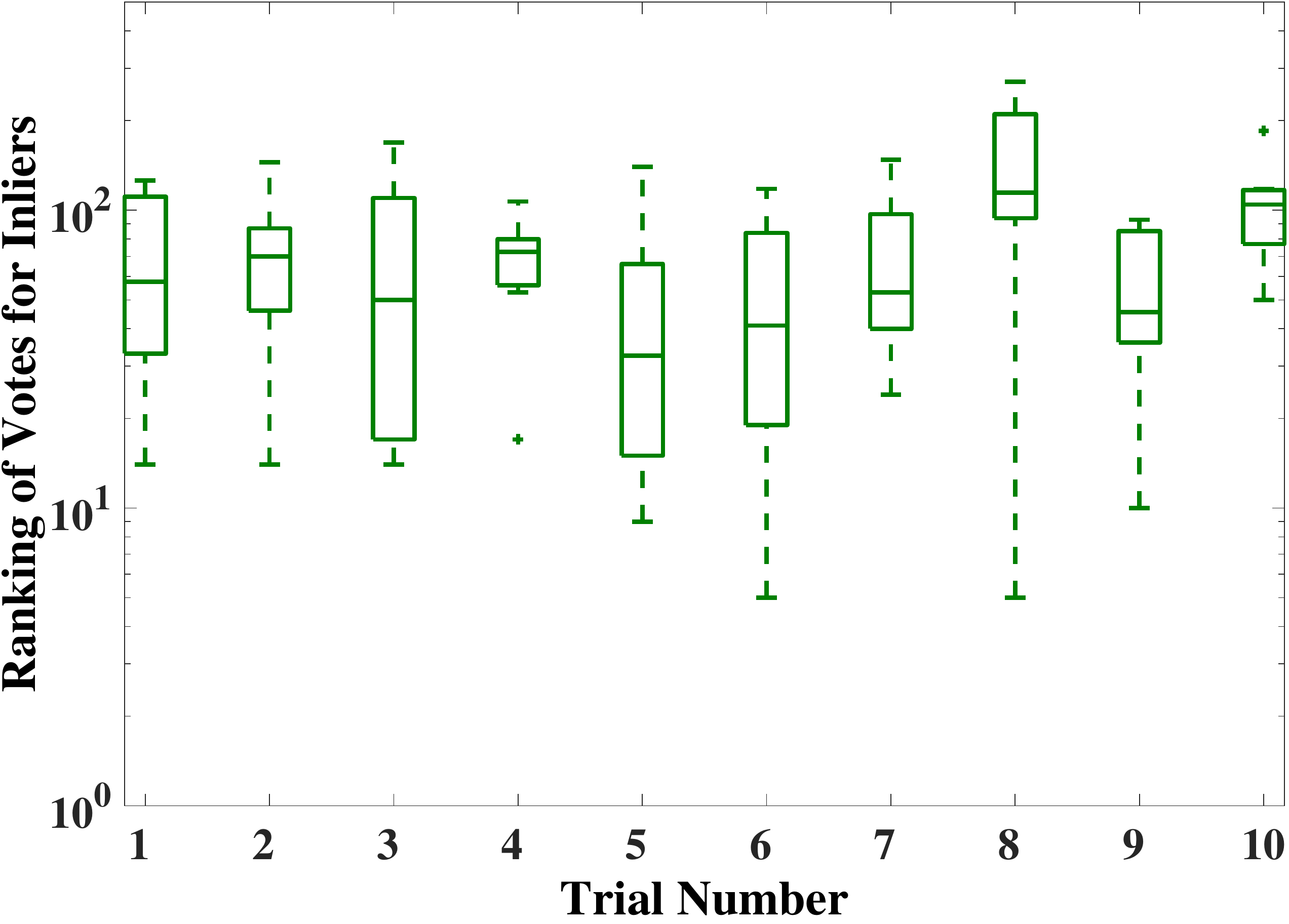}
\end{minipage}

\end{tabular}
\vspace{-2mm}
\caption{Examples of failed equal-length voting and sorting technique under extreme outlier ratios (10 random trials for each correspondence number). We report the ranking of votes obtained by the true inliers. We find that inliers are actually far from the correspondences that get the most votes, so it is infeasible to discern inliers from outliers only using voting and sorting.}
\label{extreme-voting}
\vspace{-1mm}
\end{figure}

However, simply performing such voting can poorly differentiate true inliers from outliers when the outlier ratio is too high (Fig.~\ref{extreme-voting} are typical examples where many outliers get even more votes than the true inliers). Therefore, we first decompose the selection of the 3-point set into 3 consecutive layers where only 1 point is selected in each layer, and then design a smart and fast \textit{triple-layered voting} framework to realize rapid consensus maximization.

\subsection{Main Framework: Triple-layered Voting}\label{sec-main}

At the beginning, we build the consistency matrix $\boldsymbol{M}$ via equal-length tests according to Algorithm~\ref{algo-consistency}. Then, we sort the correspondences via the voting process above, which can be swiftly operated by: (i) computing the sums of all $N$ rows (or columns) in $\boldsymbol{M}$ and (ii) sorting these sums to obtain the associated correspondence order (since each row or column corresponds to one correspondence and each entry equal to 1 corresponds to 1 vote), as illustrated in Algorithm~\ref{algo-sorting}.
\begin{algorithm}[h]
\caption{\textit{sortCorrespondences} (subroutine)}
\label{algo-sorting}
\SetKwInOut{Input}{\textbf{Input}}
\Input{consistency matrix $\boldsymbol{M}$\;}
Set all-zero vector $\boldsymbol{s}\leftarrow[0,0,\dots,0]^{\top}\in\mathbb{R}^{|\boldsymbol{M}|_{row}}$\;
\For{$i=1:|\boldsymbol{M}|_{row}$}{
$\boldsymbol{s}_{(i)}\leftarrow\sum_{j=1}^{|\boldsymbol{M}|_{col}}(\boldsymbol{M}_{(i,j)})$\;
}
Sort $\boldsymbol{s}$ in descending order to get the index vector $\boldsymbol{v}$\;
\Return index vector $\boldsymbol{v}$\;
\end{algorithm}

We can obtain the sorted correspondence index vector by: $\boldsymbol{v}_1\leftarrow$\textit{sortCorrespondences}\,($\boldsymbol{M}$), which makes up the first layer of voting in TriVoC. Then, we select a single correspondence in sequence from vector $\boldsymbol{v}_1$, say $a={\boldsymbol{v}_1}_{(1)}$, which serves as the first point  of the 3-point set, where ${\boldsymbol{v}_1}_{(1)}$ denotes the first entry of $\boldsymbol{v}_1$. After that, we find all the correspondences that can satisfy condition~\eqref{rigidity} with $a$, which can be easily operated by seeking all the entries equal to 1 from the $a_{th}$ row of $\boldsymbol{M}$ where their column numbers are the correspondence indices desired, as shown in Algorithm~\ref{algo-candidates}. Here, we call these correspondences the `\textit{inlier candidates}' w.r.t. $a$, which can be obtained by: $\boldsymbol{N}_2\leftarrow$\textit{findInlierCandidates}\,($a,\boldsymbol{M}$). Intuitively, the insight here consists in that if $a$ is indeed a true inlier, then all the other inliers must lie within $\boldsymbol{N}_2$, which greatly reduces the time cost for finding the second inlier point later. This procedure is illustrated in Fig.~\ref{demo-concrete}(b) where the yellow point denotes $a$ and green lines link $a$ to correspondences in $\boldsymbol{N}_2$.

\begin{algorithm}[h]
\caption{\textit{findInlierCandidates} (subroutine)}
\label{algo-candidates}
\SetKwInOut{Input}{\textbf{Input}}
\Input{correspondence $a$; consistency matrix $\boldsymbol{M}$\;}
Set empty vector (set) $\boldsymbol{N}^*\leftarrow\emptyset$\;
\For{$i=1:|\boldsymbol{M}|_{col}$}{
\If{$\boldsymbol{M}_{(a,i)}=1$ and $i\neq a$}{
$\boldsymbol{N}^*\leftarrow \boldsymbol{N}^* \cup \{i\}$\;
}
}
\Return inlier candidates $\boldsymbol{N}^*$\;
\end{algorithm}

As a result, in the second layer of voting, we only need to operate with the inlier candidates $\boldsymbol{N}_2$. We first obtain a reduced consistency matrix $\boldsymbol{M}_2$ w.r.t. $\boldsymbol{N}_2$ using Algorithm~\ref{algo-reduce} such that $\boldsymbol{M}_2\leftarrow$\textit{getReducedConsistency}\,($\boldsymbol{M},\boldsymbol{N}_2$). Then, we perform voting and correspondence sorting once again so as to obtain the sorted index vector in the second layer: $\boldsymbol{v}_2={\boldsymbol{N}_2}_{(\boldsymbol{v}_2^*)}$ where $\boldsymbol{v}_2^*\leftarrow$\textit{sortCorrespondences}\,($\boldsymbol{M}_2$) and ${\boldsymbol{N}_2}_{(\boldsymbol{v}_2^*)}$ denotes a re-ordered vector of ${\boldsymbol{N}_2}$ indexed by $\boldsymbol{v}_2^*$. Thus, the second point of the 3-point set can be chosen as $b={\boldsymbol{v}_2}_{(1)}$, as shown by the deep purple point in Fig.~\ref{demo-concrete}(c).

Subsequently, similar to the operations in the second layer, we further obtain the inlier candidates w.r.t. $b$ such that $\boldsymbol{N}_3\leftarrow$\textit{findInlierCandidates}\,($b,\boldsymbol{M}_2$) (the cyan lines in Fig.~\ref{demo-concrete}(c) link $b$ to the correspondences in $\boldsymbol{N}_2$), and we then sort the correspondences in set $\boldsymbol{N}_3$ for achieving the index vector in the third layer: $\boldsymbol{v}_3={\boldsymbol{N}_3}_{(\boldsymbol{v}_3^*)}$ where $\boldsymbol{v}_3^*\leftarrow$\textit{sortCorrespondences}\,($\boldsymbol{M}_3$). Subsequently, we pick the last (third) point of the 3-point set such that $c={\boldsymbol{v}_3}_{(1)}$. This procedure corresponds to Fig.~\ref{demo-concrete}(d) where $c$ is the red point.

\begin{algorithm}[h]
\caption{\textit{getReducedConsistency} (subroutine)}
\label{algo-reduce}
\SetKwInOut{Input}{\textbf{Input}}
\Input{consistency matrix $\boldsymbol{M}$; inlier candidate $\boldsymbol{N}^*$\;}
Set a $|\boldsymbol{N}^*|\times|\boldsymbol{N}^*|$ matrix $\boldsymbol{M}^*$\;
\For{$i=1:|\boldsymbol{N}^*|$}{
\For{$j=1:|\boldsymbol{N}^*|$}{
$\boldsymbol{M}^*_{(i,j)}\leftarrow \boldsymbol{M}_{(\boldsymbol{N}^*_{(i)},\boldsymbol{N}^*_{(j)})}$\;
}
}
\Return reduced consistency matrix $\boldsymbol{M}^*$\;
\end{algorithm}

Since that now we have obtained a full 3-point set $[a,b,c]$, we can use it to estimate the minimal transformation model $\boldsymbol{R}^{\circ}$ and $\boldsymbol{t}^{\circ}$ with Horn's traid-based method~\cite{horn1987closed} and build its consensus set $\boldsymbol{C}$. Note that this 3-point set only consists of the first entries of set $\boldsymbol{v}$, $\boldsymbol{v}_2$ and $\boldsymbol{v}_3$, so we need to continously select more
3-point sets in sequence from $\boldsymbol{v}_1$, $\boldsymbol{v}_2$ and $\boldsymbol{v}_3$. 

But it is apparently unnecessary to select all the possible 3-point sets from all correspondences in $\boldsymbol{N}$. The 3 points in the 3 layers are independently selected according to their respective orders of the inlier probability and the raw outliers that cannot satisfy the equal-length constraint have already been eliminated (by the \textit{findInlierCandidates} subroutine), so it is fairly easy to select a pure-inlier set during this process. Now we provide a probabilistic termination condition for each of the 3 layers in order to timely return the maximum consensus set with a strong guarantee.

We adopt the probabilistic computation of maximum iteration in RANSAC. We set 0.99 confidence and derive the maximum iteration number $T_{max}$ as:
\begin{equation}\label{max-itr}
T_{max}=\frac{\log(1-0.99)}{\log\left(1-\frac{X}{Y}\right)},
\end{equation}
where $X$ can be set as $|\boldsymbol{C}|$, $|\boldsymbol{C}|-1$ and $|\boldsymbol{C}|-2$, $Y$ can be set as $|\boldsymbol{N}|$, $|\boldsymbol{N}_2|$ and $|\boldsymbol{N}_3|$ and maximum iteration numbers are $T^{max}_1$, $T^{max}_2$ and $T^{max}_3$ in the 3 layers, respectively.

\begin{algorithm}[t]
\caption{{TriVoC} (main algorithm)}
\label{algo-main}
\SetKwInOut{Input}{\textbf{Input}}
\Input{correspondences $\{(\mathbf{p}_i,\mathbf{q}_i)\}_{i=1}^N$; noise $\sigma$\;}
$\boldsymbol{M}\leftarrow$\textit{buildConsistencyMatrix}\,($\{(\mathbf{p}_i,\mathbf{q}_i)\}_{i=1}^N,\sigma$)\;
$\boldsymbol{v}_1\leftarrow$\textit{sortCorrespondences}\,($\boldsymbol{M}$), $C^{max}\leftarrow0$\;
\For{$i=1:\min\left(N,T^{max}_1\right)$}{
$a\leftarrow{\boldsymbol{v}_1}_{(i)}$, $\boldsymbol{N}_2\leftarrow$\textit{findInlierCandidates}\,($a,\boldsymbol{M}$)\;
$\boldsymbol{M}_2\leftarrow$\textit{getReducedConsistency}\,($\boldsymbol{M},\boldsymbol{N}_2$)\;
$\boldsymbol{v}_2^*\leftarrow$\textit{sortCorrespondences}\,($\boldsymbol{M}_2$), $\boldsymbol{v}_2\leftarrow{\boldsymbol{N}_2}_{(\boldsymbol{v}_2^*)}$\;
\For{$j=1:\min\left(|\boldsymbol{N}_2|,T^{max}_2\right)$}{
$b\leftarrow{\boldsymbol{v}_2}_{(j)}$\;
$\boldsymbol{N}_3\leftarrow$\textit{findInlierCandidates}\,($b,\boldsymbol{M}_2$)\;
$\boldsymbol{M}_3\leftarrow$\textit{getReducedConsistency}\,($\boldsymbol{M},\boldsymbol{N}_3$)\;
$\boldsymbol{v}_3^*\leftarrow$\textit{sortCorrespondences}\,($\boldsymbol{M}_3$), $\boldsymbol{v}_3\leftarrow{\boldsymbol{N}_3}_{(\boldsymbol{v}_3^*)}$\;
\For{$k=1:\min\left(|\boldsymbol{N}_3|,T^{max}_3\right)$}{
$c\leftarrow{\boldsymbol{v}_3}_{(k)}$, and solve $(\boldsymbol{R}^{\circ},\boldsymbol{t}^{\circ})$ minimally\;
Find the consensust set $\boldsymbol{C}$ with $(\boldsymbol{R}^{\circ},\boldsymbol{t}^{\circ})$\;
\If{$|\boldsymbol{C}|>C^{best}$}{
$\boldsymbol{C}^{best}\leftarrow\boldsymbol{C}$, $C^{max}\leftarrow|\boldsymbol{C}|$, and update $T^{max}_1$, $T^{max}_2$ and $T^{max}_3$ with~\eqref{max-itr}\;
}
}
}
}
Solve optimal $(\boldsymbol{R}^{\star}, \boldsymbol{t}^{\star})$ with $\boldsymbol{C}^{best}$ using SVD~\cite{arun1987least}\;
\Return best transformation $\boldsymbol{R}^{\star}$ and $\boldsymbol{t}^{\star}$\;
\end{algorithm}

Note that the maximum iteration numbers here are significantly smaller than that in RANSAC, because: (i) the problem dimension is always 1 since the 3 points are selected independently in 3 layers, and (ii) the ratio $\frac{X}{Y}$ here is much greater than actual inlier ratio since a huge number of raw outliers have already been removed by the equal-length constraint. Moreover, since our correspondences are selected according to the vote numbers (from the biggest to smallest) rather than completely randomly, so the actual confidence to obtain a pure-inlier 3-point set should be essentially higher than 0.99. Consequently, though it is hard to measure the exact confidence, we can state that the guarantee of obtaining the maximum consensus from at least one pure-inlier 3-point set is rather strong (will be shown in experiments).

\subsection{Main Algorithm}

We provide the pseudocode of the main algorithm of the proposed solver TriVoC in Algorithm~\ref{algo-main}.

\section{Experiments}\label{sec-experiments}

We conduct various experiments on real datasets to evaluate the performance of TriVoC, compared with other state-of-the-art competitors. All experiments are conducted in Matlab on a laptop with an i7-7700HQ CPU and 16GB of RAM.

\begin{figure}[t]
\centering

\begin{minipage}[t]{1\linewidth}
\centering
\includegraphics[width=0.49\linewidth]{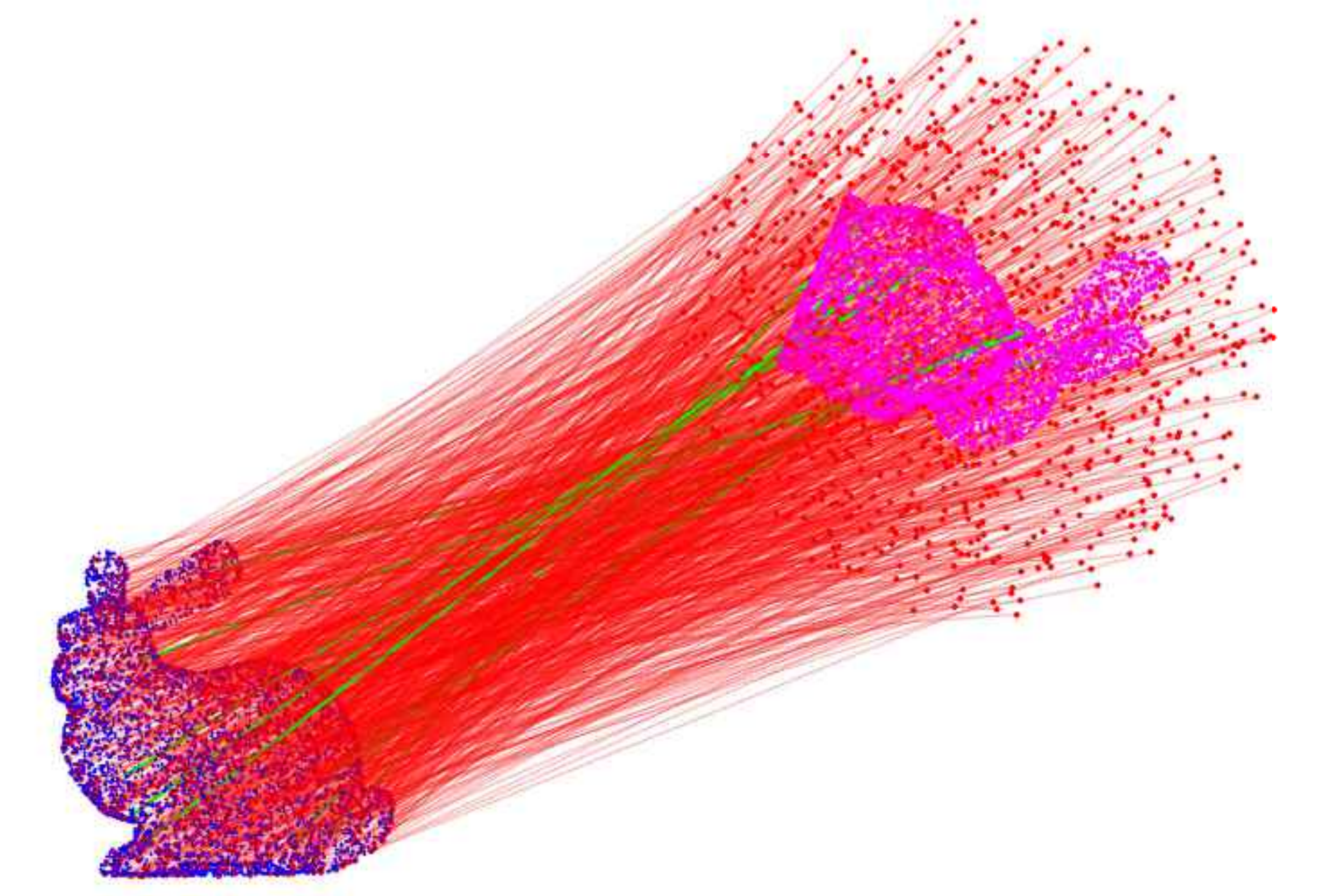}
\includegraphics[width=0.49\linewidth]{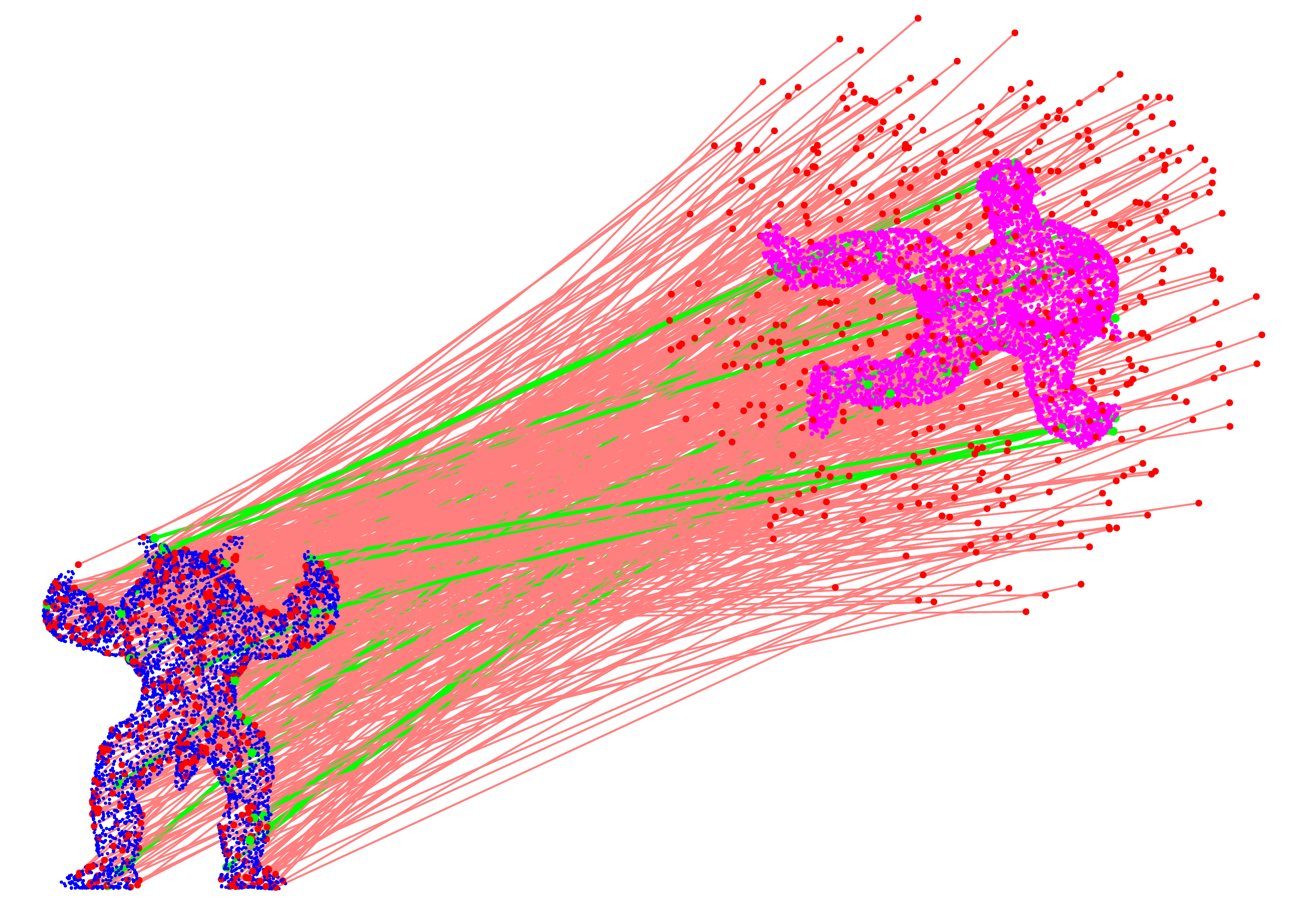}
\end{minipage}

\caption{Environmental setup of the standard benchmarking experiments. Left: A registration example with $N=1000$ and 99\% outliers on \textit{bunny}. Right: A registration example with $N=500$ and 95\% outliers on \textit{armadillo}.}
\label{Setup}
\vspace{-1mm}
\end{figure}

\begin{figure*}[ht]
\centering


\setlength\tabcolsep{1pt}
\addtolength{\tabcolsep}{0pt}

\begin{tabular}{ccc}

\footnotesize{(a) Results on \textit{bunny} with $N=100$}
&
\footnotesize{(b) Results on \textit{bunny} with $N=500$}
&
\footnotesize{(c) Results on \textit{bunny} with $N=1000$}
\\

\begin{minipage}[t]{0.3\linewidth}
\centering
\includegraphics[width=1\linewidth]{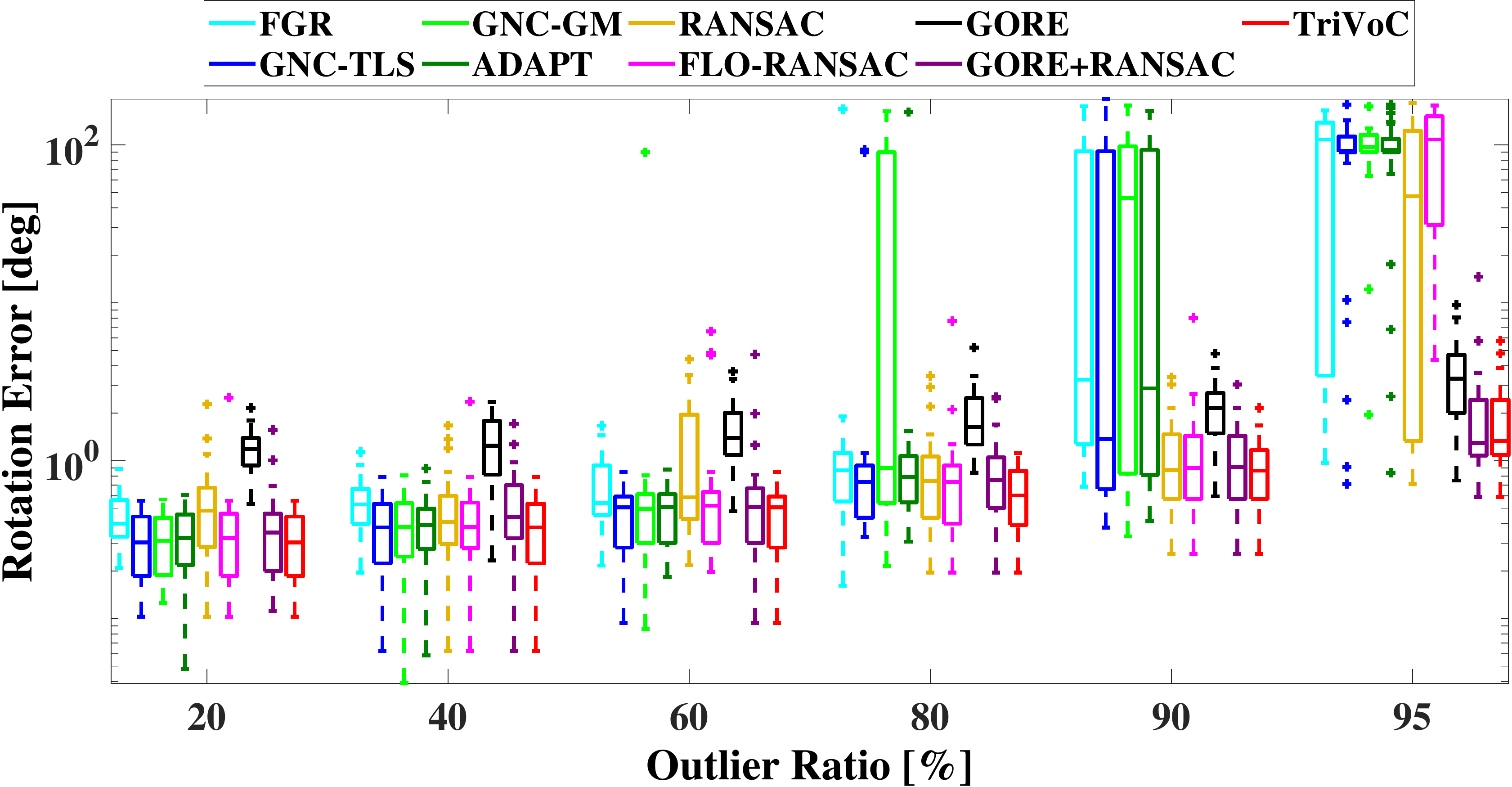}
\end{minipage}
&
\begin{minipage}[t]{0.3\linewidth}
\centering
\includegraphics[width=1\linewidth]{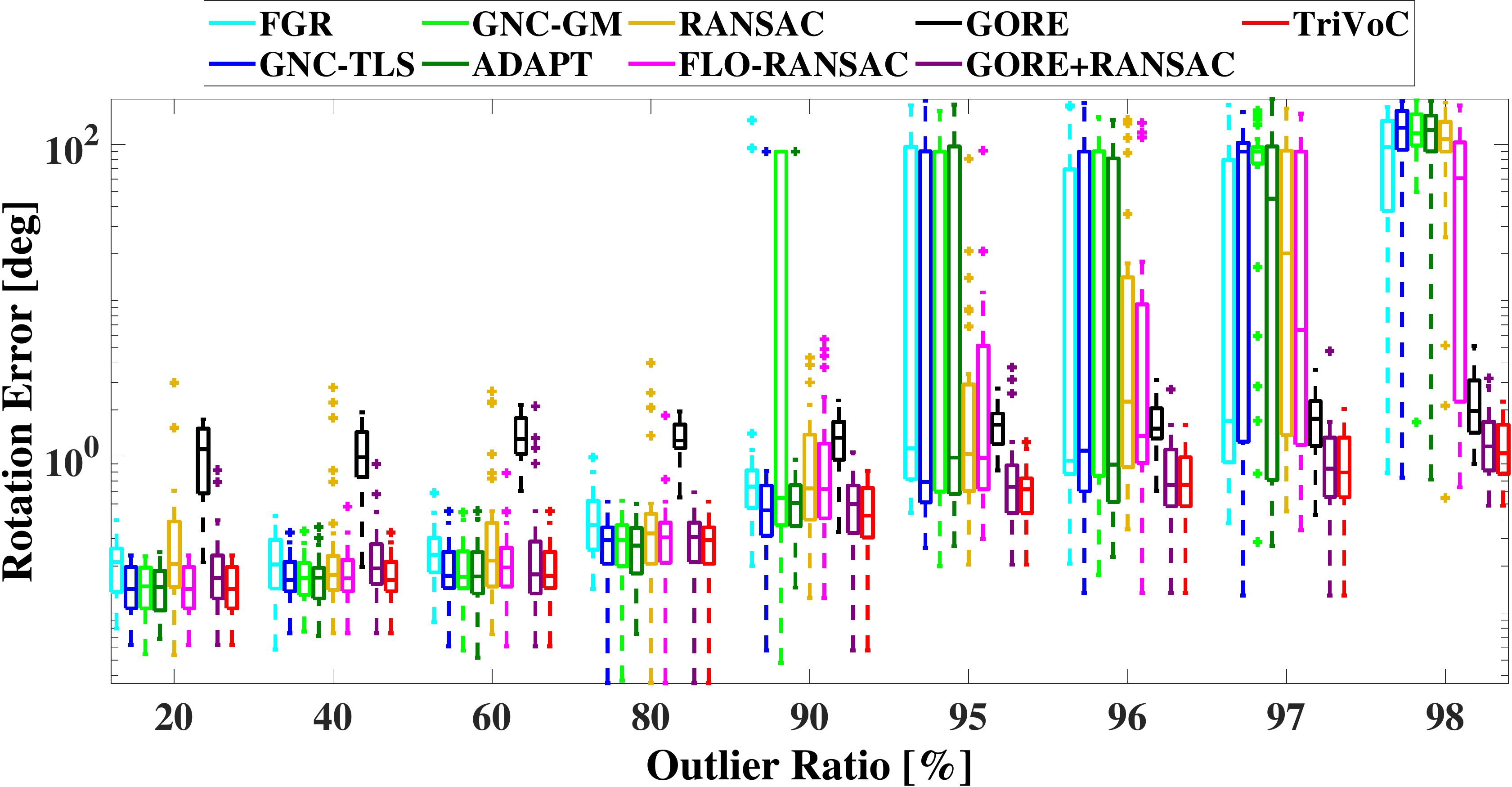}
\end{minipage}
&
\begin{minipage}[t]{0.3\linewidth}
\centering
\includegraphics[width=1\linewidth]{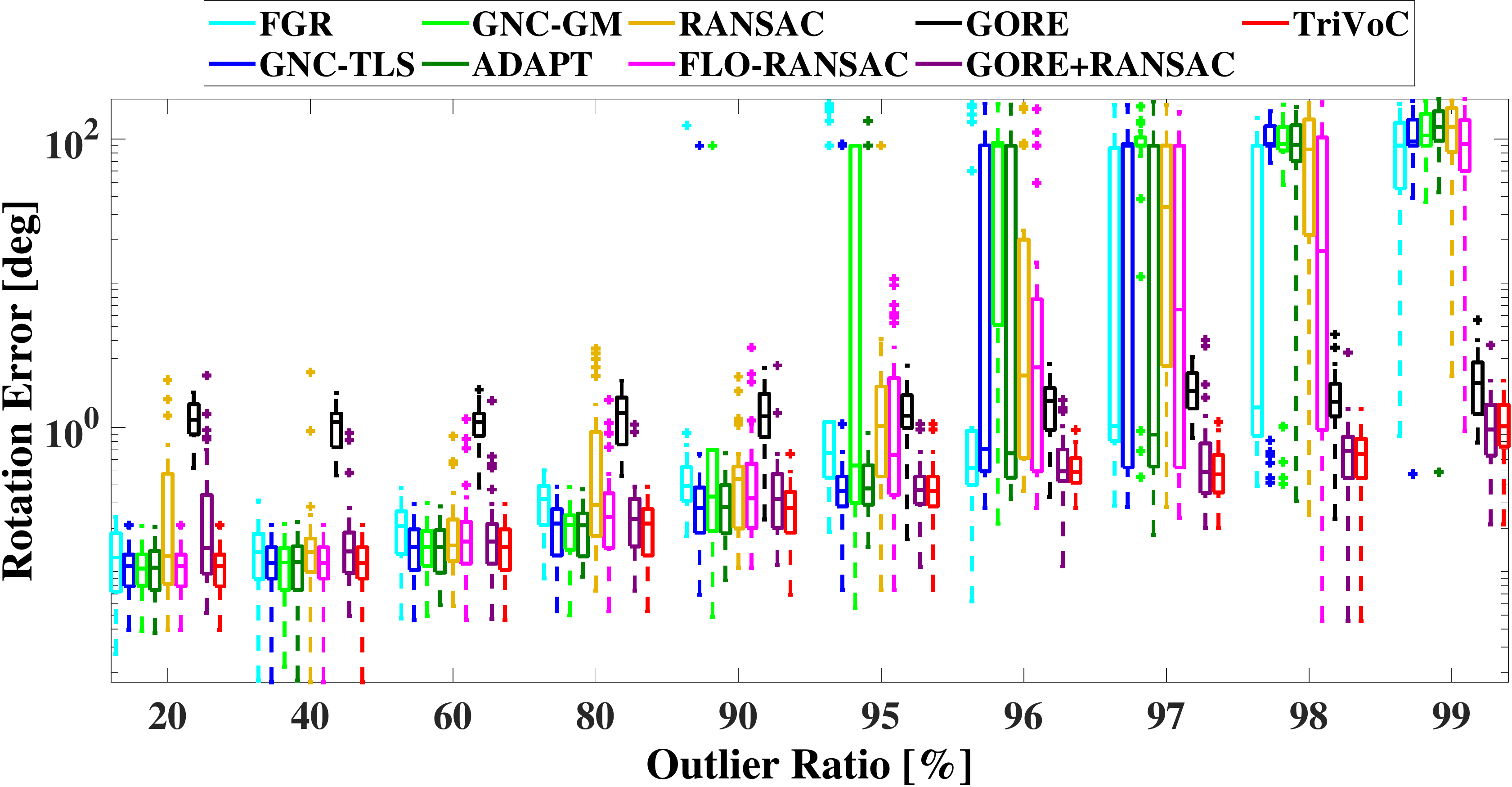}
\end{minipage}

\\
\begin{minipage}[t]{0.3\linewidth}
\includegraphics[width=1\linewidth]{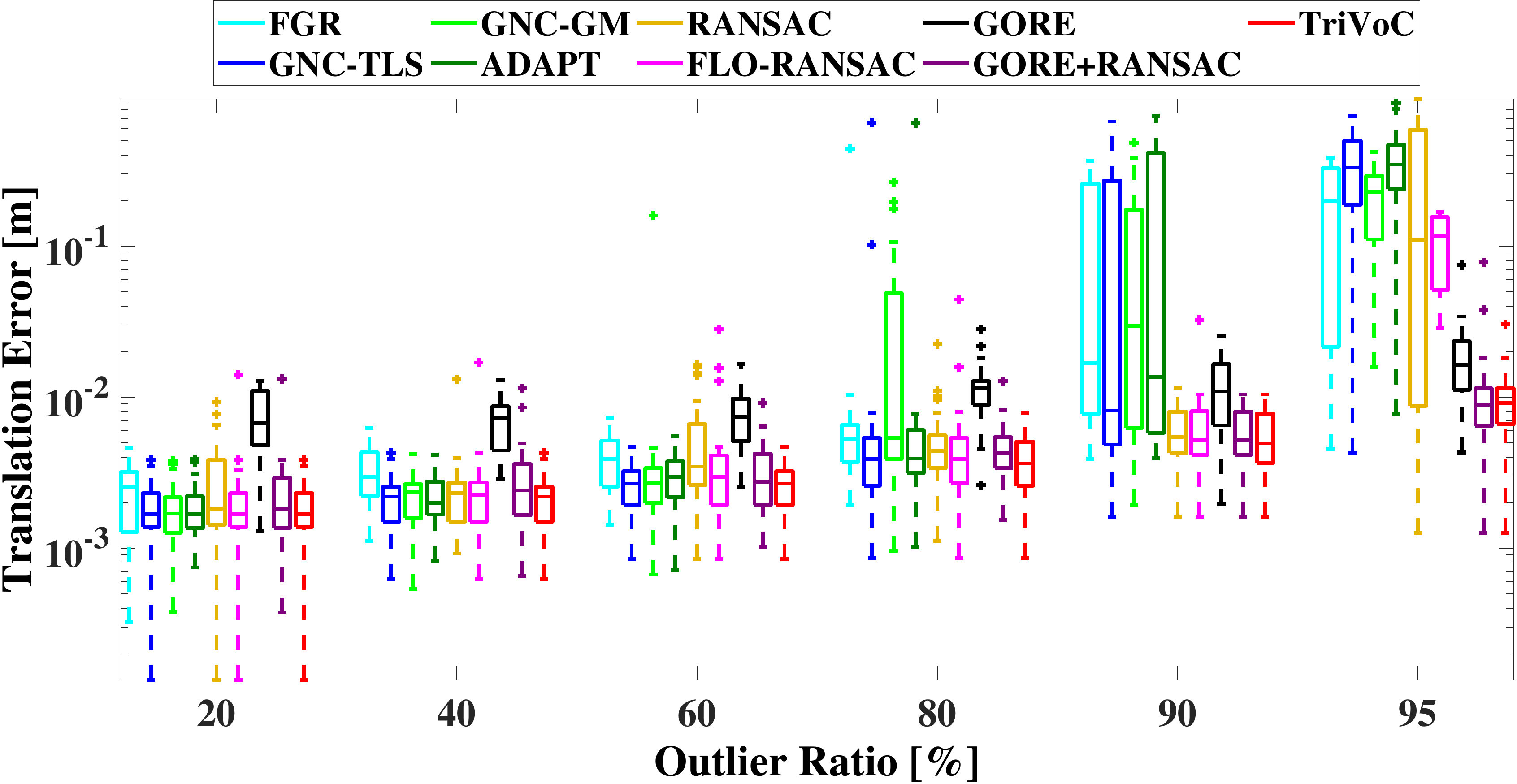}
\end{minipage}
&
\begin{minipage}[t]{0.3\linewidth}
\centering
\includegraphics[width=1\linewidth]{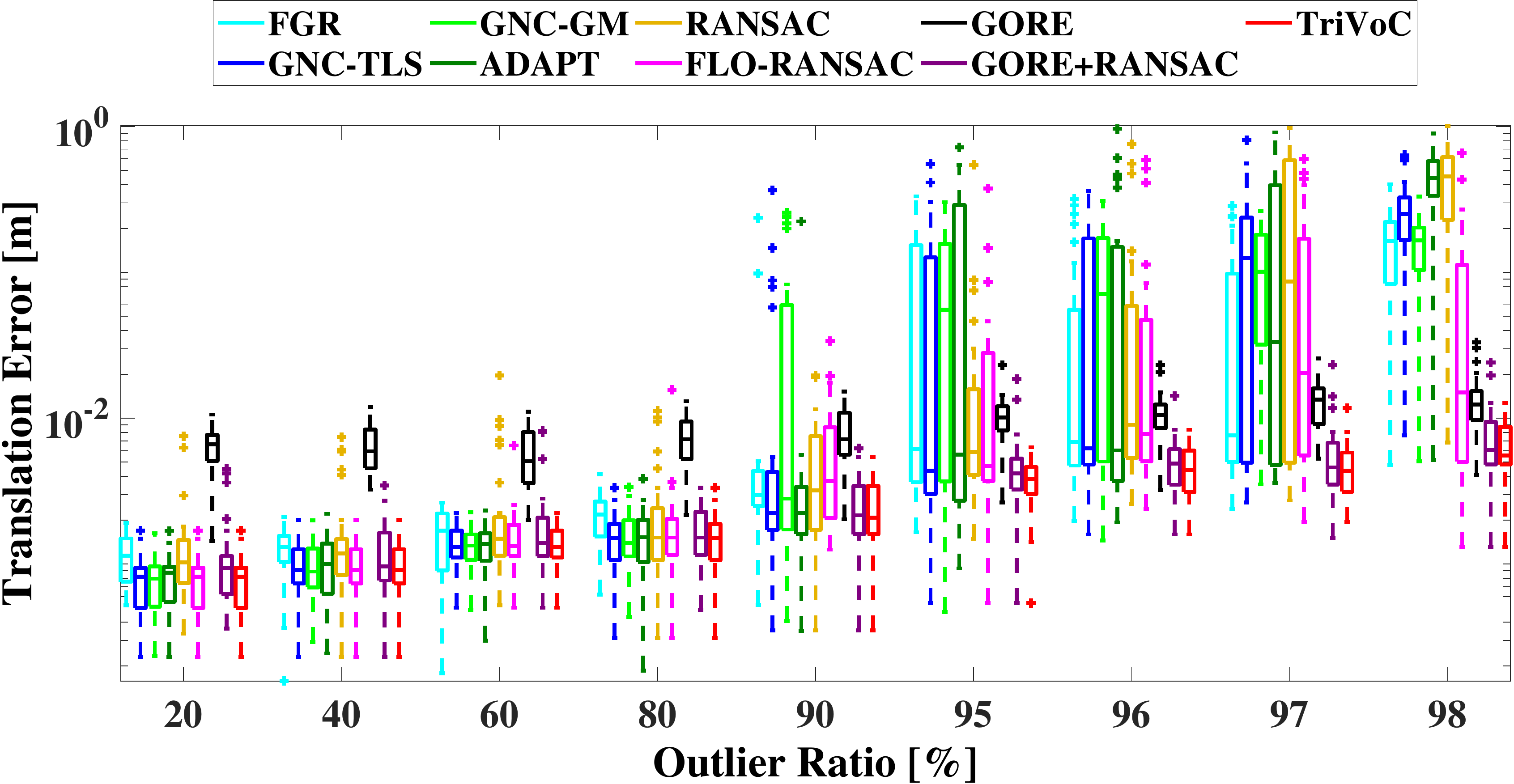}
\end{minipage}
&
\begin{minipage}[t]{0.3\linewidth}
\centering
\includegraphics[width=1\linewidth]{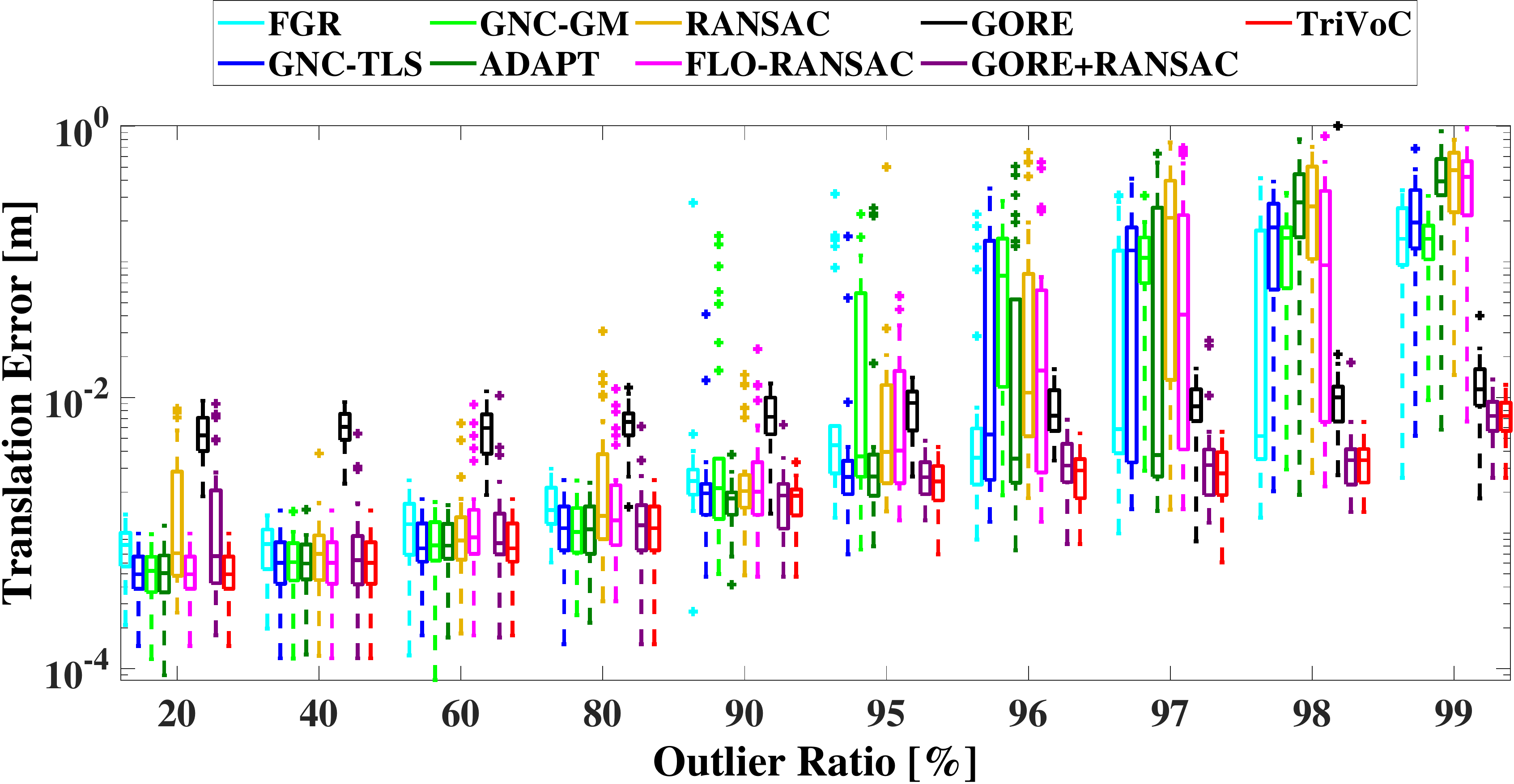}
\end{minipage}

\\
\begin{minipage}[t]{0.3\linewidth}
\includegraphics[width=1\linewidth]{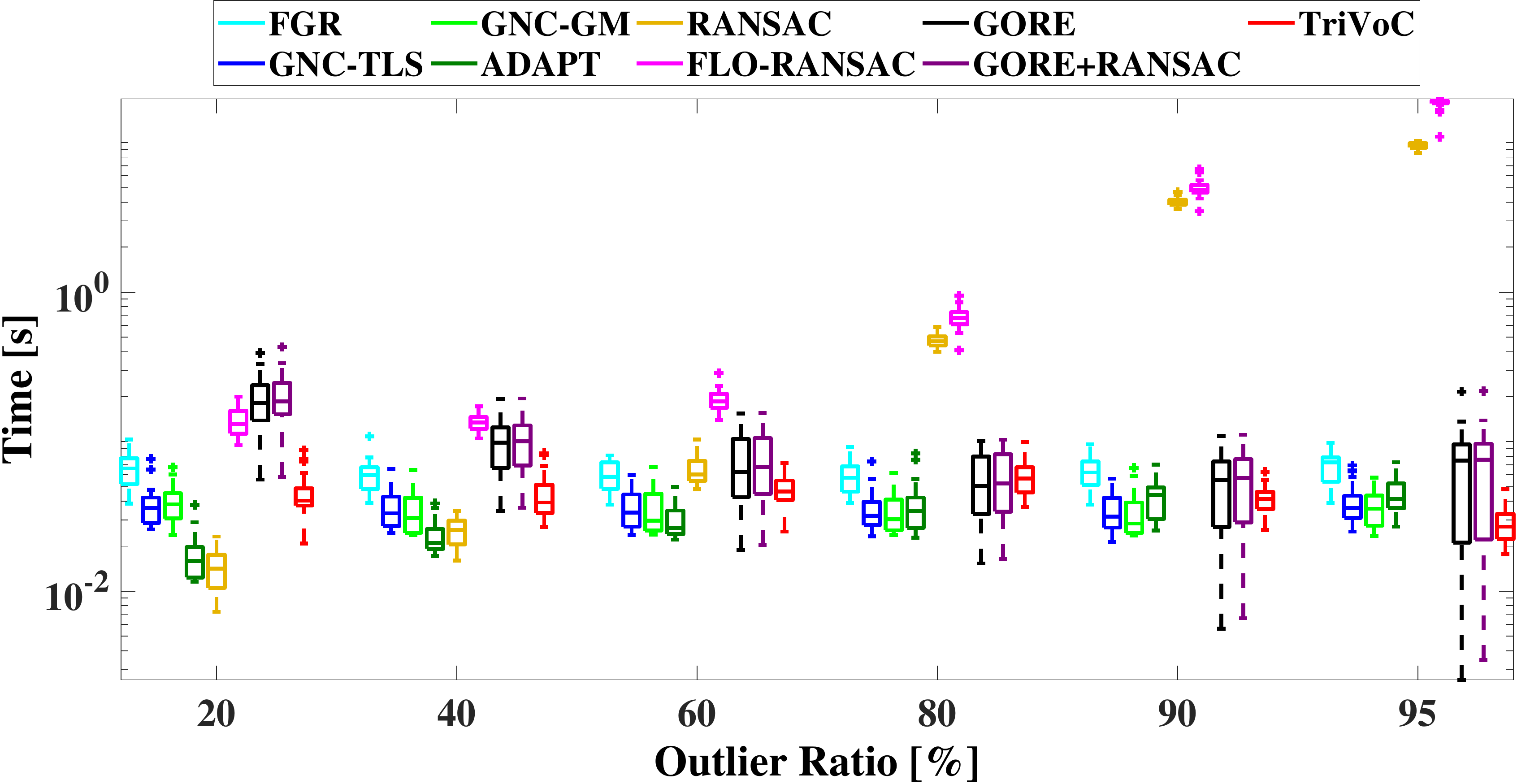}
\end{minipage}
&
\begin{minipage}[t]{0.3\linewidth}
\centering
\includegraphics[width=1\linewidth]{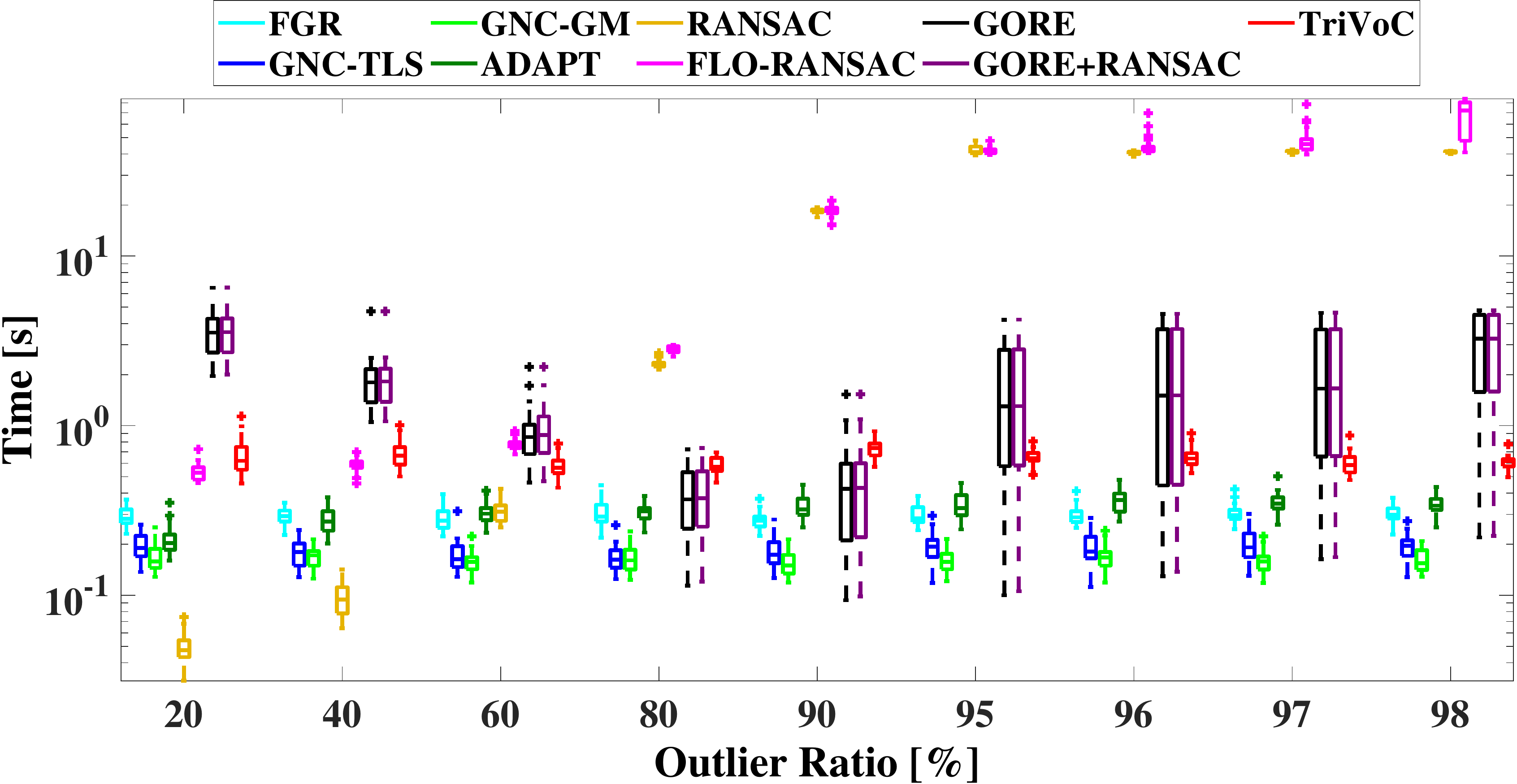}
\end{minipage}
&
\begin{minipage}[t]{0.3\linewidth}
\centering
\includegraphics[width=1\linewidth]{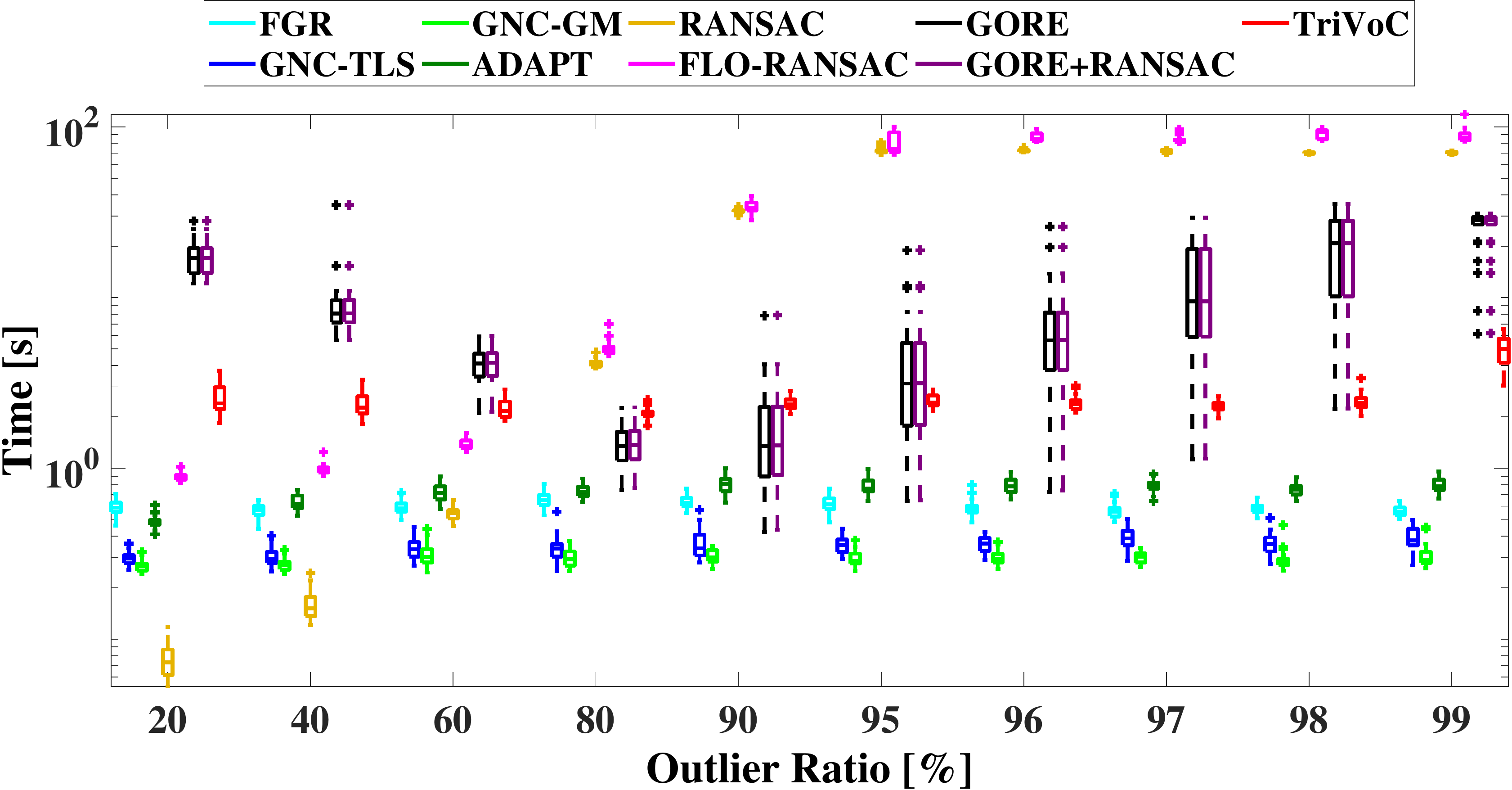}
\end{minipage}

\end{tabular}

\vspace{0.5mm}
\setlength\tabcolsep{1pt}
\addtolength{\tabcolsep}{0pt}

\begin{tabular}{ccc}

\footnotesize{(e) Results on \textit{armadillo} with $N=100$}
&
\footnotesize{(f) Results on \textit{armadillo} with $N=500$}
&
\footnotesize{(g) Results on \textit{armadillo} with $N=1000$}
\\

\begin{minipage}[t]{0.3\linewidth}
\centering
\includegraphics[width=1\linewidth]{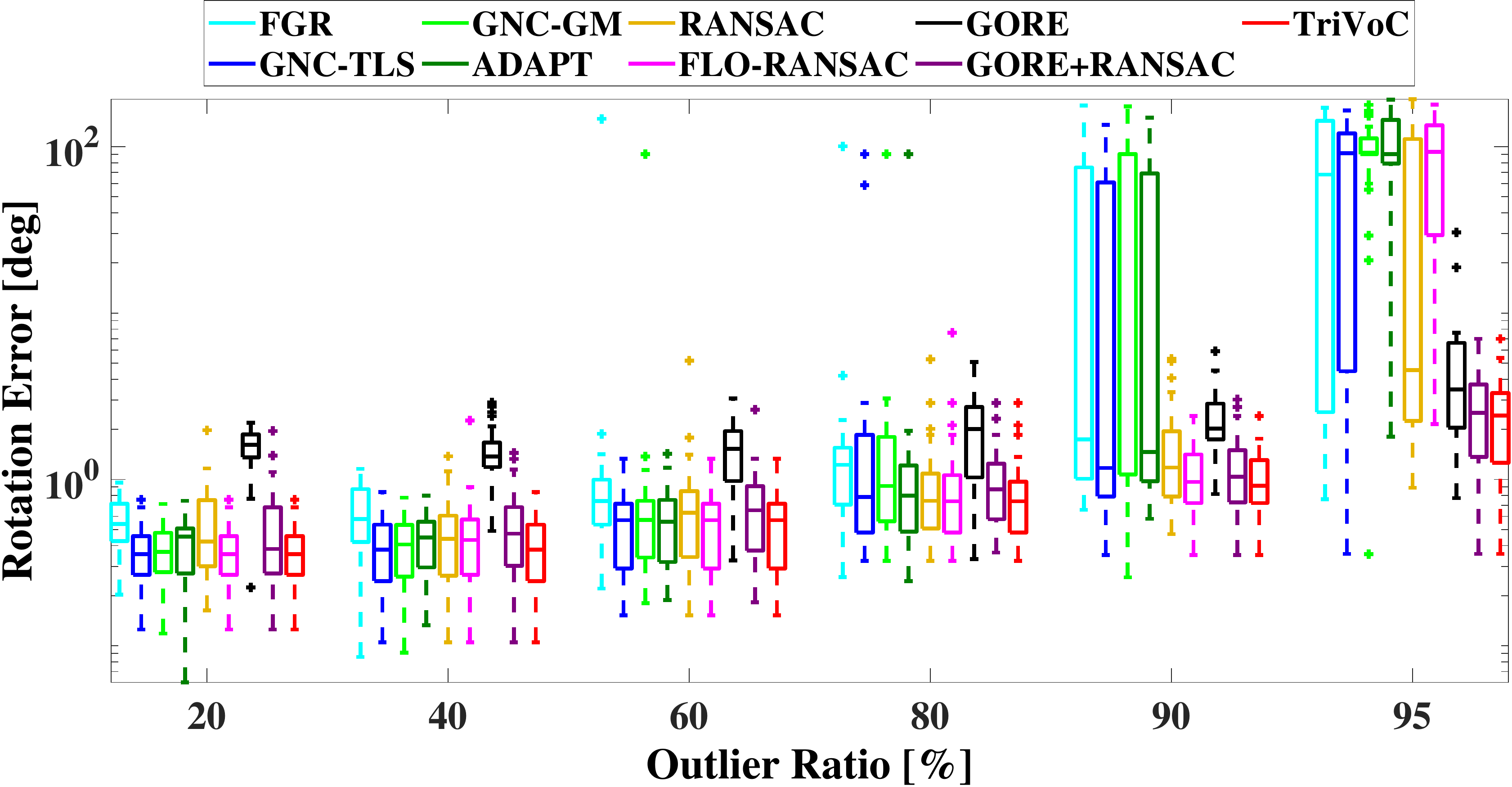}
\end{minipage}
&
\begin{minipage}[t]{0.3\linewidth}
\centering
\includegraphics[width=1\linewidth]{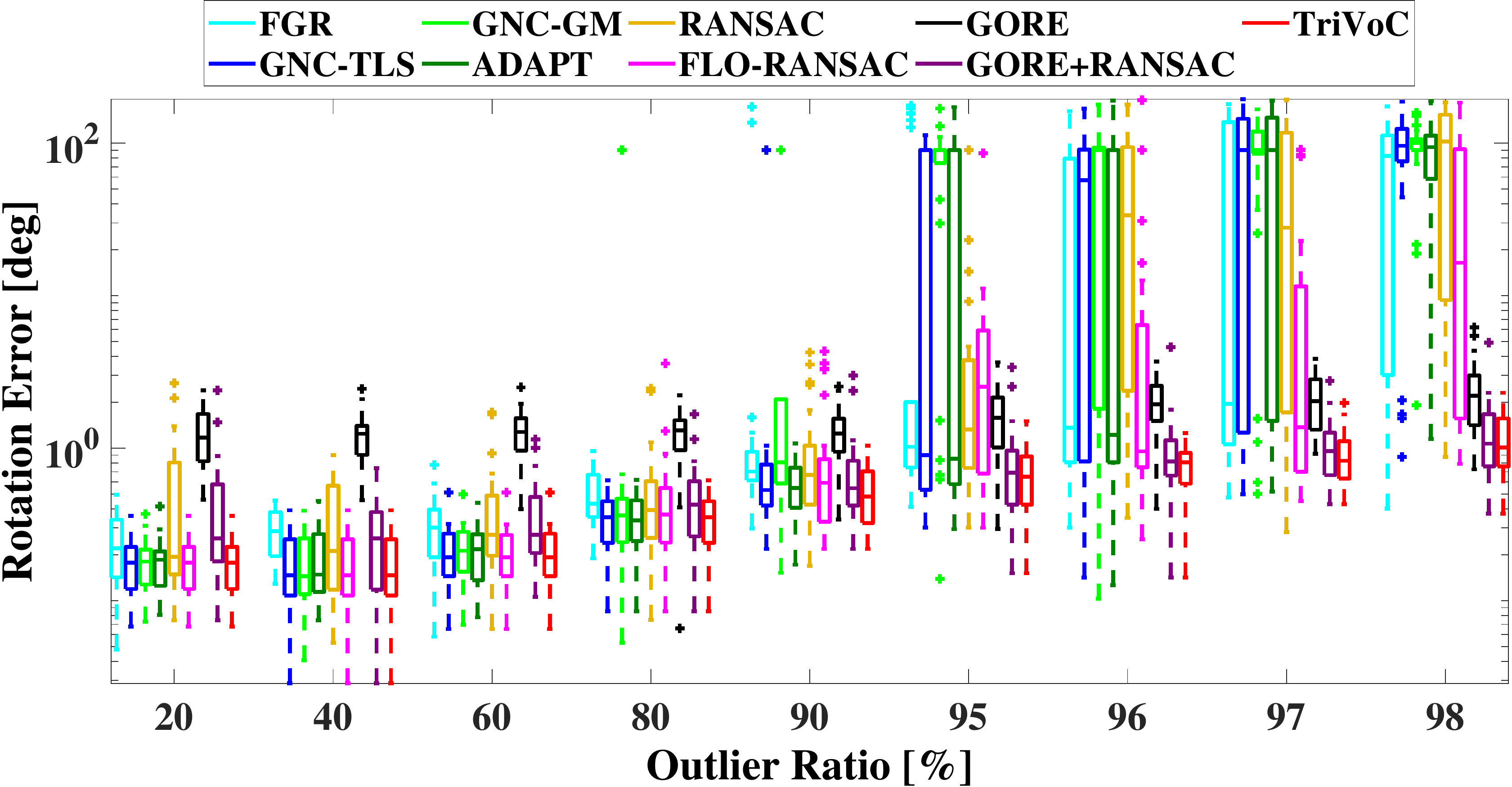}
\end{minipage}
&
\begin{minipage}[t]{0.3\linewidth}
\centering
\includegraphics[width=1\linewidth]{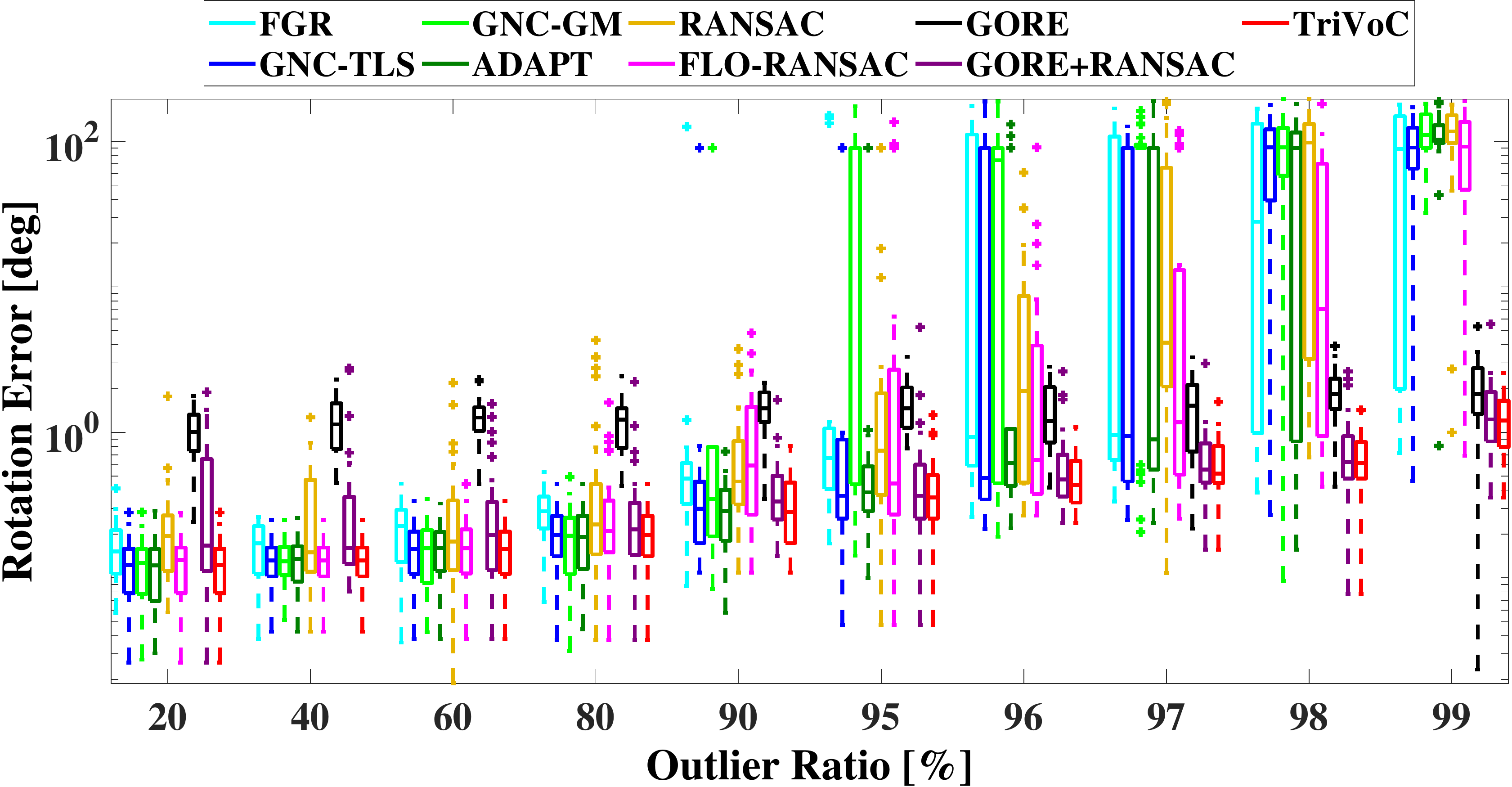}
\end{minipage}

\\
\begin{minipage}[t]{0.3\linewidth}
\includegraphics[width=1\linewidth]{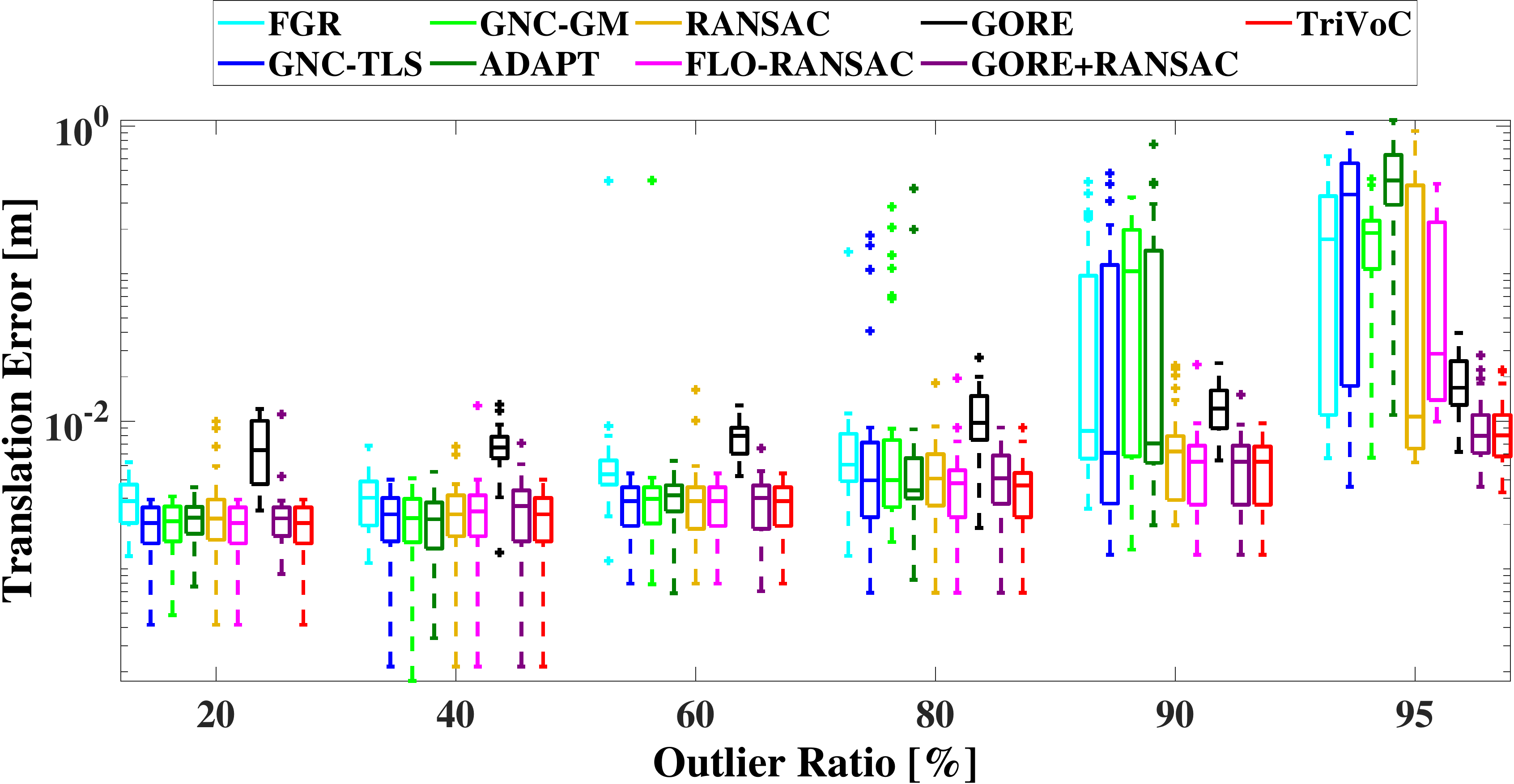}
\end{minipage}
&
\begin{minipage}[t]{0.3\linewidth}
\centering
\includegraphics[width=1\linewidth]{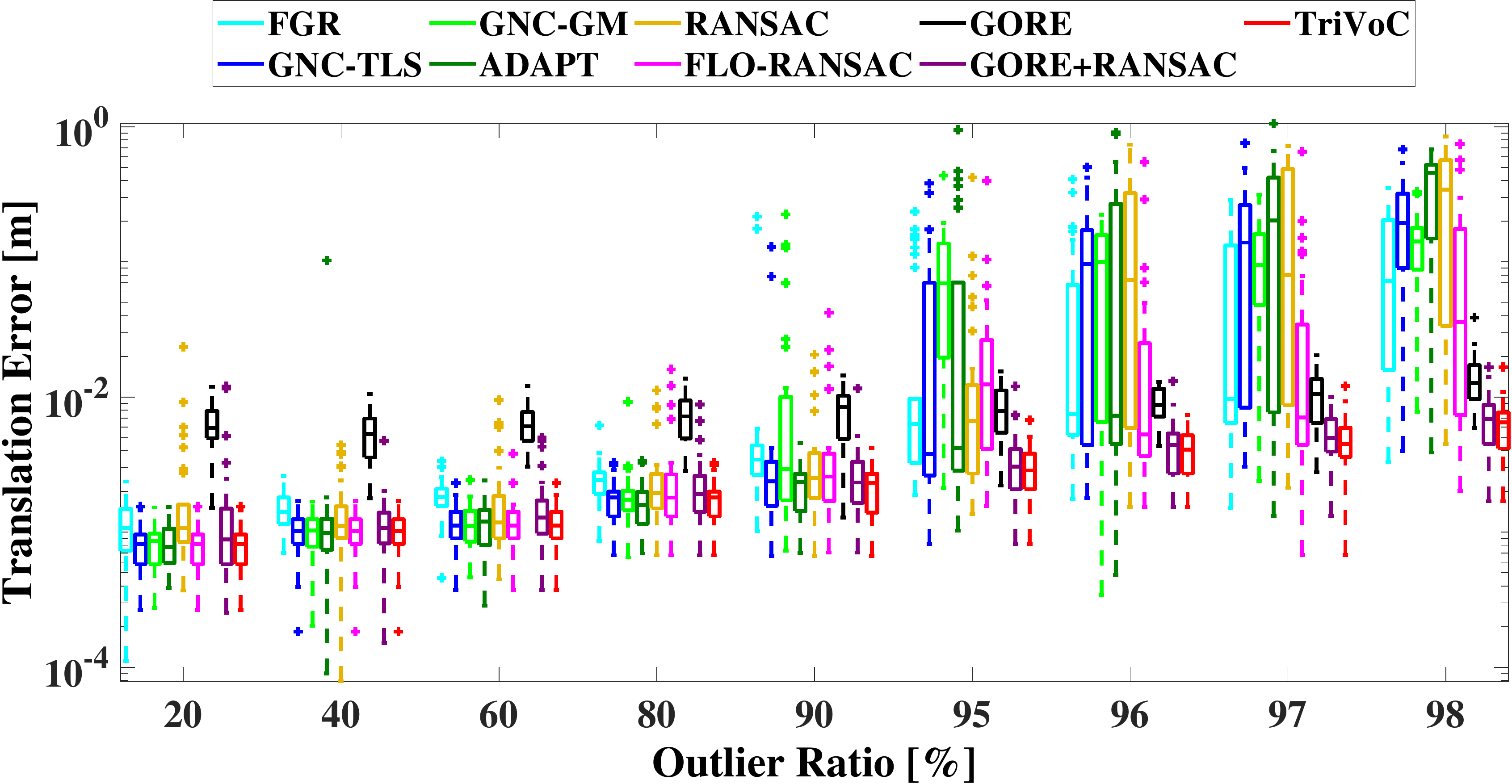}
\end{minipage}
&
\begin{minipage}[t]{0.3\linewidth}
\centering
\includegraphics[width=1\linewidth]{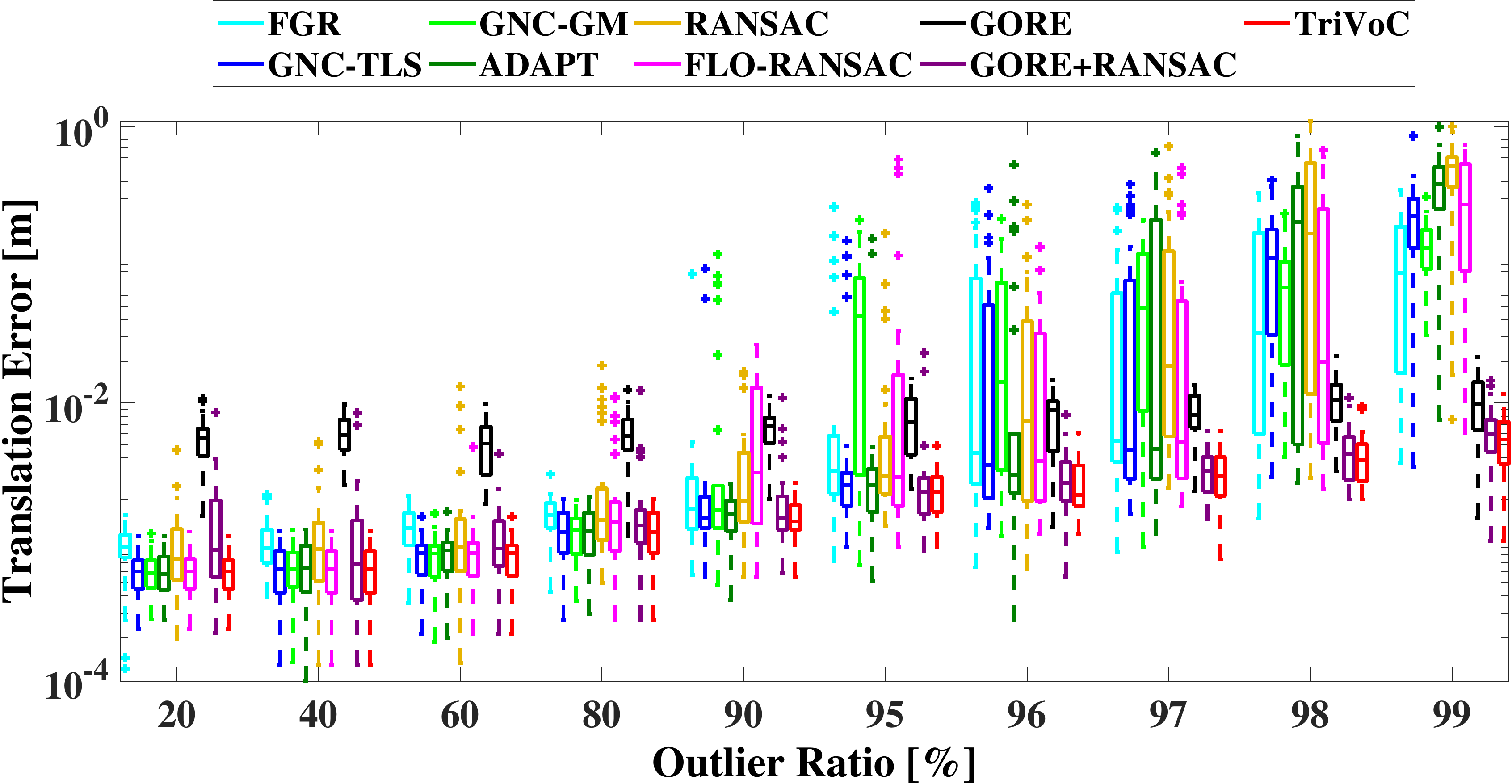}
\end{minipage}

\\
\begin{minipage}[t]{0.3\linewidth}
\includegraphics[width=1\linewidth]{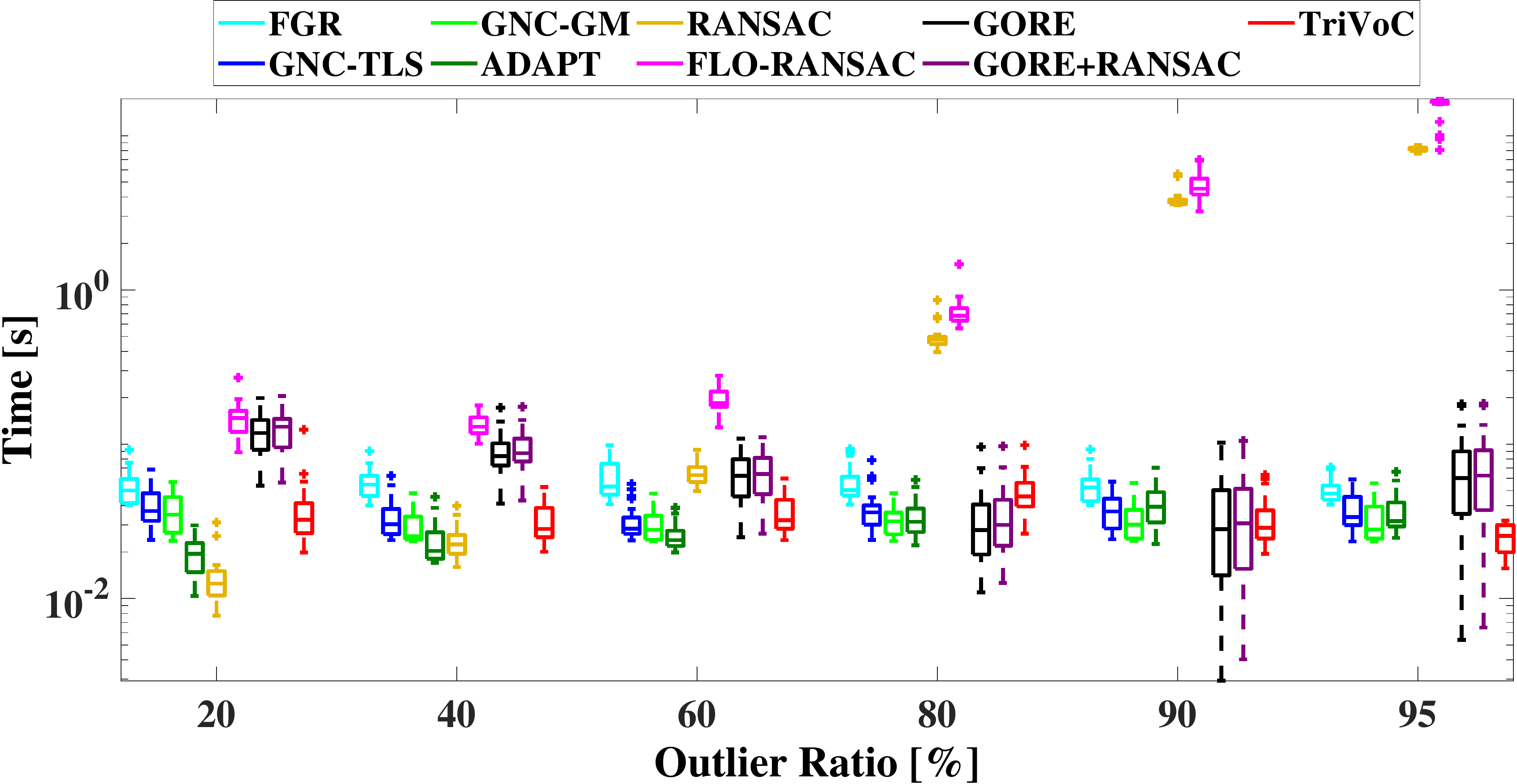}
\end{minipage}
&
\begin{minipage}[t]{0.3\linewidth}
\centering
\includegraphics[width=1\linewidth]{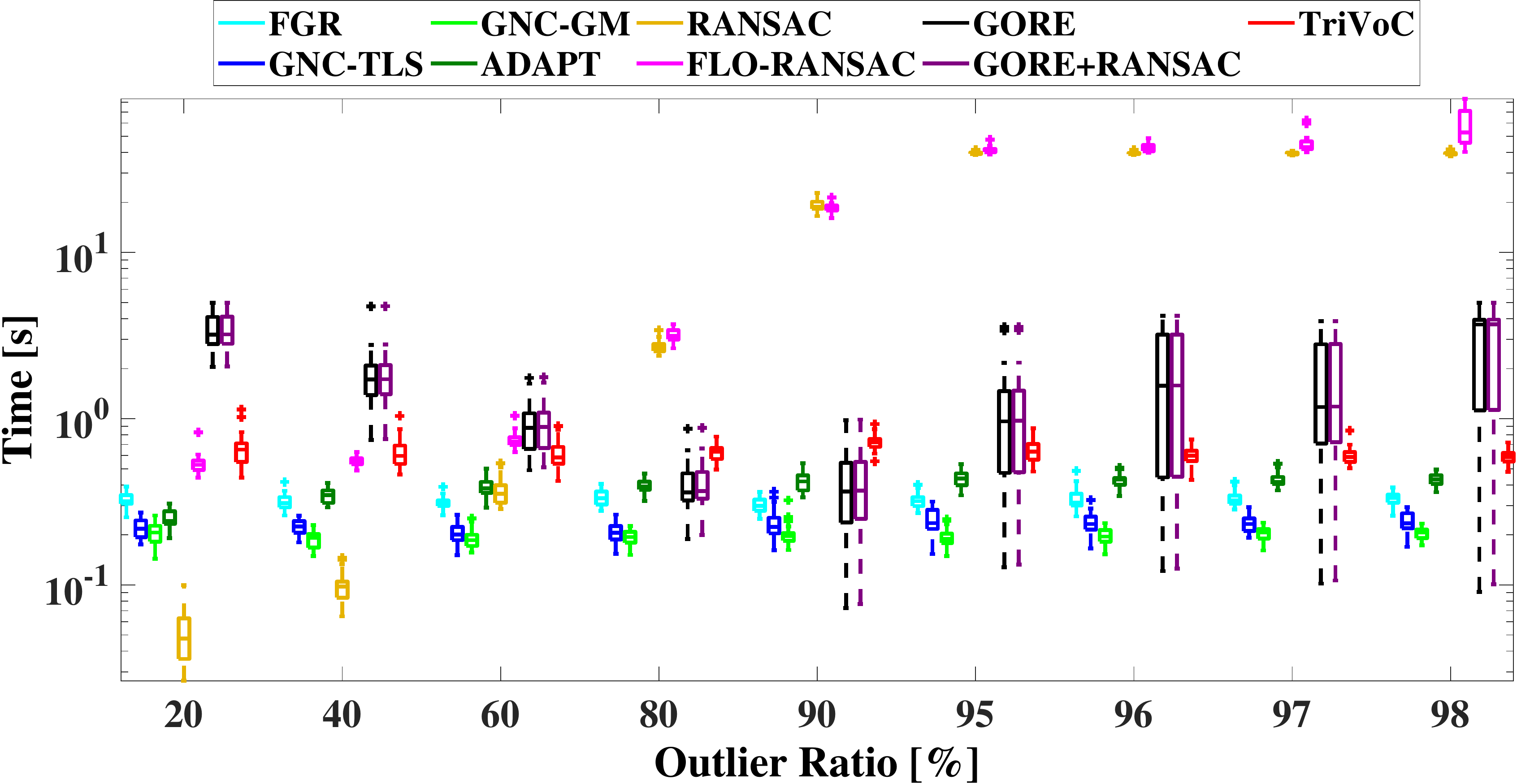}
\end{minipage}
&
\begin{minipage}[t]{0.3\linewidth}
\centering
\includegraphics[width=1\linewidth]{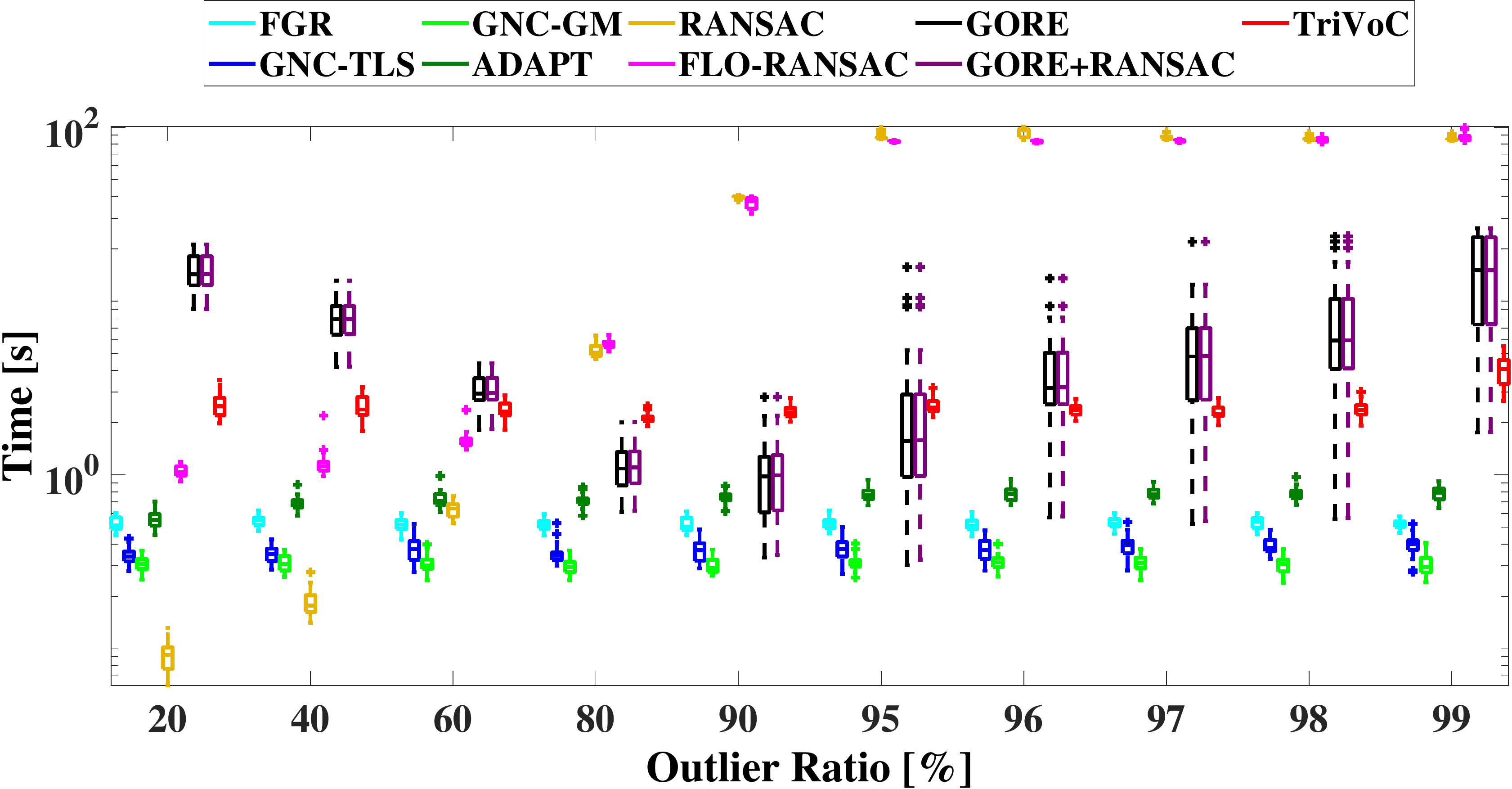}
\end{minipage}

\end{tabular}

\vspace{-2mm}
\caption{Standard benchmarking on \textit{bunny} and \textit{armadillo} with different correspondence numbers w.r.t. increasing outlier ratios (from 20\% to at most 99\%).}
\label{Benchmarking}
\vspace{-5mm}
\end{figure*}

\begin{figure}[h]
\centering

\begin{minipage}[t]{1\linewidth}
\centering
\includegraphics[width=0.8\linewidth]{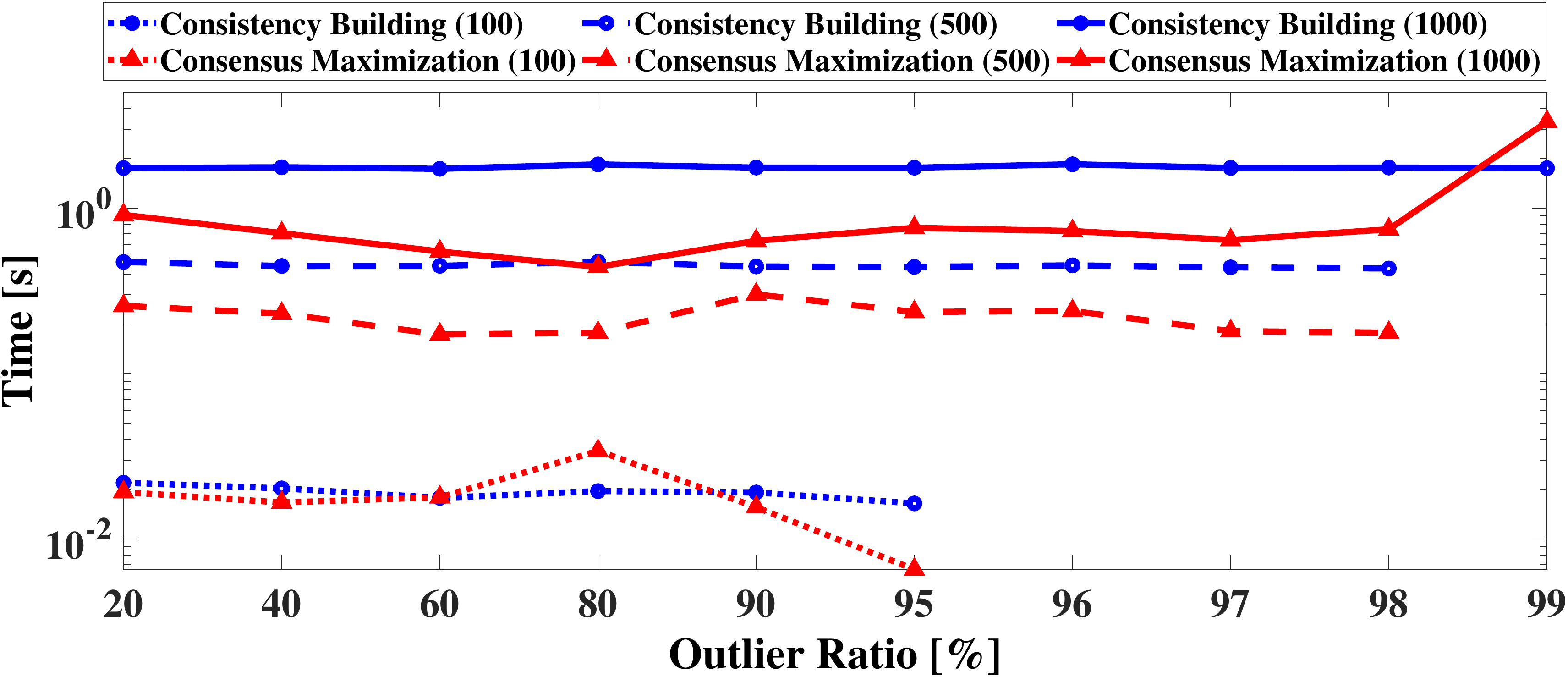}
\end{minipage}

\caption{Mean runtime of the different parts of our TriVoC.}
\label{time-TriVoC}
\vspace{-1mm}
\end{figure}

\begin{figure}[h]
\centering

\begin{minipage}[t]{1\linewidth}
\centering
\includegraphics[width=0.8\linewidth]{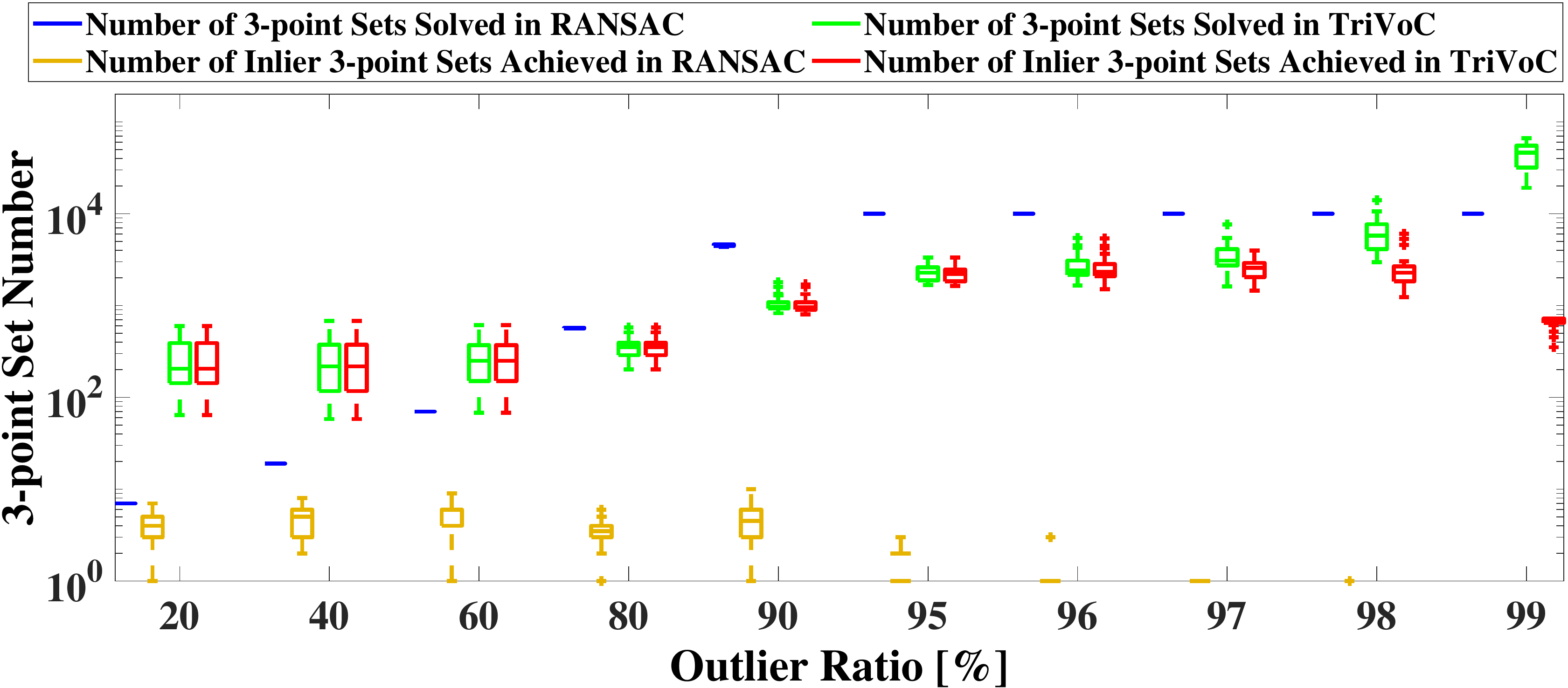}
\end{minipage}

\caption{Correctness guarantee of our TriVoC compared with RANSAC.}
\label{guarantee-TriVoC}
\vspace{-1mm}
\end{figure}

\begin{figure*}[t]
\centering
\setlength\tabcolsep{0.0pt}
\addtolength{\tabcolsep}{0pt}
\begin{tabular}{c|ccccccc c|ccccccc}

\quad &\,&\,\scriptsize{FPFH}\, &\,&  \scriptsize{GNC-TLS} & \scriptsize{FLO-RANSAC} & \scriptsize{GORE+RANSAC} & \scriptsize{TriVoC} & \quad &\,&\,\scriptsize{FPFH}\, &\,&  \scriptsize{GNC-TLS} & \scriptsize{FLO-RANSAC} & \scriptsize{GORE+RANSAC} & \scriptsize{TriVoC} \\

\hline

&&\tiny{$N$=564, 97.34\%}& & \,\tiny{74.90$^{\circ}$,14665$m$,0.09$s$}\, &\,\tiny{\textbf{4.53}$^{\circ}$,\textbf{0.13}$m$,15.41$s$}\, &\,\tiny{\textbf{4.53}$^{\circ}$,\textbf{0.13}$m$,0.91$s$}\,&\,\tiny{\textbf{4.53}$^{\circ}$,\textbf{0.13}$m$,\textbf{0.37}$s$}\,

&

&&\tiny{$N$=1053, 97.34\%}& & \,\tiny{90.60$^{\circ}$,0.44$m$,0.18$s$}\, &\,\tiny{130.26$^{\circ}$,2.24$m$,28.25$s$}\, &\,\tiny{\textbf{1.22}$^{\circ}$,\textbf{0.02}$m$,7.73$s$}\,&\,\tiny{\textbf{1.22}$^{\circ}$,\textbf{0.02}$m$,\textbf{1.33}$s$}\,

\\

\rotatebox{90}{\,\,\scriptsize{\textit{Scene-01}}\,}\,

& &

\begin{minipage}[t]{0.09\linewidth}
\centering
\includegraphics[width=1\linewidth]{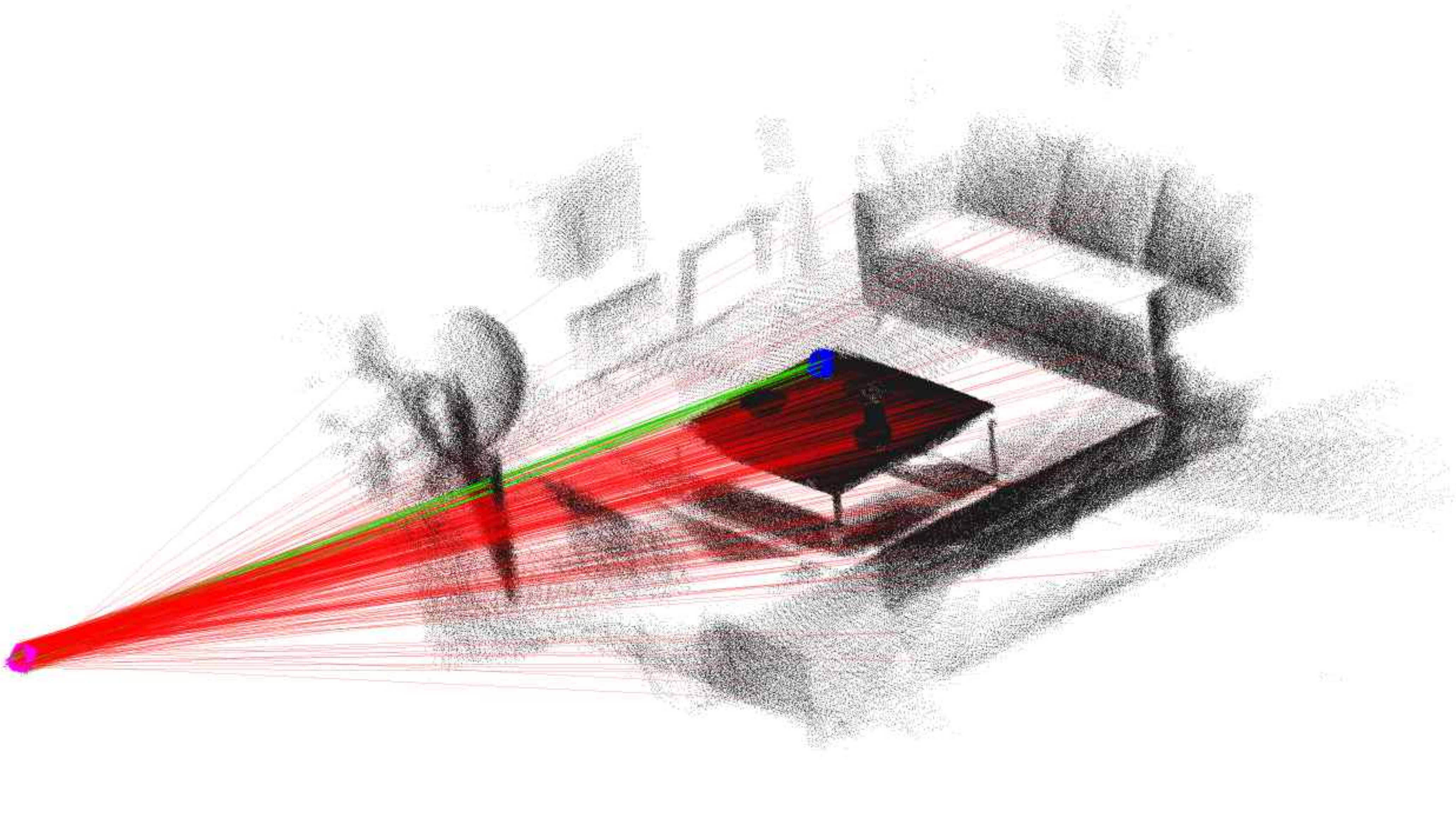}
\end{minipage}

& &

\begin{minipage}[t]{0.09\linewidth}
\centering
\includegraphics[width=1\linewidth]{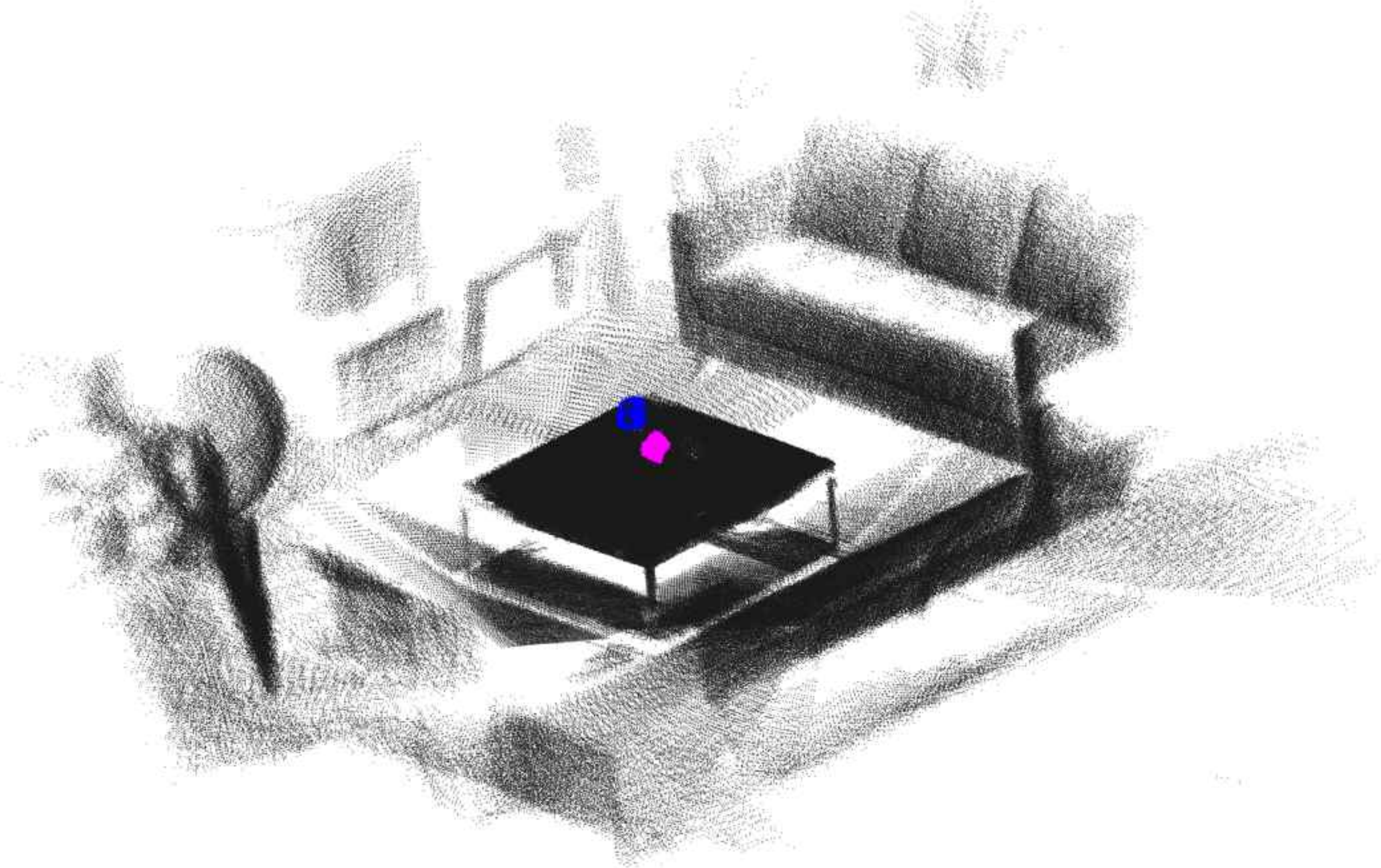}
\end{minipage}

&

\begin{minipage}[t]{0.09\linewidth}
\centering
\includegraphics[width=1\linewidth]{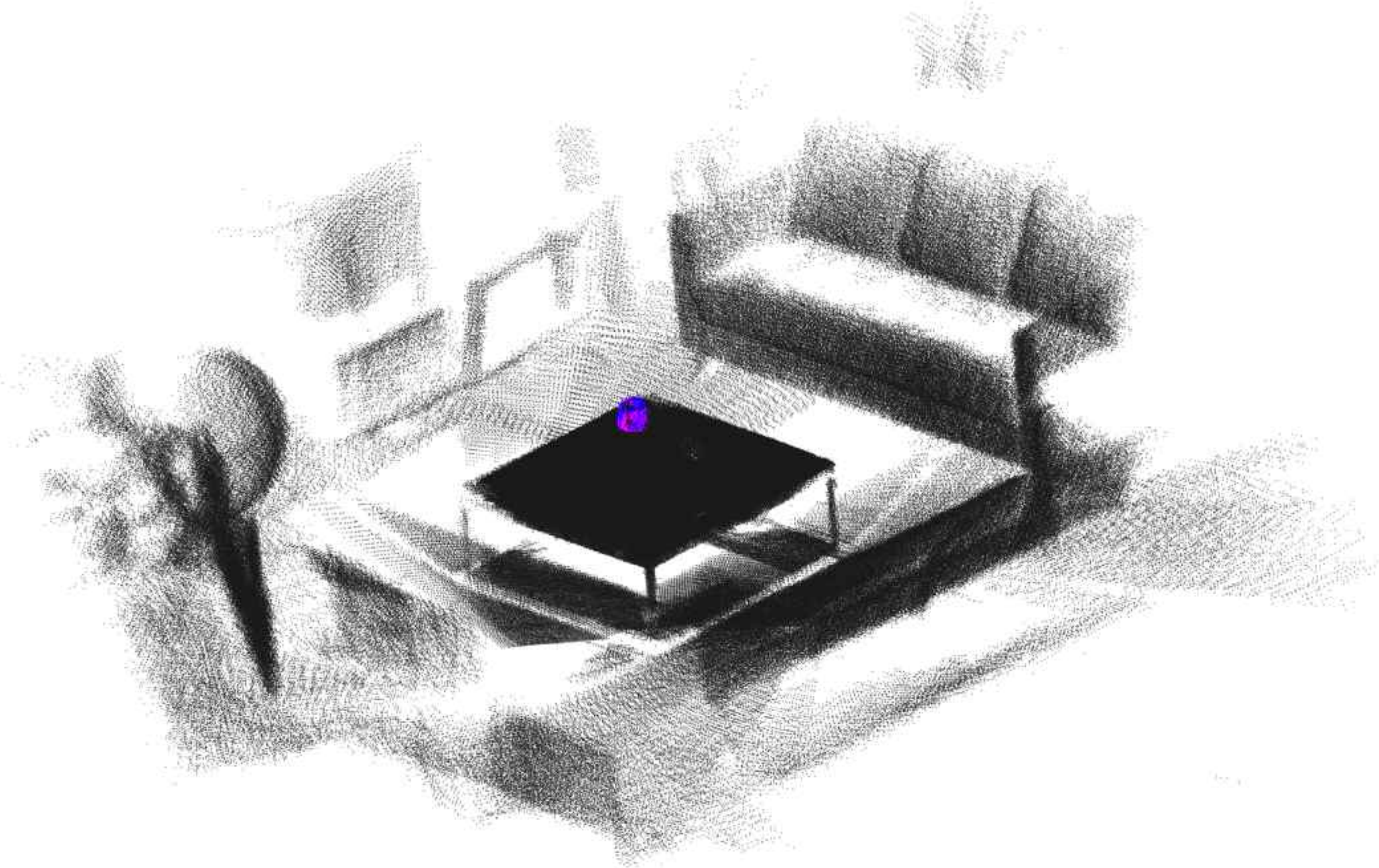}
\end{minipage}

&

\begin{minipage}[t]{0.09\linewidth}
\centering
\includegraphics[width=1\linewidth]{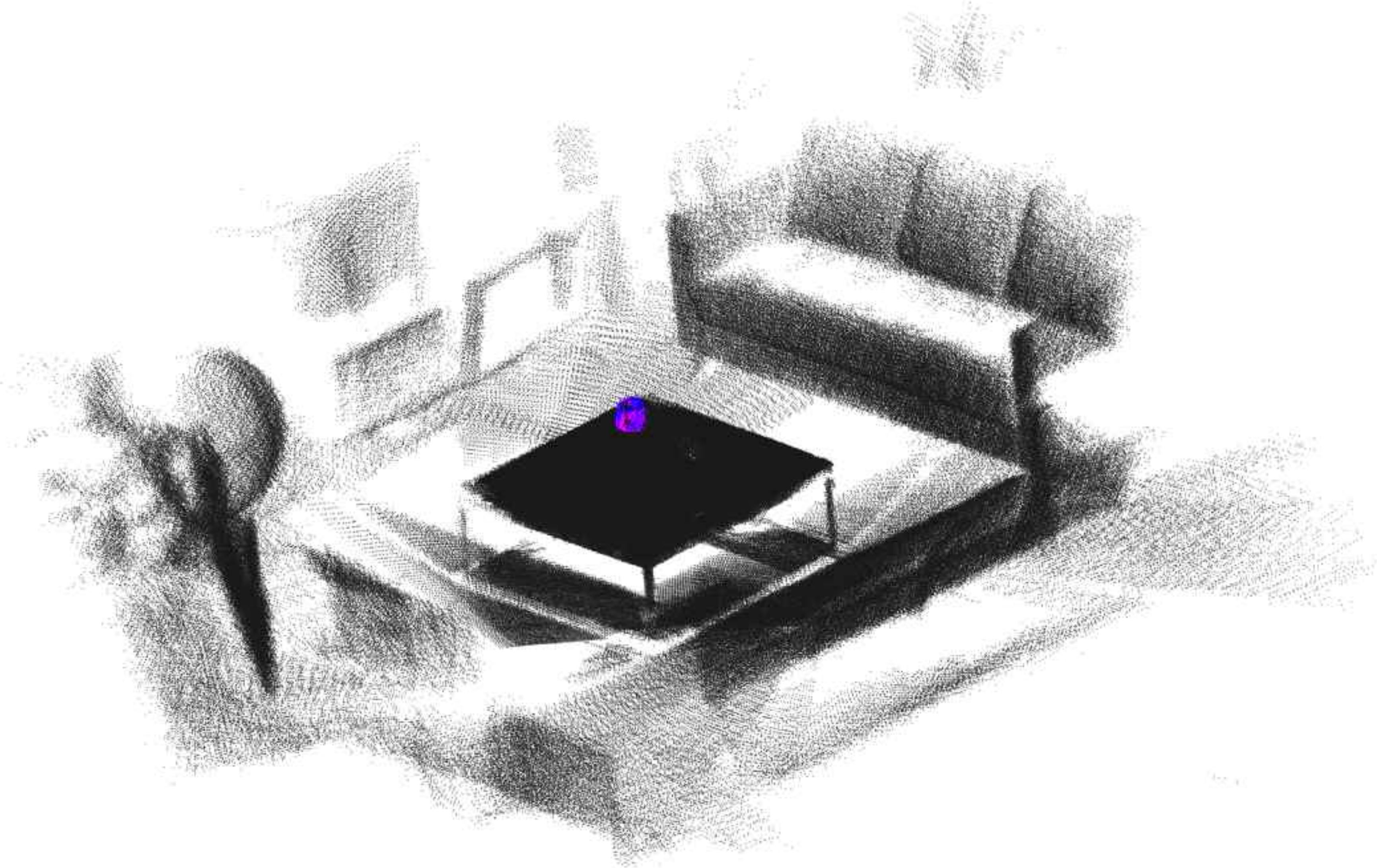}
\end{minipage}

&

\begin{minipage}[t]{0.09\linewidth}
\centering
\includegraphics[width=1\linewidth]{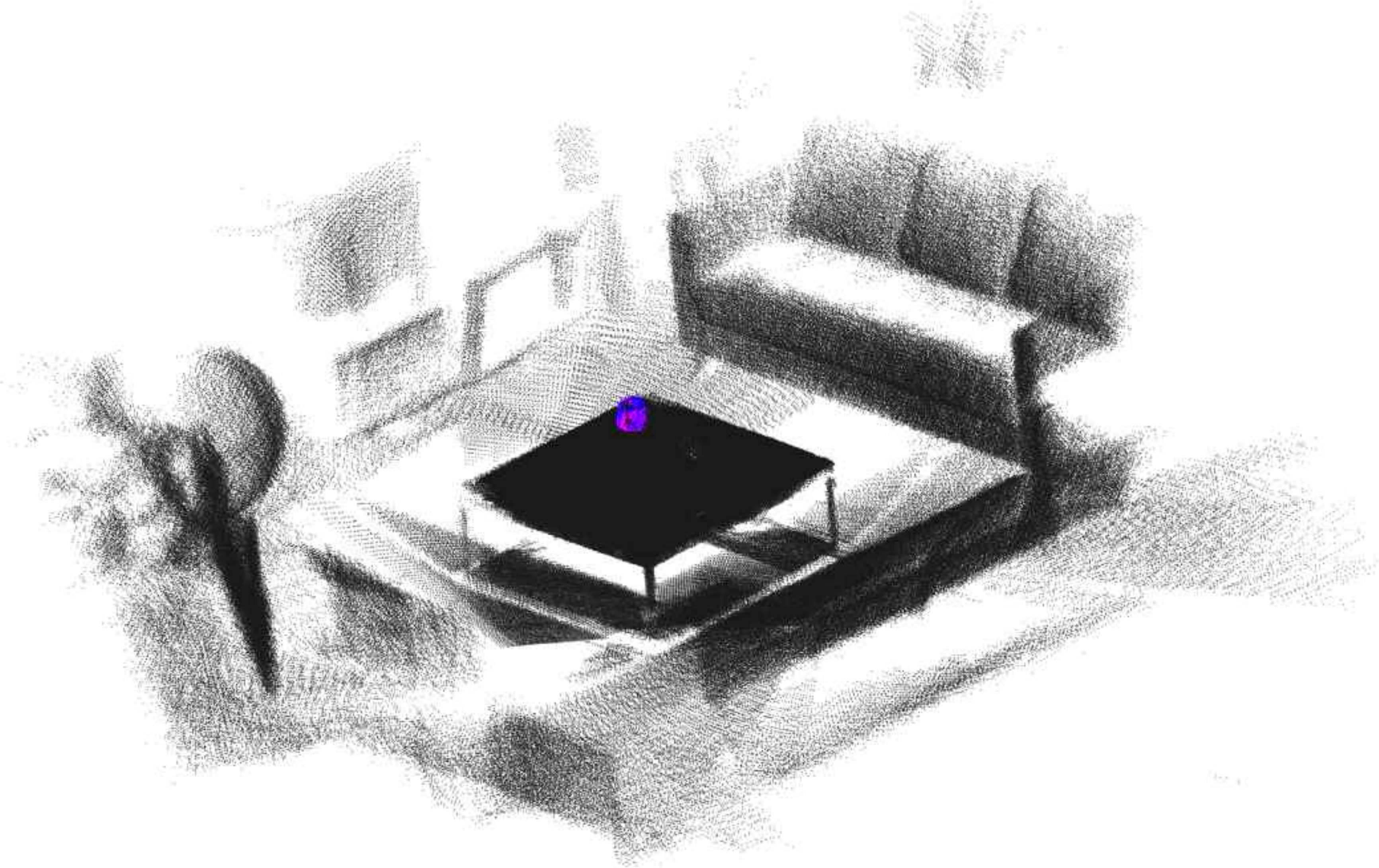}
\end{minipage}

&

\rotatebox{90}{\,\,\scriptsize{\textit{Scene-02}}\,}\,

& &

\begin{minipage}[t]{0.09\linewidth}
\centering
\includegraphics[width=1\linewidth]{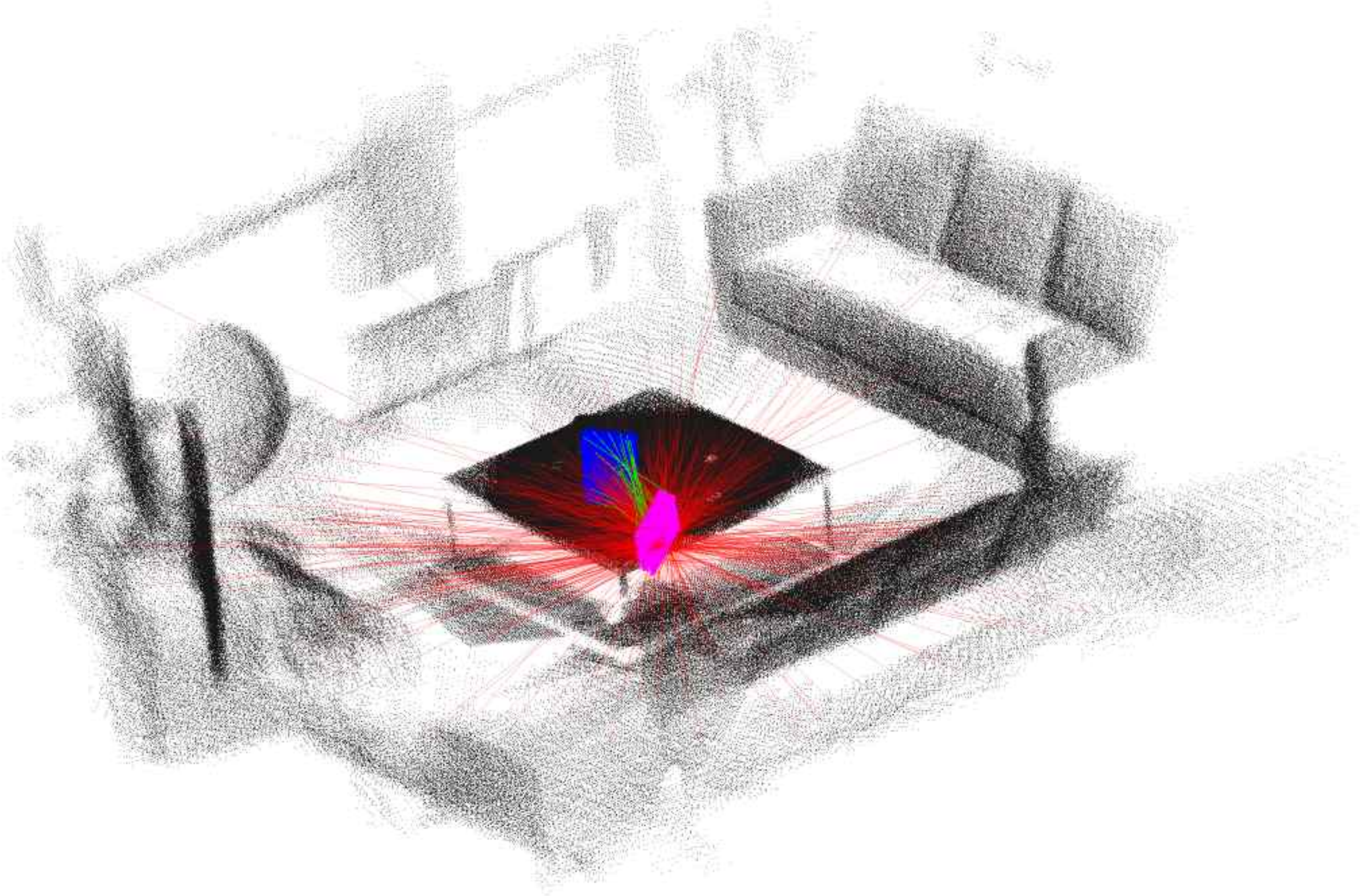}
\end{minipage}

& &

\begin{minipage}[t]{0.09\linewidth}
\centering
\includegraphics[width=1\linewidth]{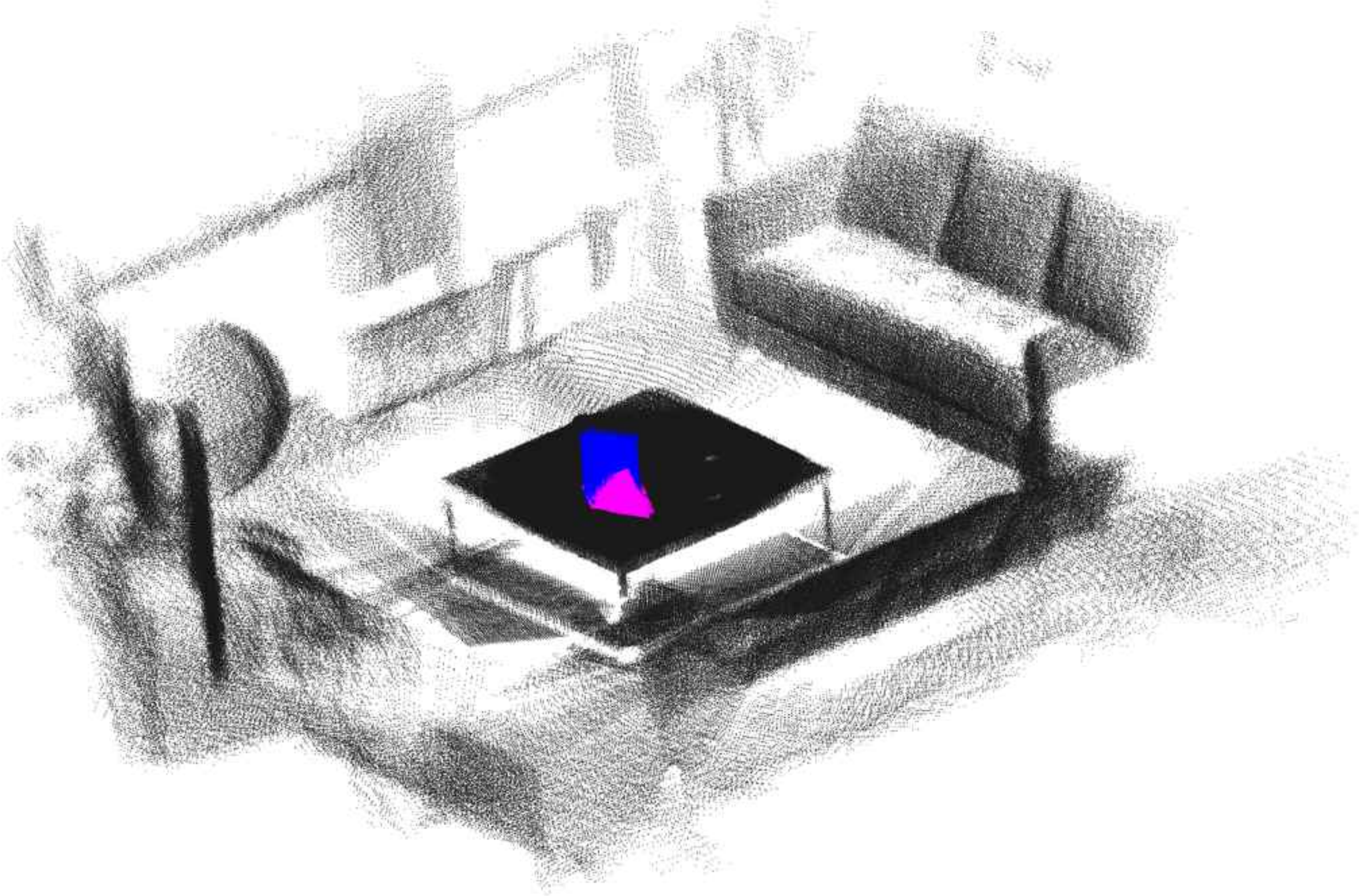}
\end{minipage}

&

\begin{minipage}[t]{0.09\linewidth}
\centering
\includegraphics[width=1\linewidth]{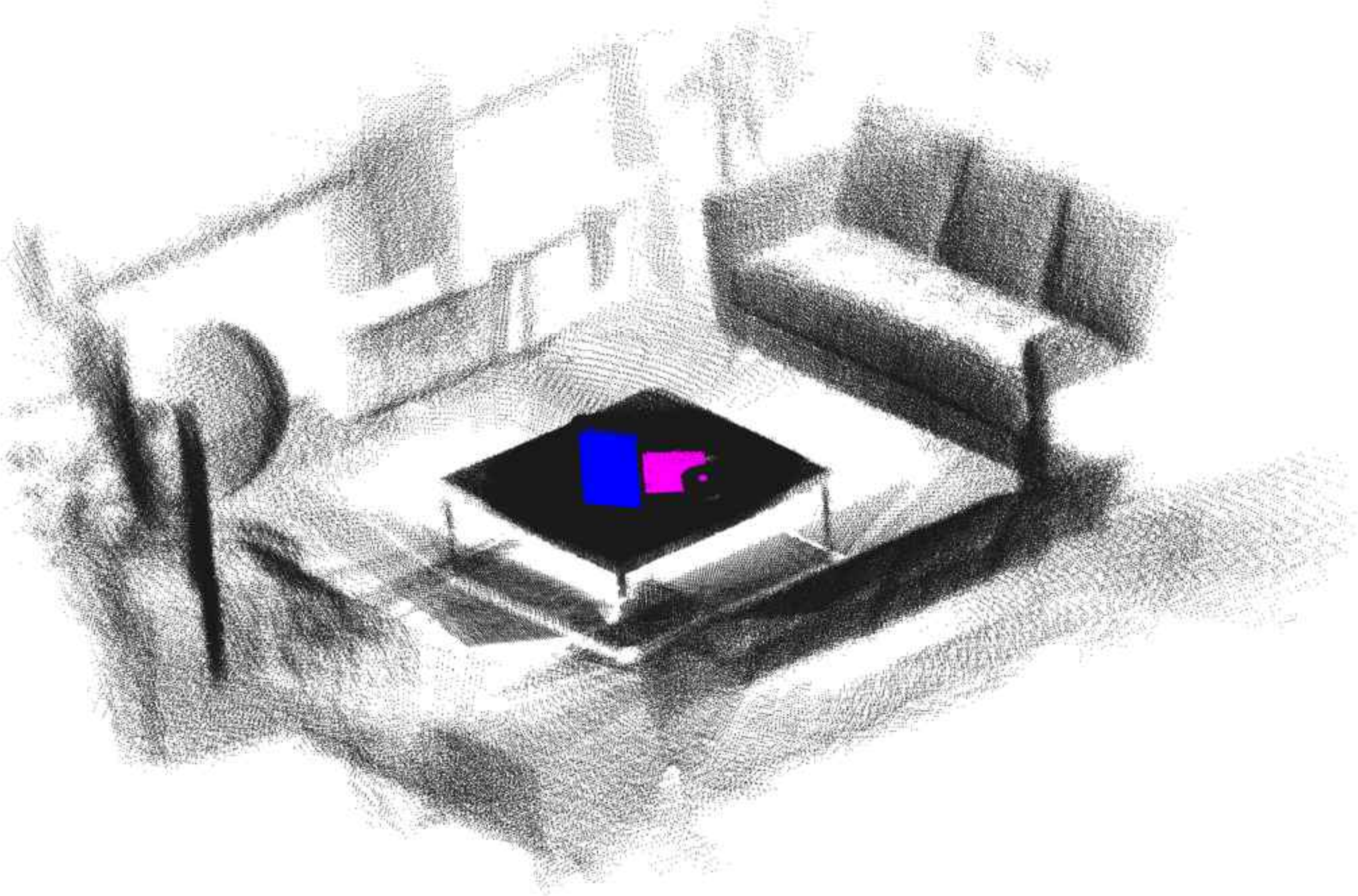}
\end{minipage}

&

\begin{minipage}[t]{0.09\linewidth}
\centering
\includegraphics[width=1\linewidth]{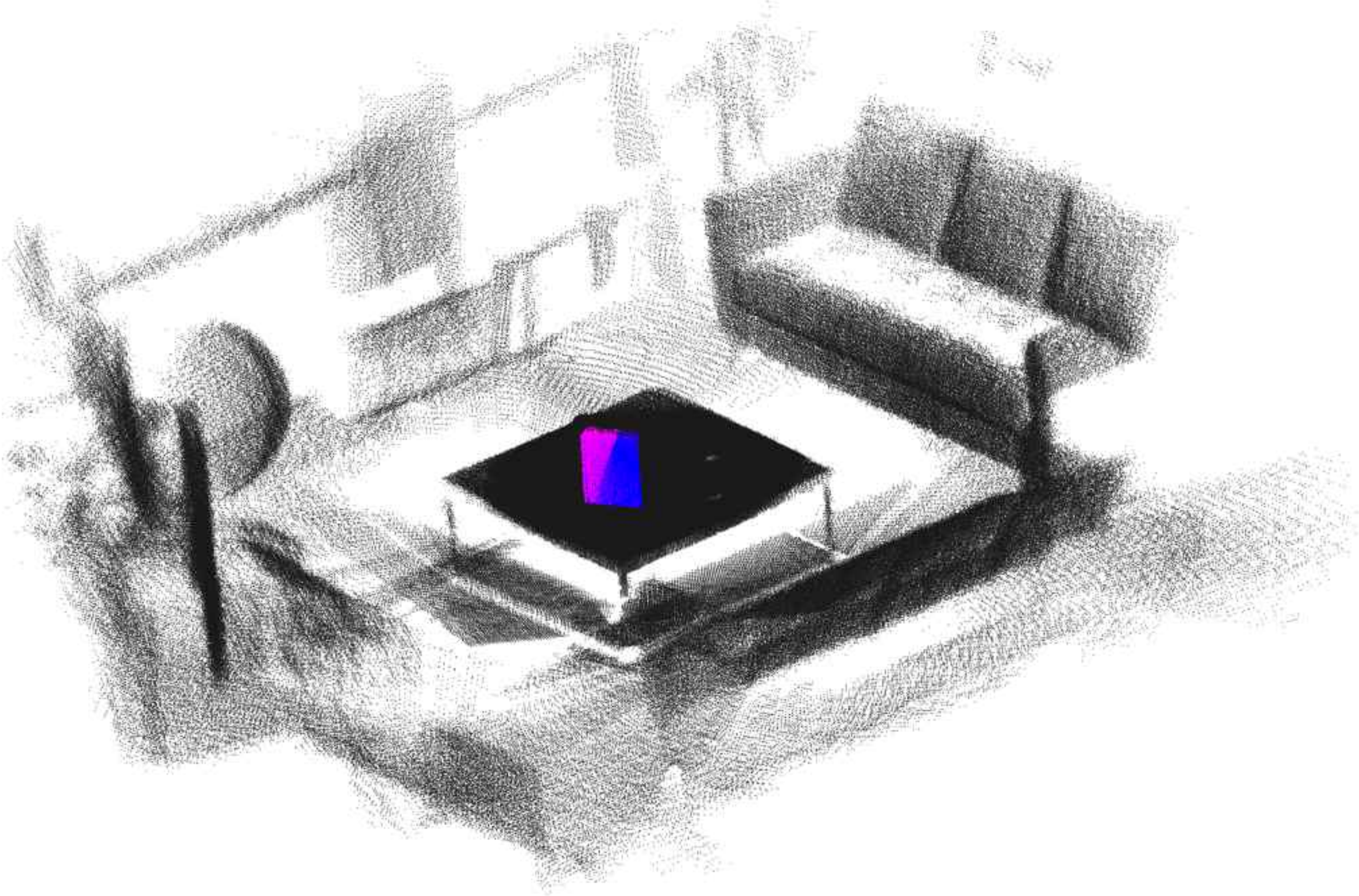}
\end{minipage}

&

\begin{minipage}[t]{0.09\linewidth}
\centering
\includegraphics[width=1\linewidth]{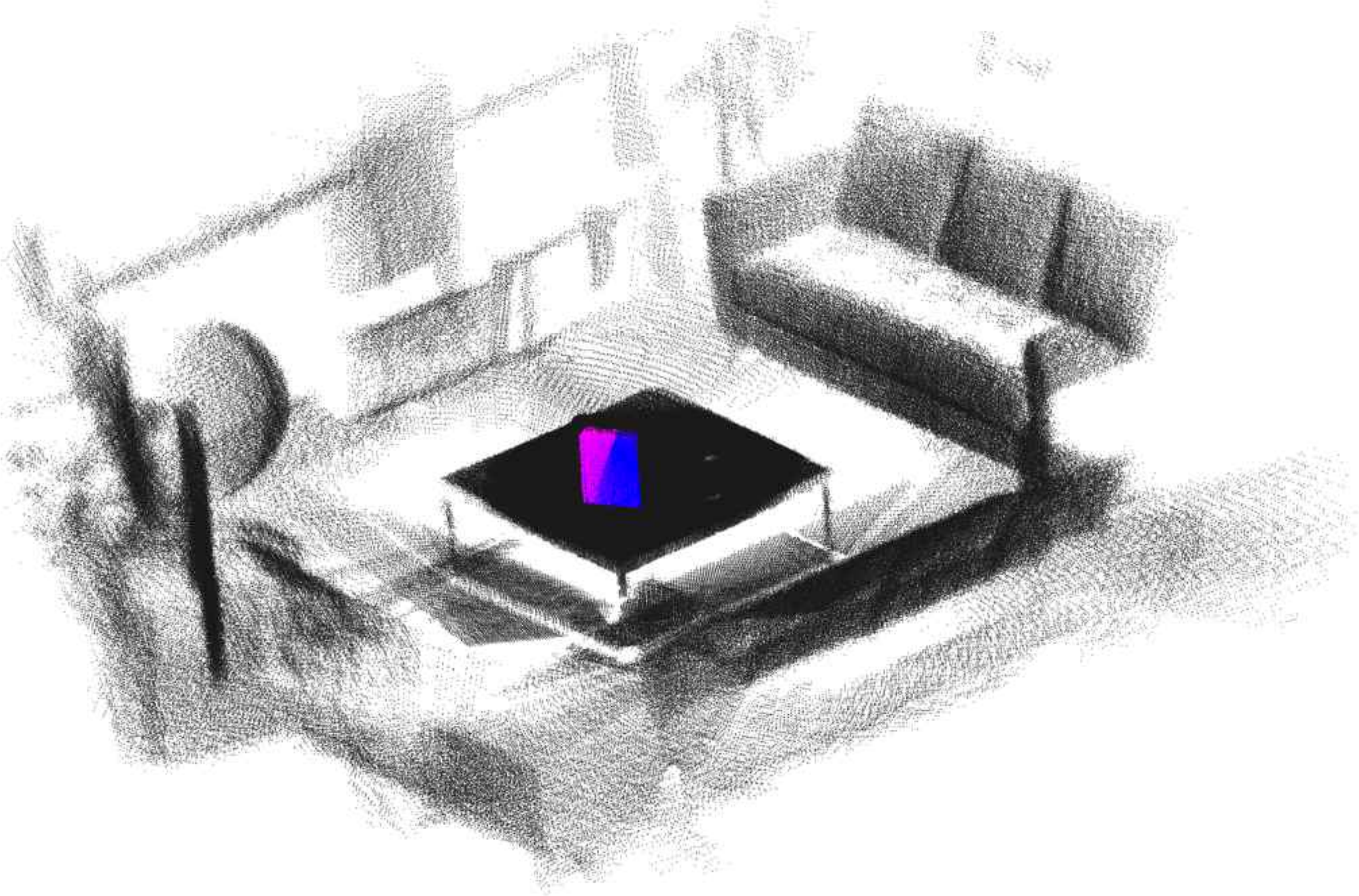}
\end{minipage}

\\

&&\tiny{$N$=325, 95.38\%}& & \,\tiny{90.55$^{\circ}$,0.59$m$,0.06$s$}\, &\,\tiny{90.05$^{\circ}$,1.11$m$,8.72$s$}\, &\,\tiny{\textbf{3.17}$^{\circ}$,\textbf{0.05}$m$,{0.25}$s$}\,&\,\tiny{\textbf{3.17}$^{\circ}$,\textbf{0.05}$m$,\textbf{0.13}$s$}\,

&

&&\tiny{$N$=510, 97.84\%}& & \,\tiny{113.73$^{\circ}$,1.66$m$,0.07$s$}\, &\,\tiny{98.23$^{\circ}$,2.71$m$,14.06$s$}\, &\,\tiny{\textbf{4.33}$^{\circ}$,\textbf{0.09}$m$,\textbf{0.35}$s$}\,&\,\tiny{\textbf{4.33}$^{\circ}$,\textbf{0.09}$m$,\textbf{0.27}$s$}\,

\\

\rotatebox{90}{\,\,\scriptsize{\textit{Scene-03}}\,}\,

& &

\begin{minipage}[t]{0.09\linewidth}
\centering
\includegraphics[width=1\linewidth]{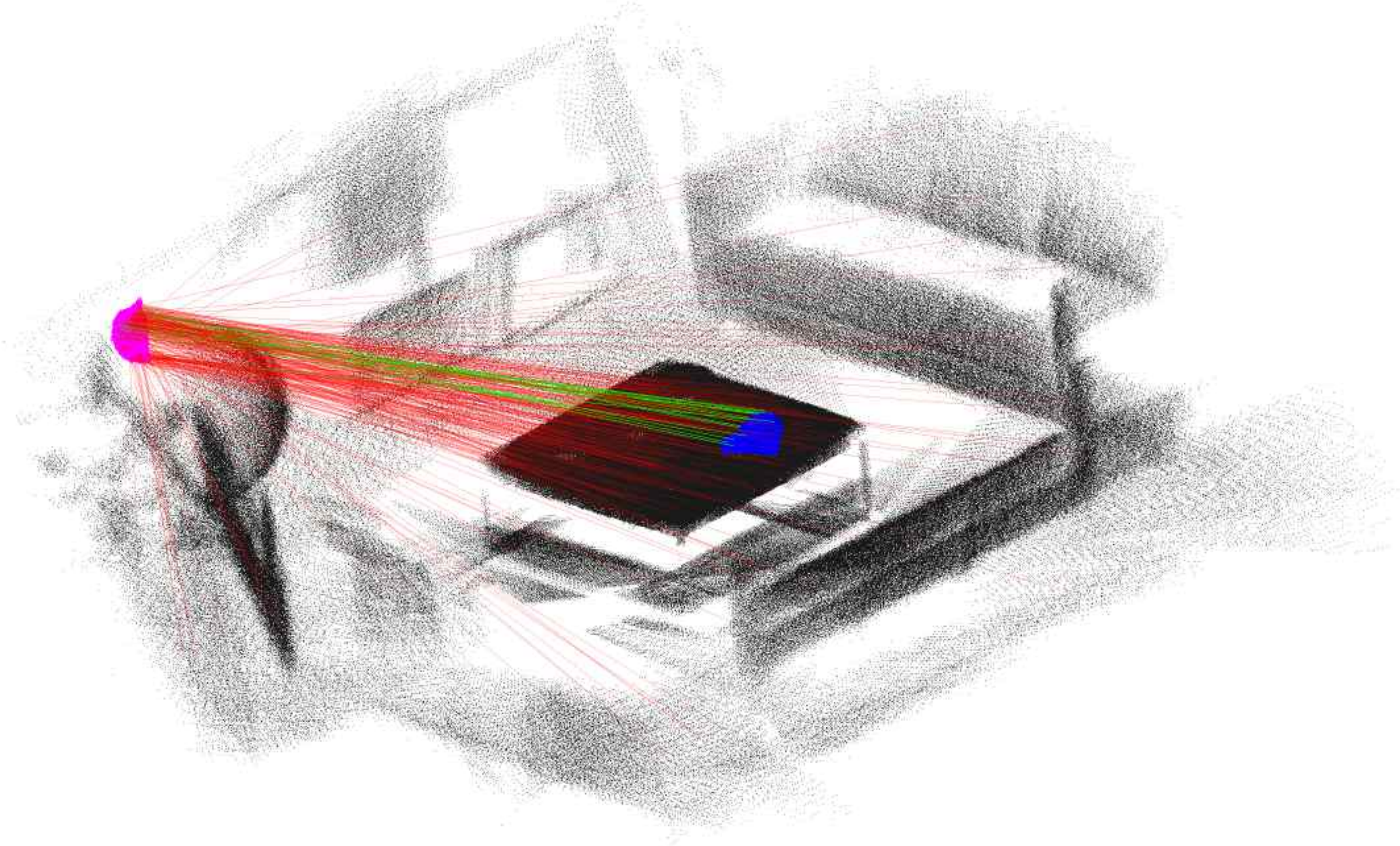}
\end{minipage}

& &

\begin{minipage}[t]{0.09\linewidth}
\centering
\includegraphics[width=1\linewidth]{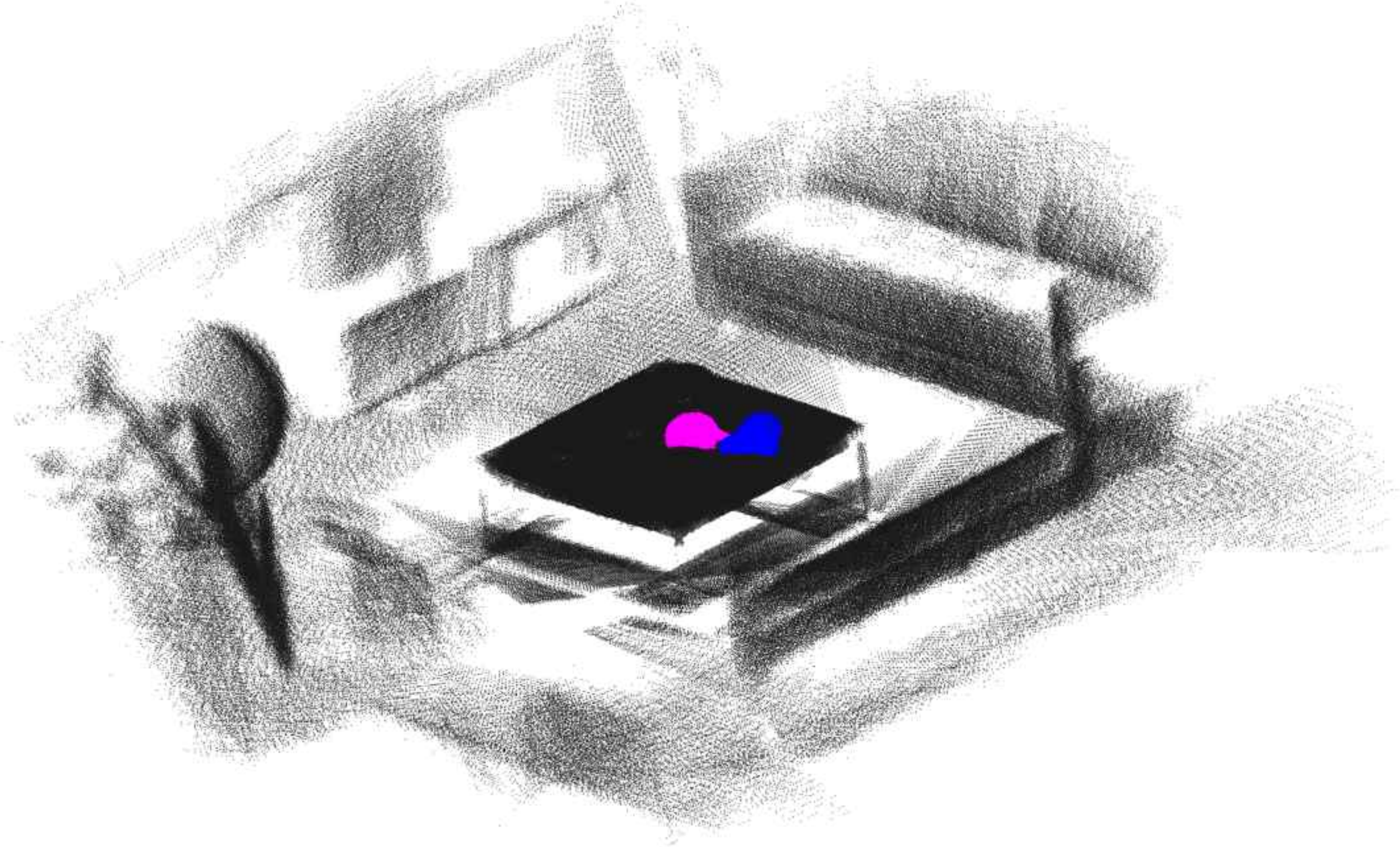}
\end{minipage}

&

\begin{minipage}[t]{0.09\linewidth}
\centering
\includegraphics[width=1\linewidth]{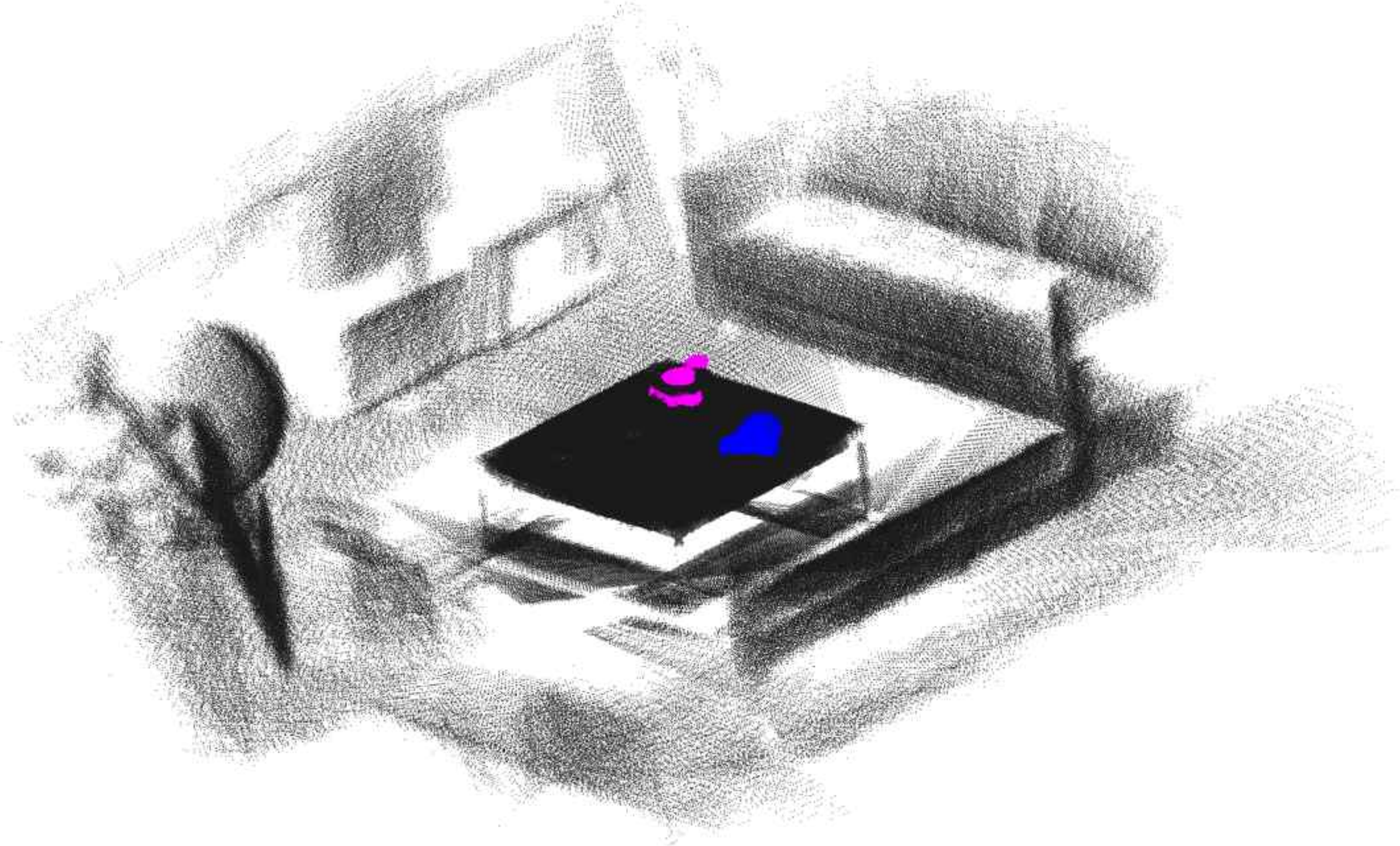}
\end{minipage}

&

\begin{minipage}[t]{0.09\linewidth}
\centering
\includegraphics[width=1\linewidth]{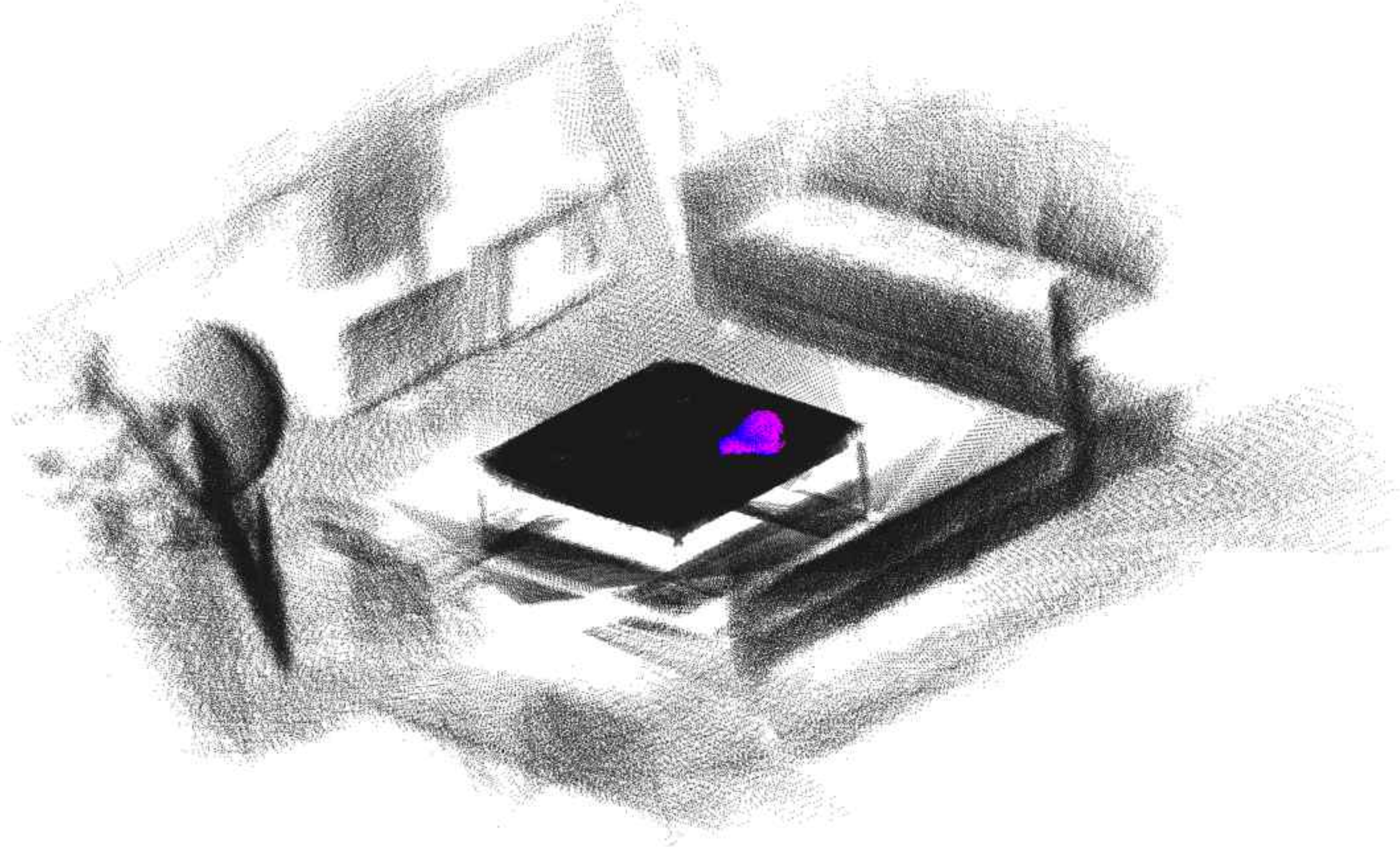}
\end{minipage}

&

\begin{minipage}[t]{0.09\linewidth}
\centering
\includegraphics[width=1\linewidth]{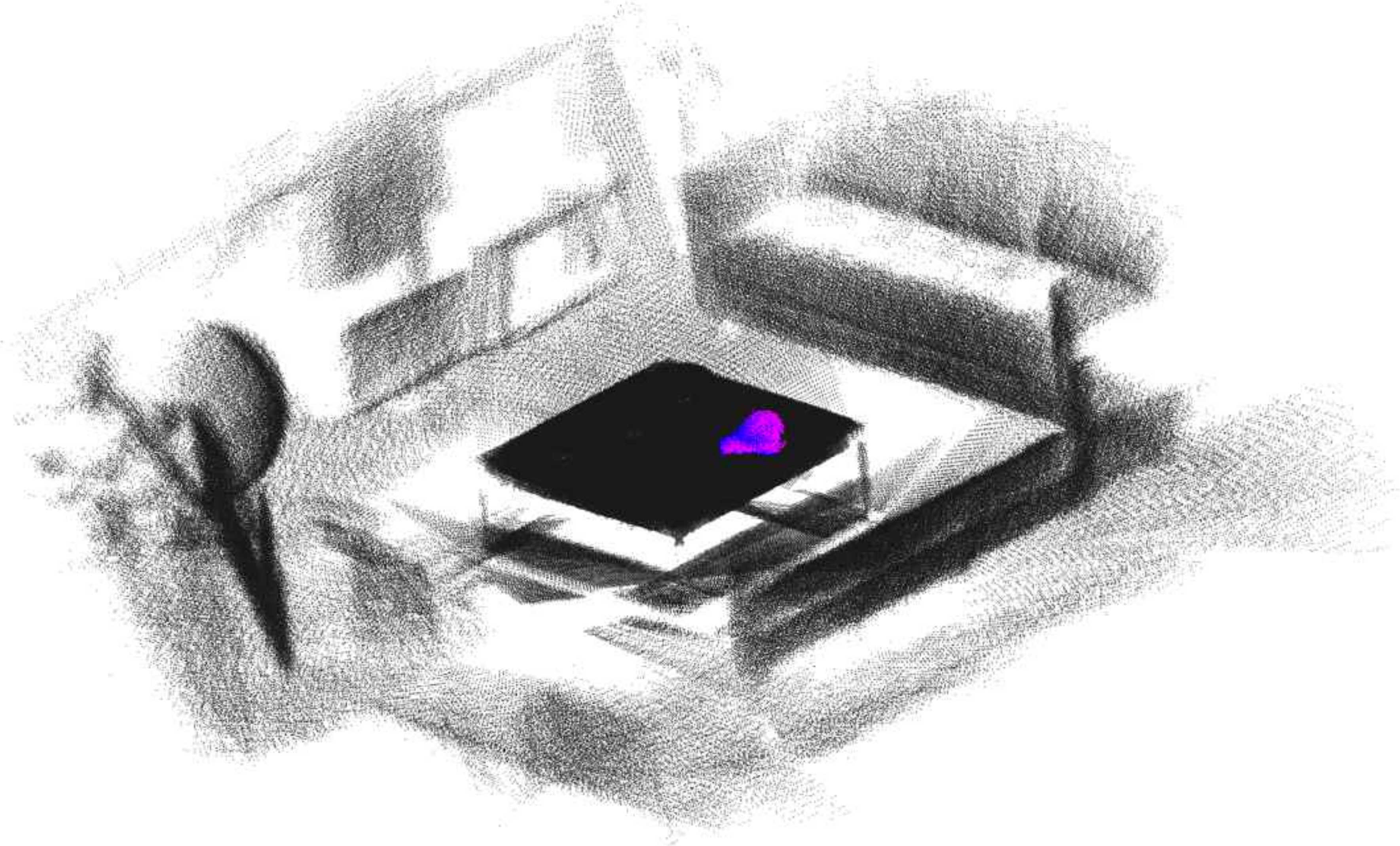}
\end{minipage}

&

\rotatebox{90}{\,\,\scriptsize{\textit{Scene-04}}\,}\,

& &

\begin{minipage}[t]{0.09\linewidth}
\centering
\includegraphics[width=1\linewidth]{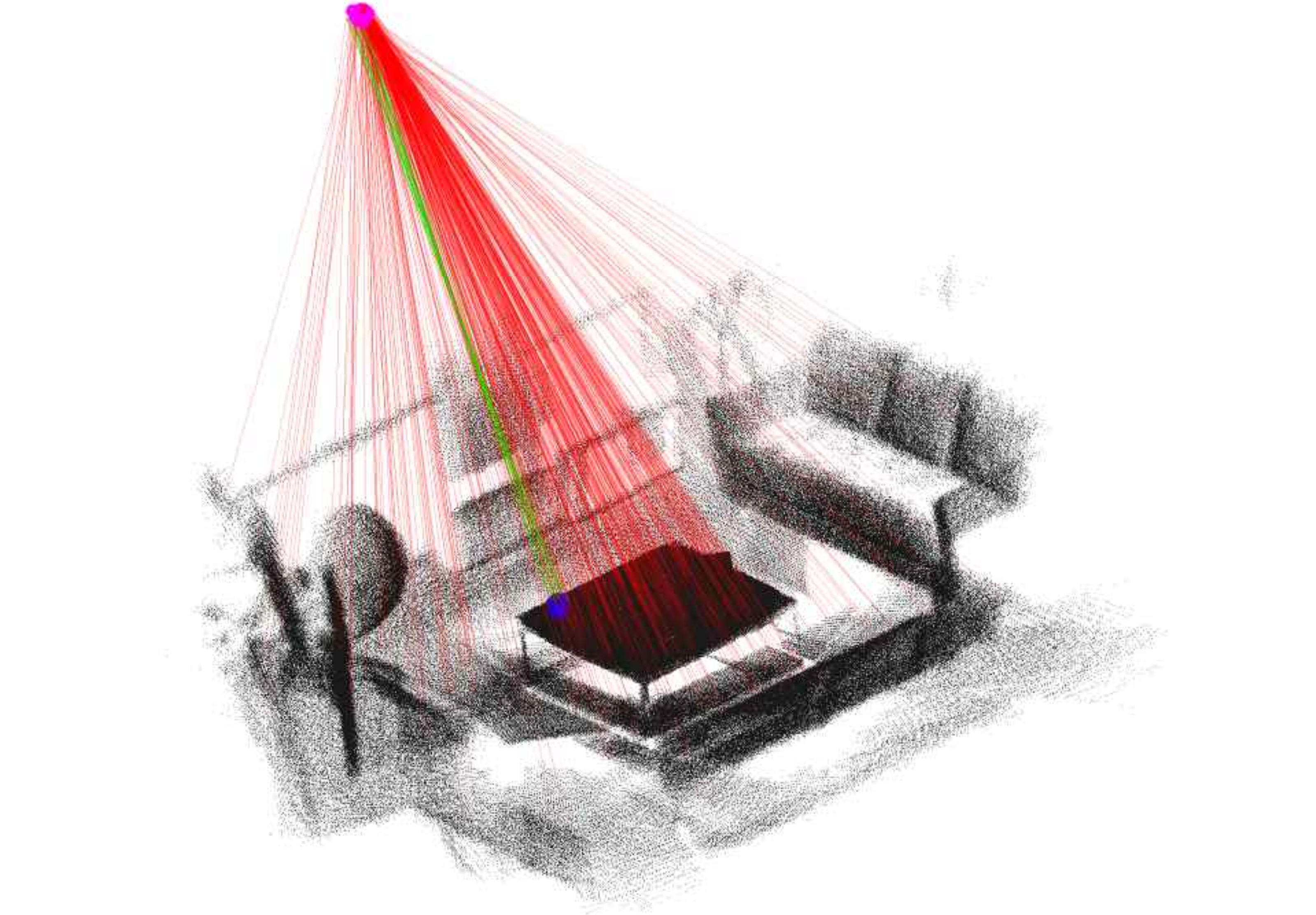}
\end{minipage}

& &

\begin{minipage}[t]{0.09\linewidth}
\centering
\includegraphics[width=1\linewidth]{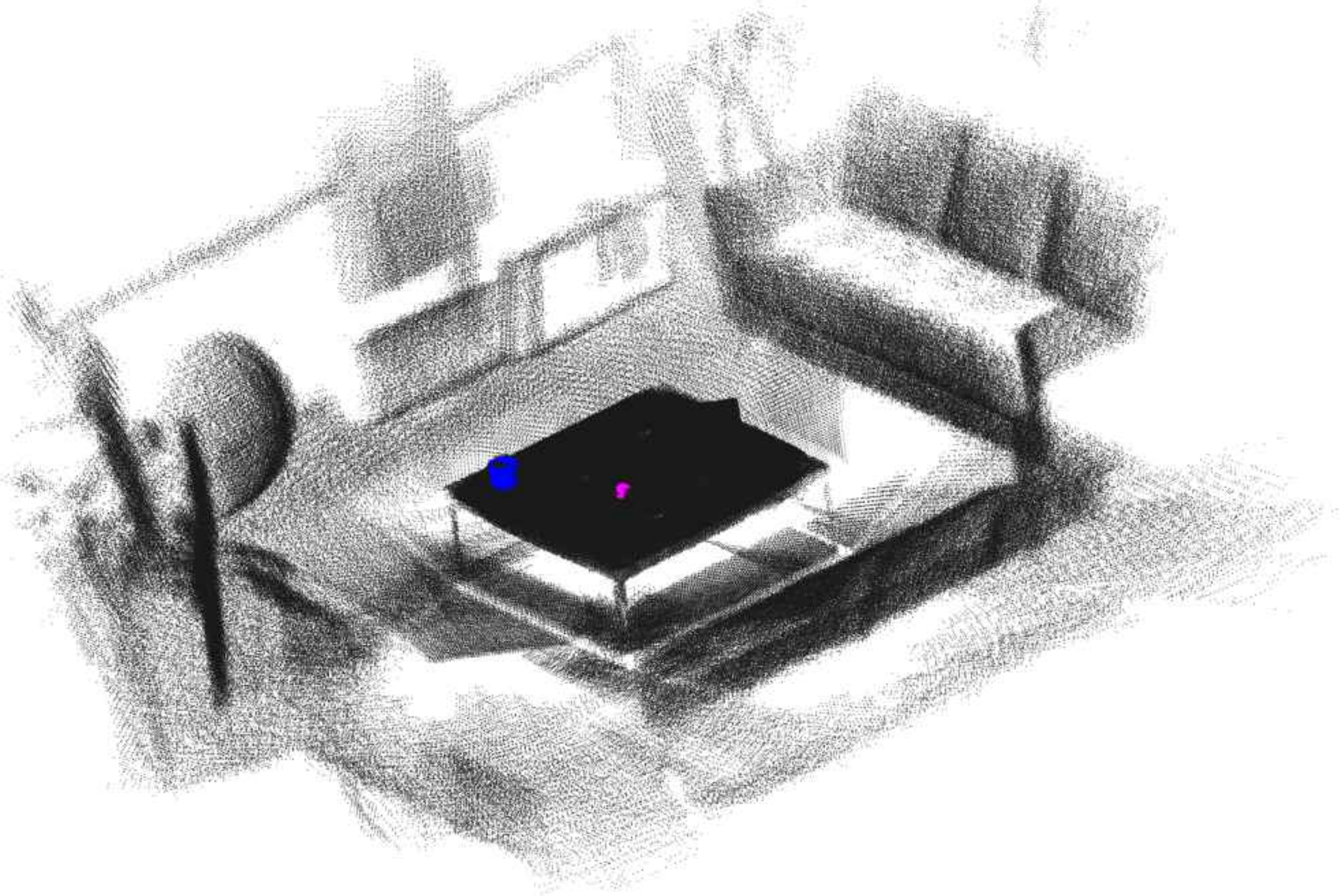}
\end{minipage}

&

\begin{minipage}[t]{0.09\linewidth}
\centering
\includegraphics[width=1\linewidth]{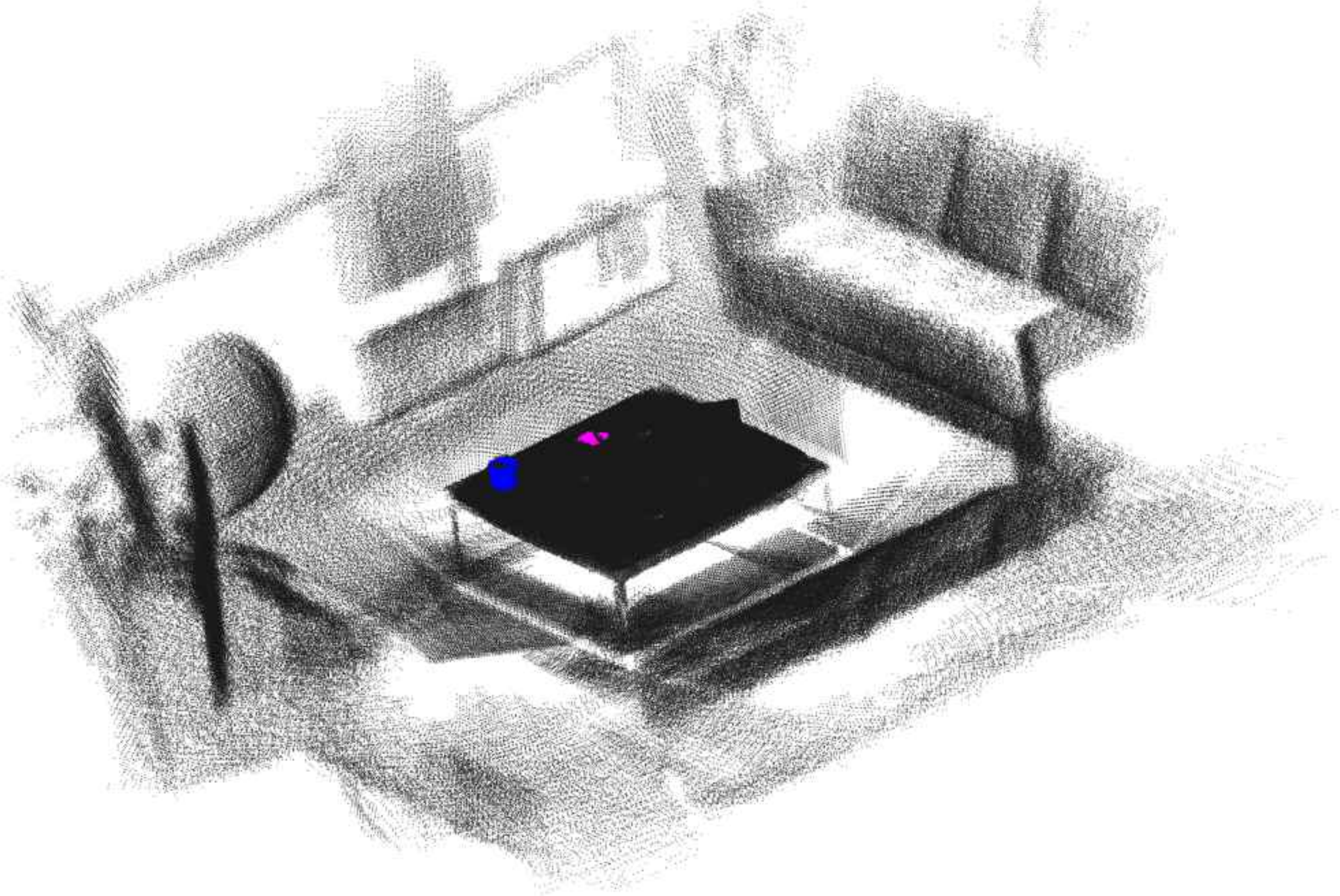}
\end{minipage}

&

\begin{minipage}[t]{0.09\linewidth}
\centering
\includegraphics[width=1\linewidth]{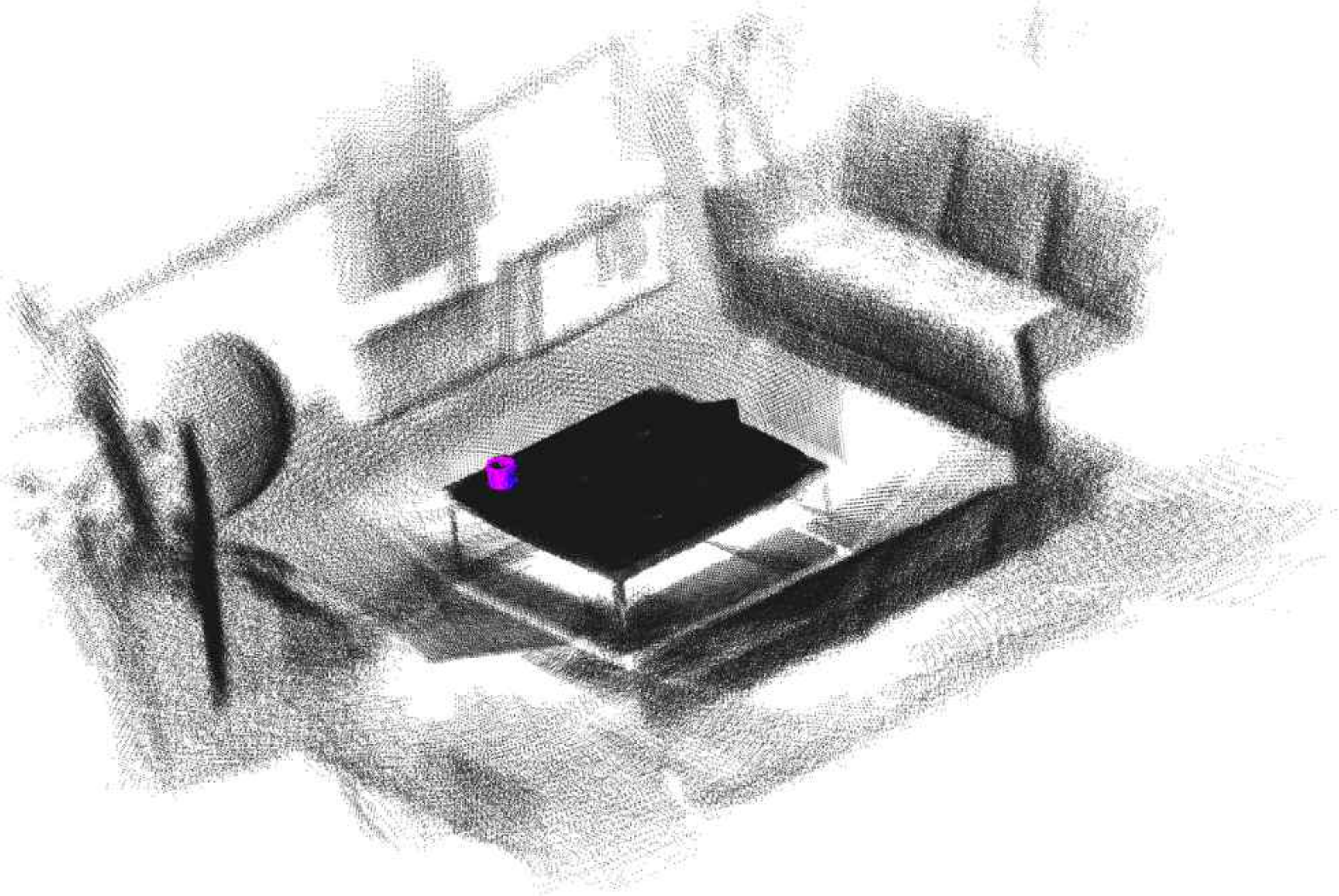}
\end{minipage}

&

\begin{minipage}[t]{0.09\linewidth}
\centering
\includegraphics[width=1\linewidth]{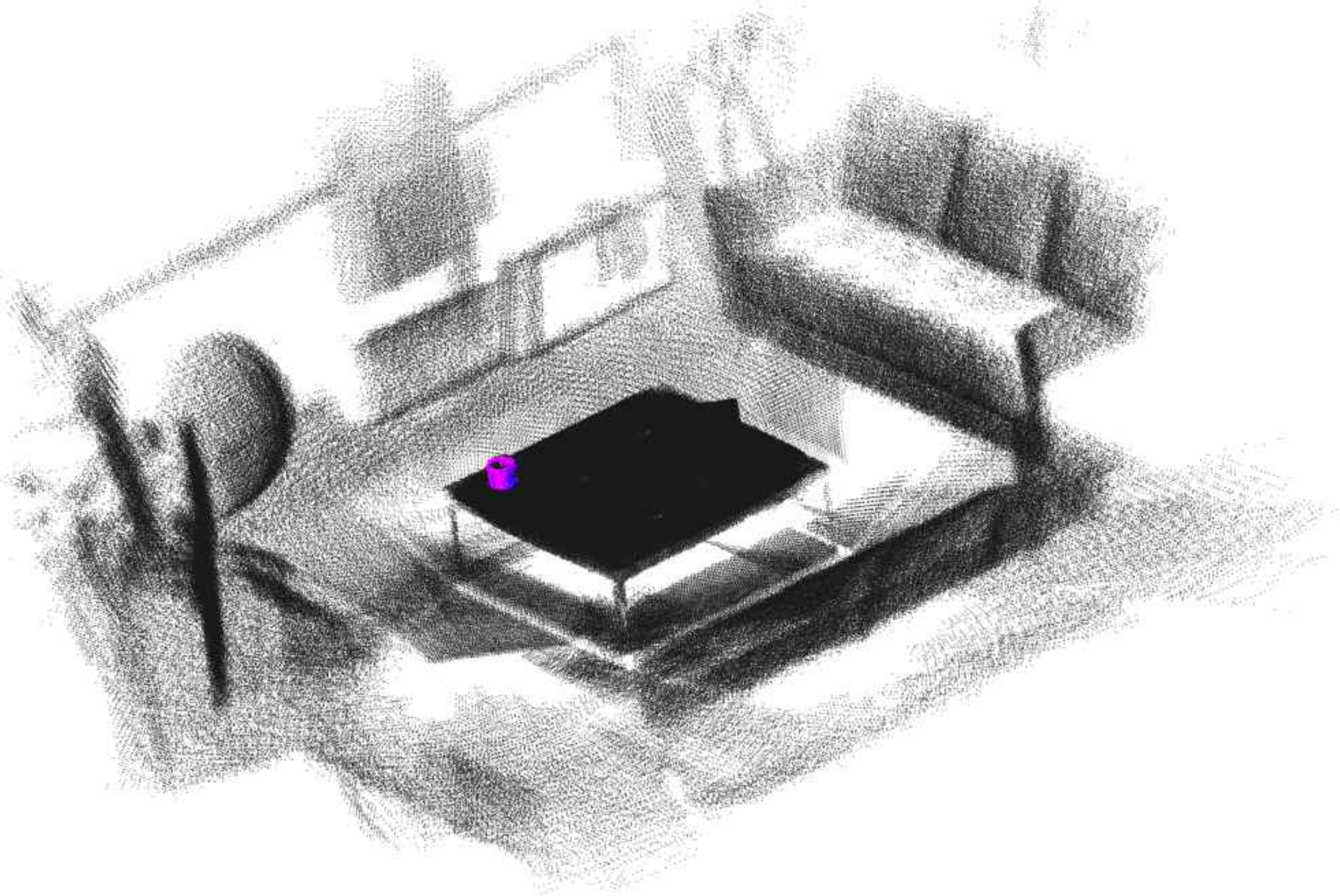}
\end{minipage}

\\

&&\tiny{$N$=463, 96.76\%}& & \,\tiny{129.99$^{\circ}$,2.50$m$,0.08$s$}\, &\,\tiny{{3.15}$^{\circ}$,\textbf{0.06}$m$,12.35$s$}\, &\,\tiny{{3.49}$^{\circ}$,\textbf{0.06}$m$,{0.32}$s$}\,&\,\tiny{\textbf{3.01}$^{\circ}$,\textbf{0.05}$m$,\textbf{0.24}$s$}\,

&

&&\tiny{$N$=348, 97.13\%}& & \,\tiny{109.33$^{\circ}$,1.12$m$,0.08$s$}\, &\,\tiny{{92.16}$^{\circ}$,{1.11}$m$,9.61$s$}\, &\,\tiny{{5.49}$^{\circ}$,{0.10}$m$,0.16$s$}\,&\,\tiny{\textbf{1.38}$^{\circ}$,\textbf{0.01}$m$,\textbf{0.12}$s$}\,

\\

\rotatebox{90}{\,\,\scriptsize{\textit{Scene-06}}\,}\,

& &

\begin{minipage}[t]{0.09\linewidth}
\centering
\includegraphics[width=1\linewidth]{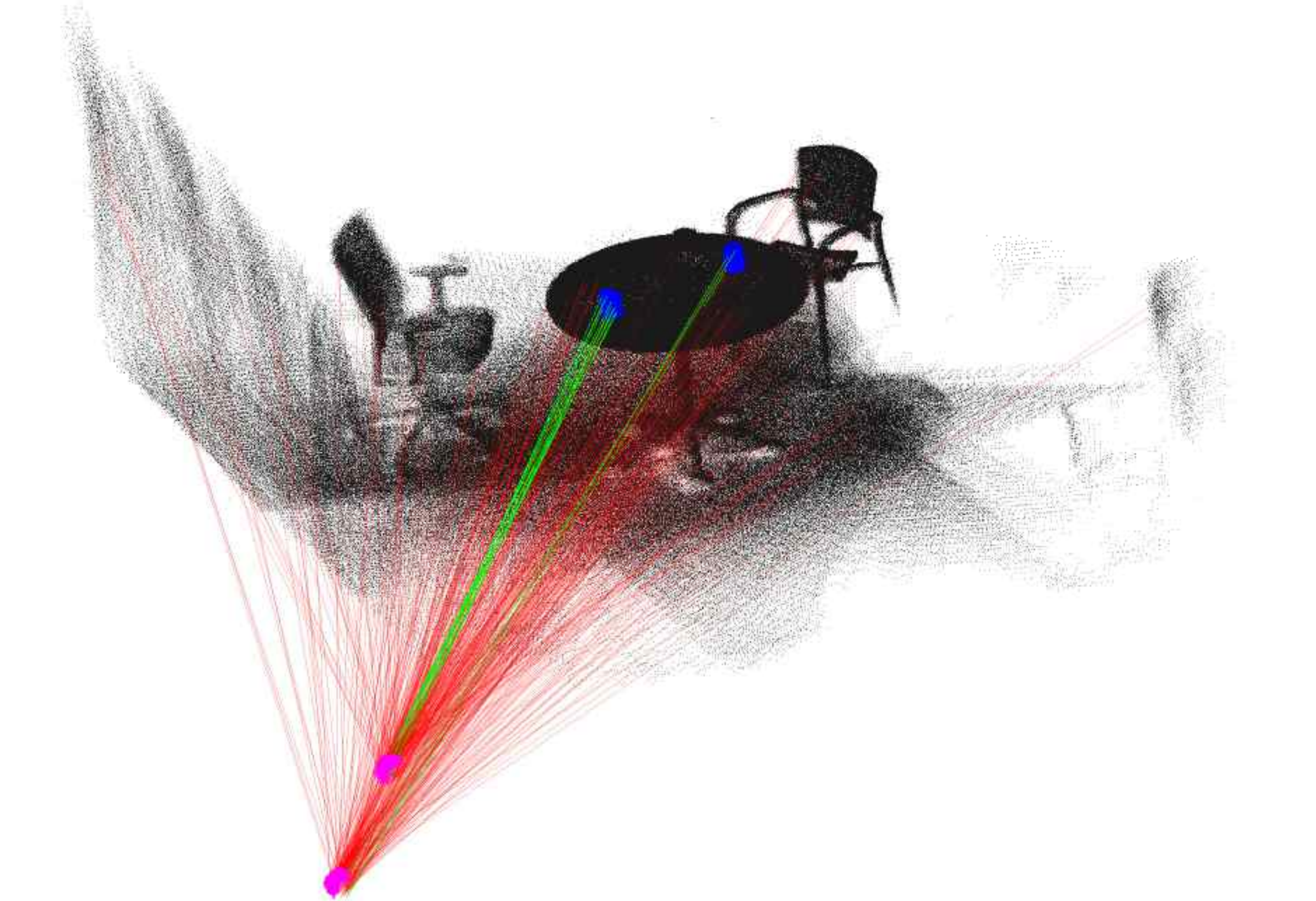}
\end{minipage}

& &

\begin{minipage}[t]{0.09\linewidth}
\centering
\includegraphics[width=1\linewidth]{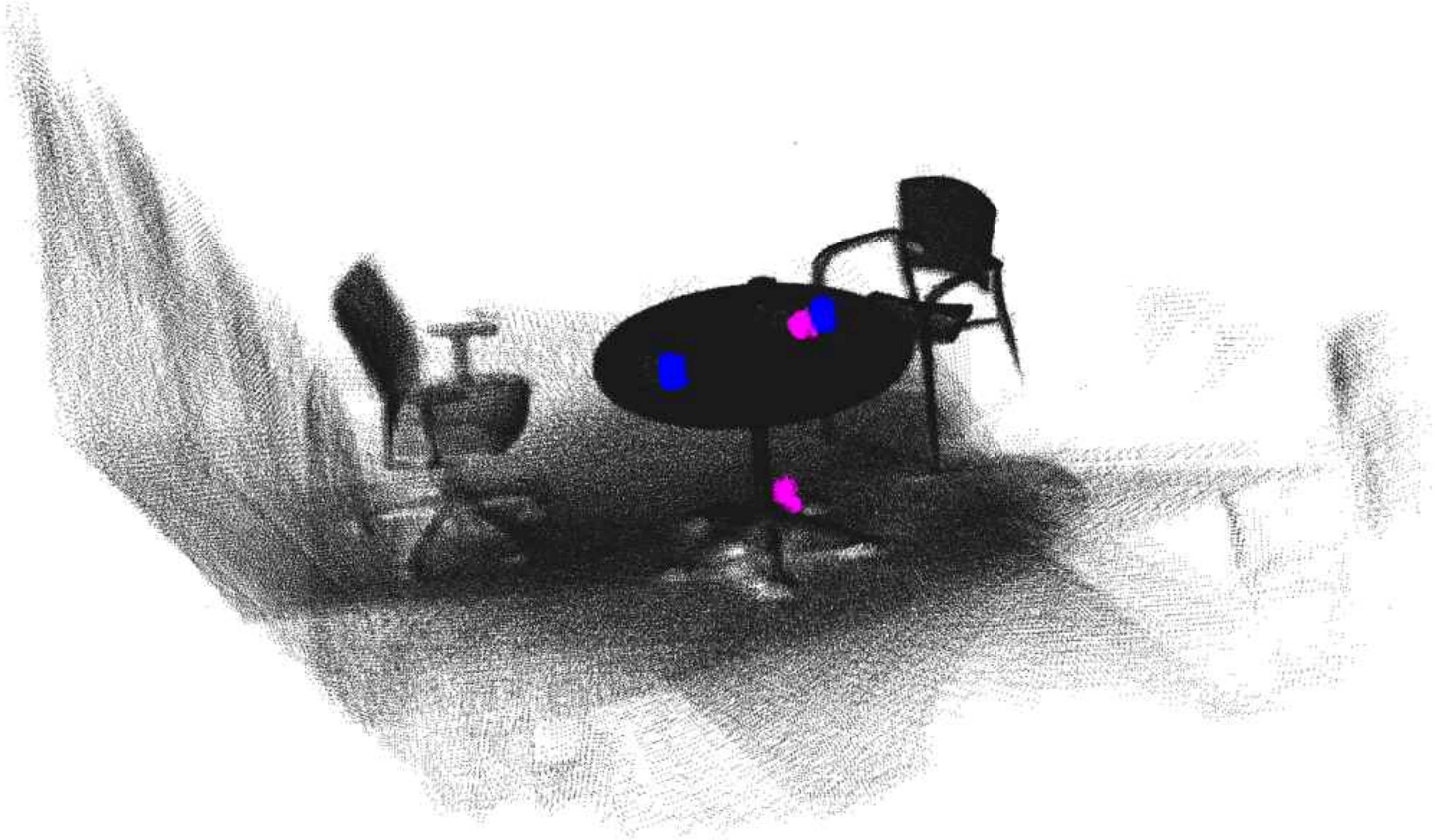}
\end{minipage}

&

\begin{minipage}[t]{0.09\linewidth}
\centering
\includegraphics[width=1\linewidth]{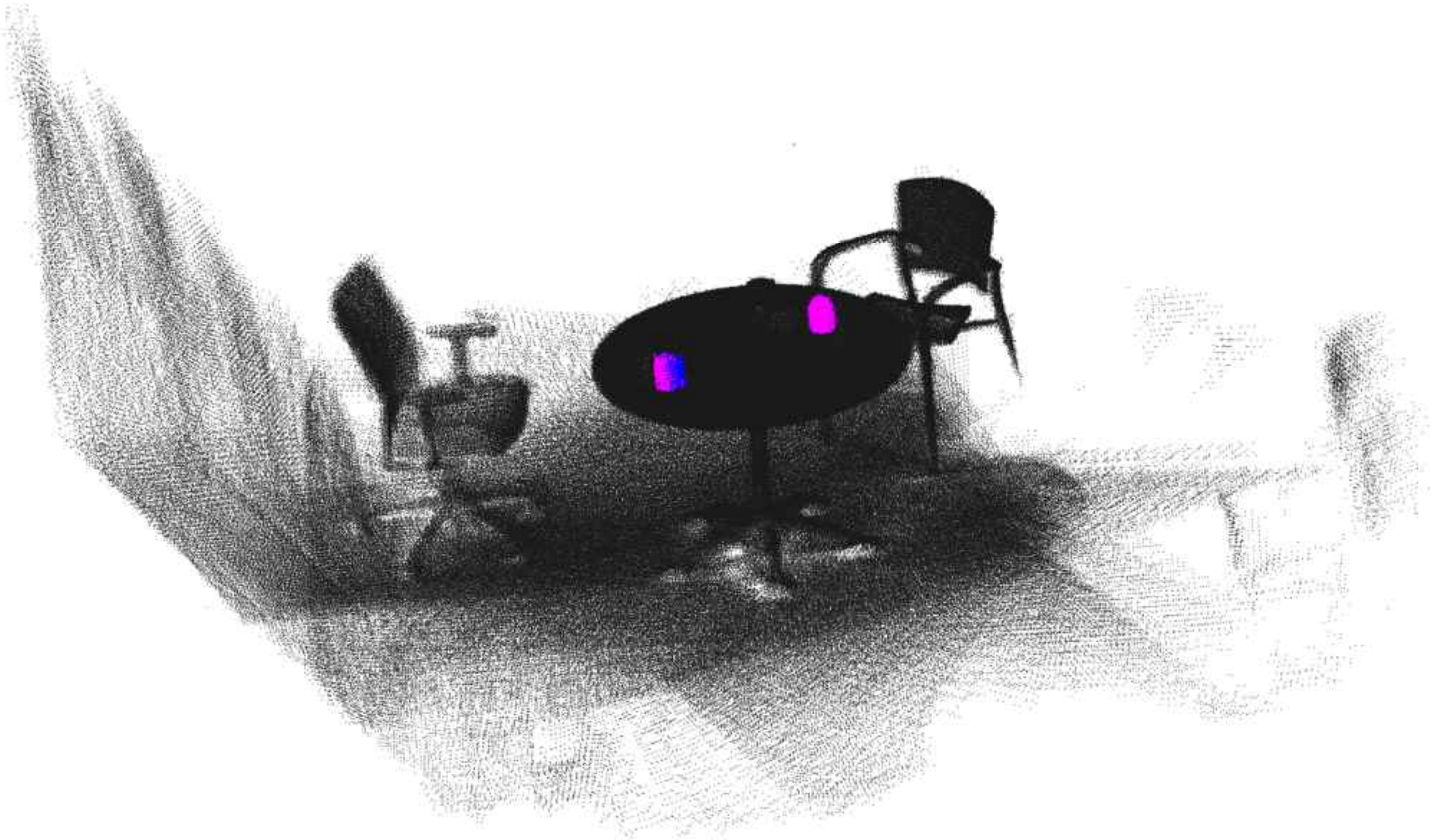}
\end{minipage}

&

\begin{minipage}[t]{0.09\linewidth}
\centering
\includegraphics[width=1\linewidth]{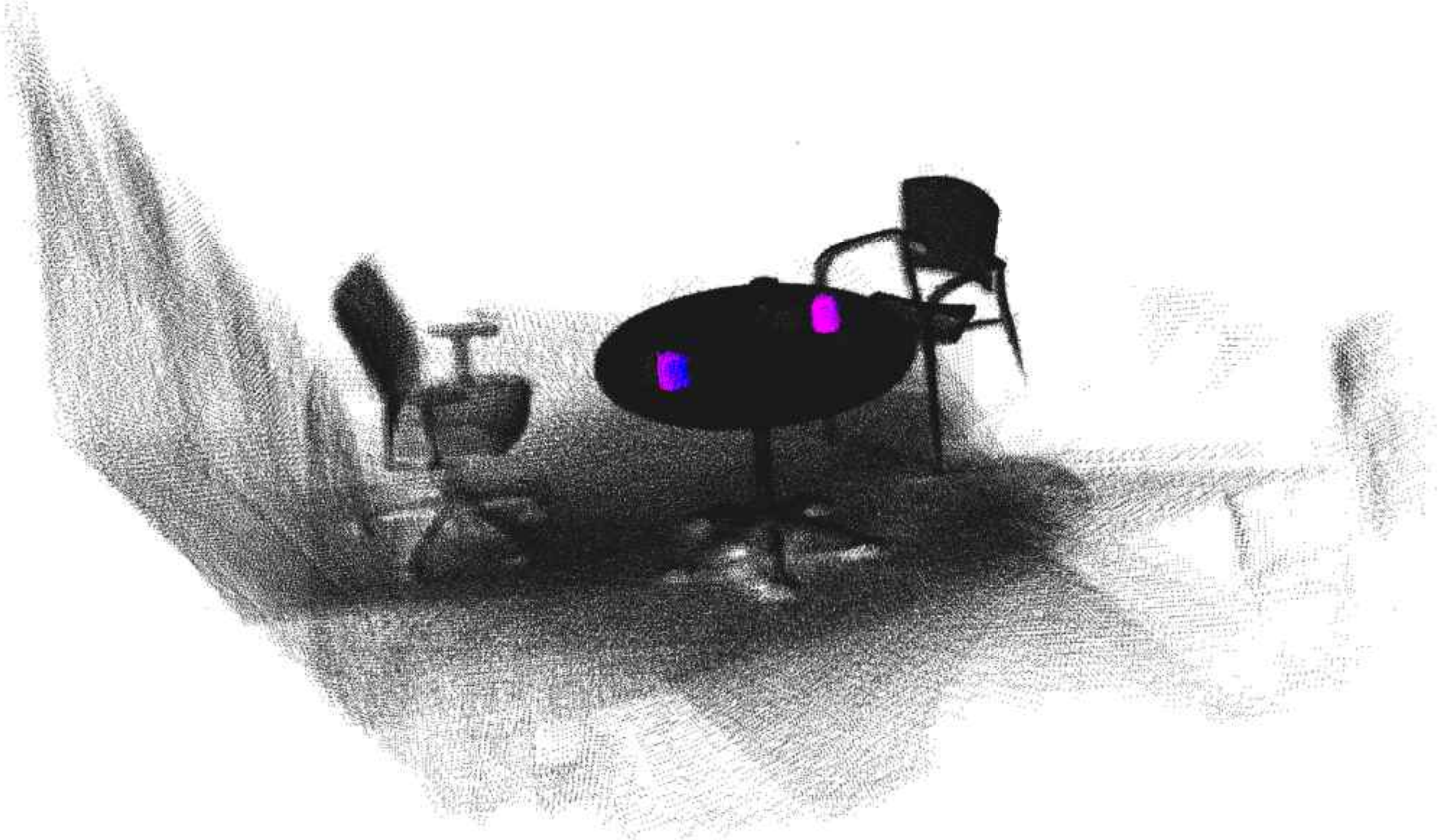}
\end{minipage}

&

\begin{minipage}[t]{0.09\linewidth}
\centering
\includegraphics[width=1\linewidth]{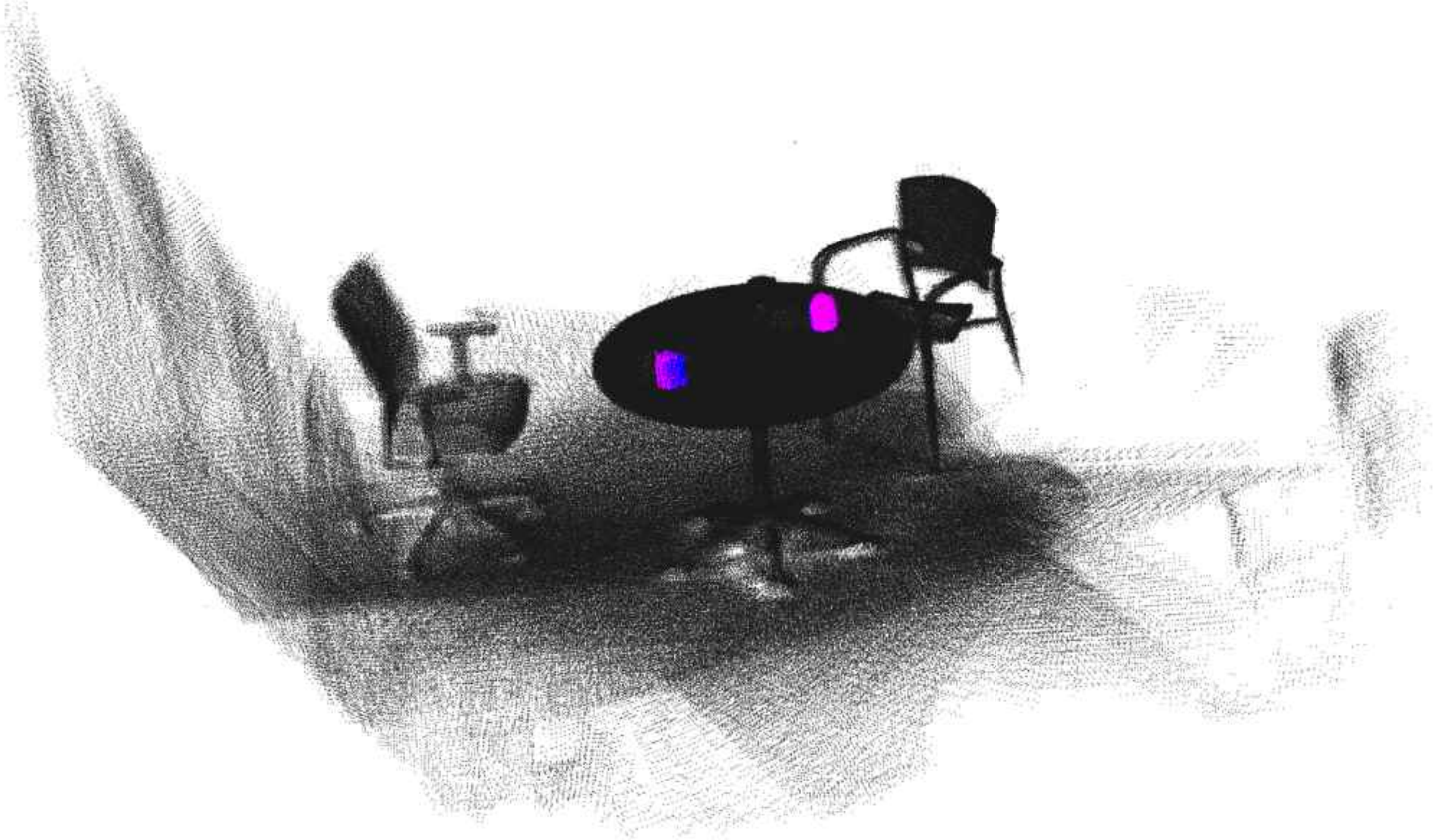}
\end{minipage}

&

\rotatebox{90}{\,\,\scriptsize{\textit{Scene-07}}\,}\,

& &

\begin{minipage}[t]{0.09\linewidth}
\centering
\includegraphics[width=1\linewidth]{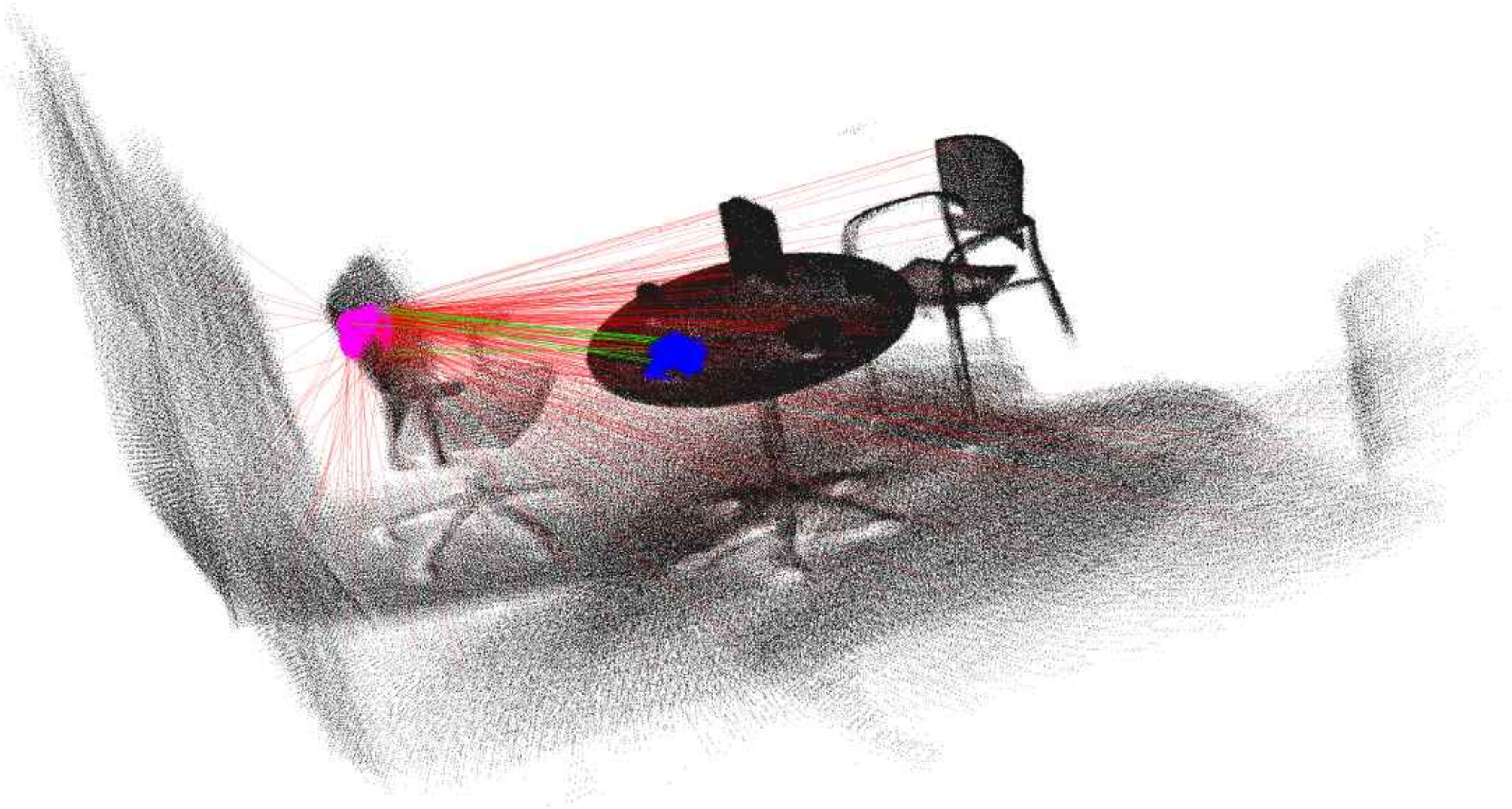}
\end{minipage}

& &

\begin{minipage}[t]{0.09\linewidth}
\centering
\includegraphics[width=1\linewidth]{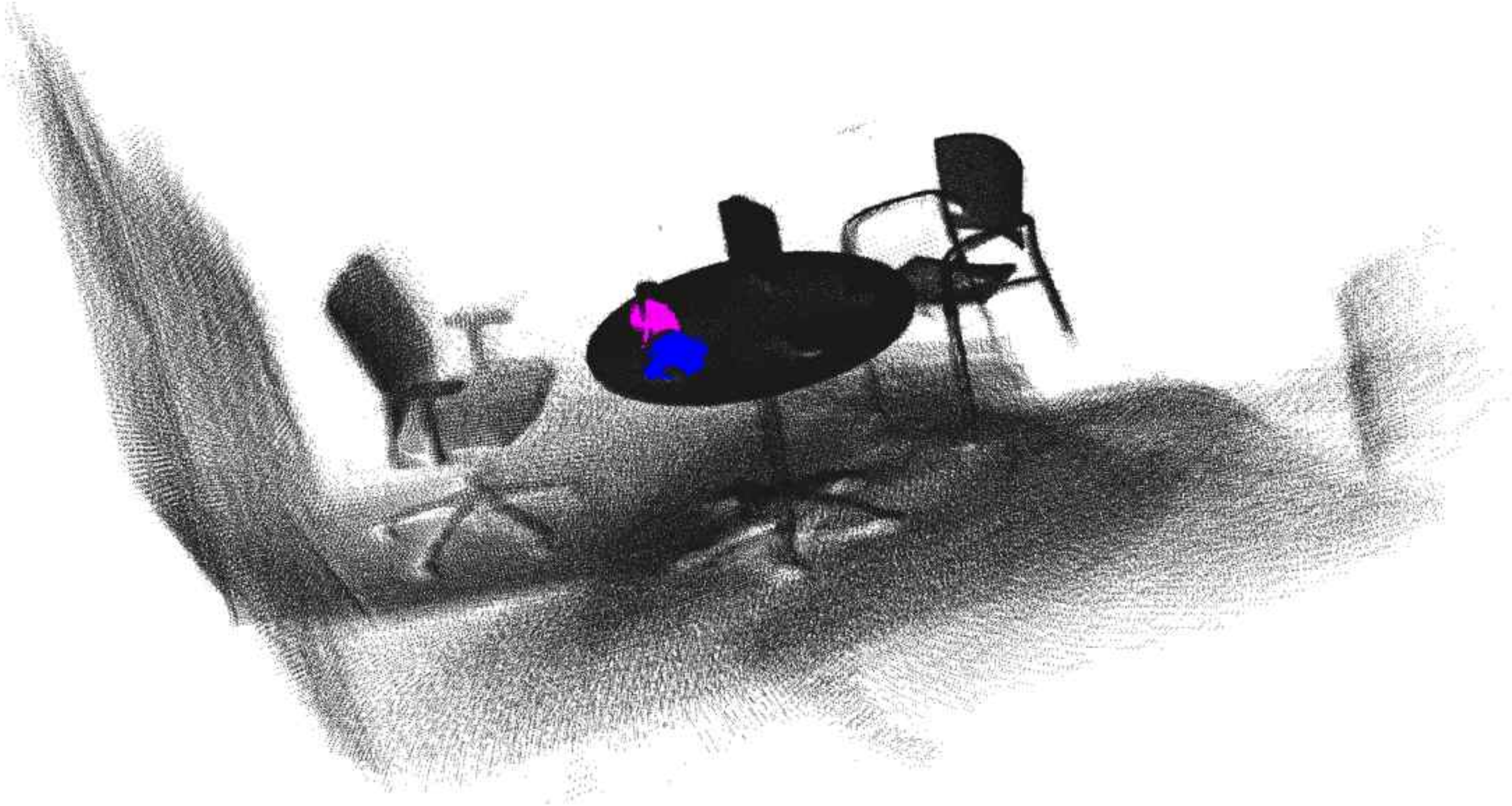}
\end{minipage}

&

\begin{minipage}[t]{0.09\linewidth}
\centering
\includegraphics[width=1\linewidth]{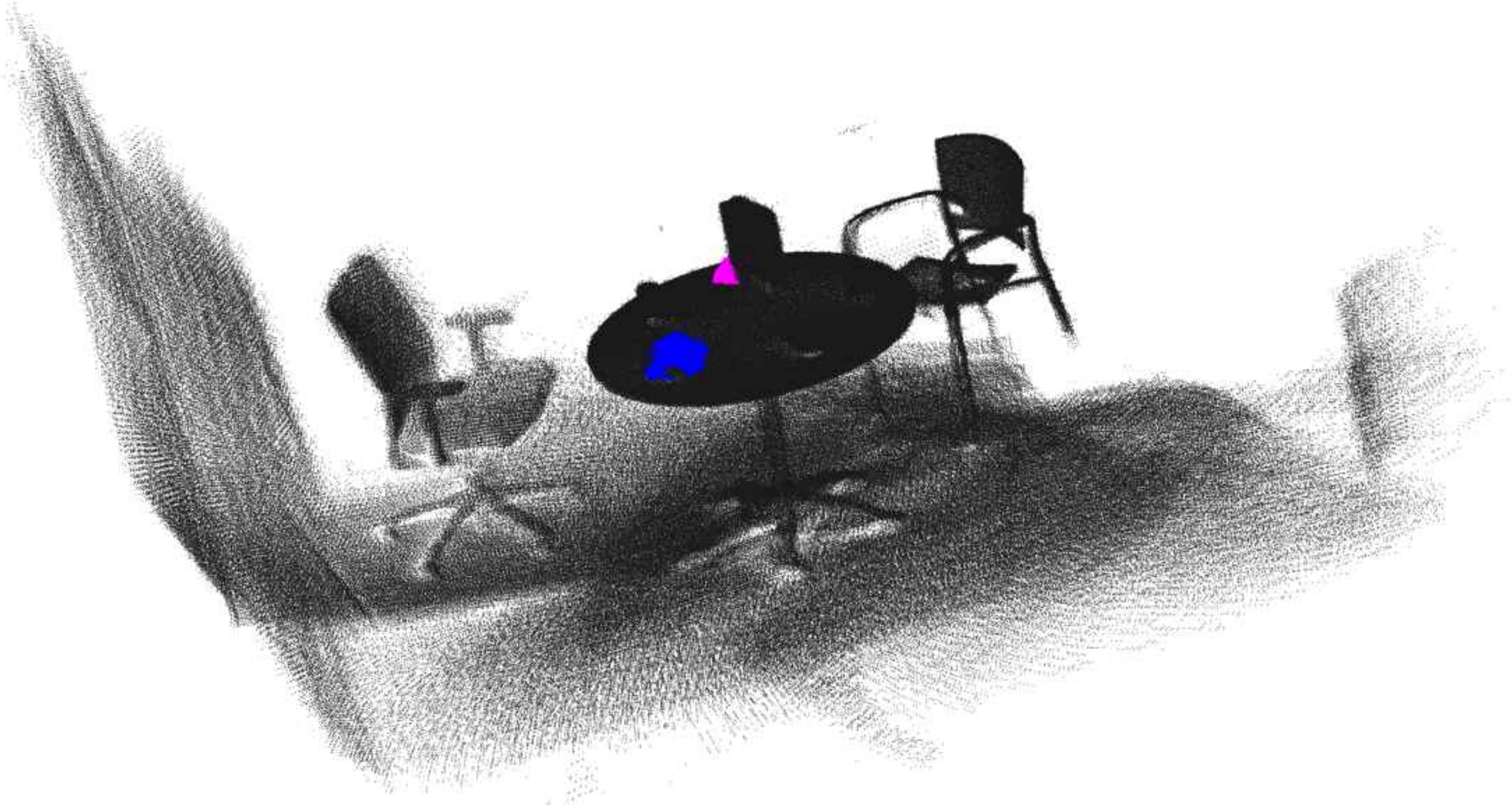}
\end{minipage}

&

\begin{minipage}[t]{0.09\linewidth}
\centering
\includegraphics[width=1\linewidth]{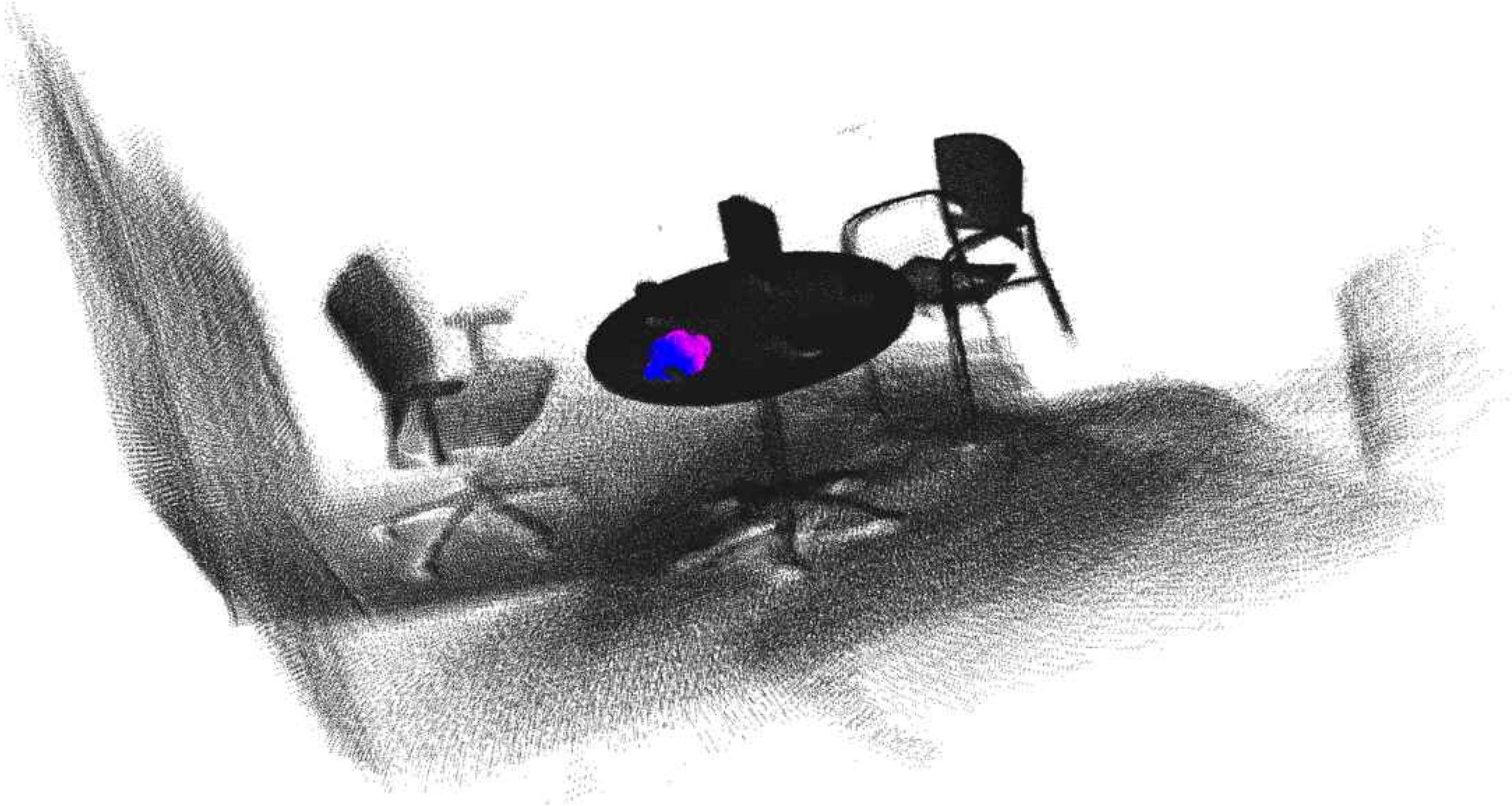}
\end{minipage}

&

\begin{minipage}[t]{0.09\linewidth}
\centering
\includegraphics[width=1\linewidth]{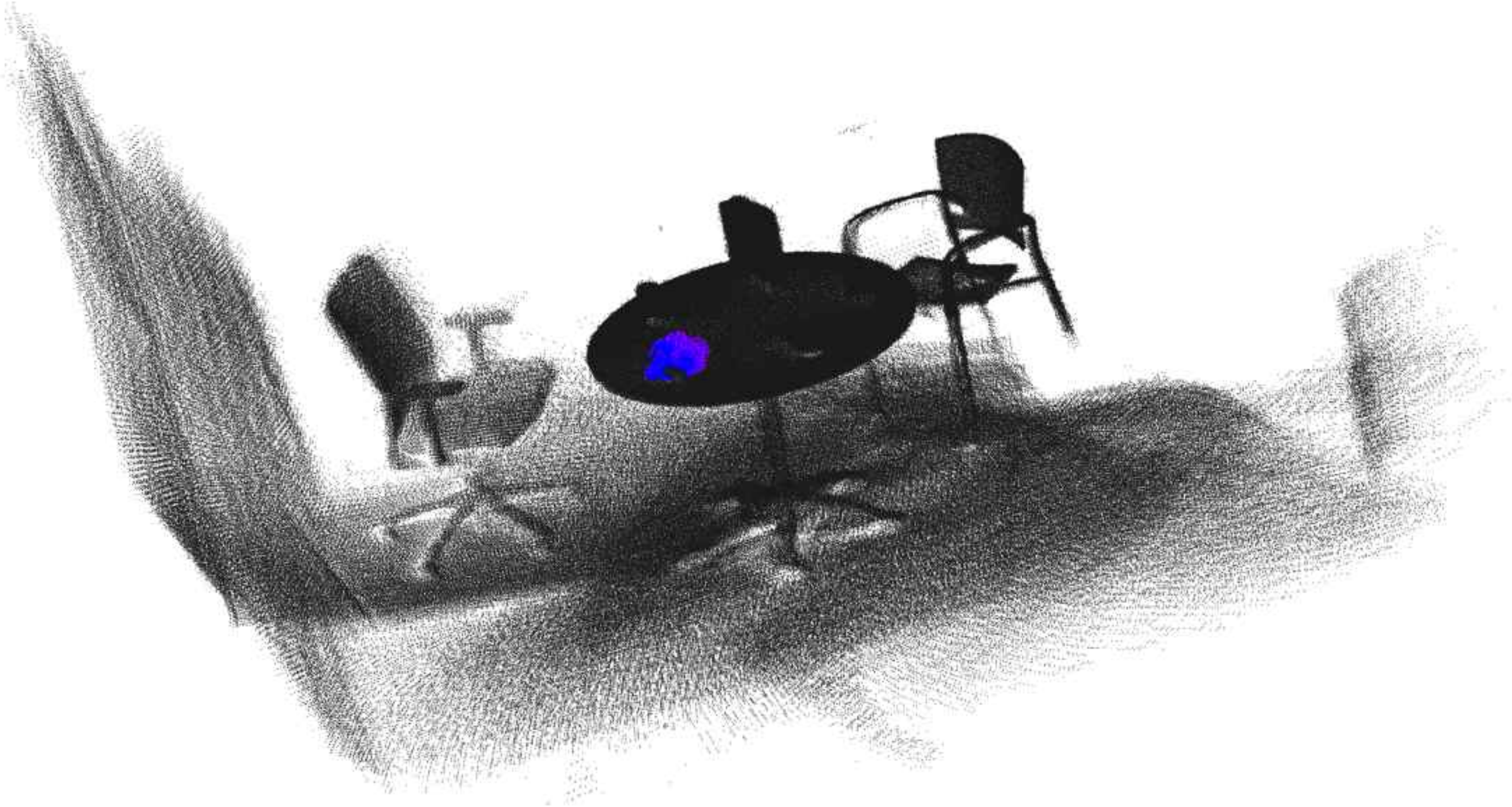}
\end{minipage}

\\

&&\tiny{$N$=592, 97.30\%}& & \,\tiny{116.46$^{\circ}$,1.97$m$,0.10$s$}\, &\,\tiny{{187.88}$^{\circ}$,{2.55}$m$,15.83$s$}\, &\,\tiny{{3.30}$^{\circ}$,{0.07}$m$,0.67$s$}\,&\,\tiny{\textbf{3.25}$^{\circ}$,\textbf{0.05}$m$,\textbf{0.41}$s$}\,

&

&&\tiny{$N$=504, 96.43\%}& & \,\tiny{175.44$^{\circ}$,2.39$m$,0.07$s$}\, &\,\tiny{{121.73}$^{\circ}$,{1.32}$m$,13.34$s$}\, &\,\tiny{\textbf{3.20}$^{\circ}$,\textbf{0.07}$m$,{1.11}$s$}\,&\,\tiny{\textbf{3.20}$^{\circ}$,\textbf{0.07}$m$,\textbf{0.31}$s$}\,

\\

\rotatebox{90}{\,\,\scriptsize{\textit{Scene-11}}\,}\,

& &

\begin{minipage}[t]{0.09\linewidth}
\centering
\includegraphics[width=1\linewidth]{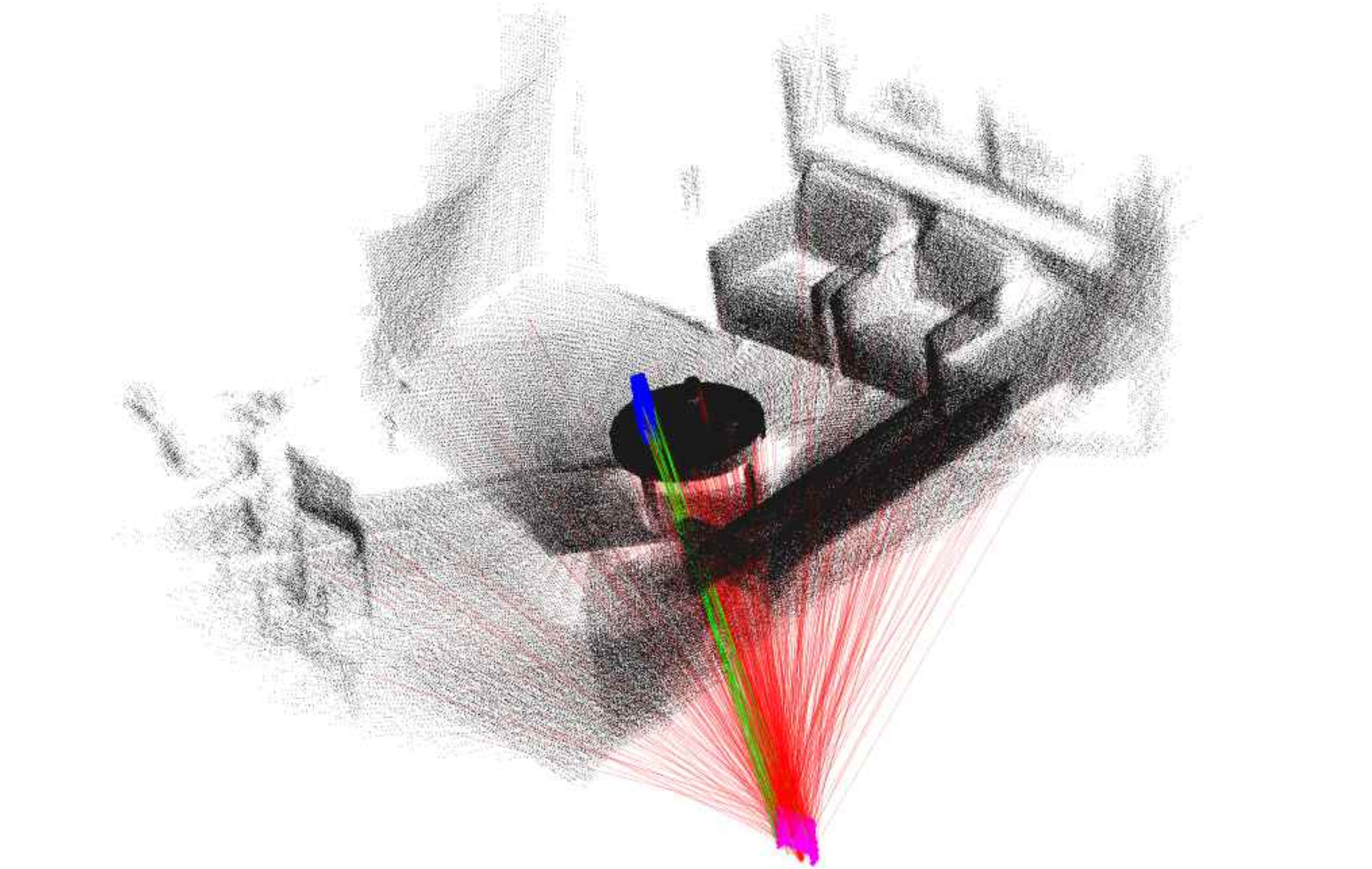}
\end{minipage}

& &

\begin{minipage}[t]{0.09\linewidth}
\centering
\includegraphics[width=1\linewidth]{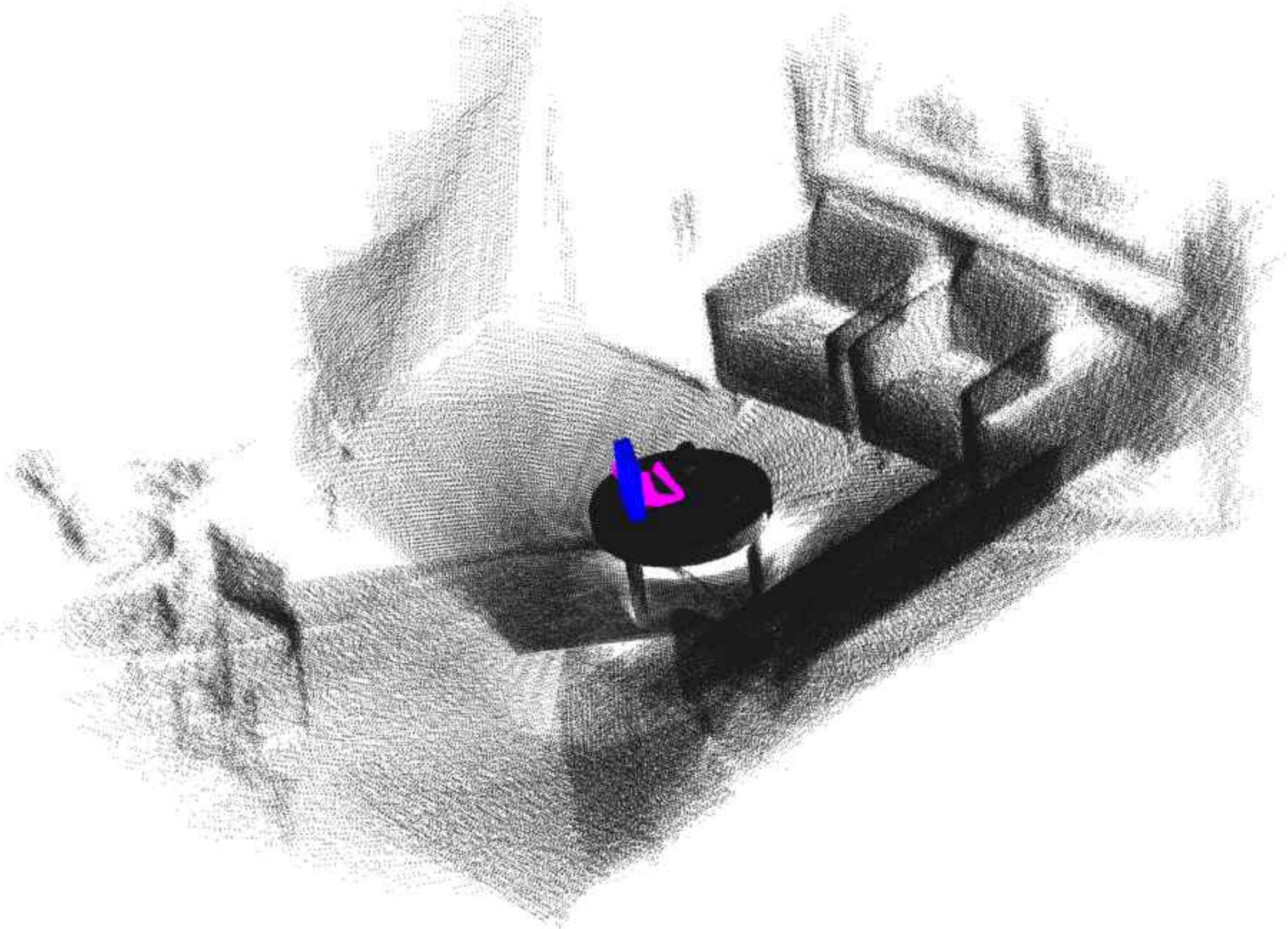}
\end{minipage}

&

\begin{minipage}[t]{0.09\linewidth}
\centering
\includegraphics[width=1\linewidth]{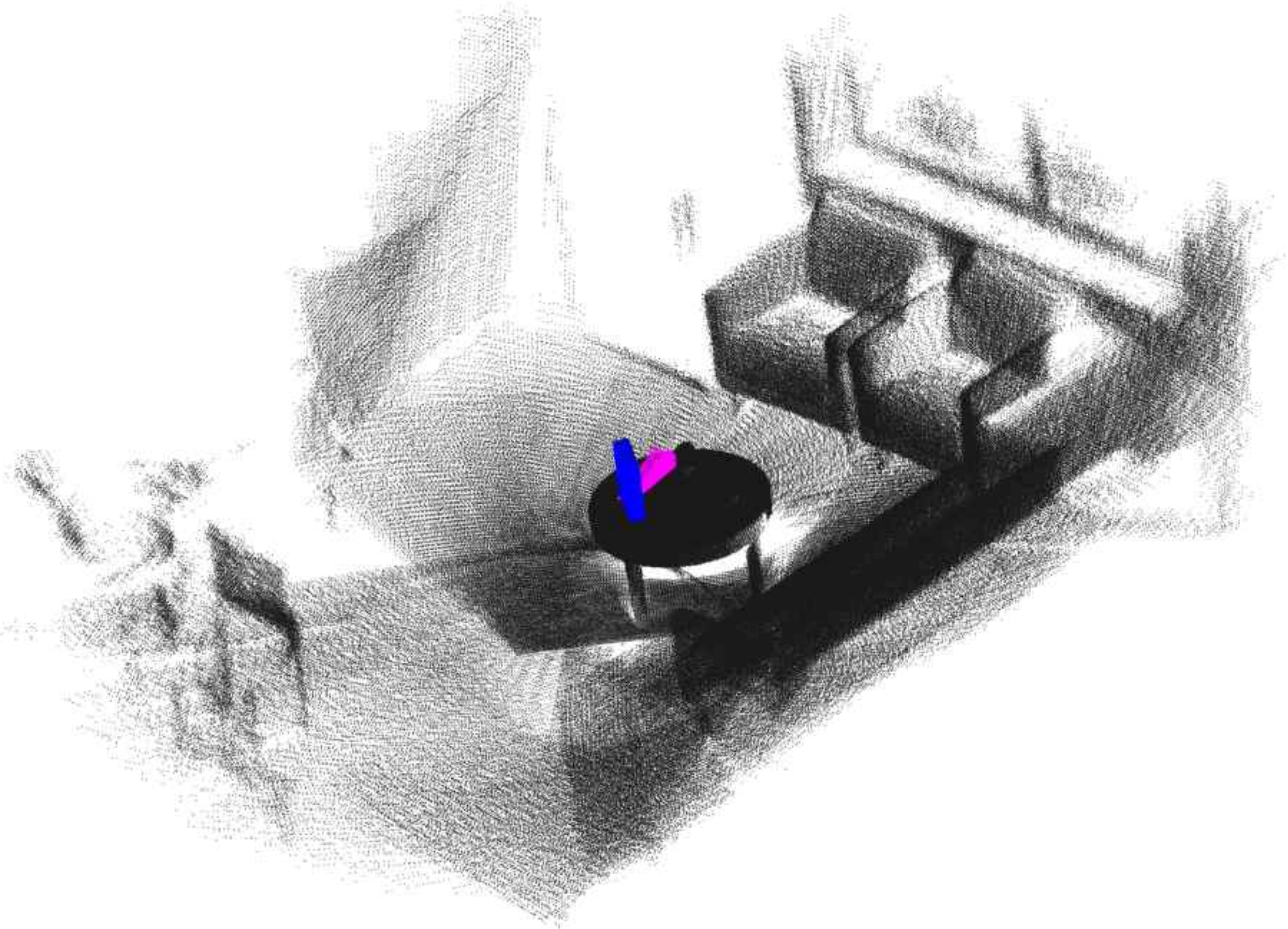}
\end{minipage}

&

\begin{minipage}[t]{0.09\linewidth}
\centering
\includegraphics[width=1\linewidth]{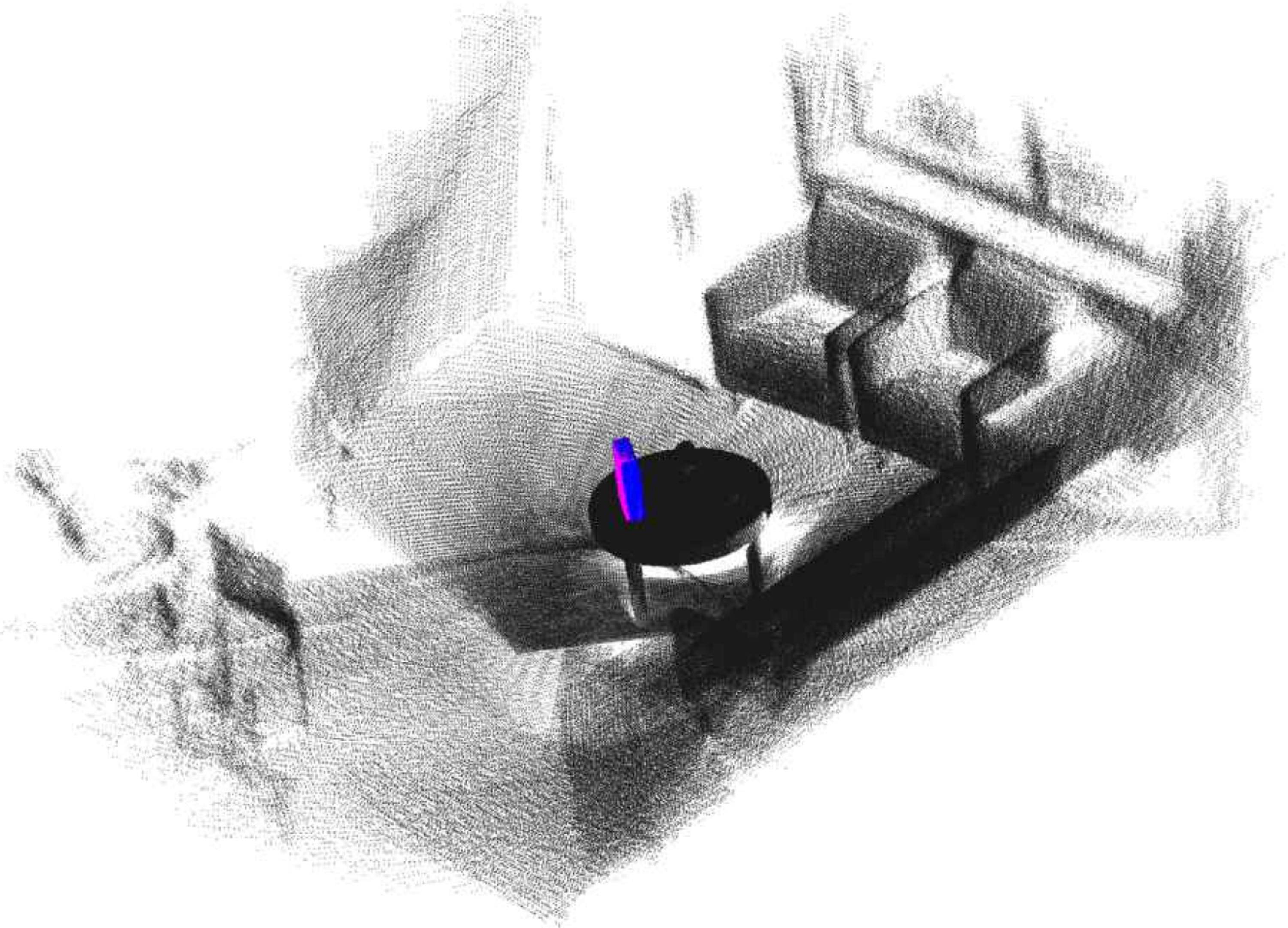}
\end{minipage}

&

\begin{minipage}[t]{0.09\linewidth}
\centering
\includegraphics[width=1\linewidth]{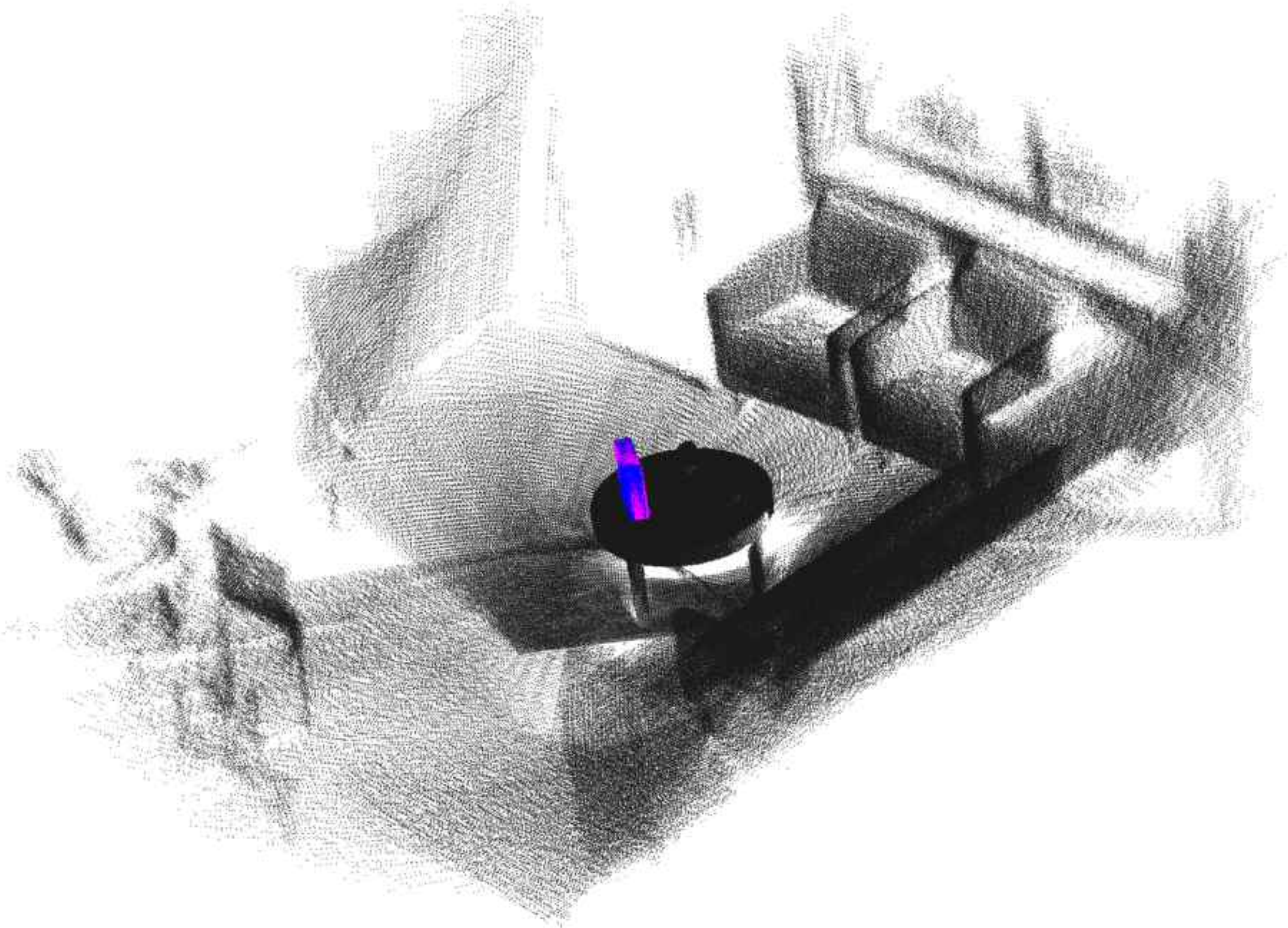}
\end{minipage}

&

\rotatebox{90}{\,\,\scriptsize{\textit{Scene-14}}\,}\,

& &

\begin{minipage}[t]{0.07\linewidth}
\centering
\includegraphics[width=1\linewidth]{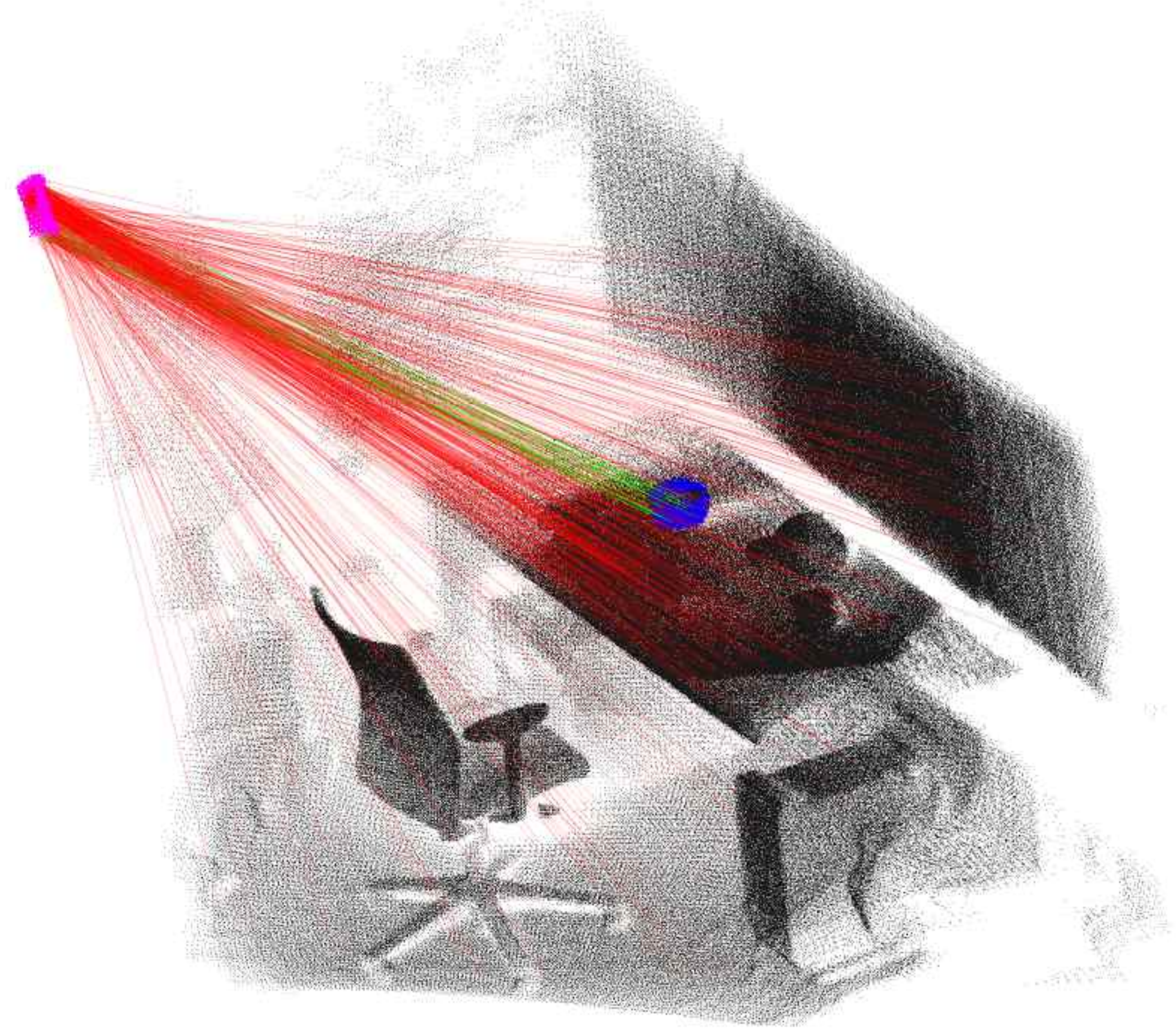}
\end{minipage}

& &

\begin{minipage}[t]{0.07\linewidth}
\centering
\includegraphics[width=1\linewidth]{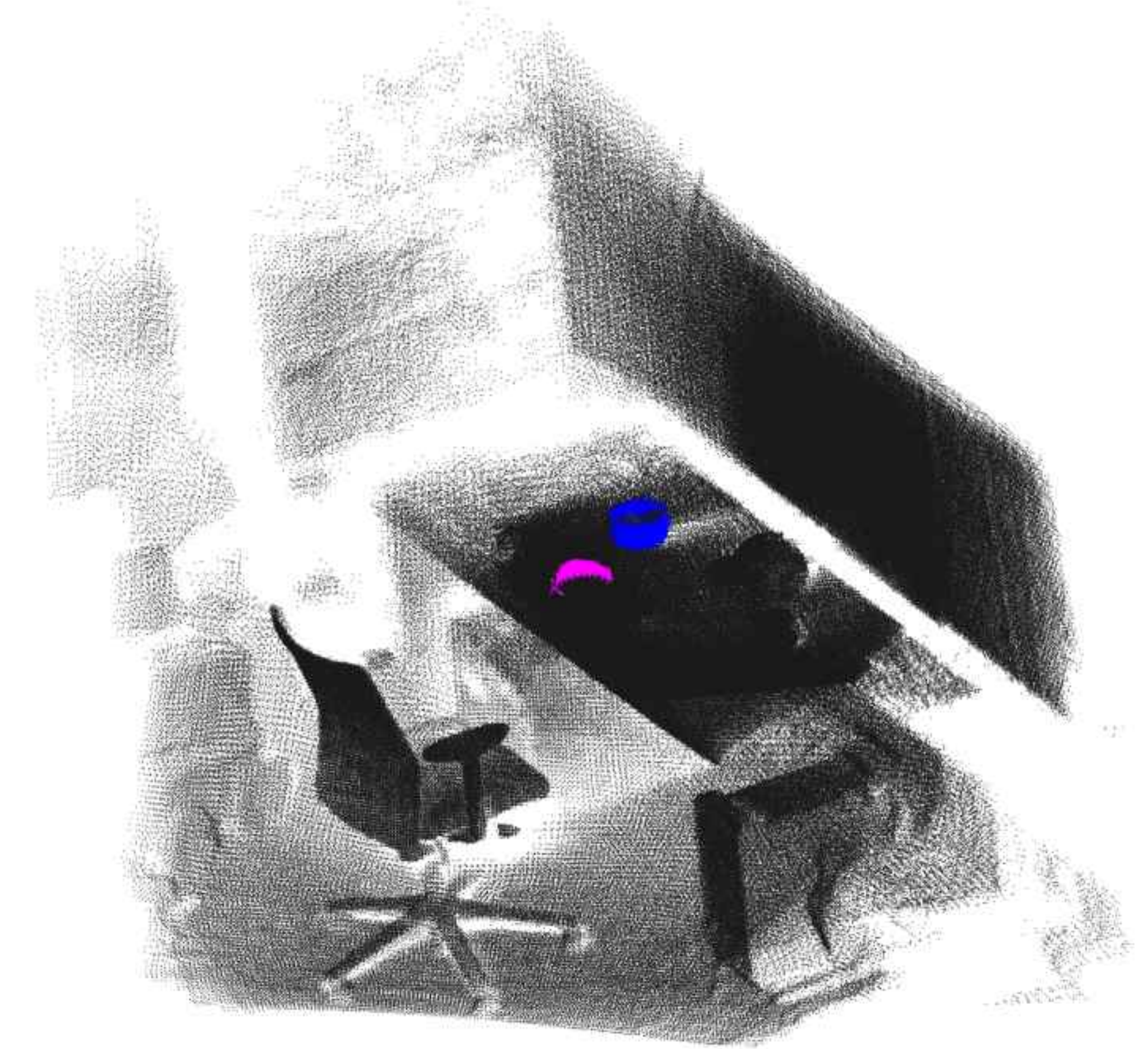}
\end{minipage}

&

\begin{minipage}[t]{0.07\linewidth}
\centering
\includegraphics[width=1\linewidth]{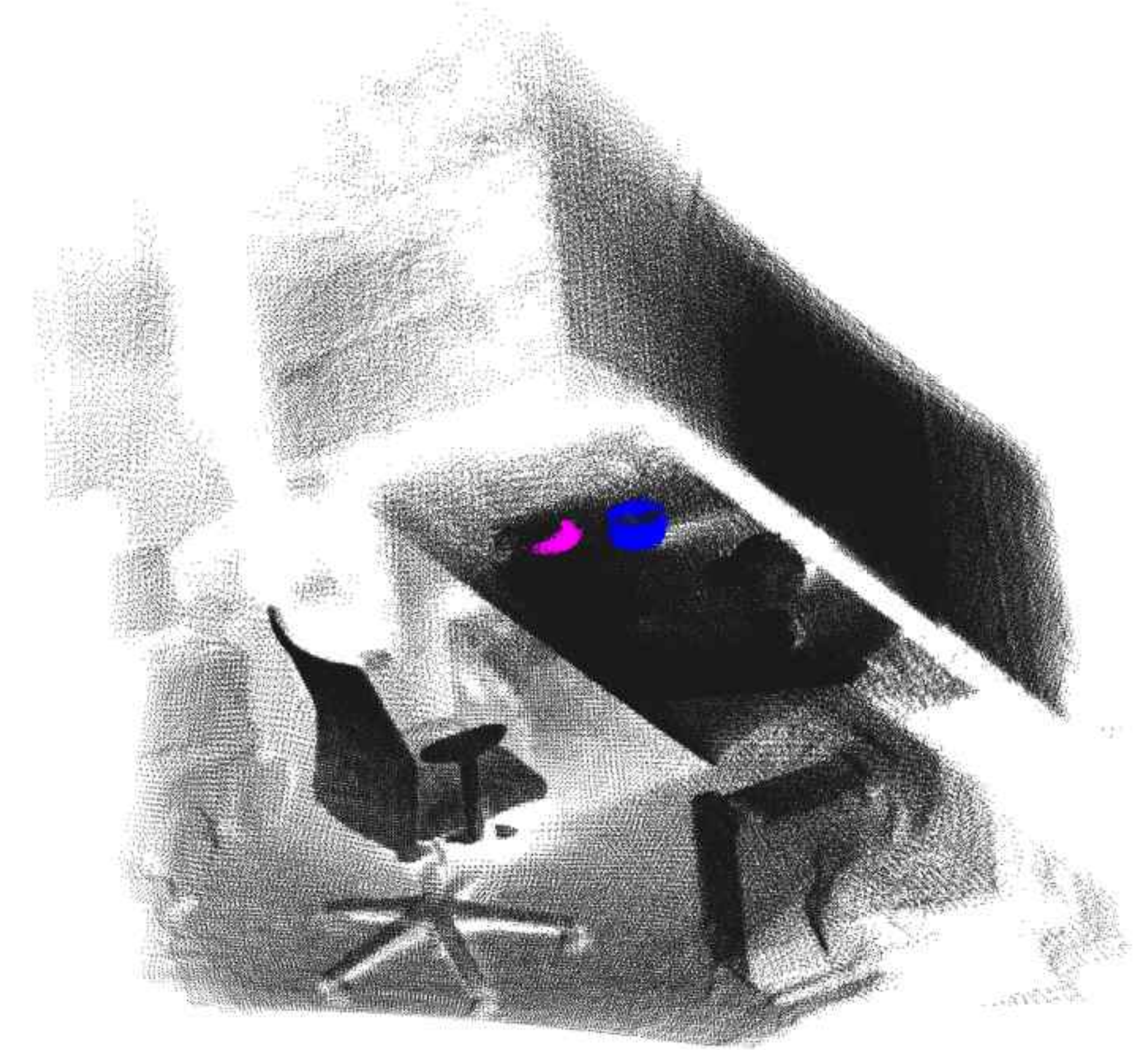}
\end{minipage}

&

\begin{minipage}[t]{0.07\linewidth}
\centering
\includegraphics[width=1\linewidth]{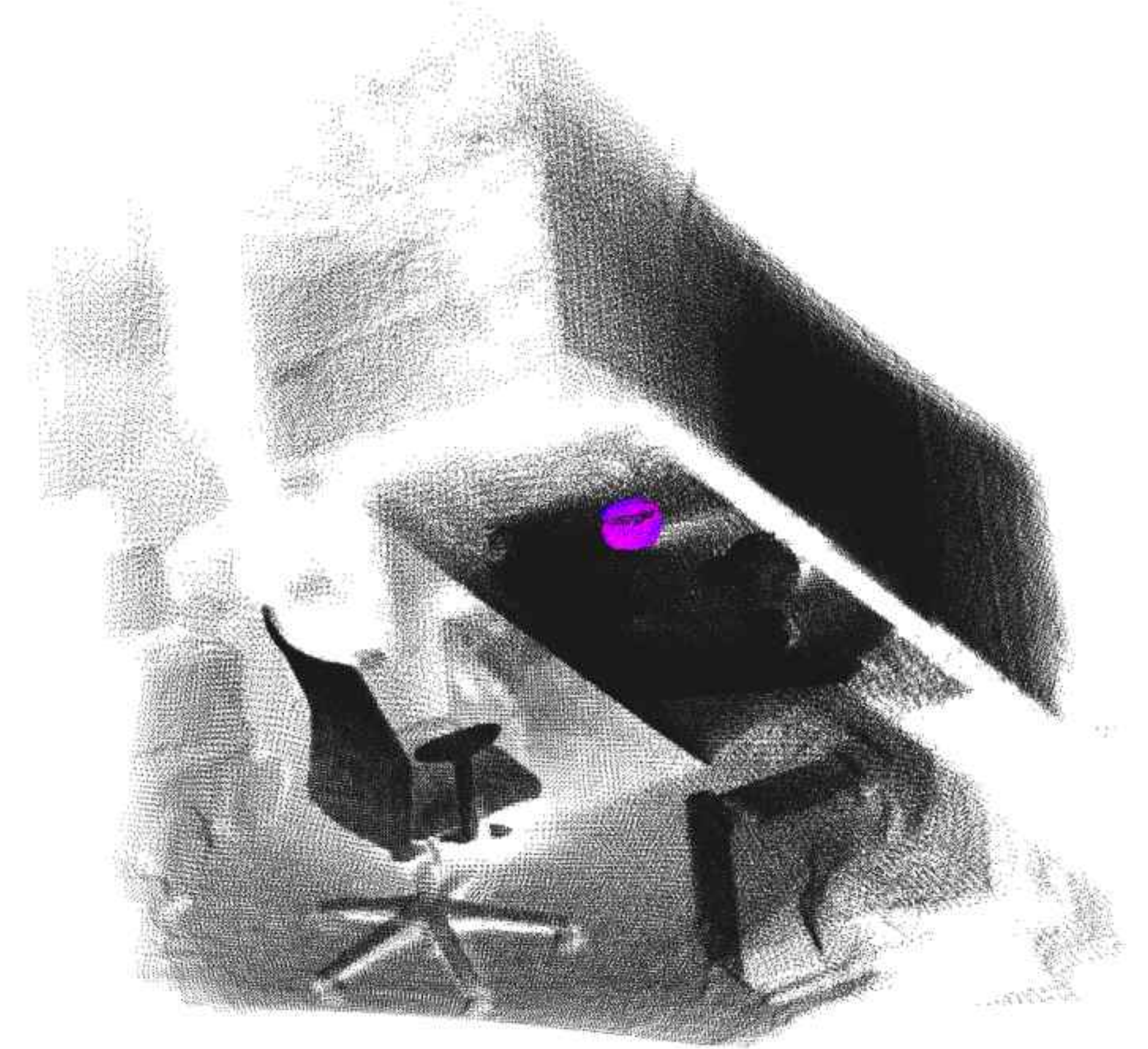}
\end{minipage}

&

\begin{minipage}[t]{0.07\linewidth}
\centering
\includegraphics[width=1\linewidth]{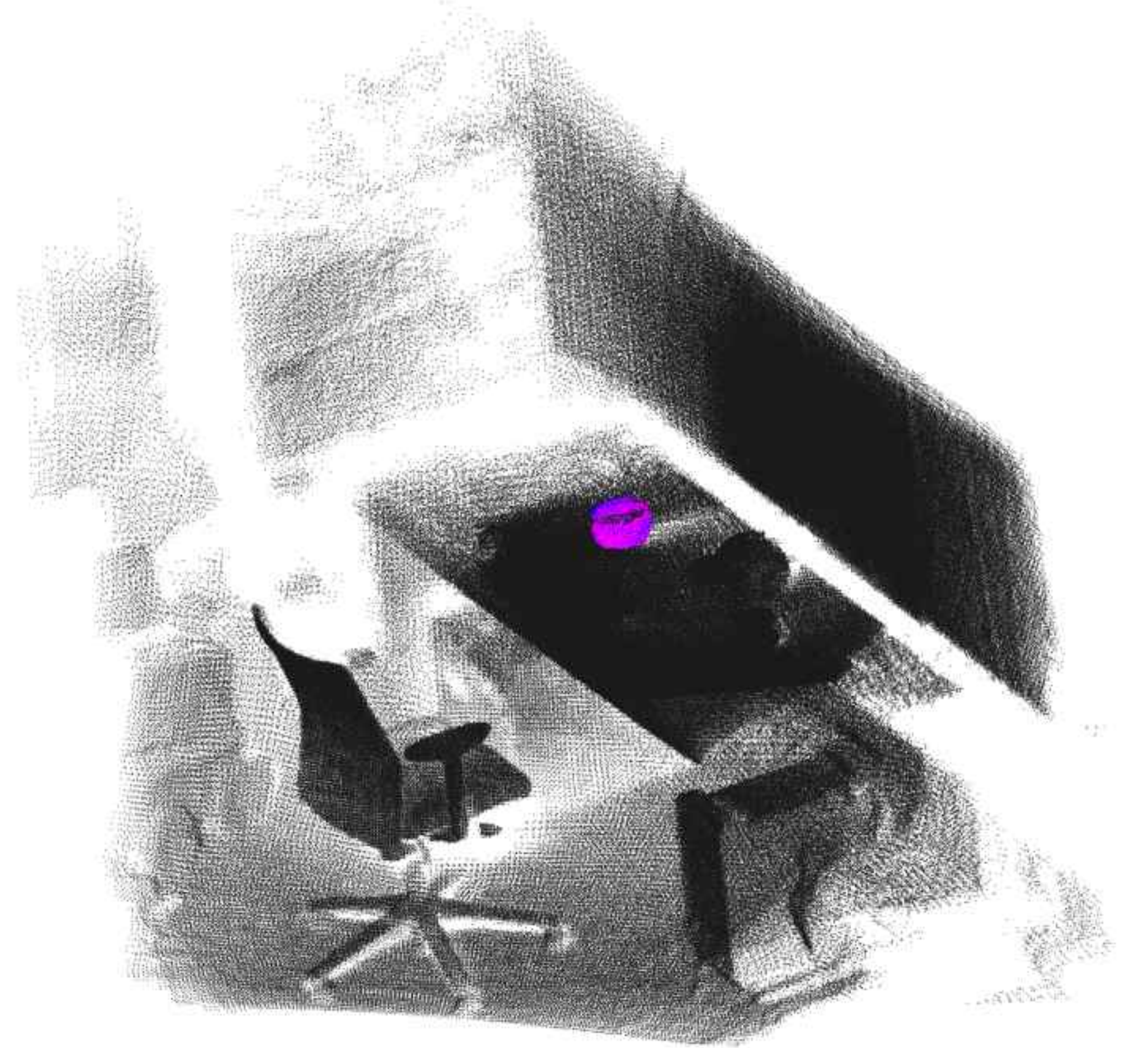}
\end{minipage}

\end{tabular}

\caption{Object localization results on RGB-D scenes dataset~\cite{lai2011large}. The first column shows FPFH correspondences (correspondence number and outlier ratio is given on top), and the rest columns show the registration results (reprojecting the object back to the scene with the transformation estimated) using GNC-TLS, FLO-RANSAC, GORE+RANSAC and TriVoC. On top of each result, we show the $\boldsymbol{R}$ error, $\boldsymbol{t}$ error and runtime. Best results are in \textbf{bold} font.}
\label{obj-local}
\vspace{-2mm}
\end{figure*}

\begin{figure*}[h]
\centering
\setlength\tabcolsep{0pt}
\addtolength{\tabcolsep}{-0.2pt}
\begin{tabular}{c|cc|c|c|c|c}
\quad &\,\scriptsize{Correspondences}\, &\,&  \scriptsize{GNC-TLS} & \scriptsize{FLO-RANSAC} & \scriptsize{GORE+RANSAC} & \scriptsize{TriVoC}
\\
\hline

 & \scriptsize{$N$=1000} && \scriptsize{\textcolor[rgb]{1,0,0}{Failed},\,\verb|\|,\,0.10$s$} &\scriptsize{\textcolor[rgb]{0,0.7,0}{Successful},\,0.28,\,8.46$s$}&\scriptsize{\textcolor[rgb]{0,0.7,0}{Successful},\,{0.34},\,5.59$s$}&\scriptsize{\textcolor[rgb]{0,0.7,0}{Successful},\,\textbf{0.24},\,\textbf{1.29}$s$}

\\

\rotatebox{90}{\,\,\footnotesize{\textit{red kitchen}}\,}\,
&
\,\,
\begin{minipage}[t]{0.1\linewidth}
\centering
\includegraphics[width=1\linewidth]{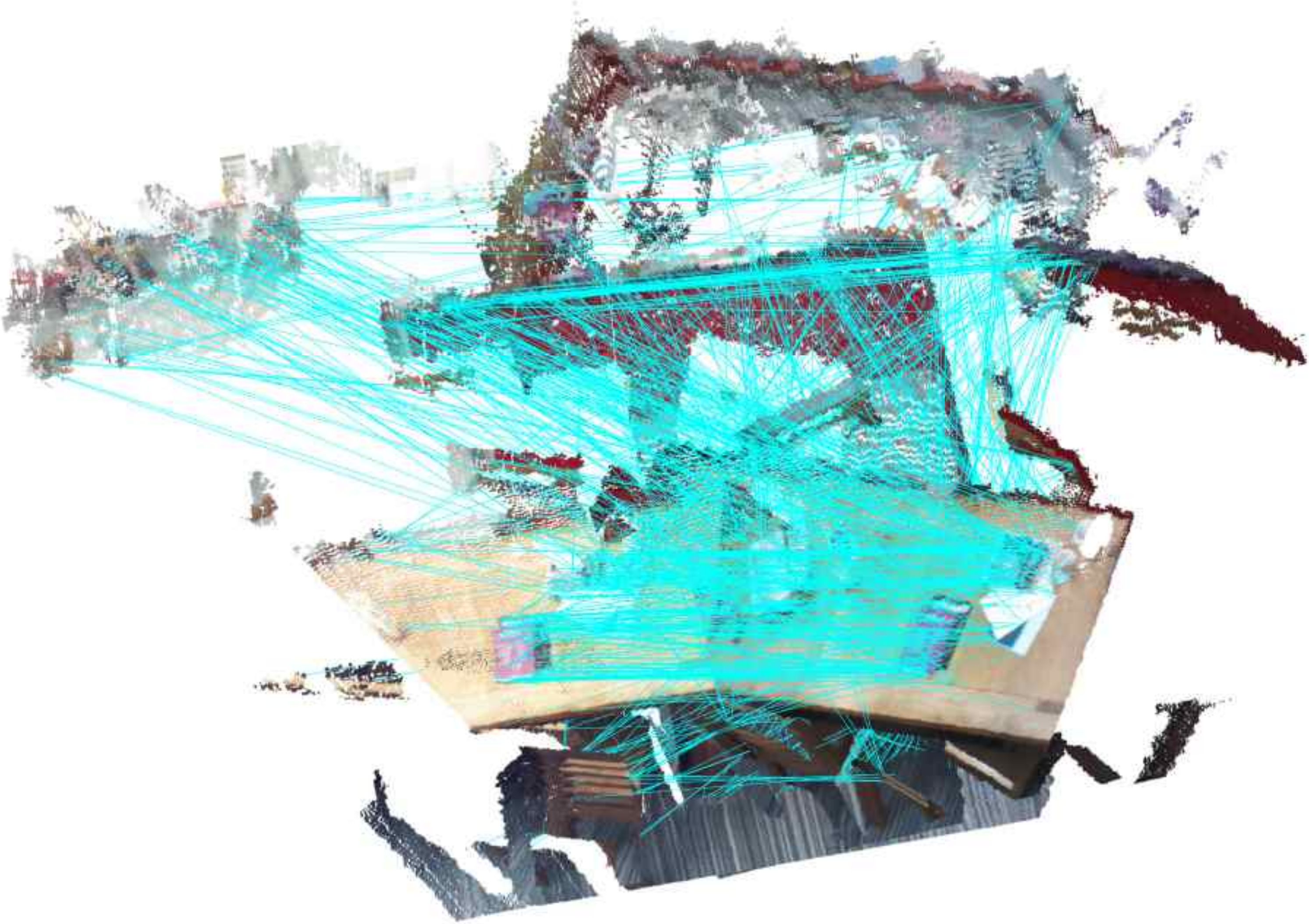}
\end{minipage}\,\,
& &
\,\,
\begin{minipage}[t]{0.19\linewidth}
\centering
\includegraphics[width=.48\linewidth]{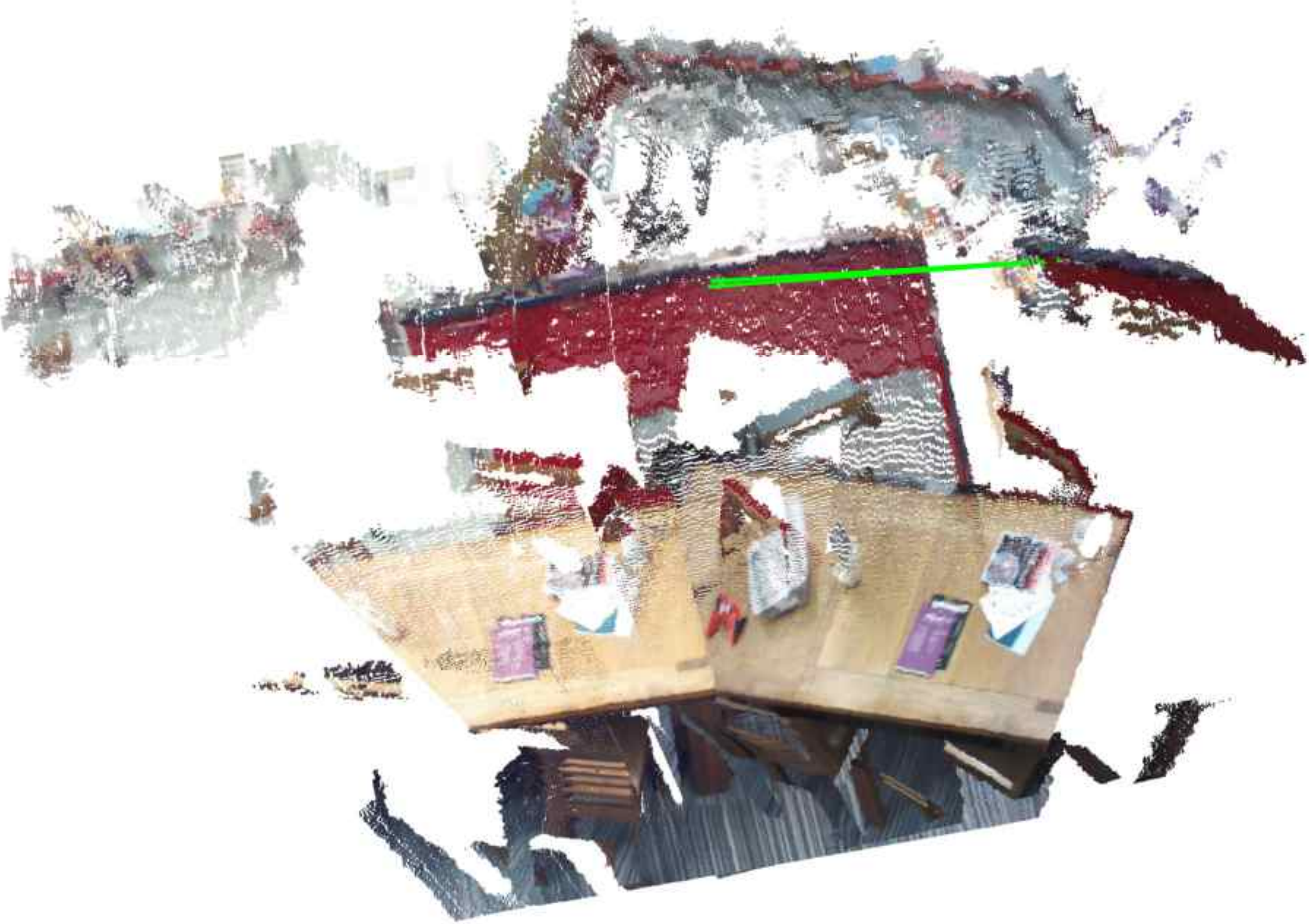}
\includegraphics[width=.48\linewidth]{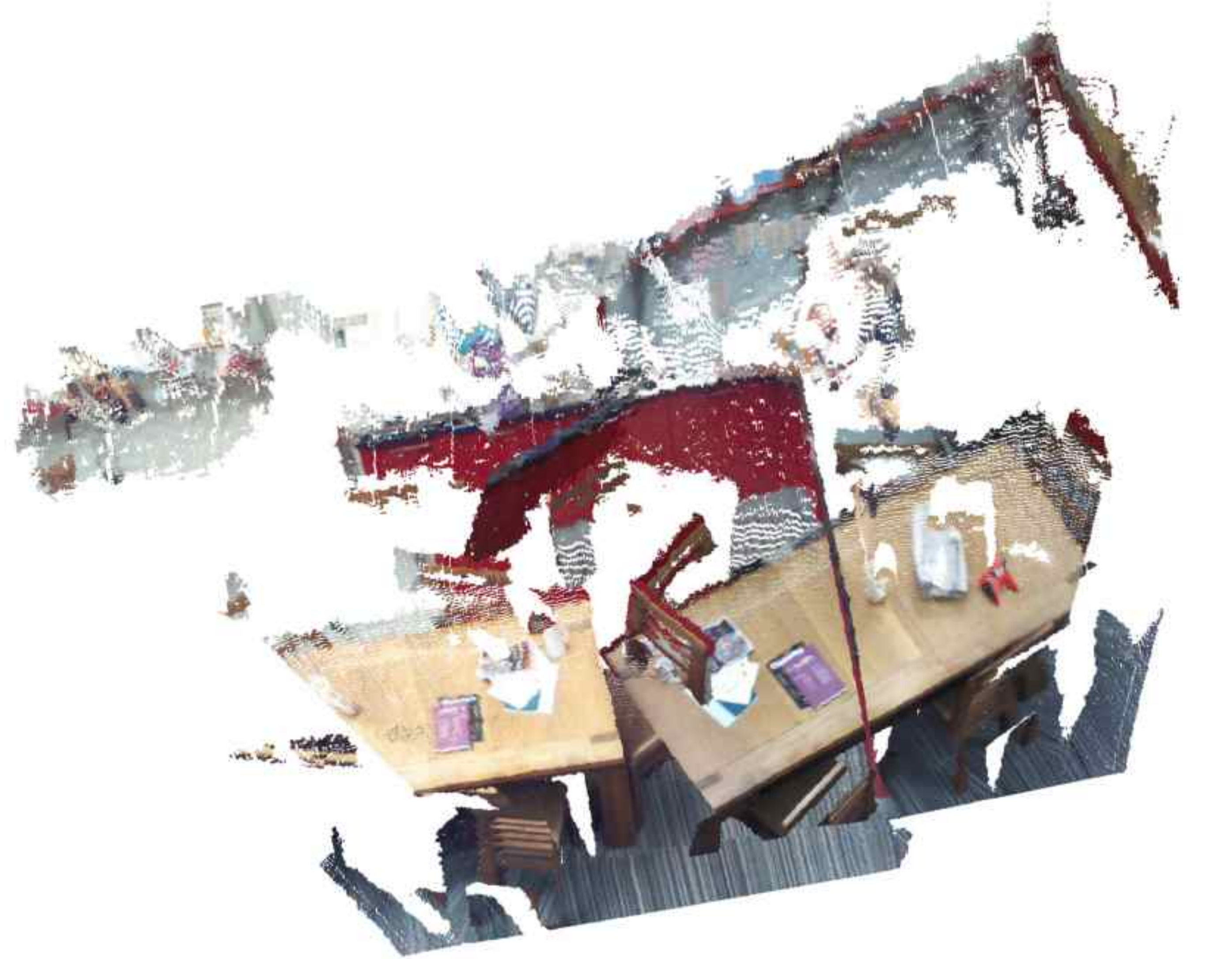}
\end{minipage}\,\,
&
\,\,
\begin{minipage}[t]{0.19\linewidth}
\centering
\includegraphics[width=.48\linewidth]{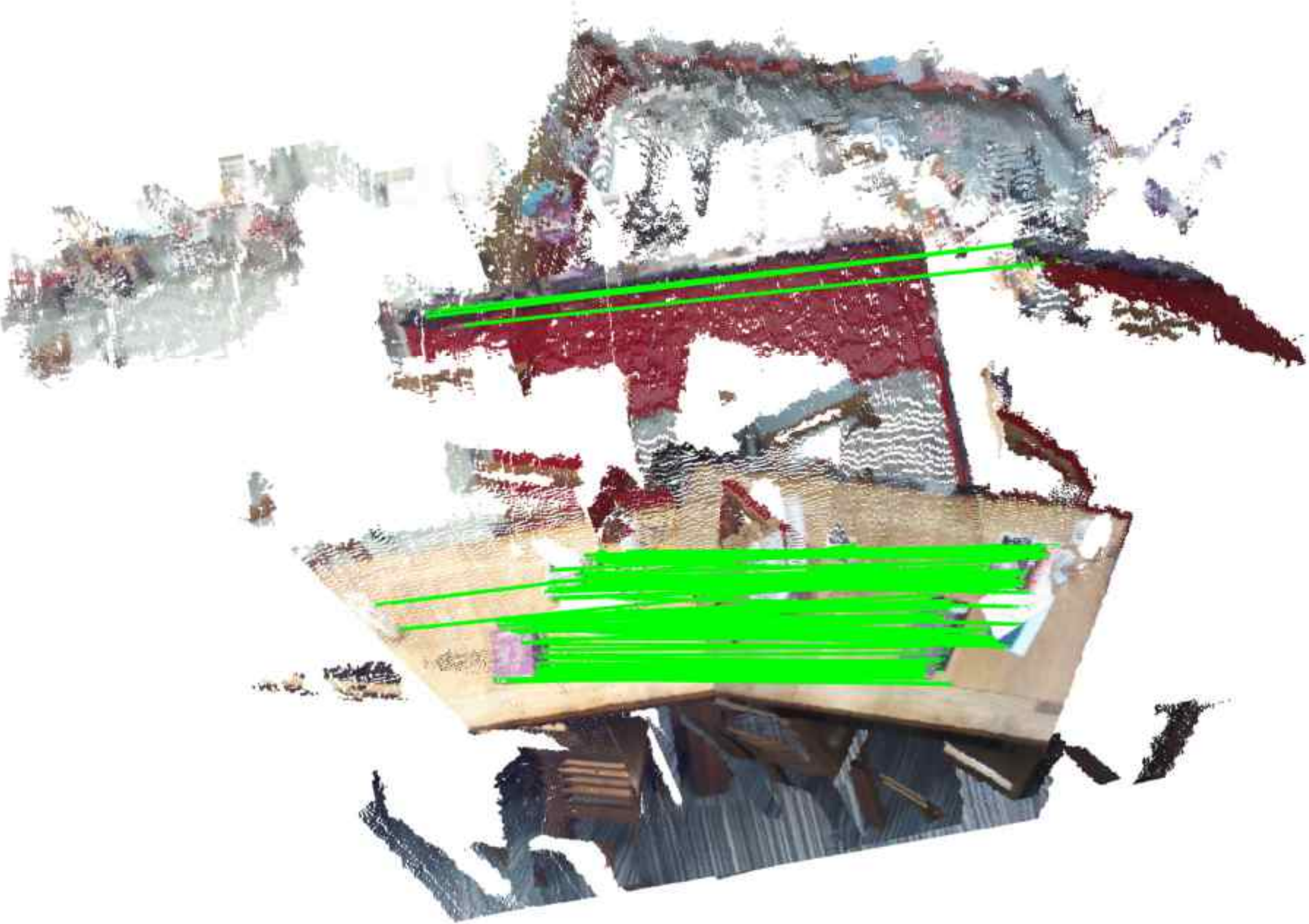}
\includegraphics[width=.48\linewidth]{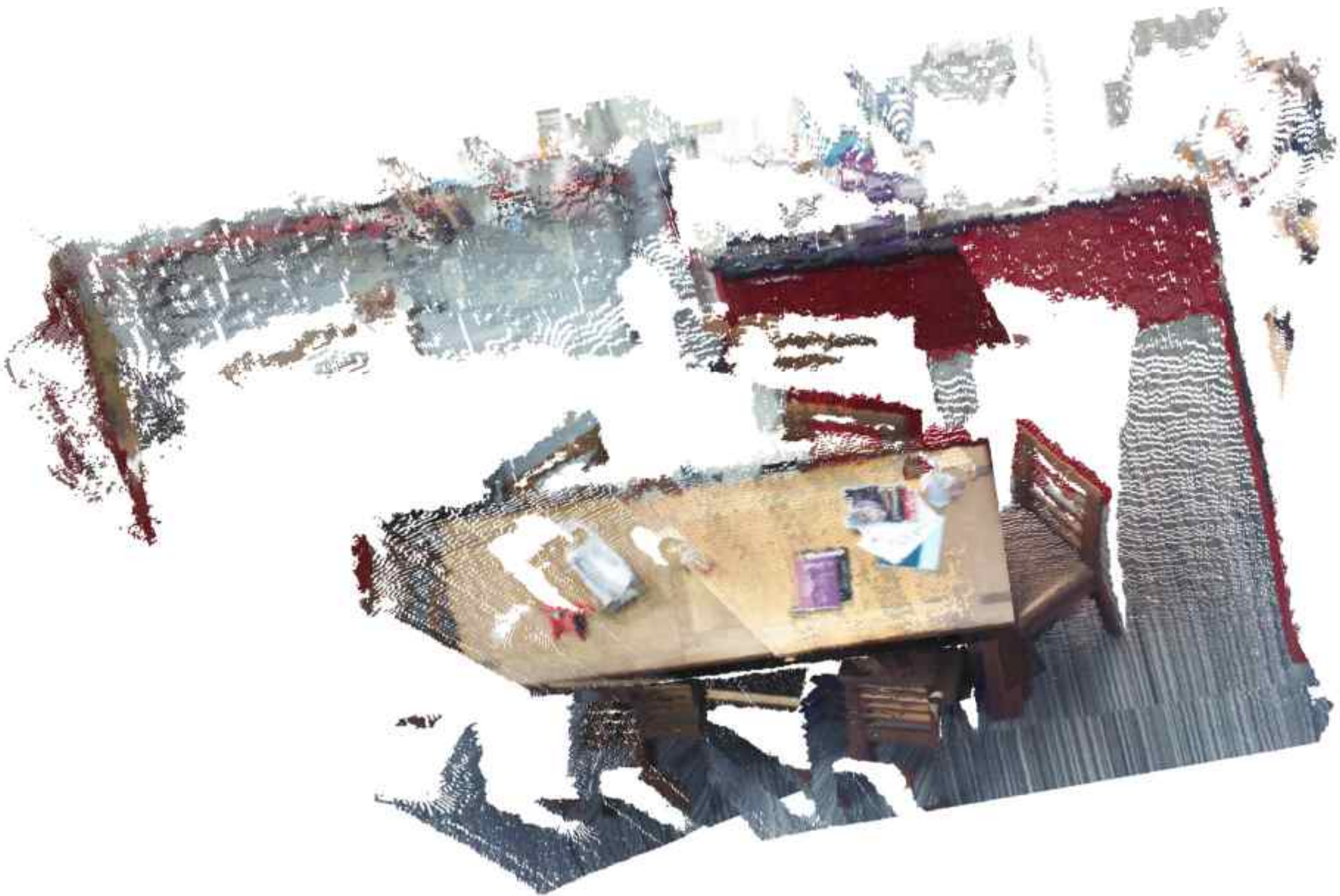}
\end{minipage}\,\,
&
\,\,
\begin{minipage}[t]{0.19\linewidth}
\centering
\includegraphics[width=.48\linewidth]{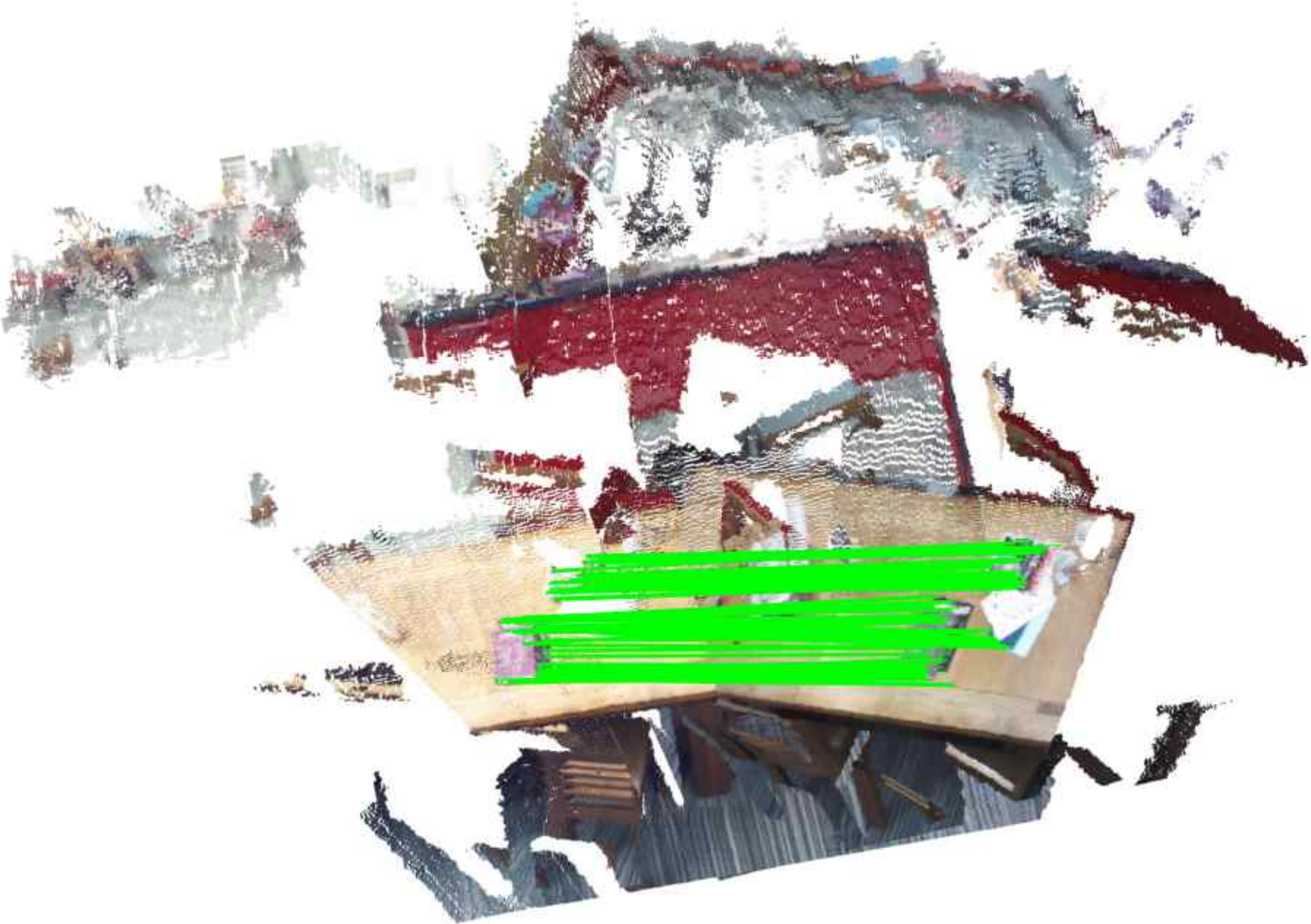}
\includegraphics[width=.48\linewidth]{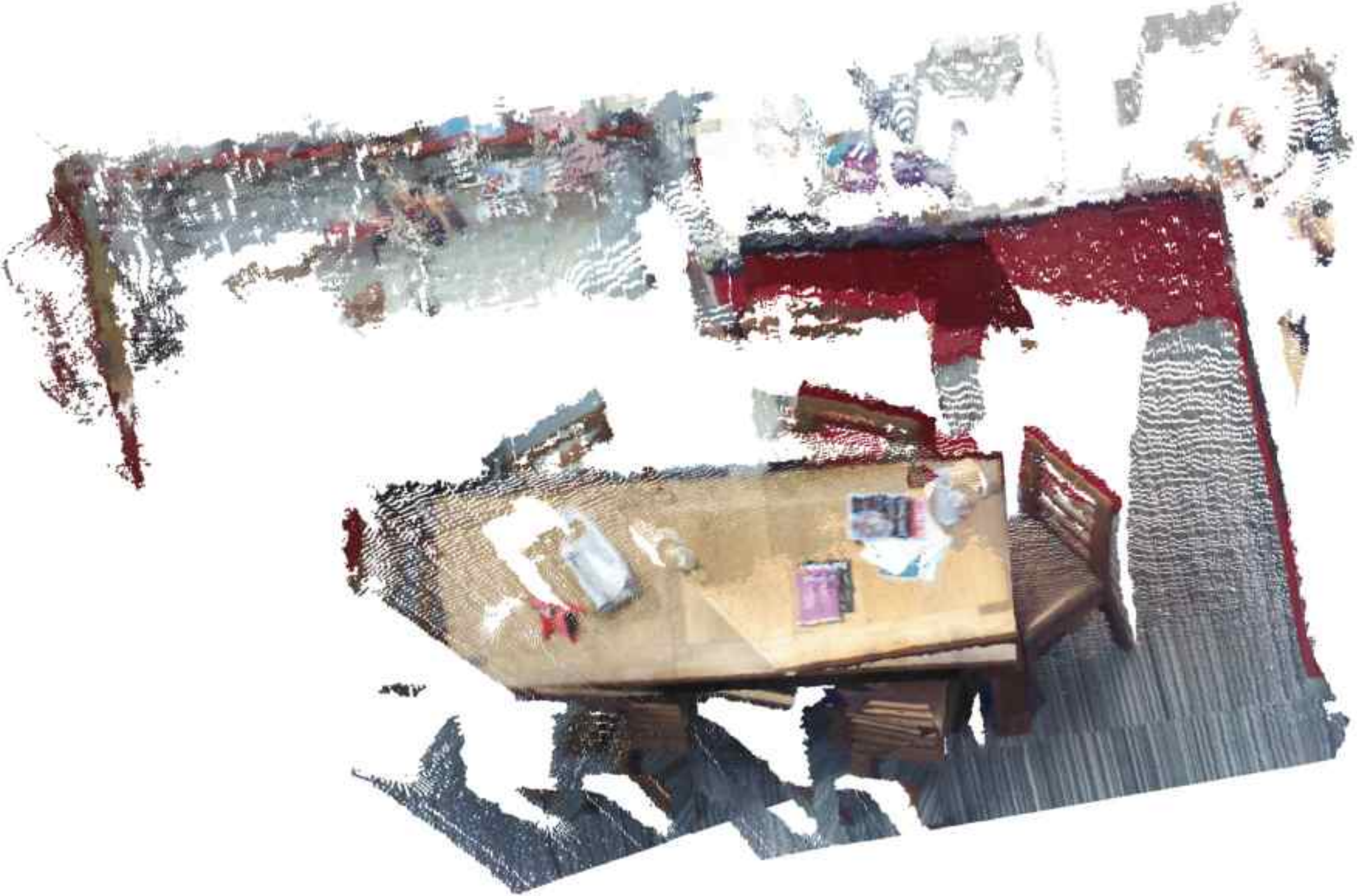}
\end{minipage}\,\,
&
\,\,
\begin{minipage}[t]{0.19\linewidth}
\centering
\includegraphics[width=.48\linewidth]{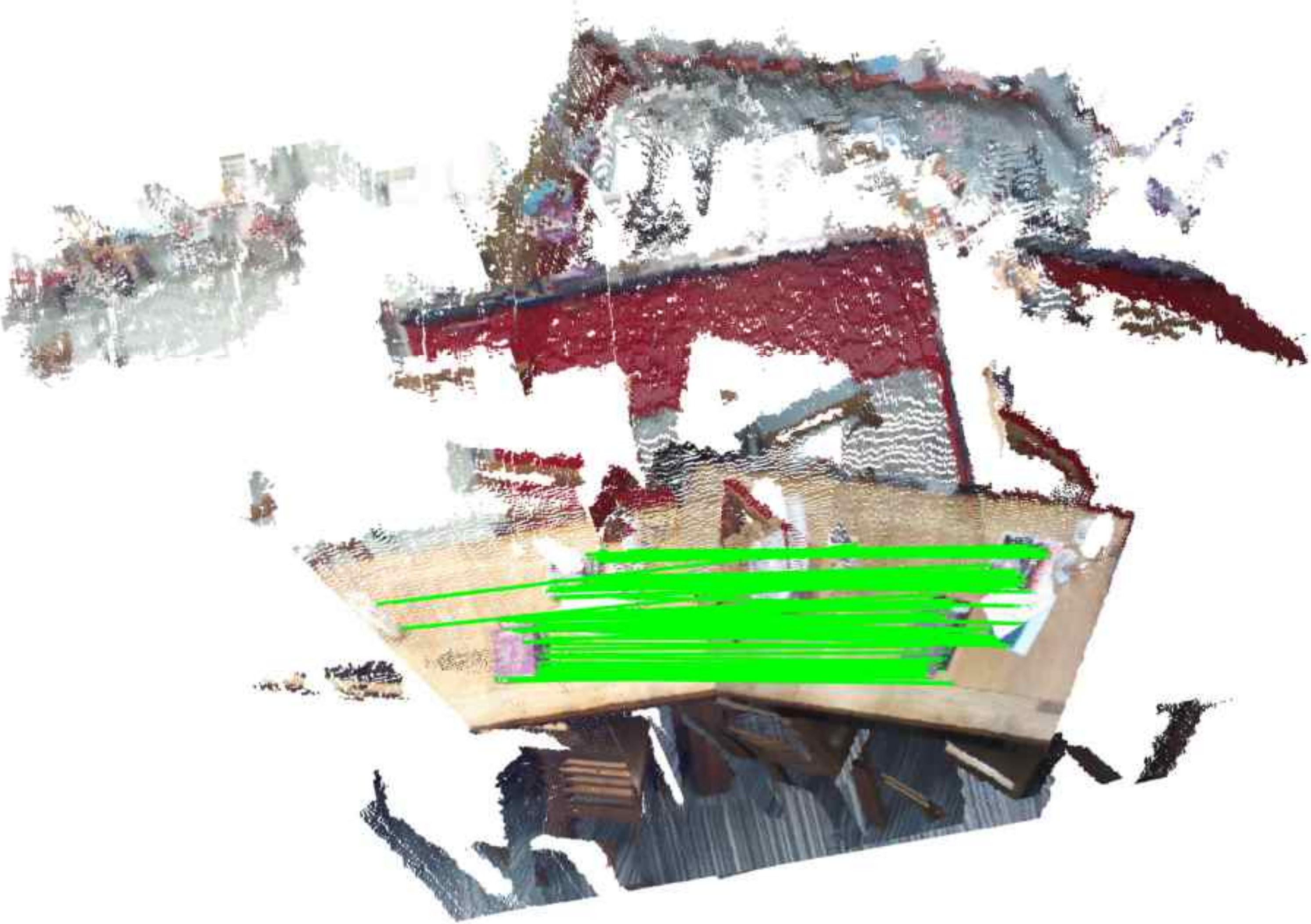}
\includegraphics[width=.48\linewidth]{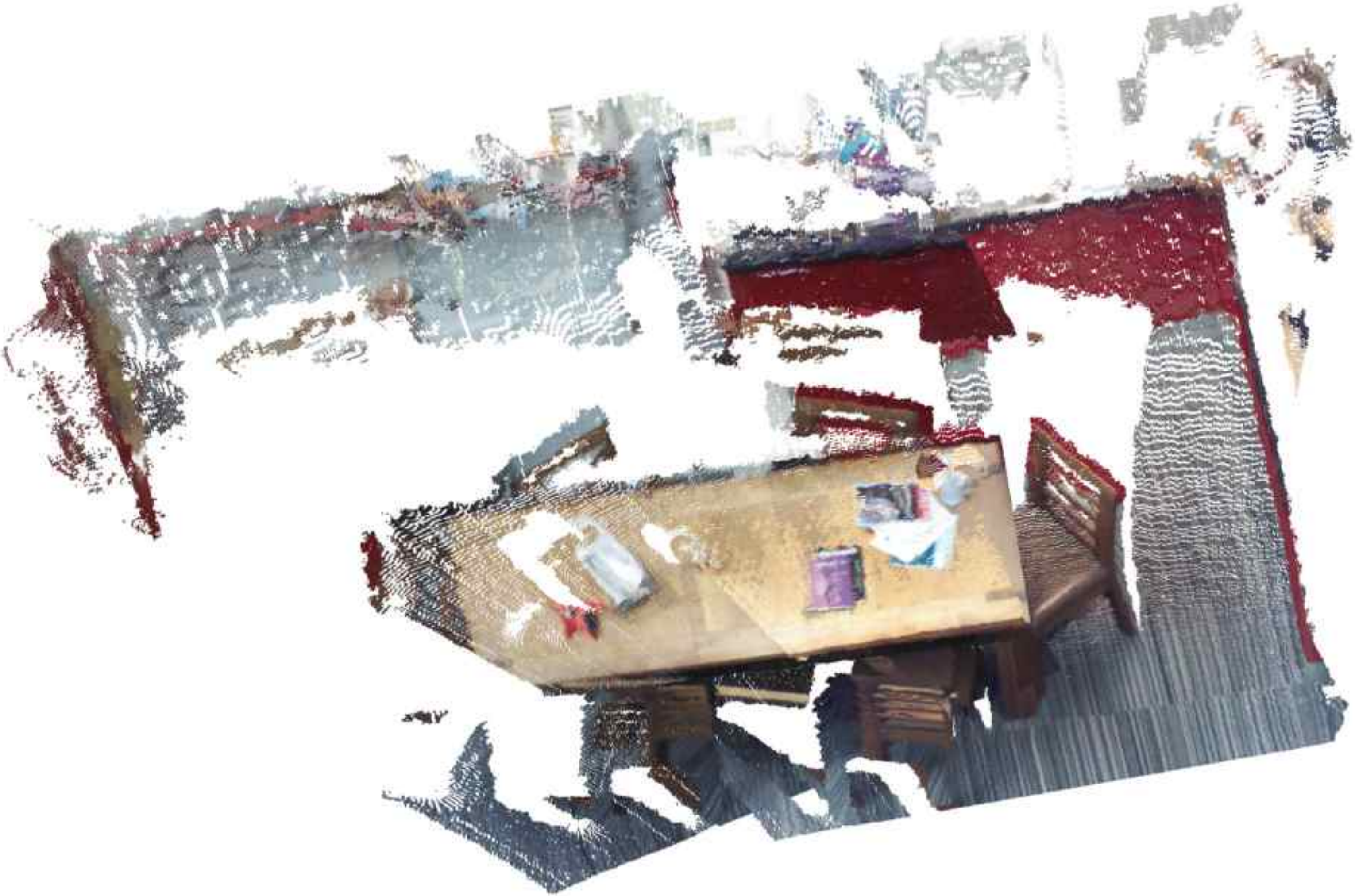}
\end{minipage}\,\,

\\

 & \scriptsize{$N$=900} && \scriptsize{\textcolor[rgb]{1,0,0}{Failed},\,\verb|\|,\,0.09$s$} &\scriptsize{\textcolor[rgb]{0,0.7,0}{Successful},\,0.49,\,8.43$s$}&\scriptsize{\textcolor[rgb]{0,0.7,0}{Successful},\,{0.33},\,4.31$s$}&\scriptsize{\textcolor[rgb]{0,0.7,0}{Successful},\,\textbf{0.33},\,\textbf{1.59}$s$}

\\

\rotatebox{90}{\,\,\footnotesize{\textit{red kitchen}}\,}\,
&
\,\,
\begin{minipage}[t]{0.1\linewidth}
\centering
\includegraphics[width=1\linewidth]{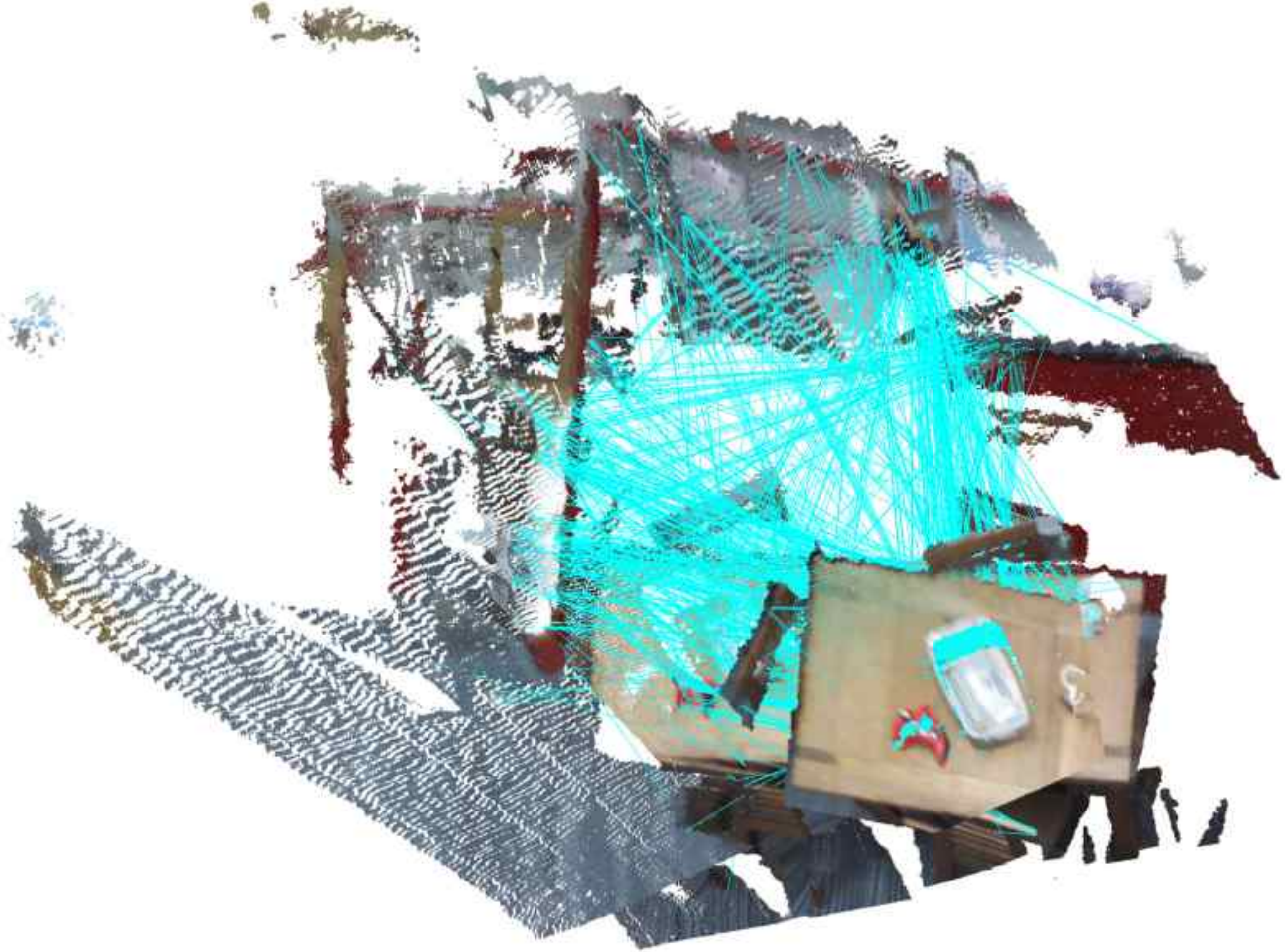}
\end{minipage}\,\,
& &
\,\,
\begin{minipage}[t]{0.19\linewidth}
\centering
\includegraphics[width=.48\linewidth]{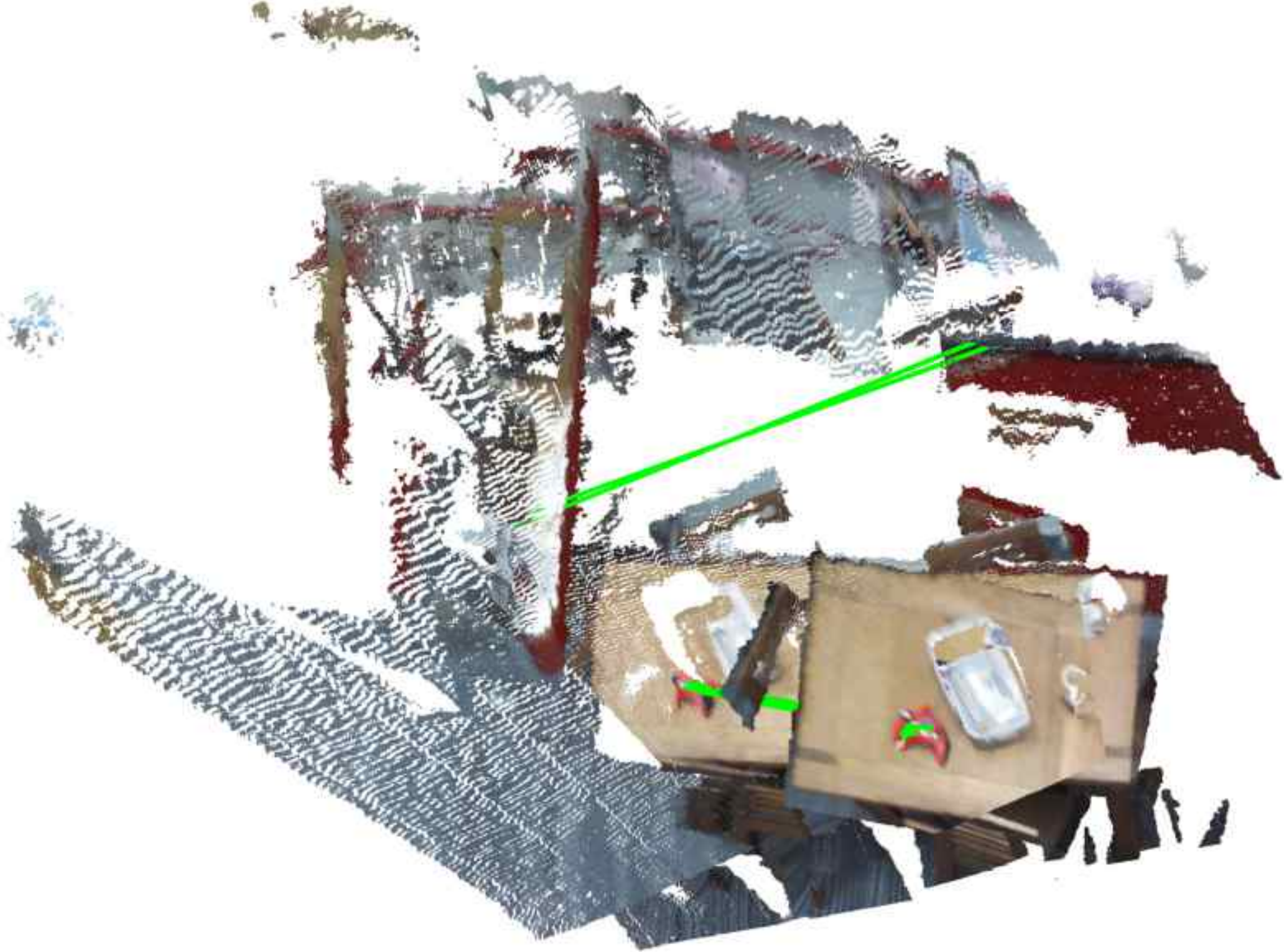}
\includegraphics[width=.48\linewidth]{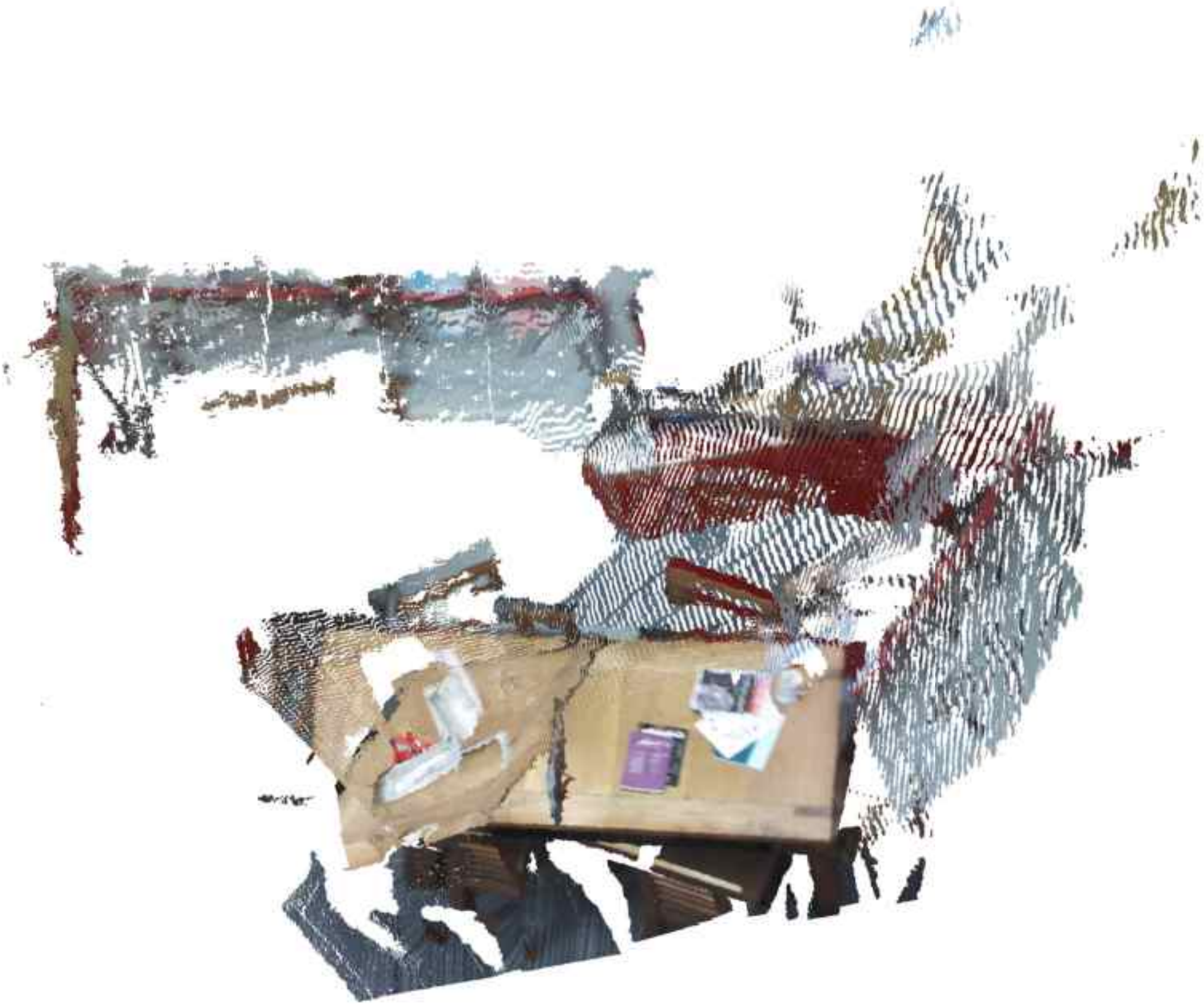}
\end{minipage}\,\,
&
\,\,
\begin{minipage}[t]{0.19\linewidth}
\centering
\includegraphics[width=.48\linewidth]{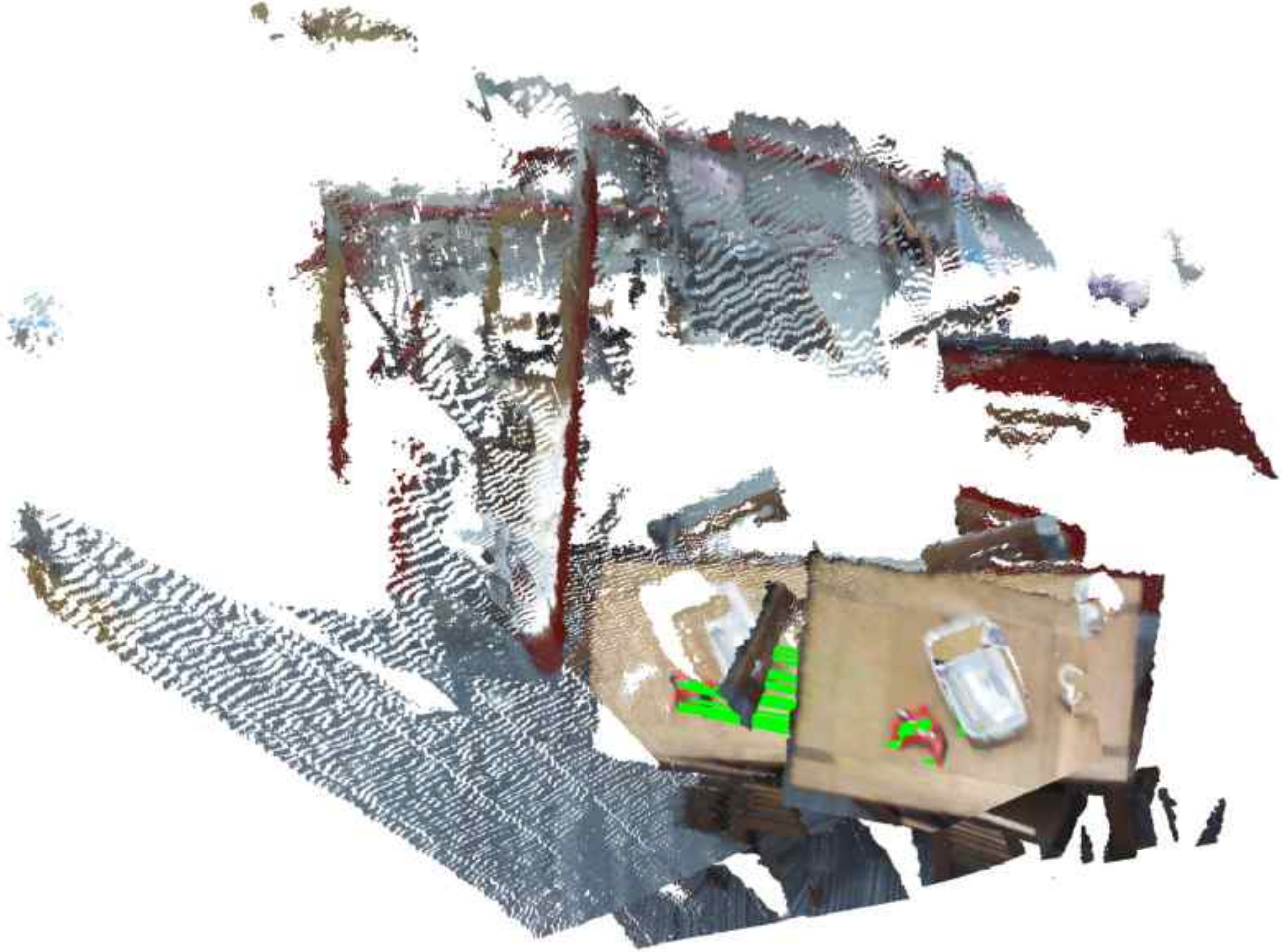}
\includegraphics[width=.48\linewidth]{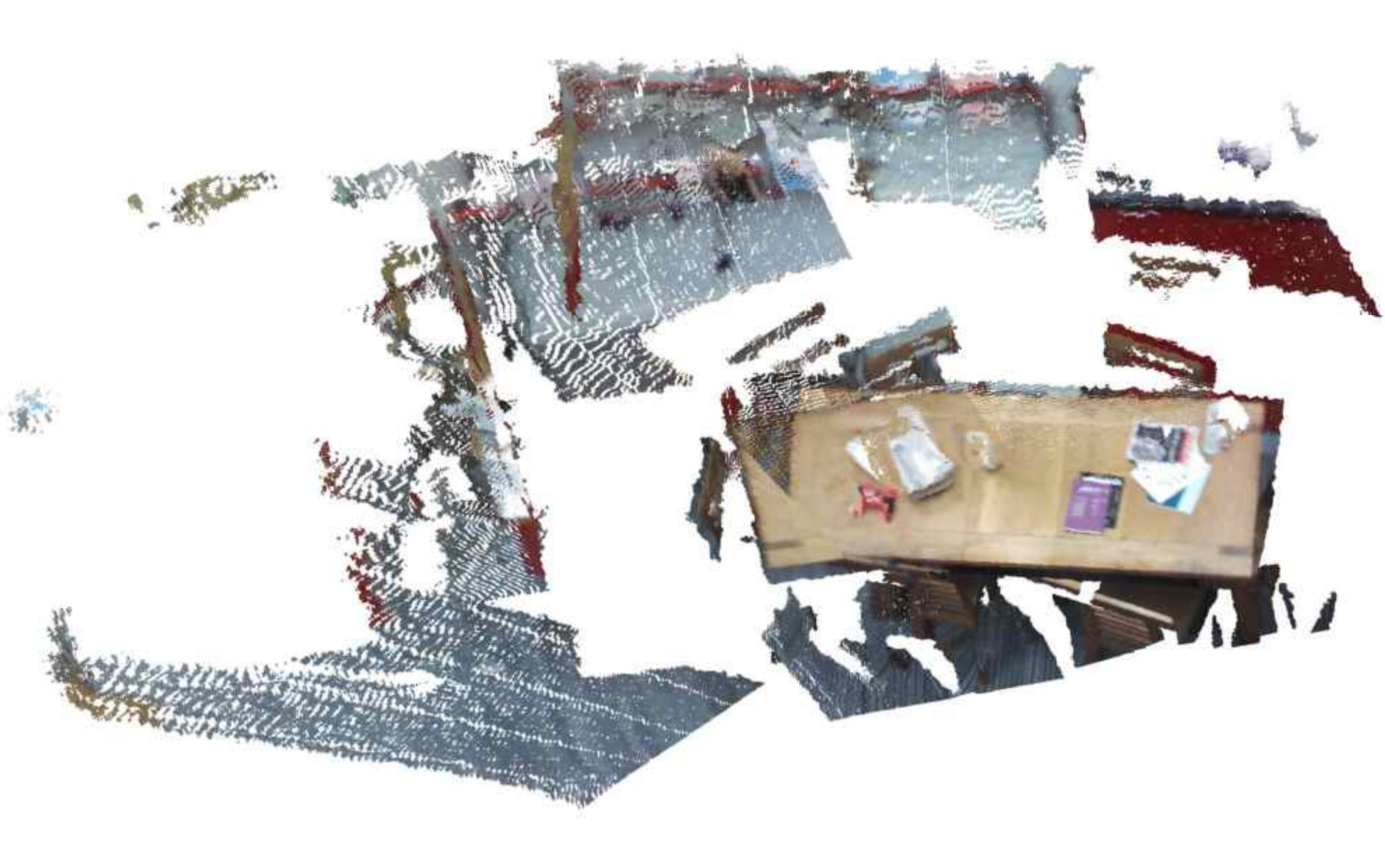}
\end{minipage}\,\,
&
\,\,
\begin{minipage}[t]{0.19\linewidth}
\centering
\includegraphics[width=.48\linewidth]{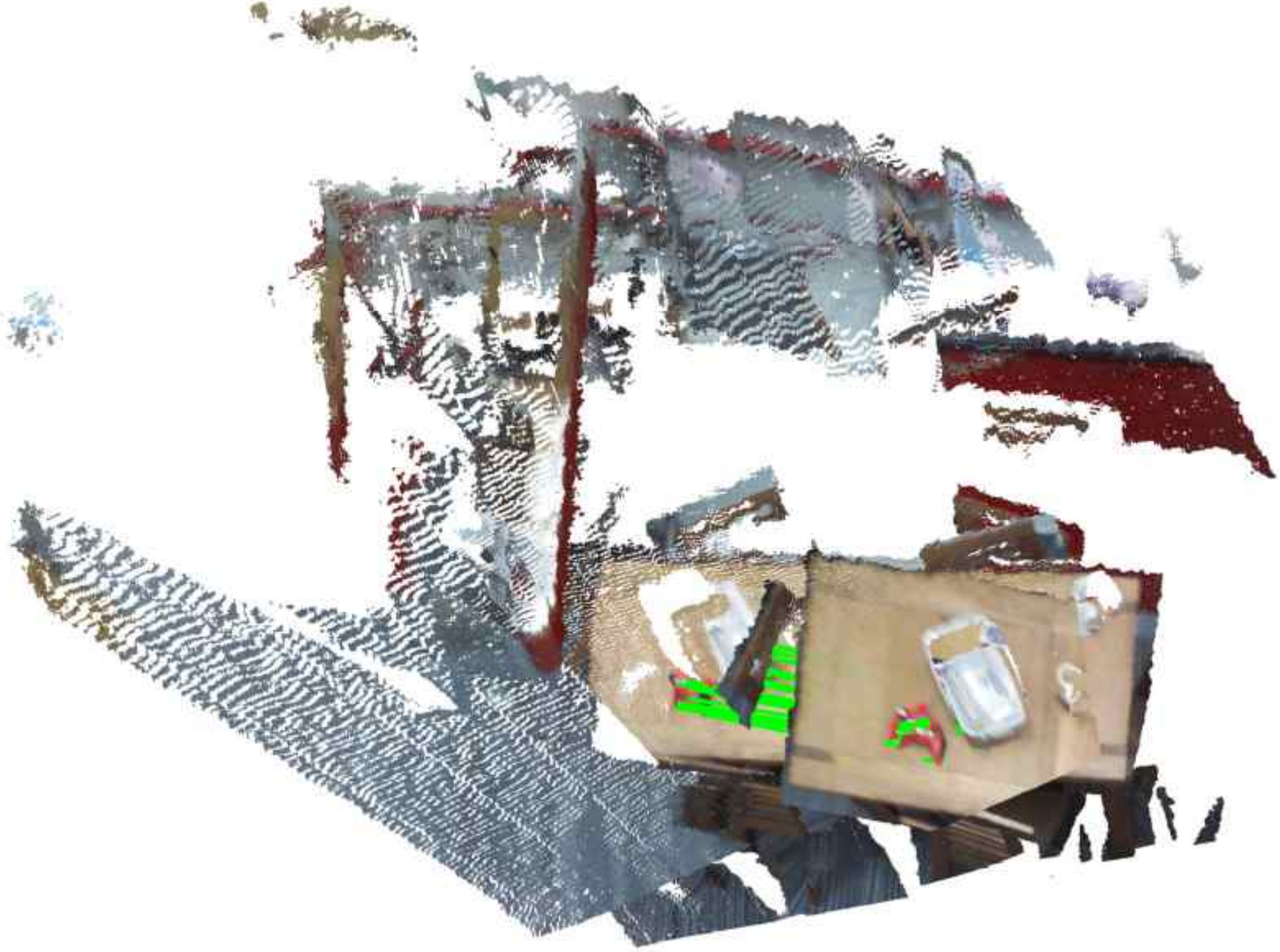}
\includegraphics[width=.48\linewidth]{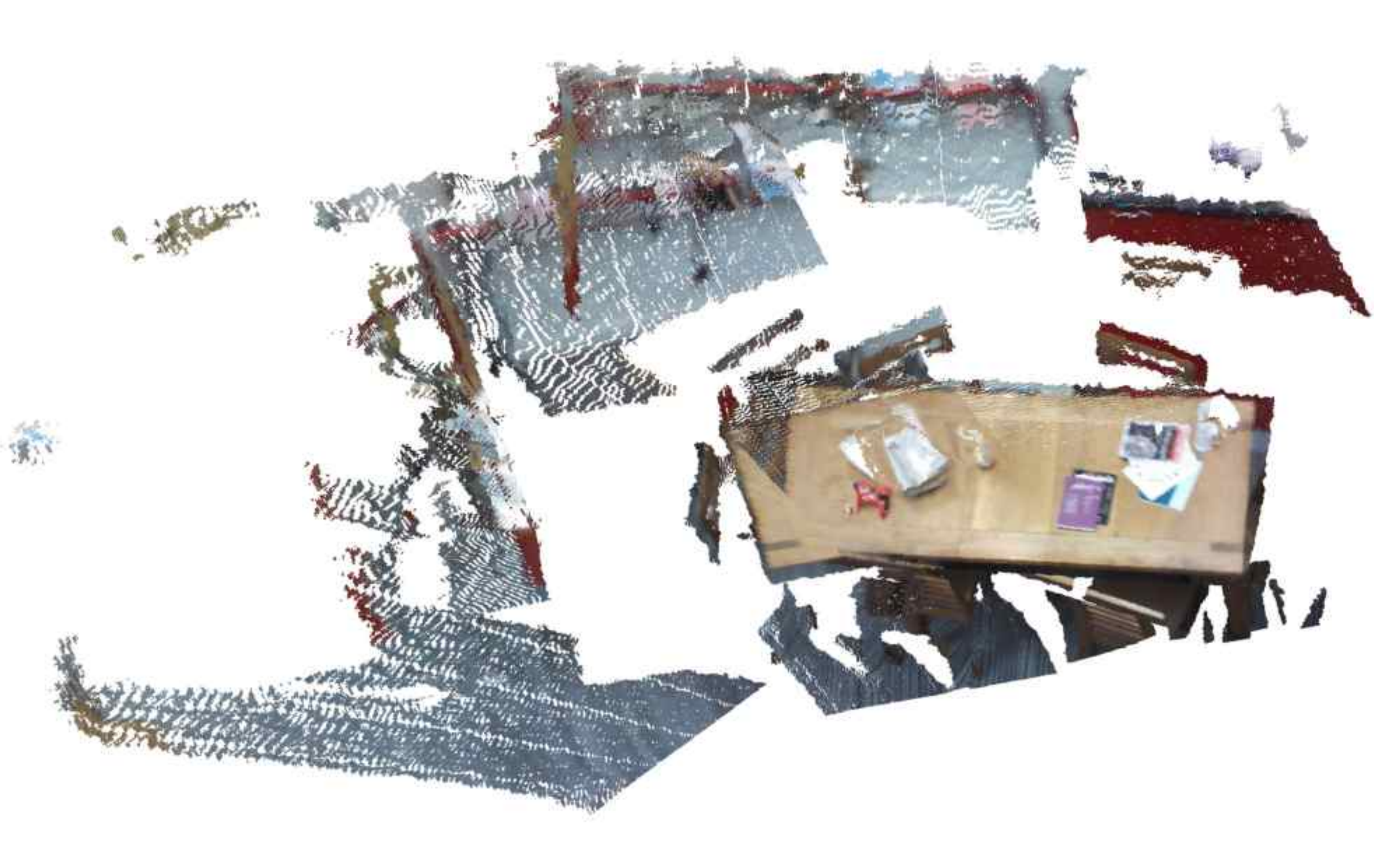}
\end{minipage}\,\,
&
\,\,
\begin{minipage}[t]{0.19\linewidth}
\centering
\includegraphics[width=.48\linewidth]{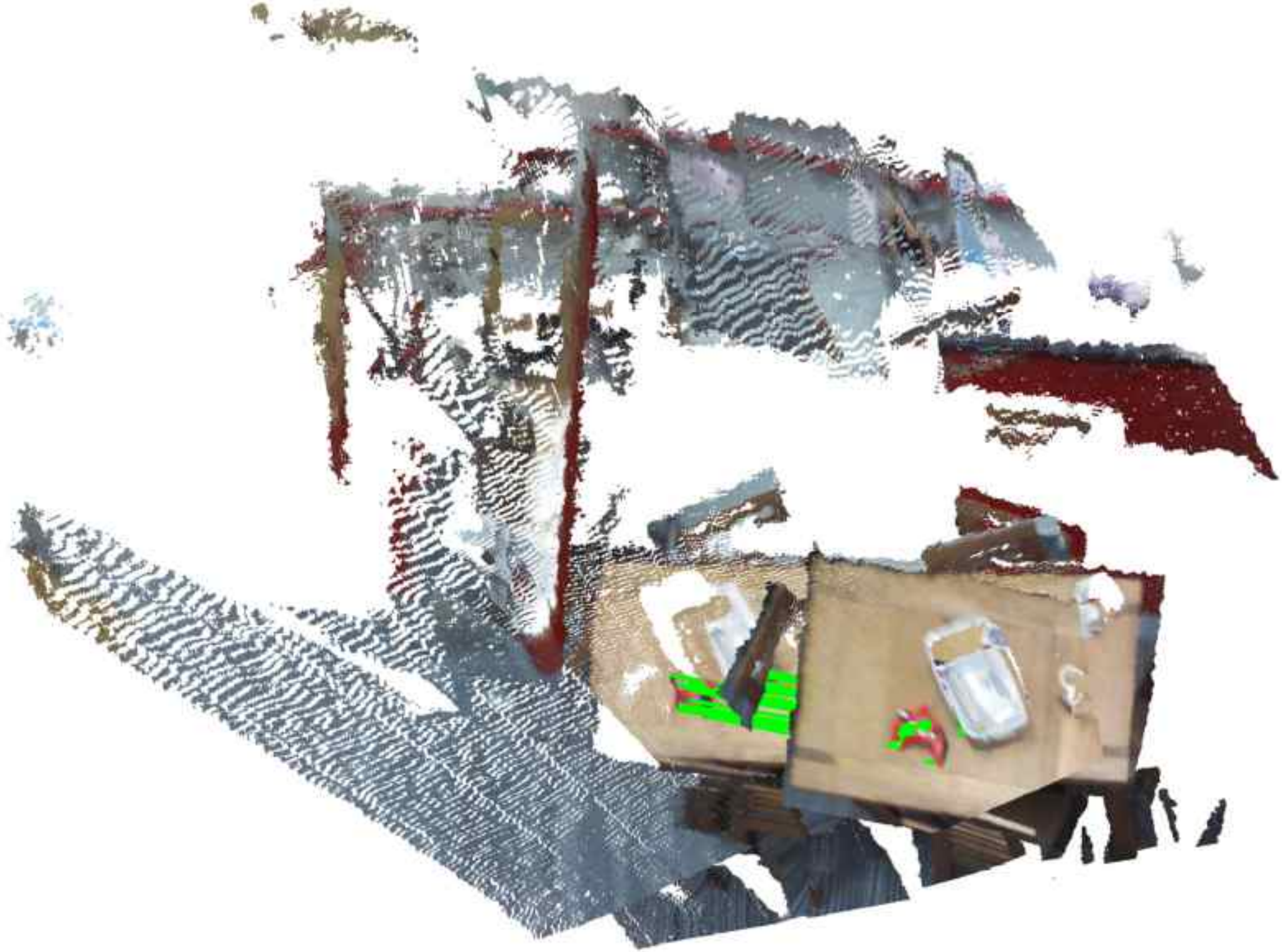}
\includegraphics[width=.48\linewidth]{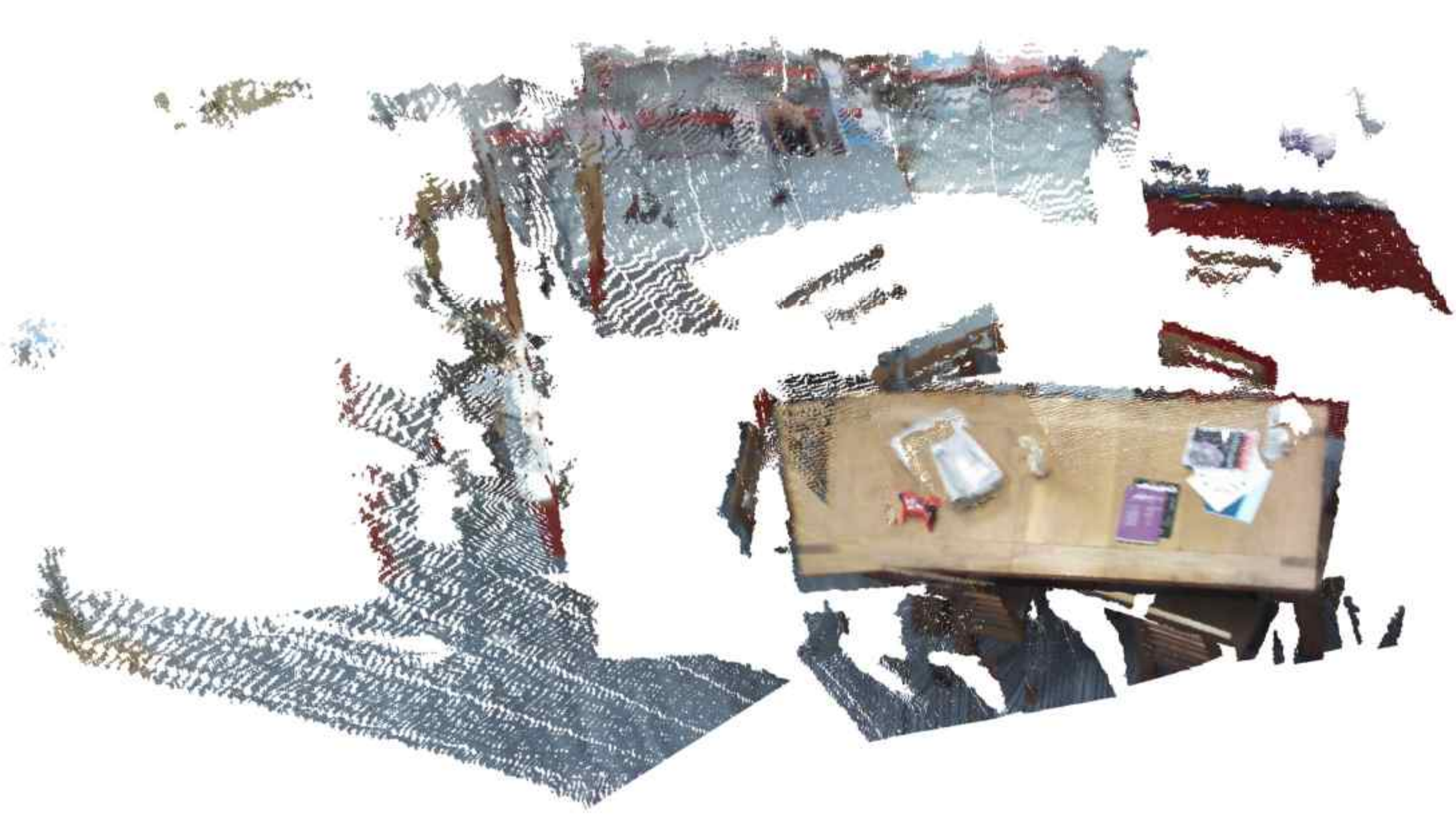}
\end{minipage}\,\,

\\

 & \scriptsize{$N$=917} && \scriptsize{\textcolor[rgb]{1,0,0}{Failed},\,\verb|\|,\,0.11$s$} &\scriptsize{\textcolor[rgb]{1,0,0}{Failed},\,1.73,\,25.48$s$}&\scriptsize{\textcolor[rgb]{0,0.7,0}{Successful},\,{0.30},\,64.55$s$}&\scriptsize{\textcolor[rgb]{0,0.7,0}{Successful},\,\textbf{0.29},\,\textbf{1.92}$s$}

\\

\rotatebox{90}{\,\,\footnotesize{\textit{red kitchen}}\,}\,
&
\,\,
\begin{minipage}[t]{0.1\linewidth}
\centering
\includegraphics[width=1\linewidth]{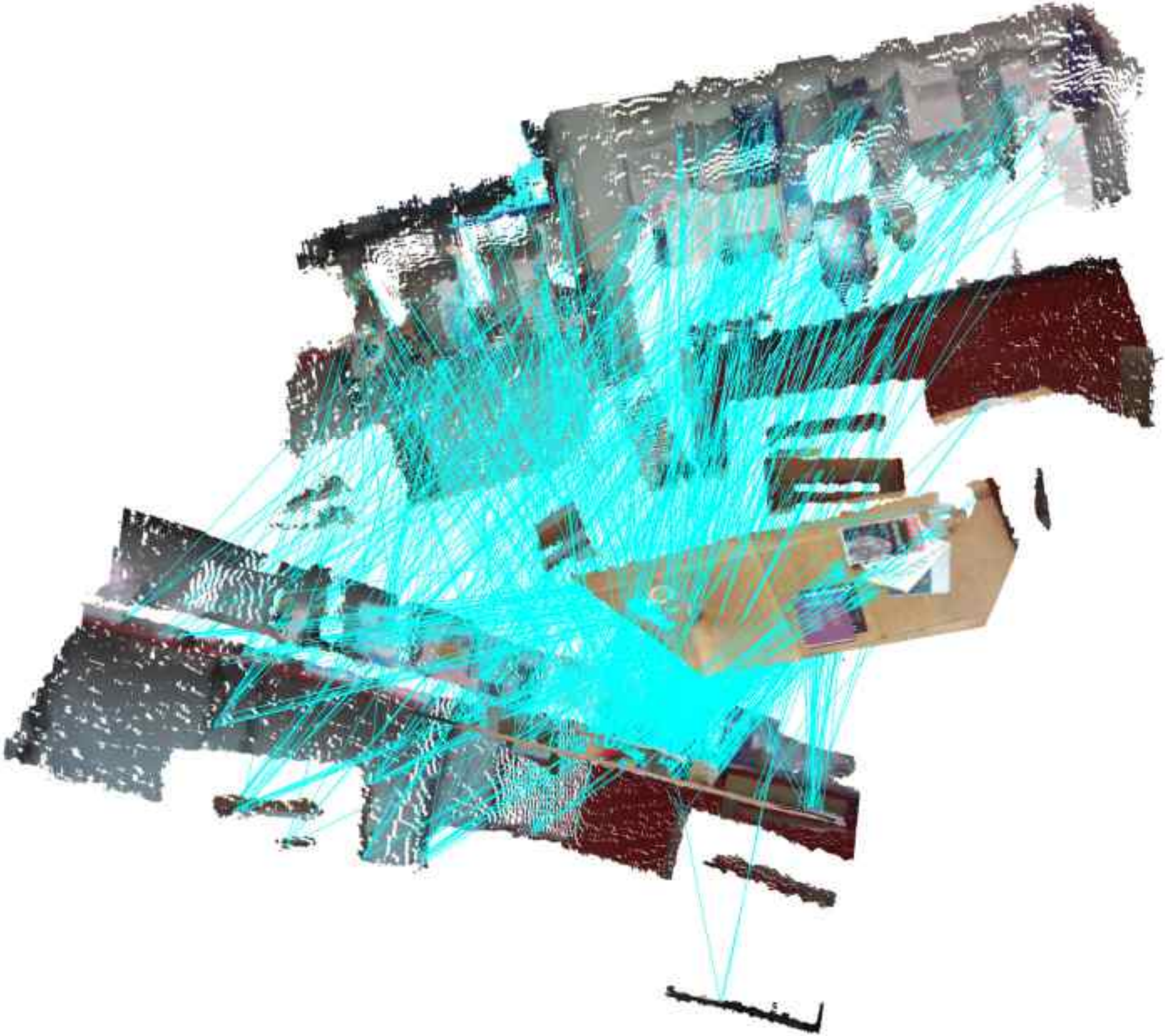}
\end{minipage}\,\,
& &
\,\,
\begin{minipage}[t]{0.19\linewidth}
\centering
\includegraphics[width=.48\linewidth]{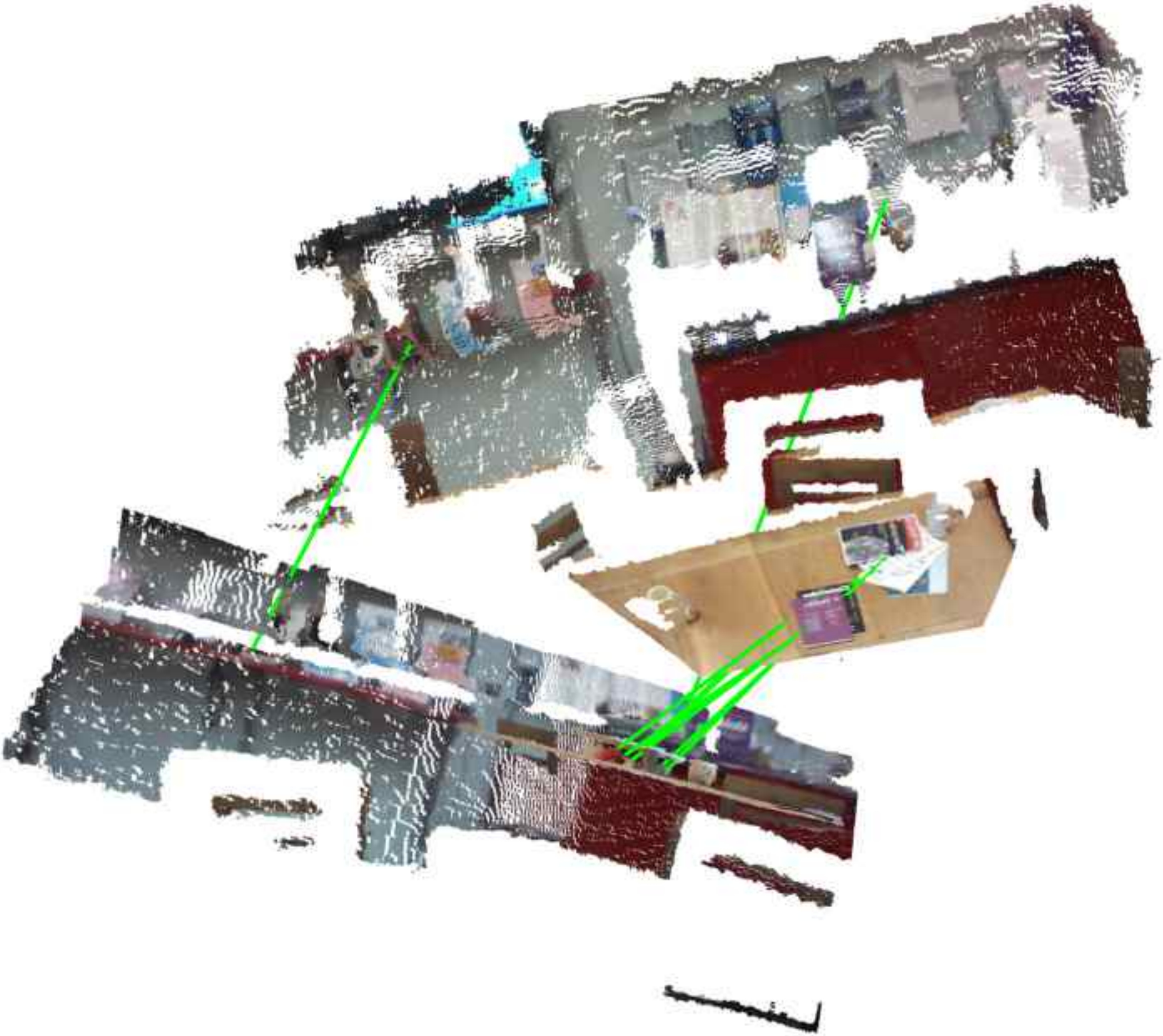}
\includegraphics[width=.48\linewidth]{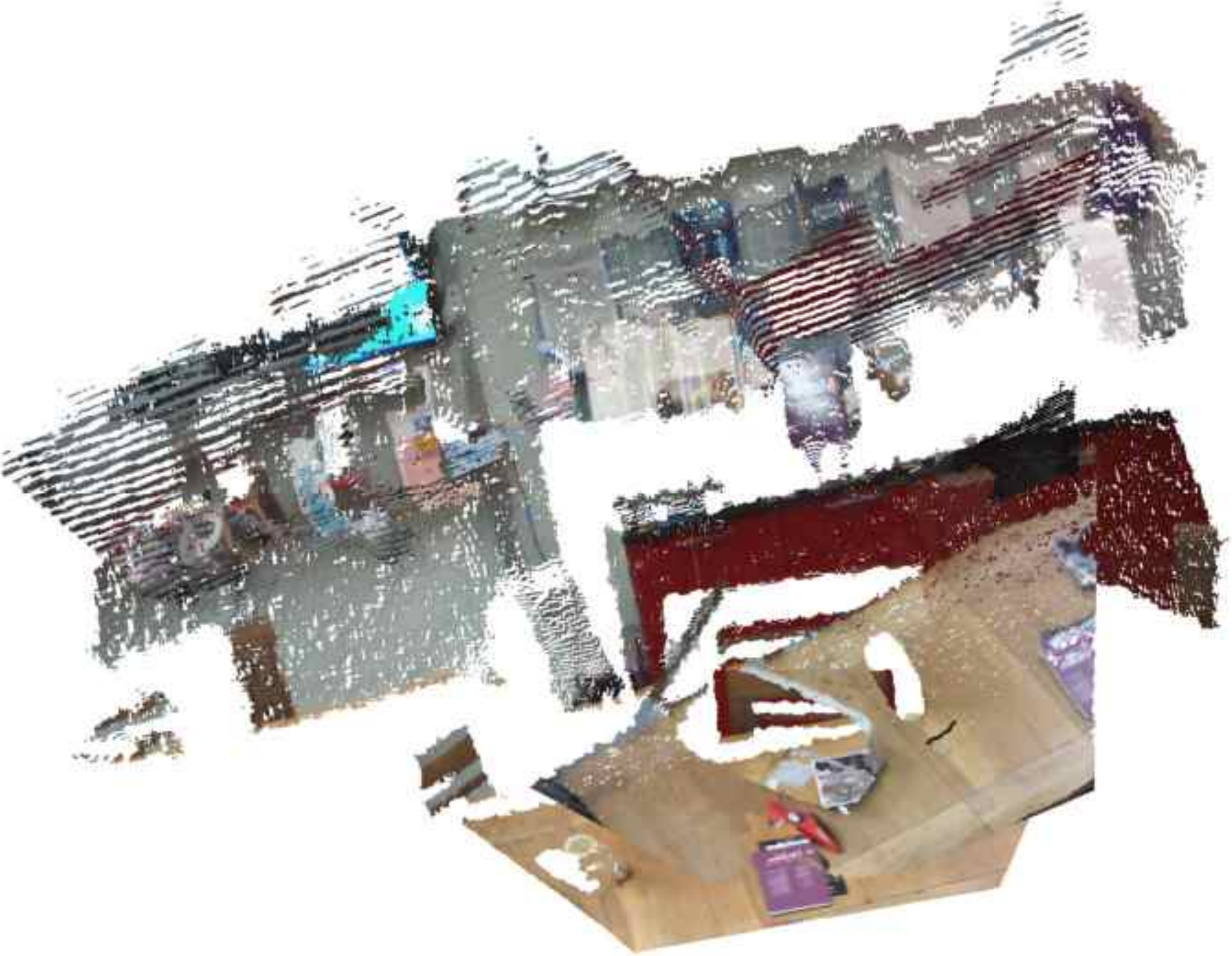}
\end{minipage}\,\,
&
\,\,
\begin{minipage}[t]{0.19\linewidth}
\centering
\includegraphics[width=.48\linewidth]{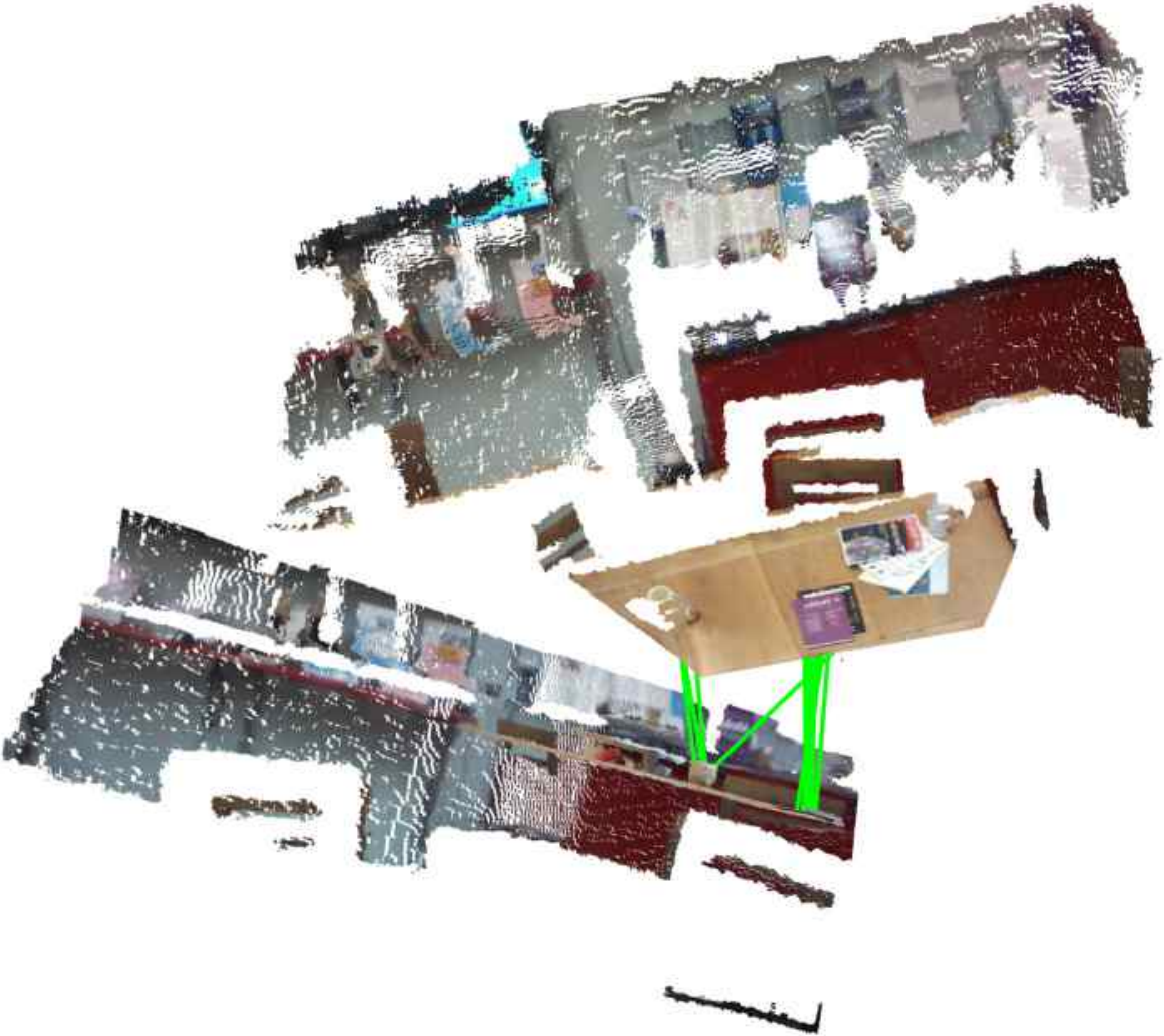}
\includegraphics[width=.48\linewidth]{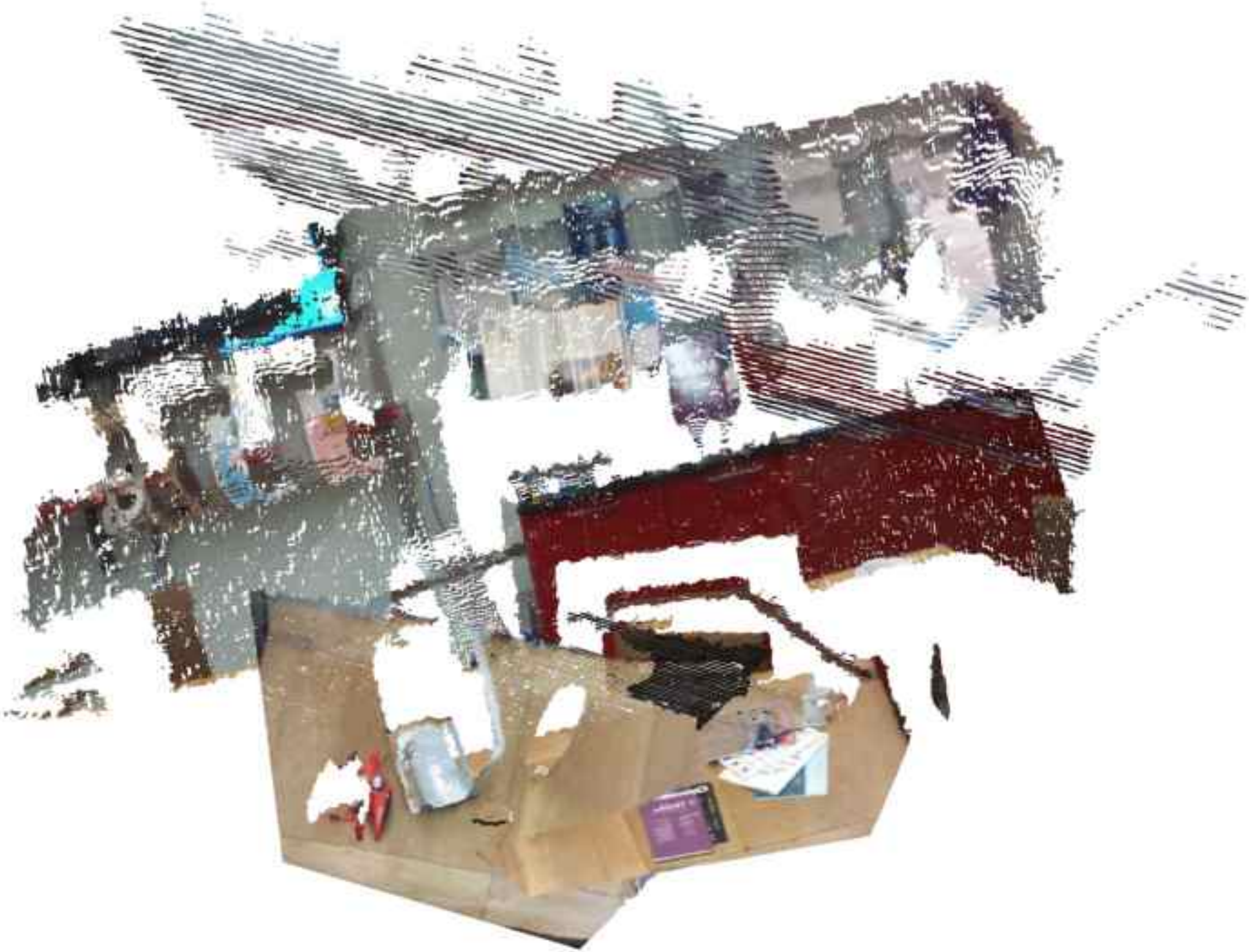}
\end{minipage}\,\,
&
\,\,
\begin{minipage}[t]{0.19\linewidth}
\centering
\includegraphics[width=.48\linewidth]{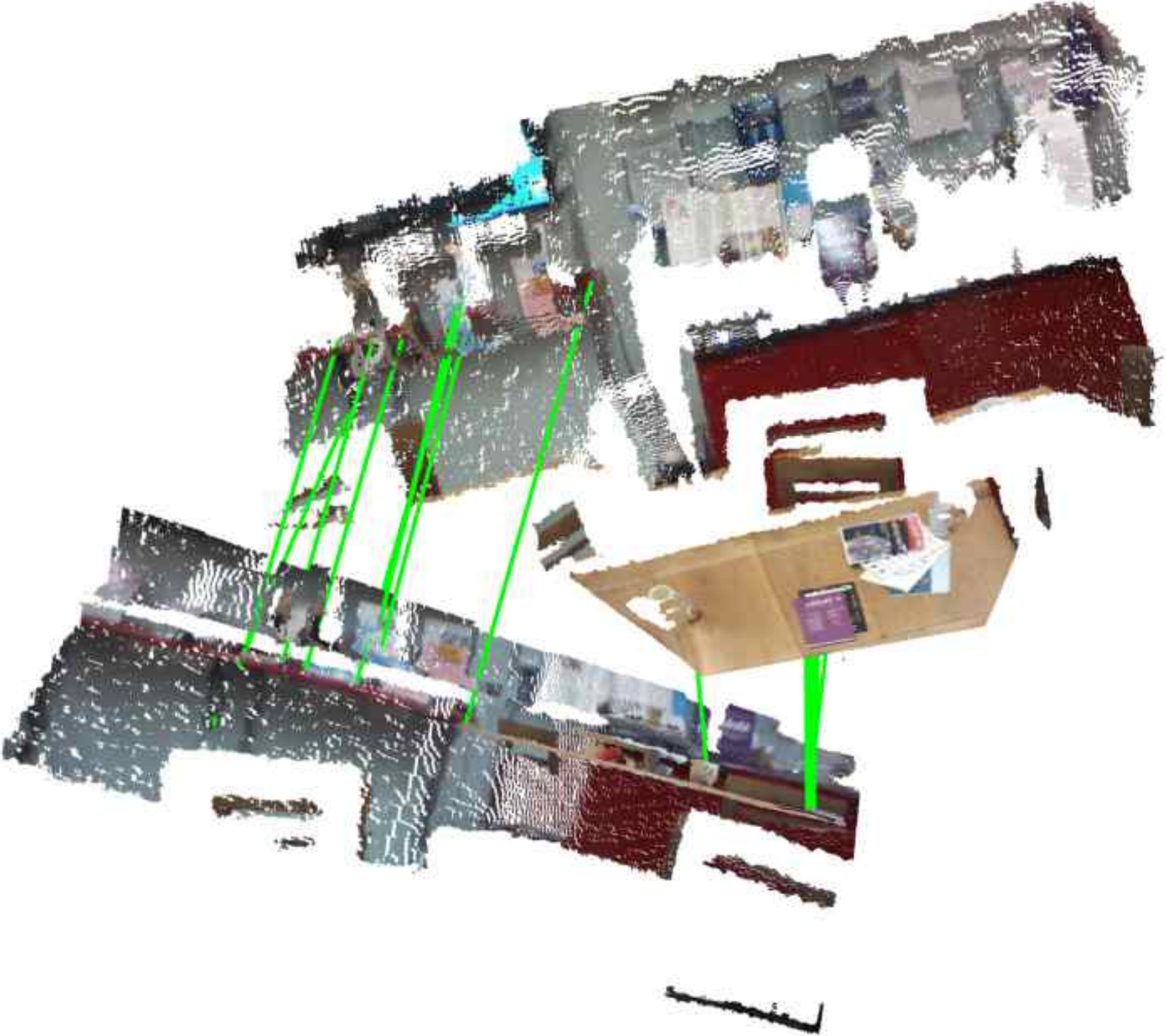}
\includegraphics[width=.48\linewidth]{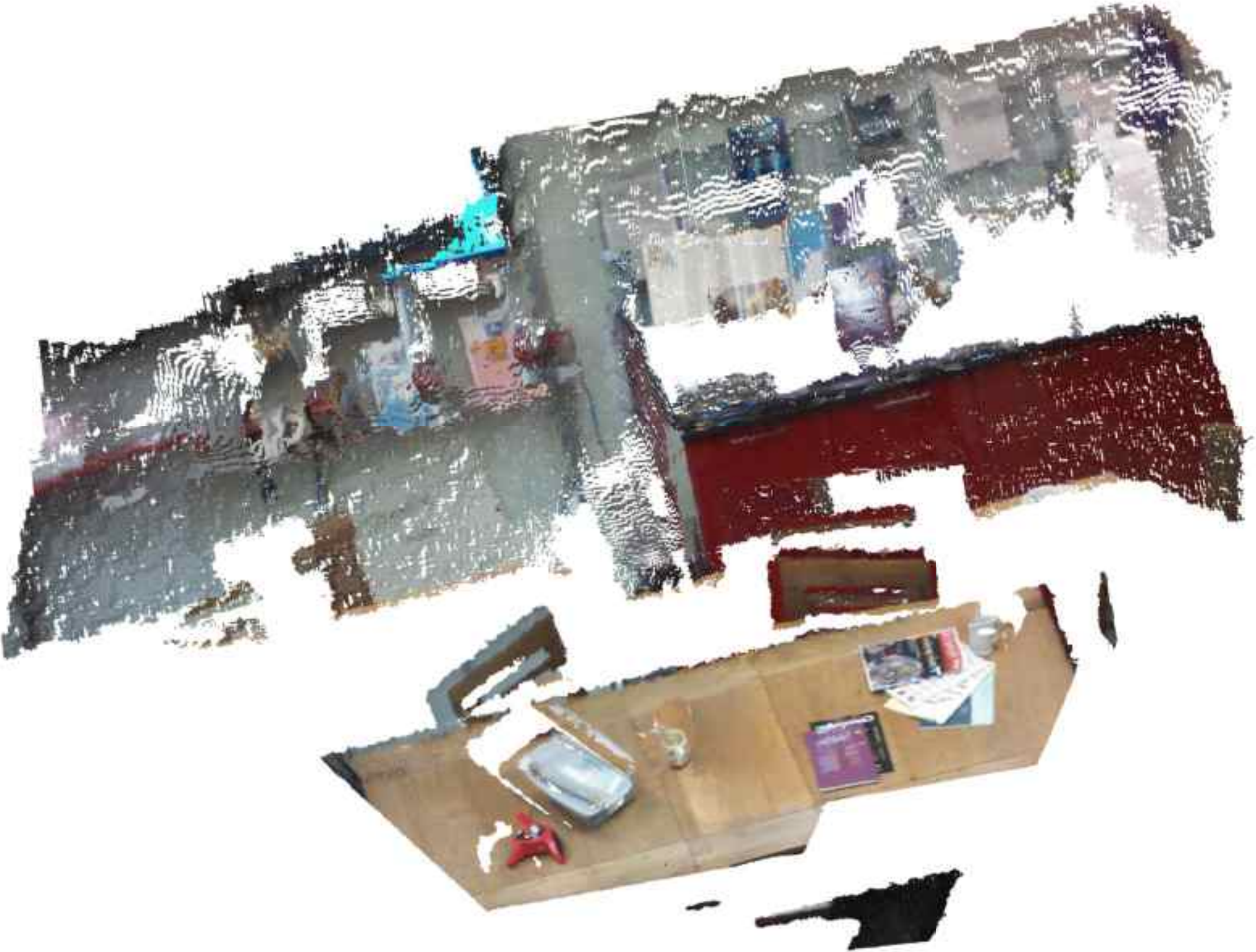}
\end{minipage}\,\,
&
\,\,
\begin{minipage}[t]{0.19\linewidth}
\centering
\includegraphics[width=.48\linewidth]{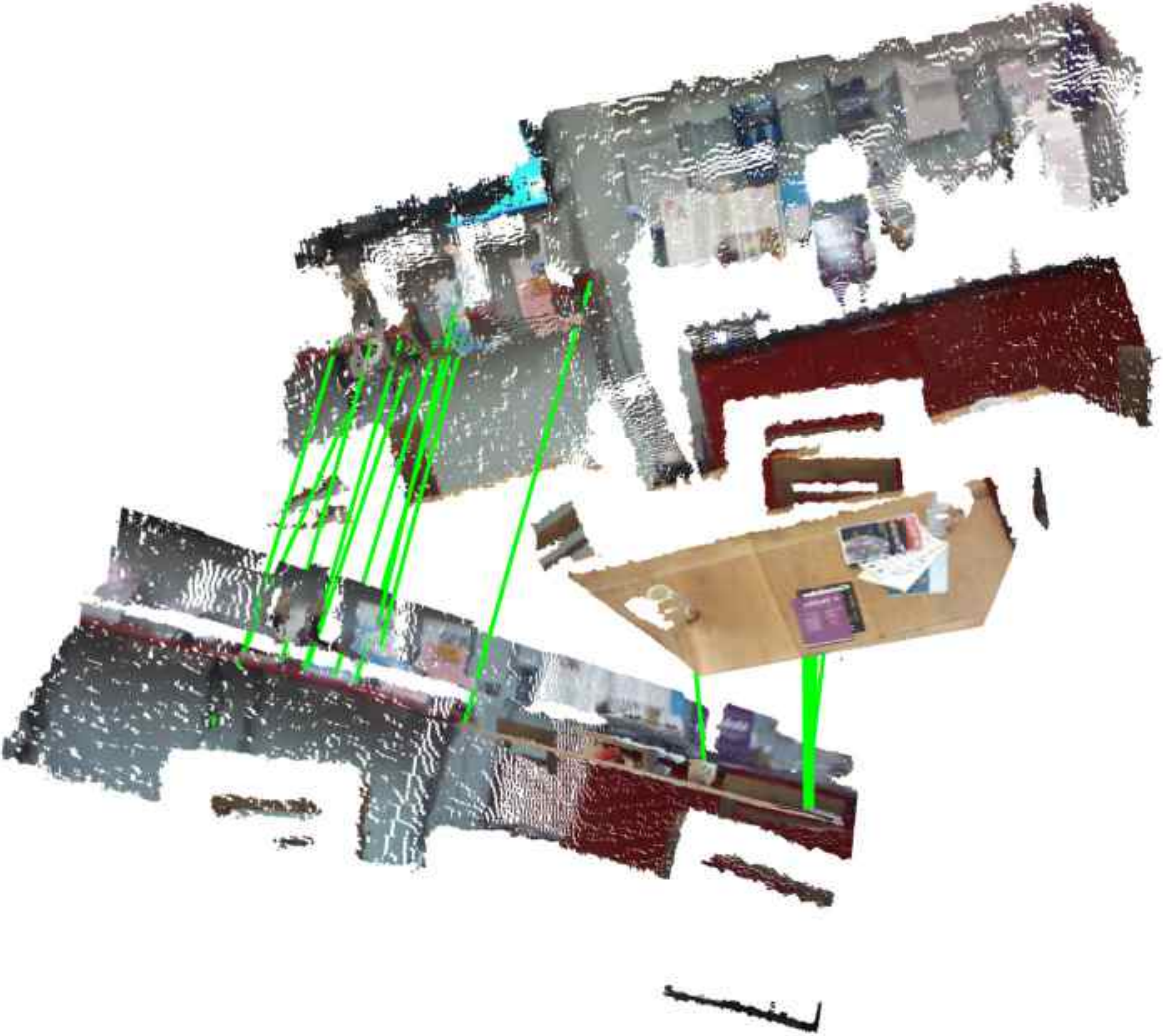}
\includegraphics[width=.48\linewidth]{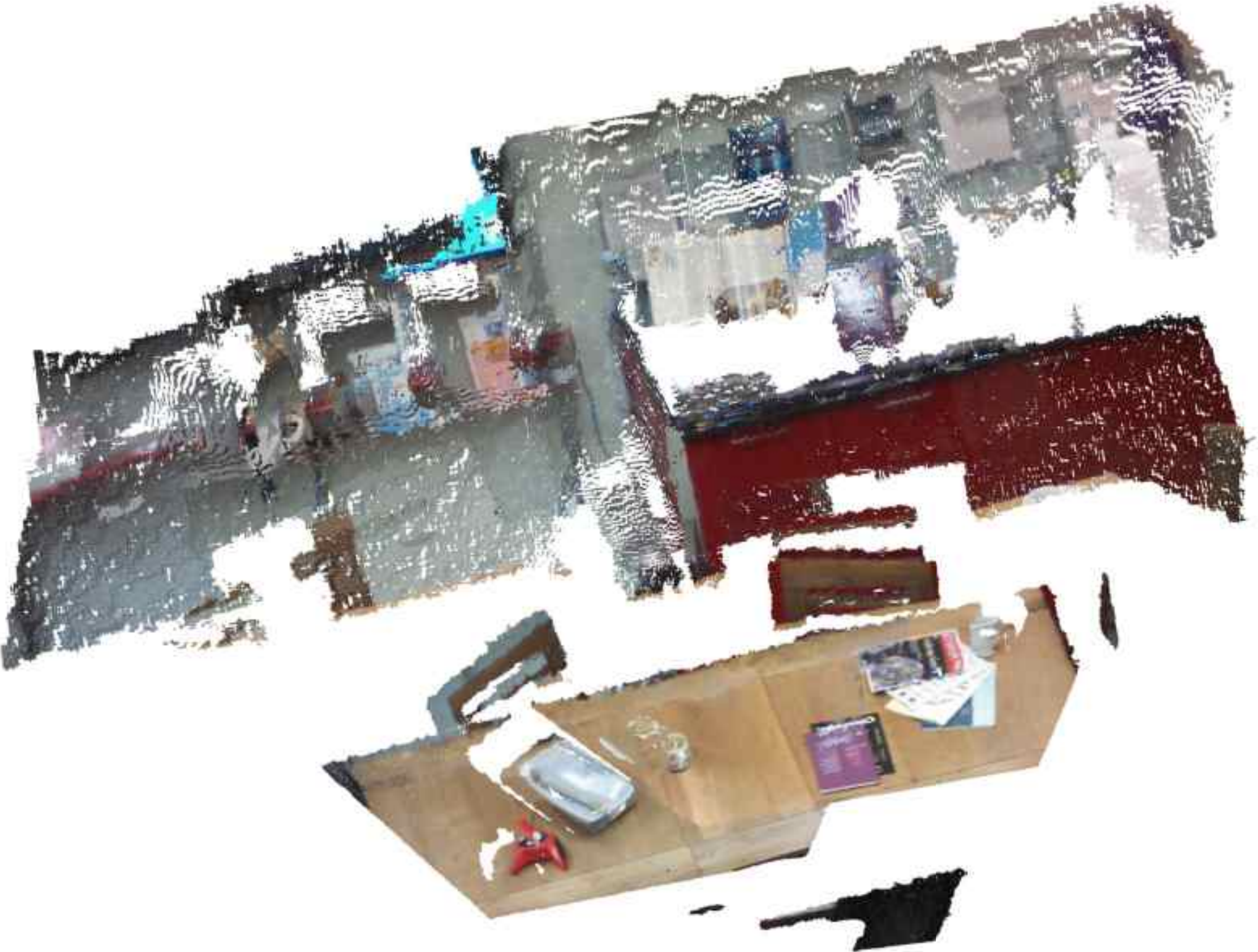}
\end{minipage}\,\,

\\

 & \scriptsize{$N$=917} && \scriptsize{\textcolor[rgb]{1,0,0}{Failed},\,\verb|\|,\,0.10$s$} &\scriptsize{\textcolor[rgb]{1,0,0}{Failed},\,0.67\,,23.25$s$}&\scriptsize{\textcolor[rgb]{0,0.7,0}{Successful},\,\textbf{0.29},\,2.87$s$}&\scriptsize{\textcolor[rgb]{0,0.7,0}{Successful},\,{0.29},\,\textbf{1.31}$s$}

\\

\rotatebox{90}{\,\,\footnotesize{\textit{red kitchen}}\,}\,
&
\,\,
\begin{minipage}[t]{0.1\linewidth}
\centering
\includegraphics[width=1\linewidth]{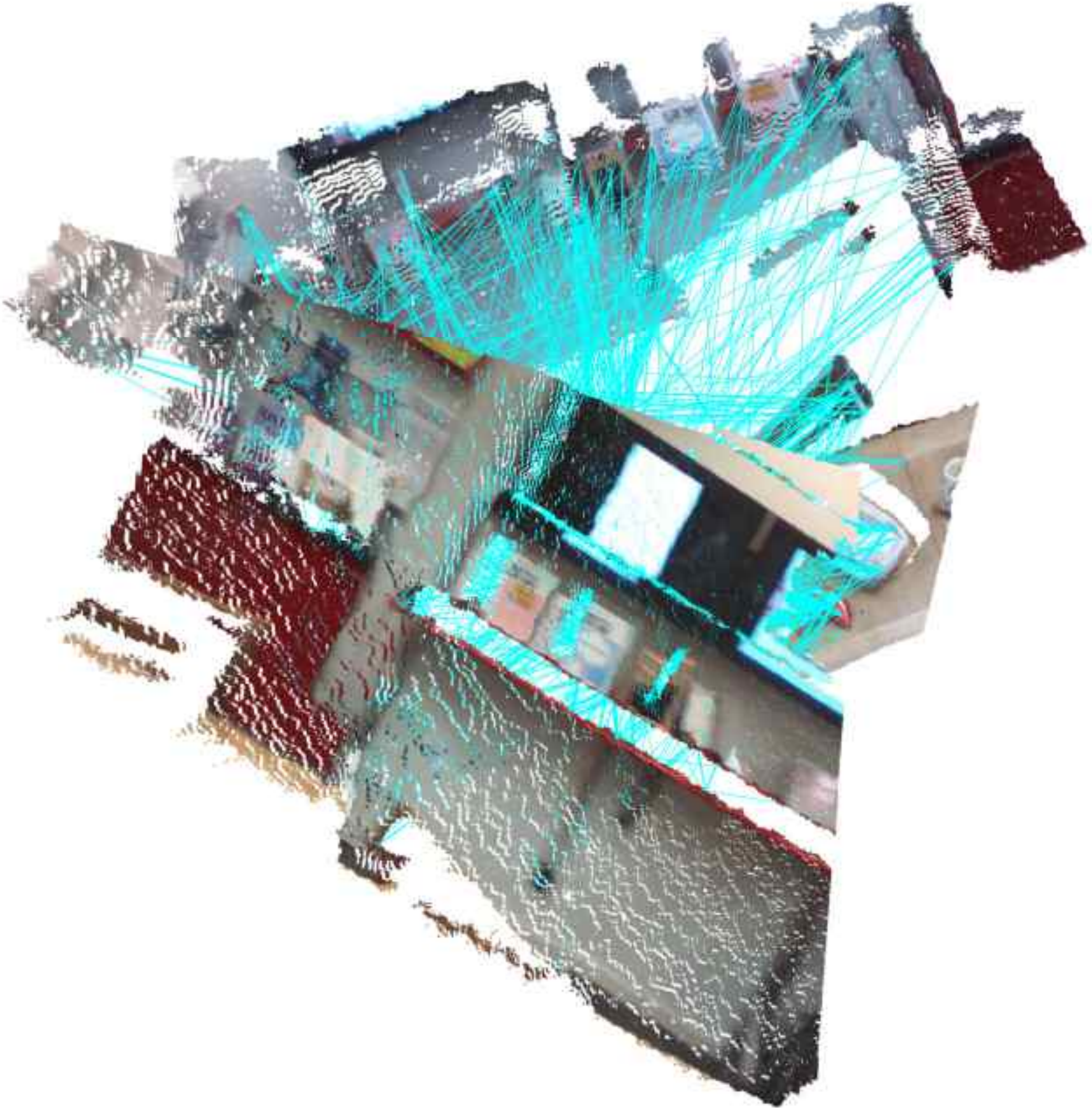}
\end{minipage}\,\,
& &
\,\,
\begin{minipage}[t]{0.19\linewidth}
\centering
\includegraphics[width=.48\linewidth]{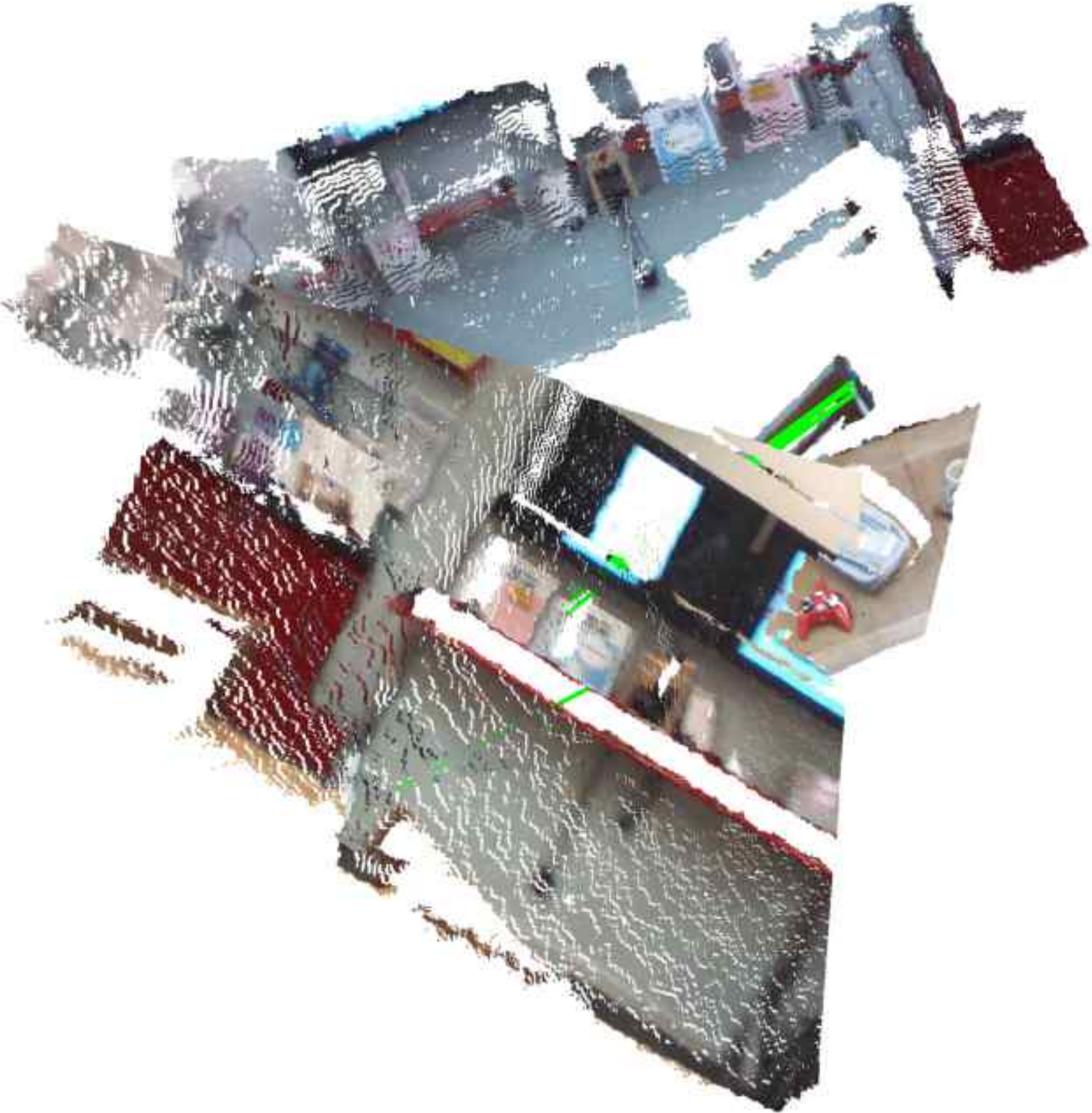}
\includegraphics[width=.48\linewidth]{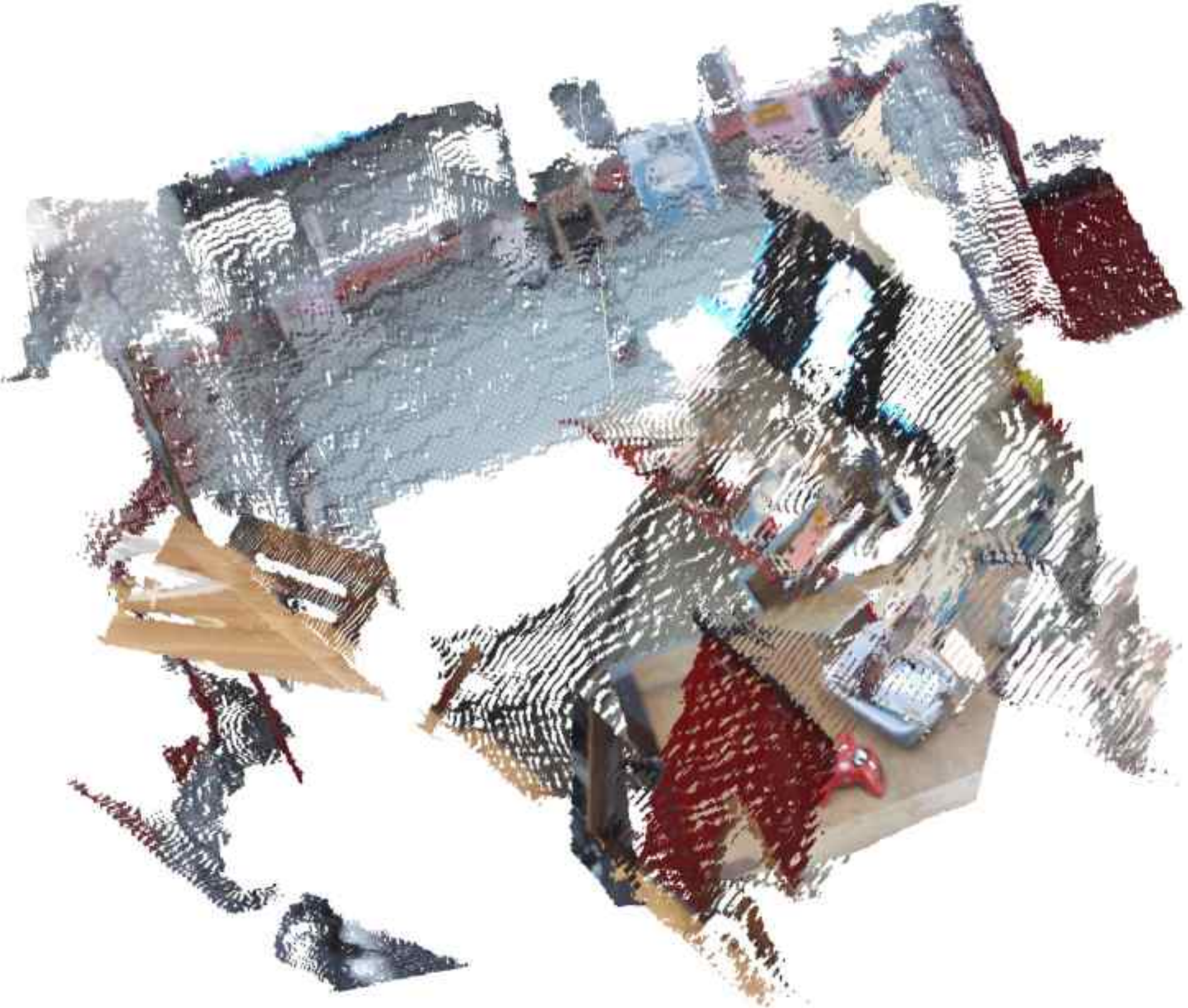}
\end{minipage}\,\,
&
\,\,
\begin{minipage}[t]{0.19\linewidth}
\centering
\includegraphics[width=.48\linewidth]{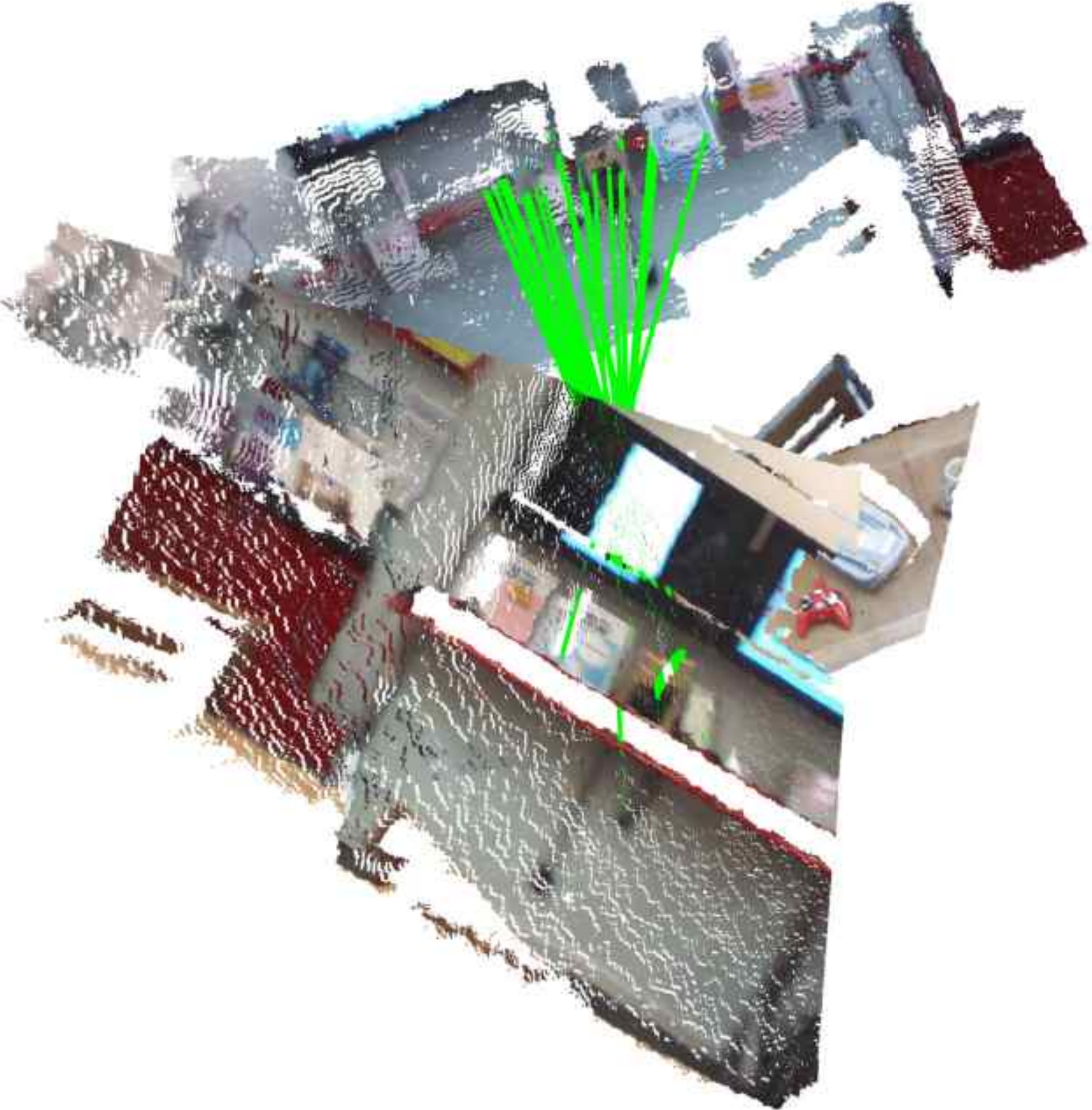}
\includegraphics[width=.48\linewidth]{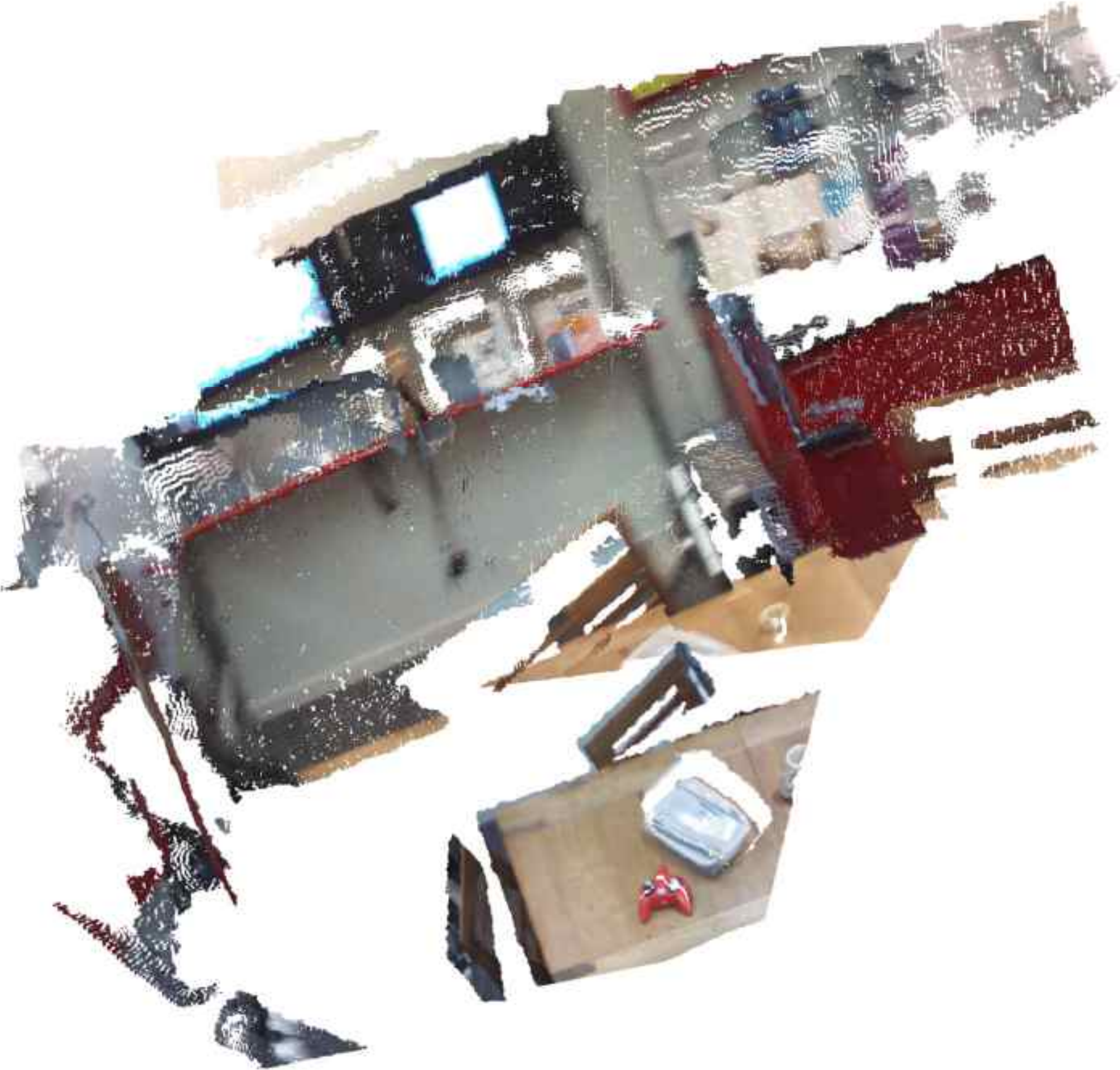}
\end{minipage}\,\,
&
\,\,
\begin{minipage}[t]{0.19\linewidth}
\centering
\includegraphics[width=.48\linewidth]{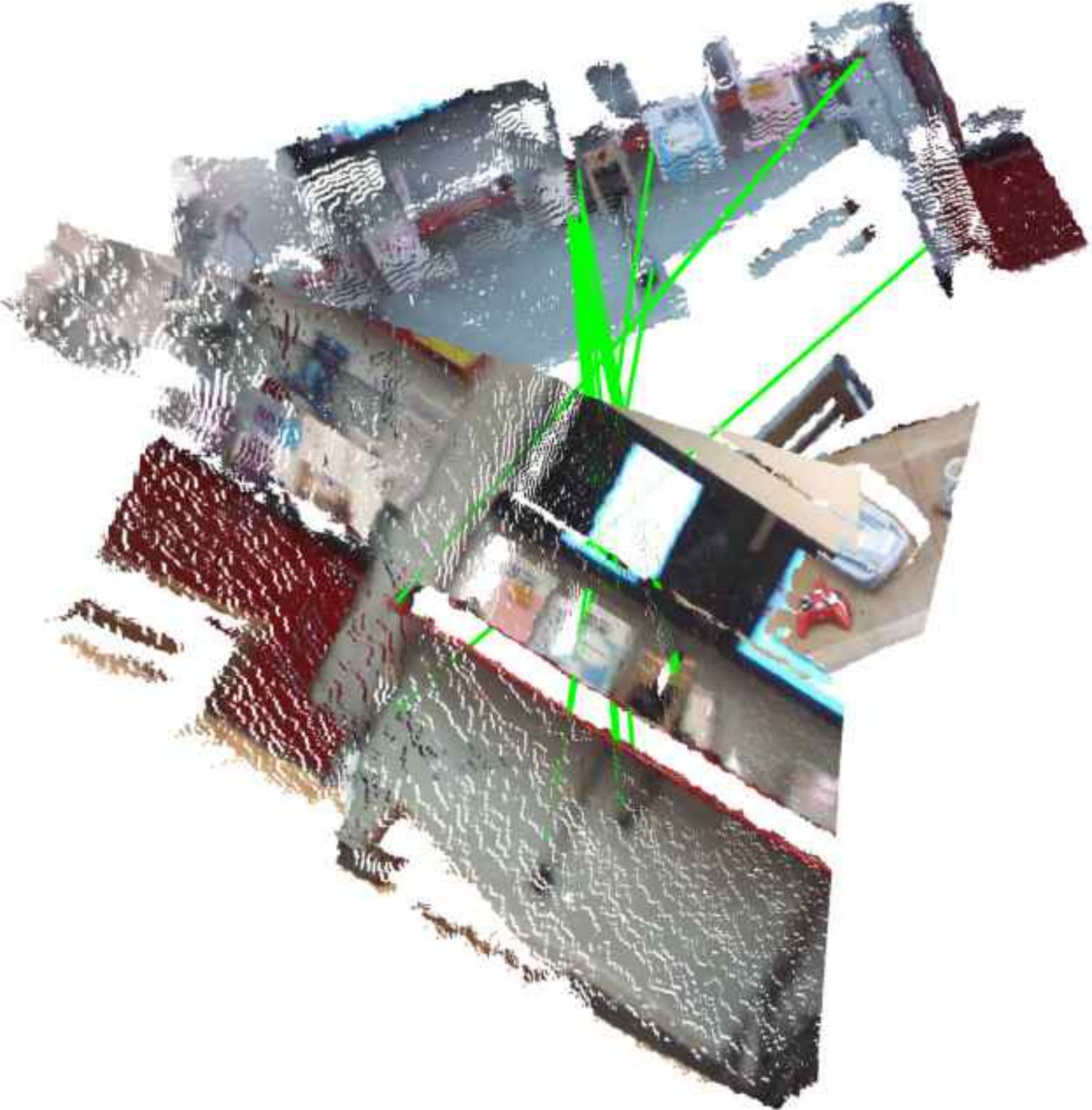}
\includegraphics[width=.48\linewidth]{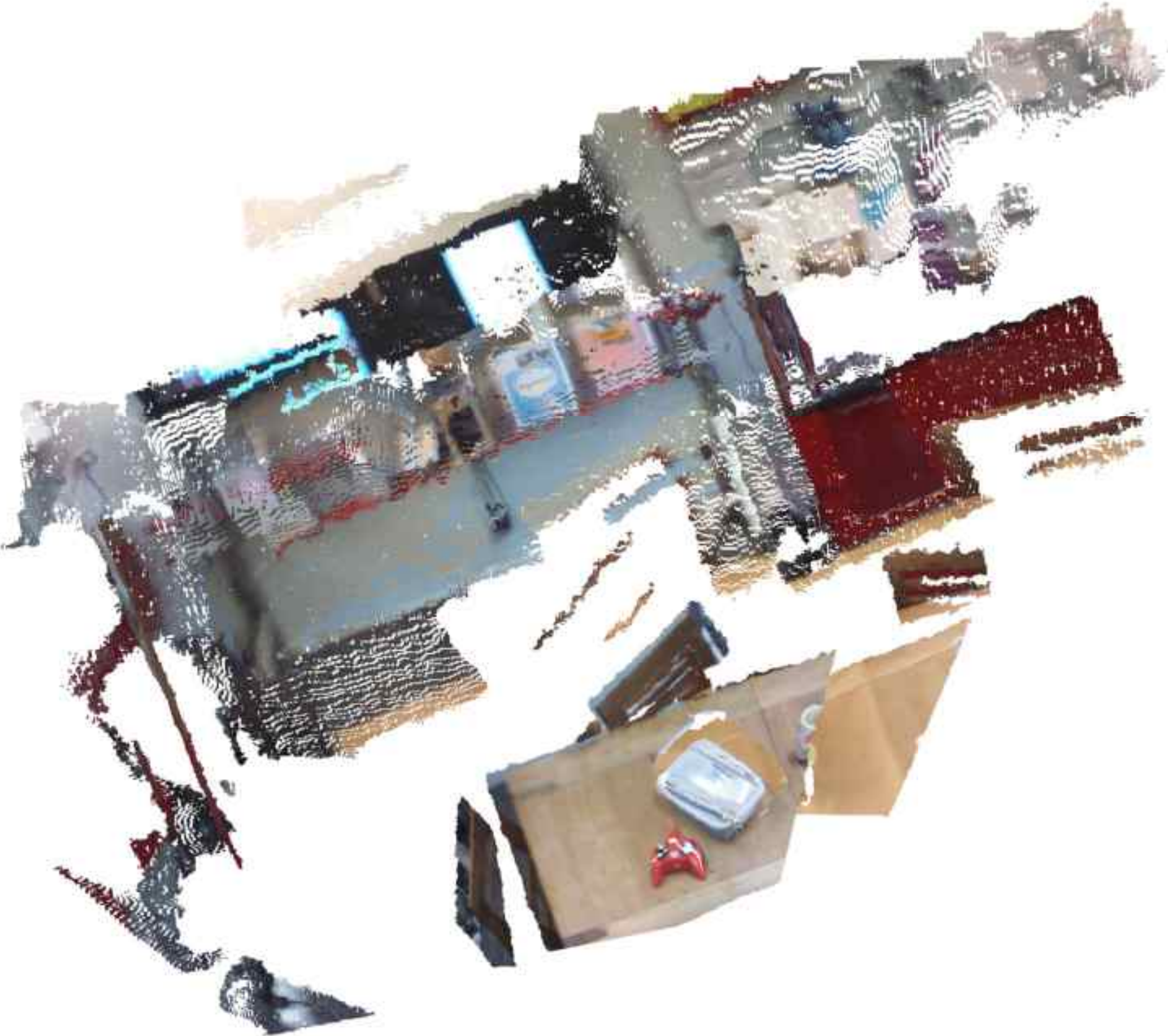}
\end{minipage}\,\,
&
\,\,
\begin{minipage}[t]{0.19\linewidth}
\centering
\includegraphics[width=.48\linewidth]{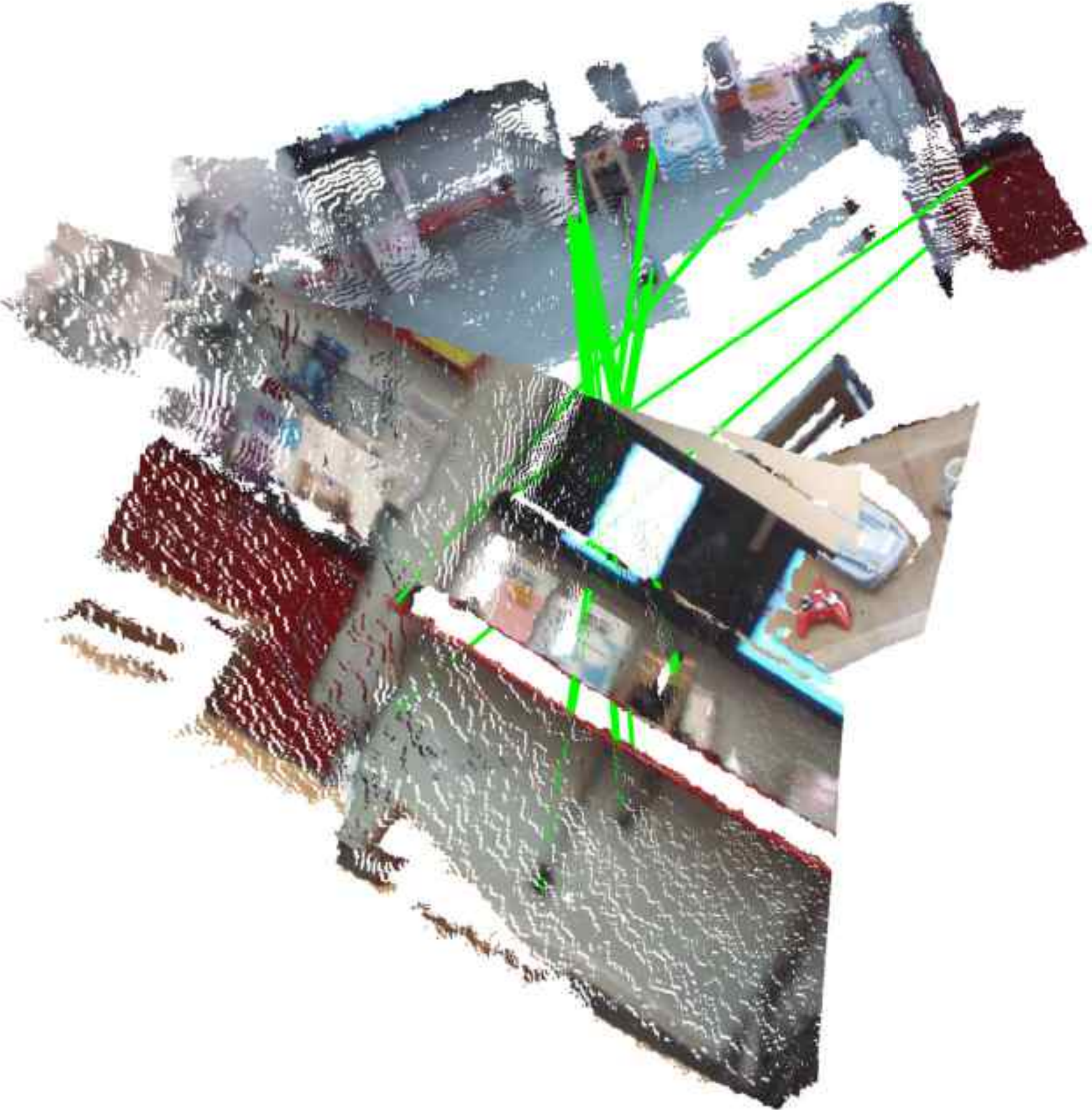}
\includegraphics[width=.48\linewidth]{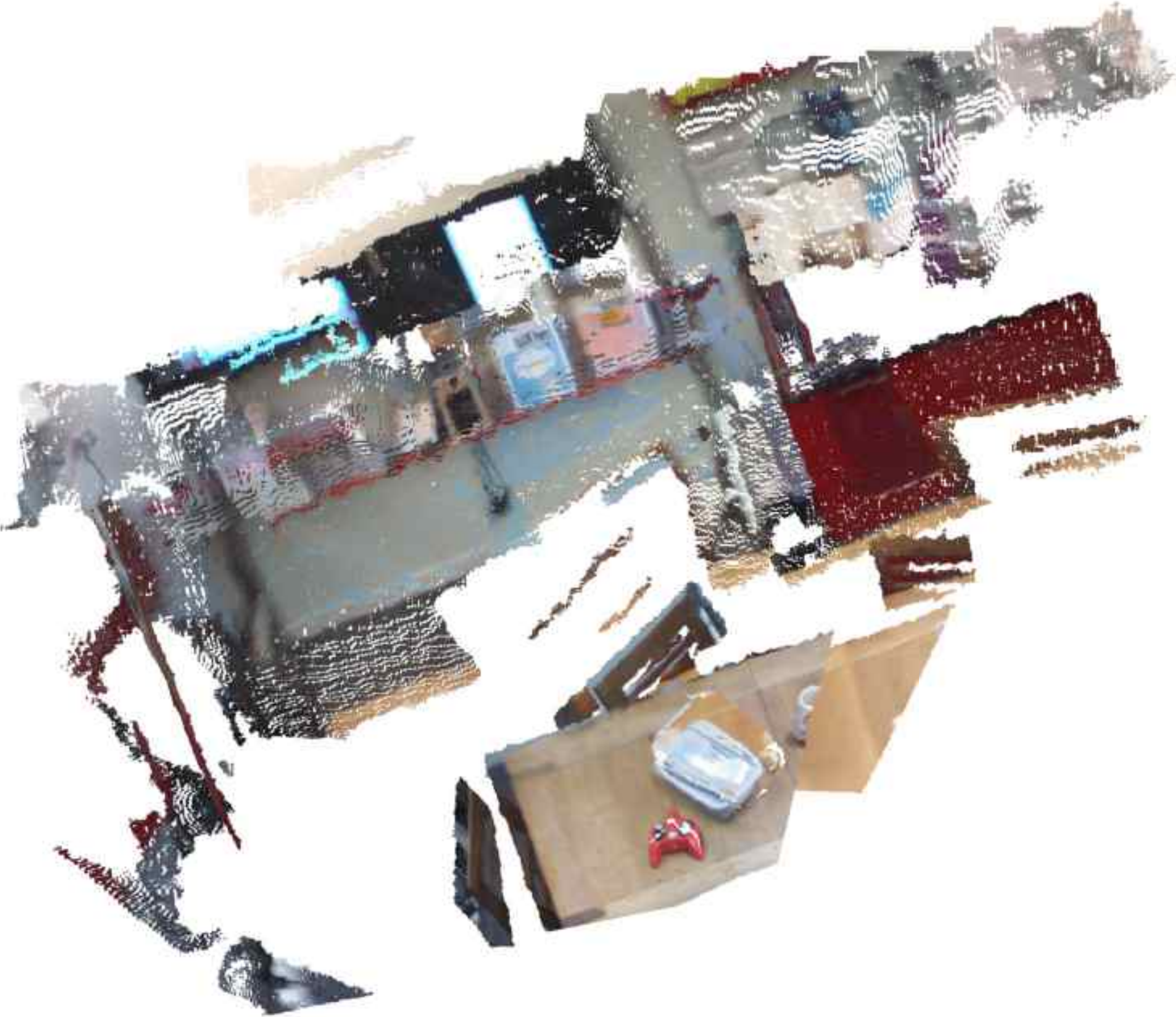}
\end{minipage}\,\,

\\

 & \scriptsize{$N$=1000} && \scriptsize{\textcolor[rgb]{1,0,0}{Failed},\,\verb|\|,\,0.11$s$} &\scriptsize{\textcolor[rgb]{0,0.7,0}{Successful},\,0.33,\,25.94$s$}&\scriptsize{\textcolor[rgb]{0,0.7,0}{Successful},\,\textbf{0.32},\,4.16$s$}&\scriptsize{\textcolor[rgb]{0,0.7,0}{Successful},\,\textbf{0.32},\,\textbf{1.45}$s$}

\\

\rotatebox{90}{\,\,\footnotesize{\textit{office}}\,}\,
&
\,\,
\begin{minipage}[t]{0.1\linewidth}
\centering
\includegraphics[width=0.5\linewidth]{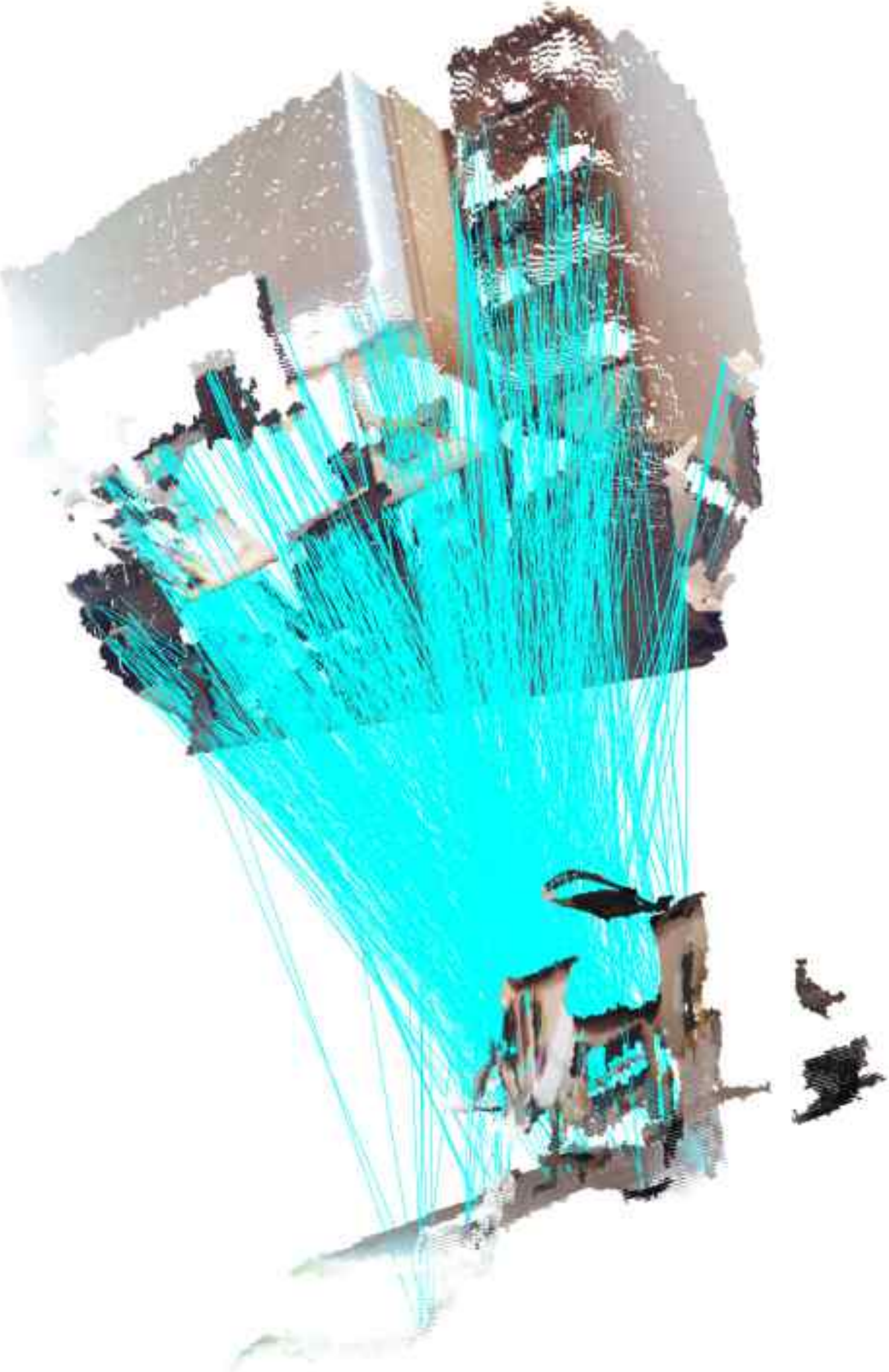}
\end{minipage}\,\,
& &
\,\,
\begin{minipage}[t]{0.19\linewidth}
\centering
\includegraphics[width=.27\linewidth]{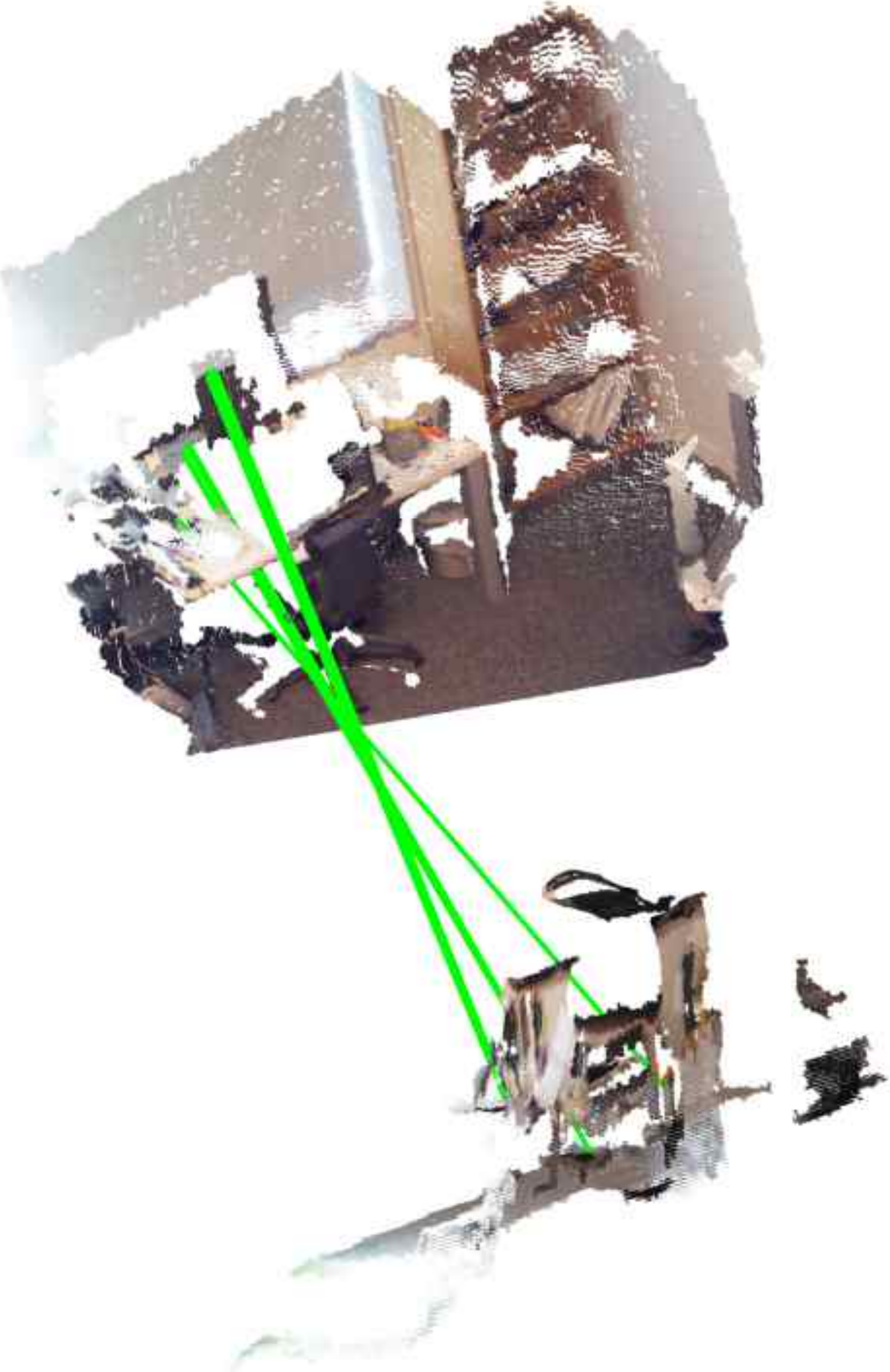}
\includegraphics[width=.4\linewidth]{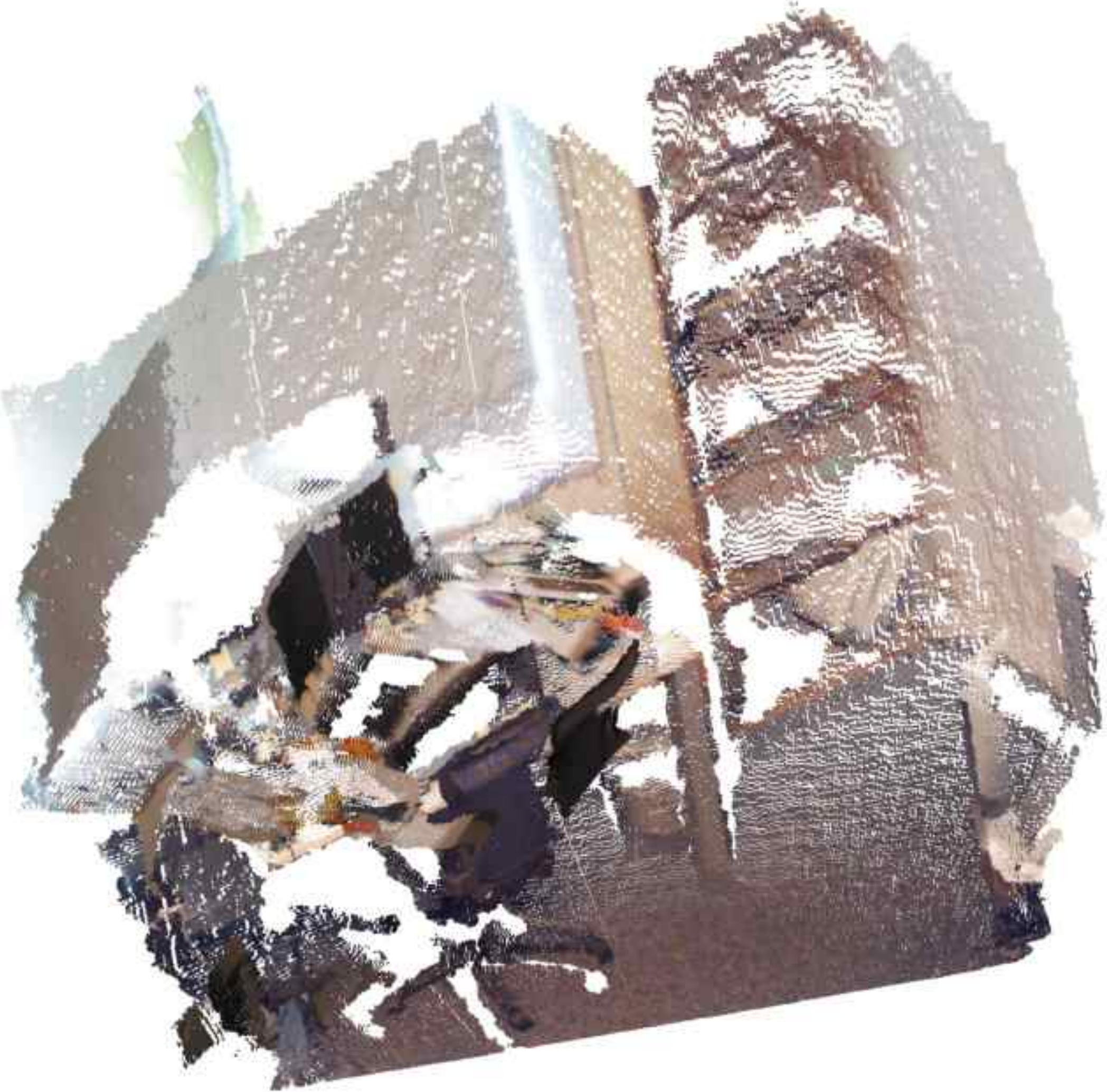}
\end{minipage}\,\,
&
\,\,
\begin{minipage}[t]{0.19\linewidth}
\centering
\includegraphics[width=.27\linewidth]{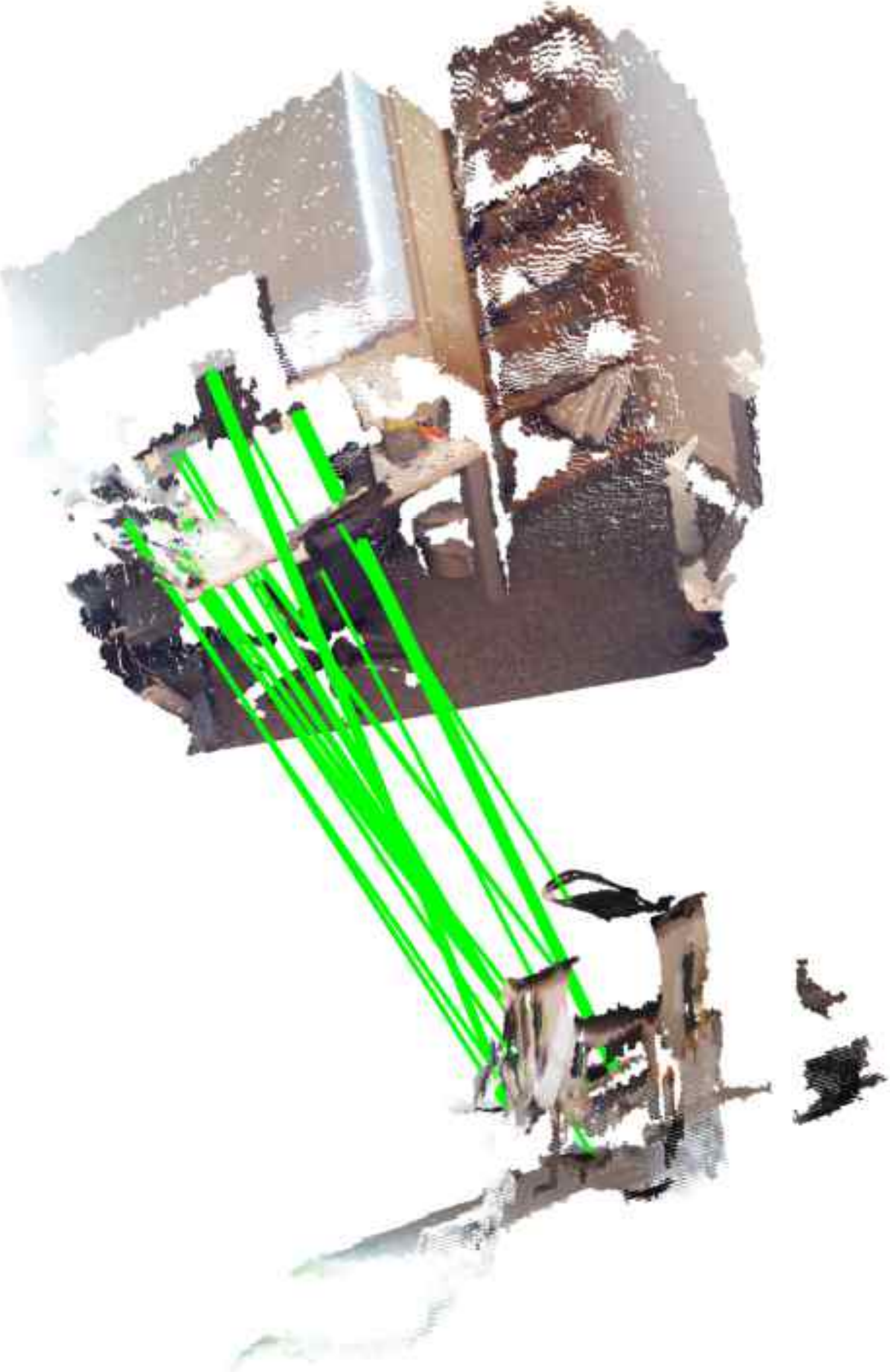}
\includegraphics[width=.4\linewidth]{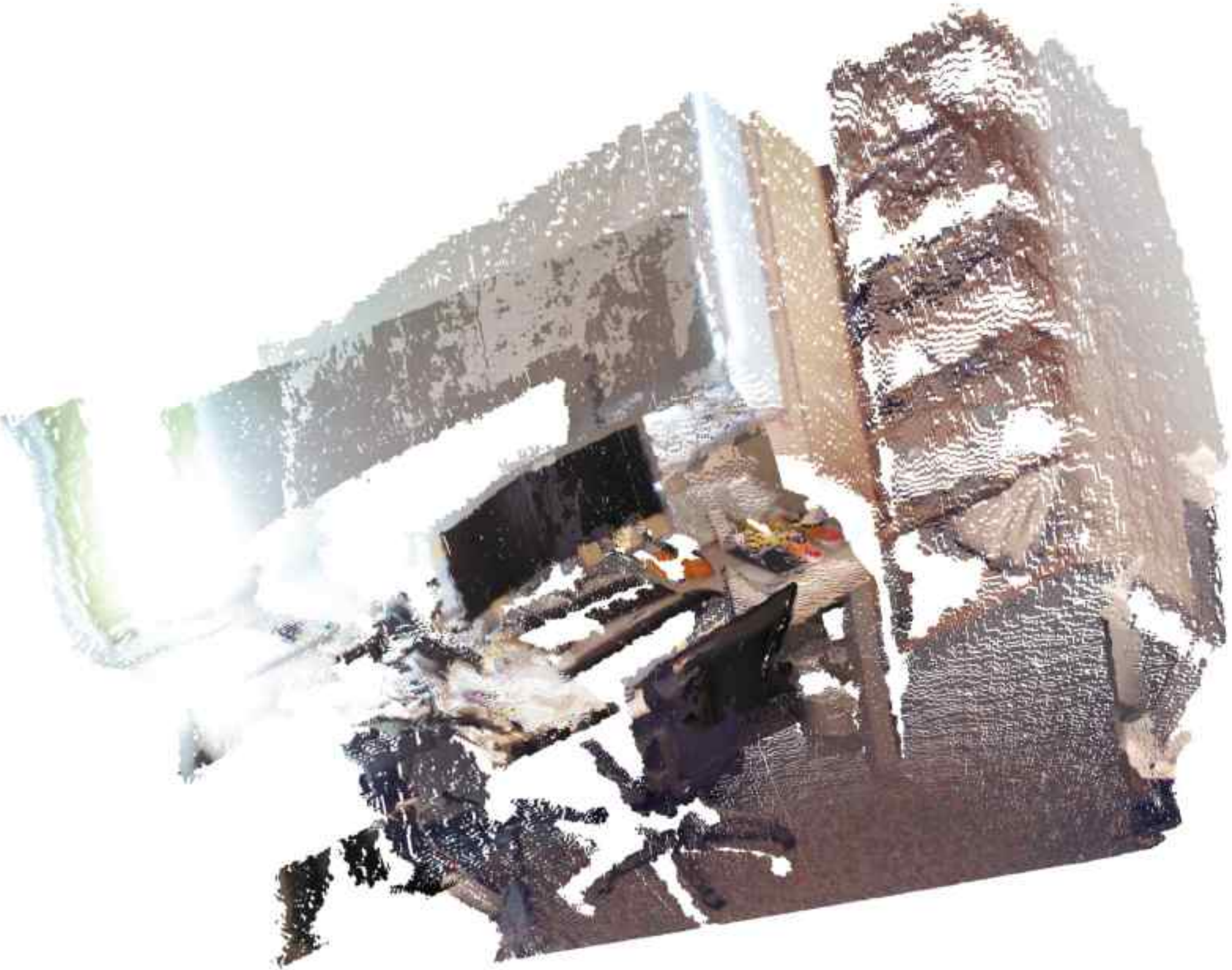}
\end{minipage}\,\,
&
\,\,
\begin{minipage}[t]{0.19\linewidth}
\centering
\includegraphics[width=.27\linewidth]{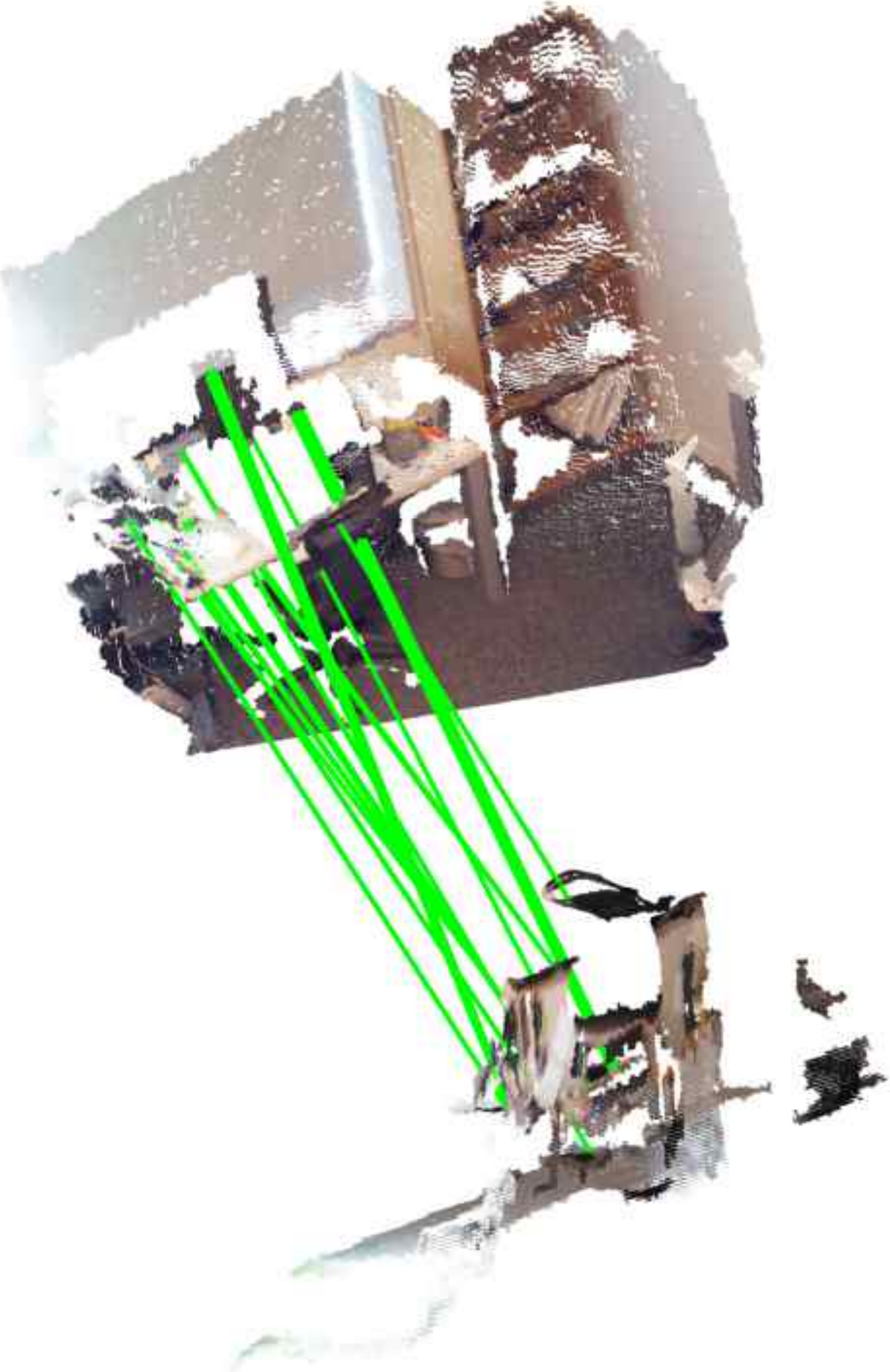}
\includegraphics[width=.4\linewidth]{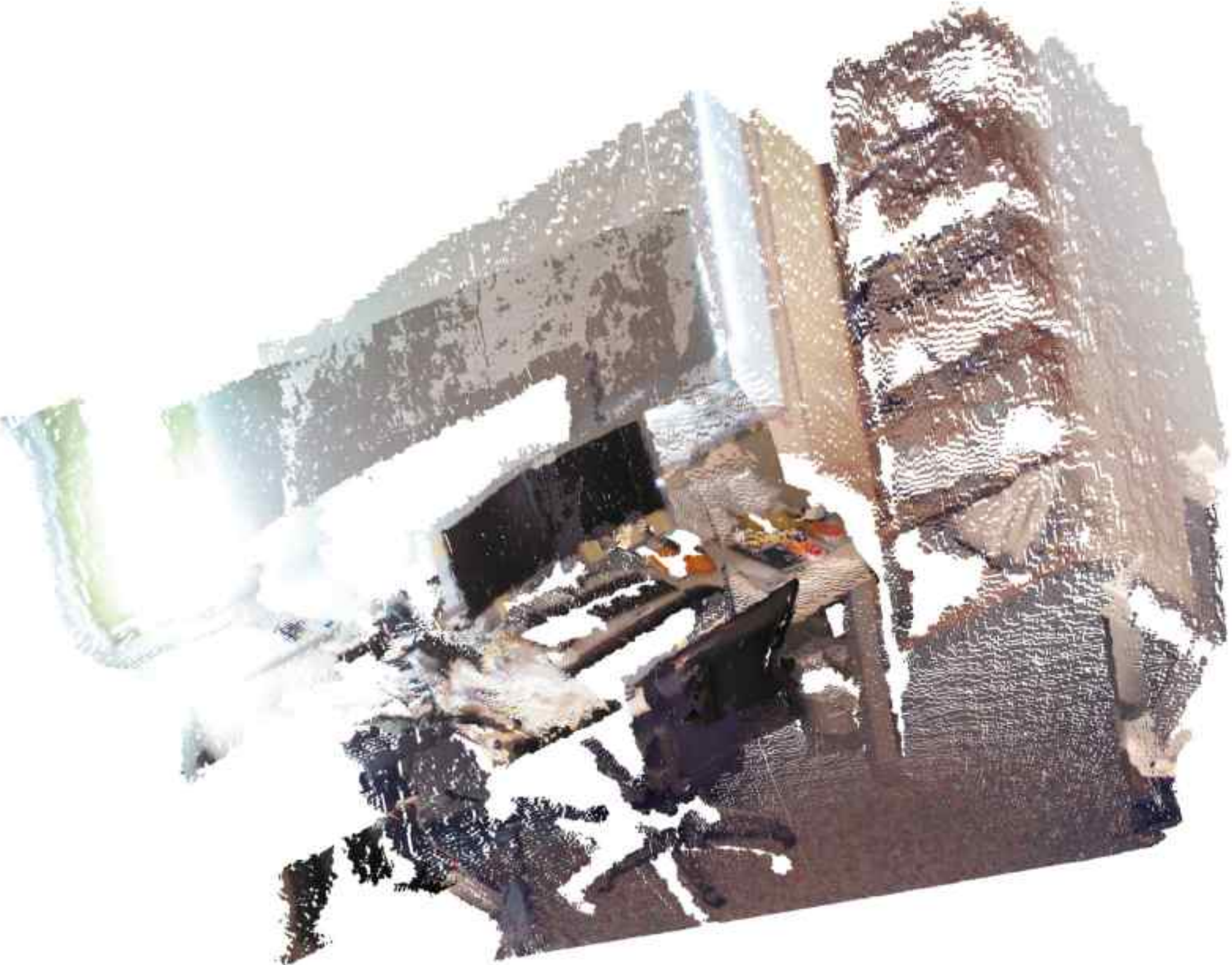}
\end{minipage}\,\,
&
\,\,
\begin{minipage}[t]{0.19\linewidth}
\centering
\includegraphics[width=.27\linewidth]{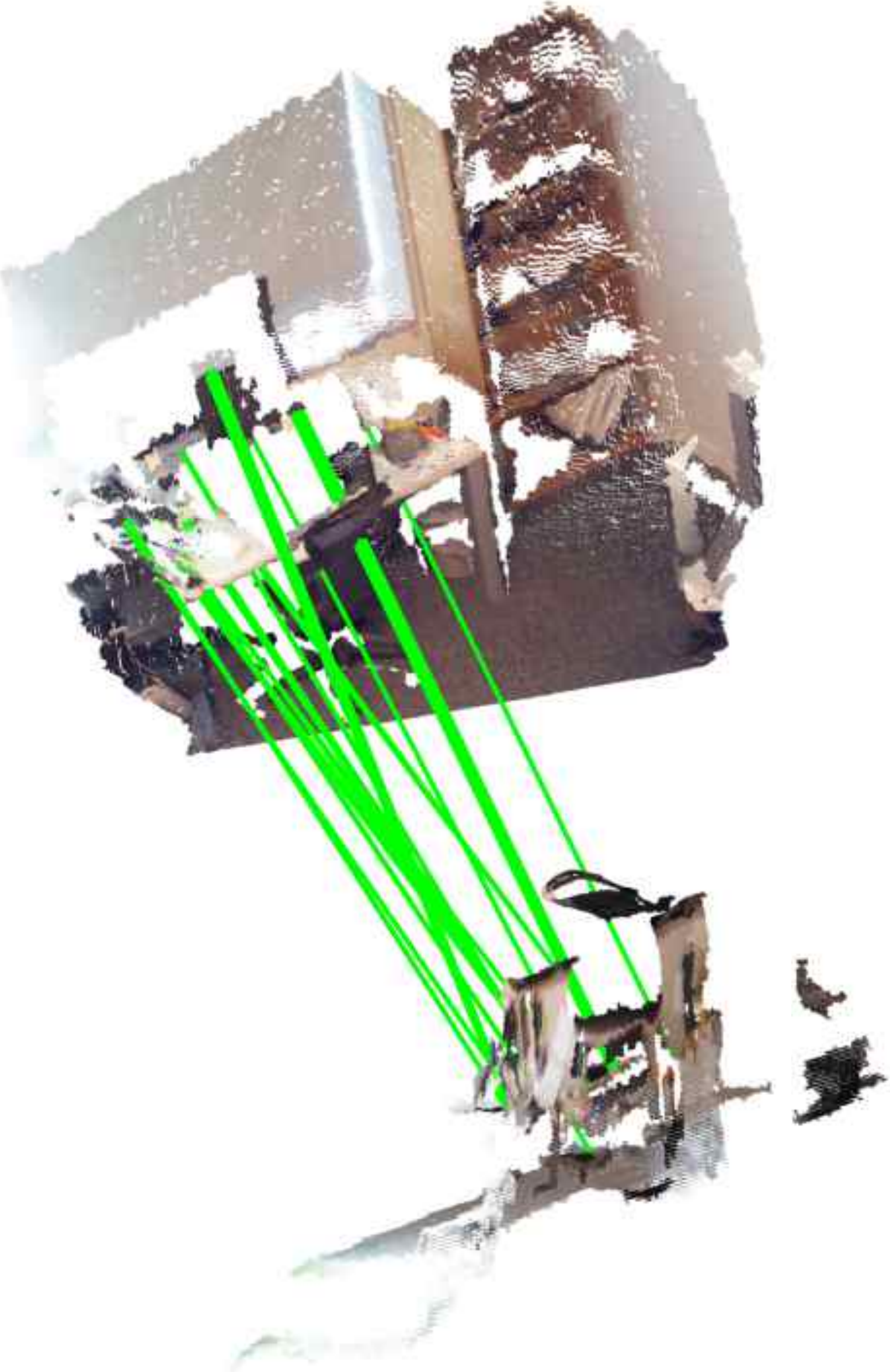}
\includegraphics[width=.4\linewidth]{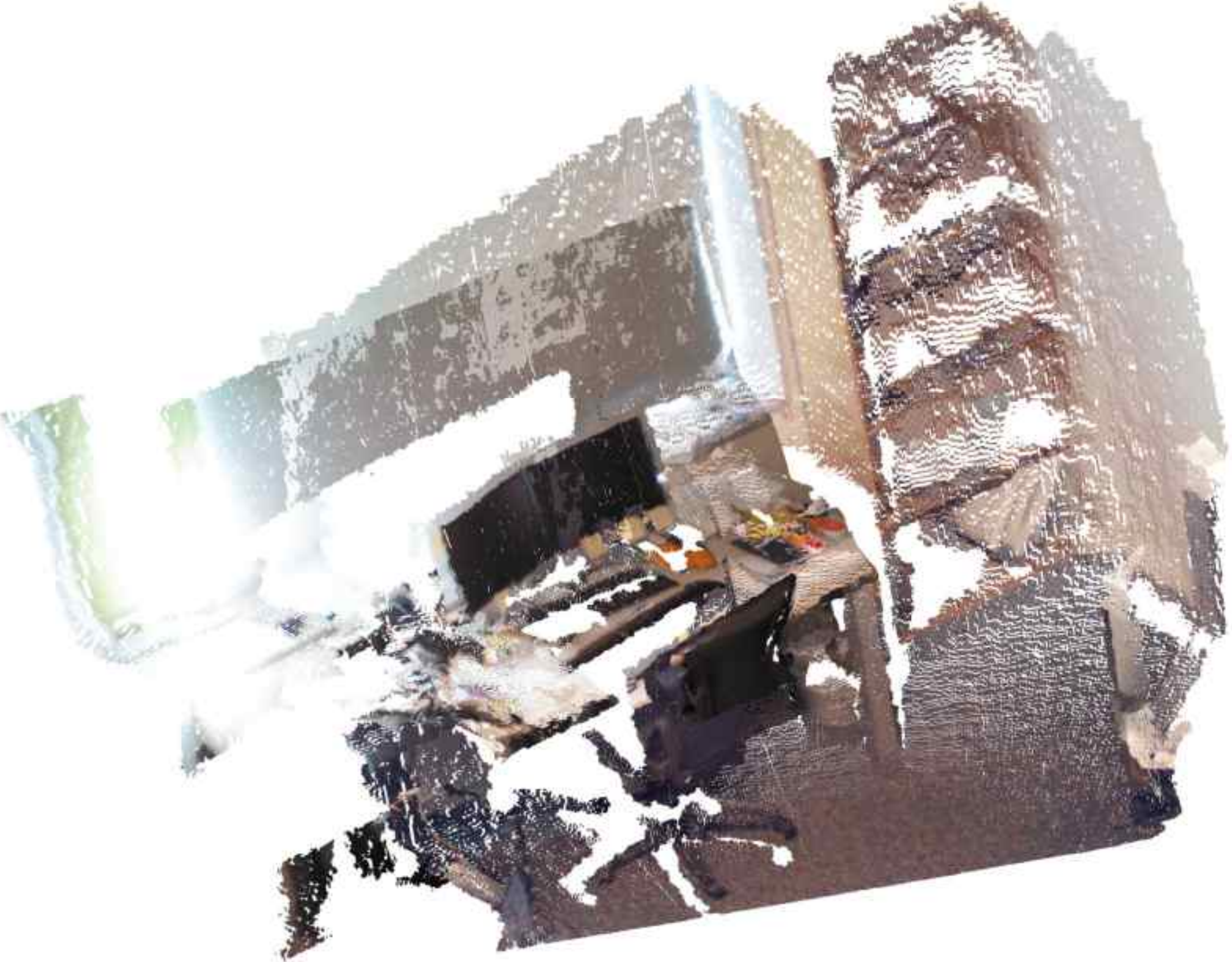}
\end{minipage}\,\,

\\

 & \scriptsize{$N$=985} && \scriptsize{\textcolor[rgb]{1,0,0}{Failed},\,\verb|\|,\,0.13$s$} &\scriptsize{\textcolor[rgb]{0,0.7,0}{Successful},\,0.37,\,25.18$s$}&\scriptsize{\textcolor[rgb]{0,0.7,0}{Successful},\,{0.28},\,17.03$s$}&\scriptsize{\textcolor[rgb]{0,0.7,0}{Successful},\,\textbf{0.26},\,\textbf{2.95}$s$}

\\

\rotatebox{90}{\,\,\footnotesize{\textit{office}}\,}\,
&
\,\,
\begin{minipage}[t]{0.1\linewidth}
\centering
\includegraphics[width=.5\linewidth]{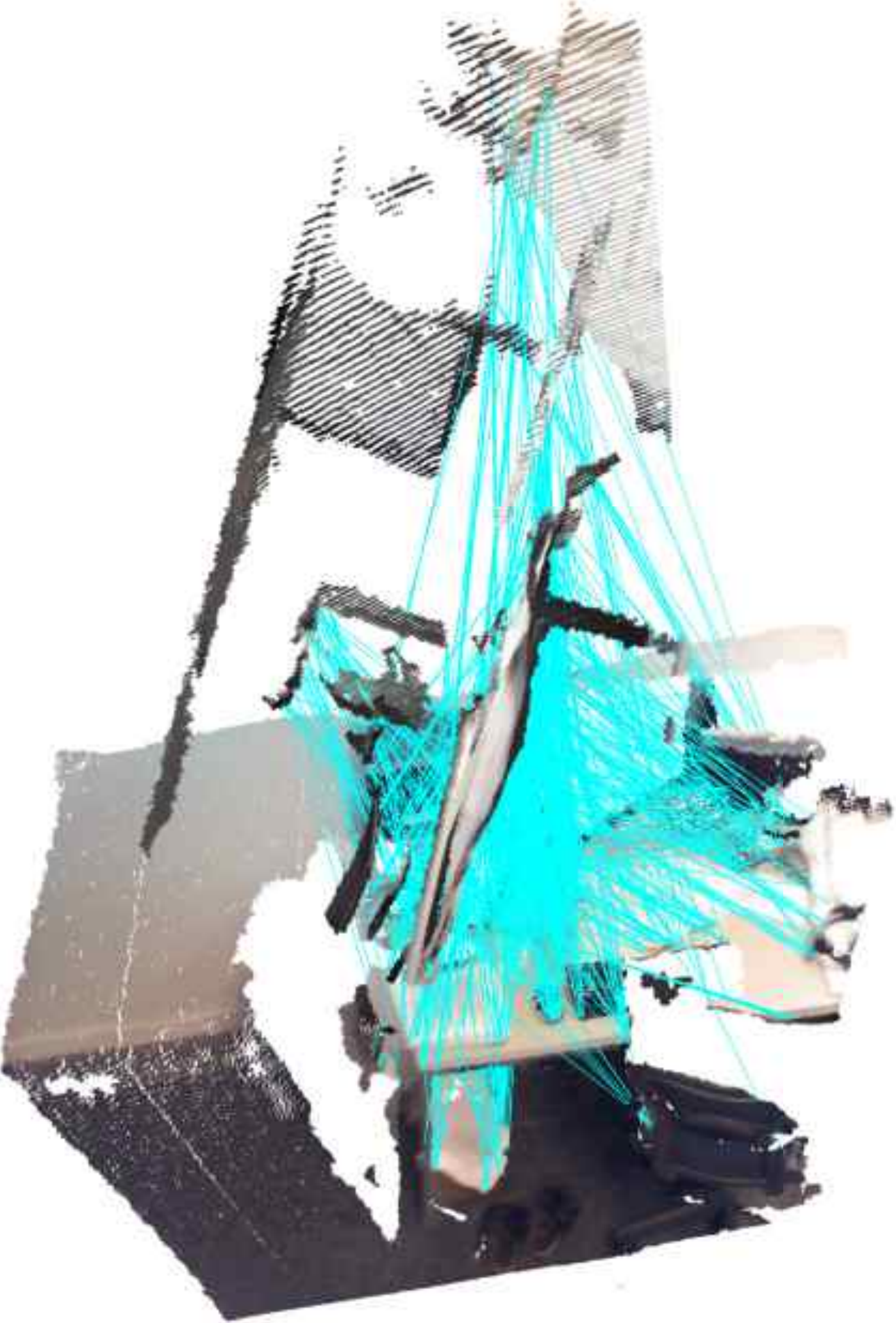}
\end{minipage}\,\,
& &
\,\,
\begin{minipage}[t]{0.19\linewidth}
\centering
\includegraphics[width=.27\linewidth]{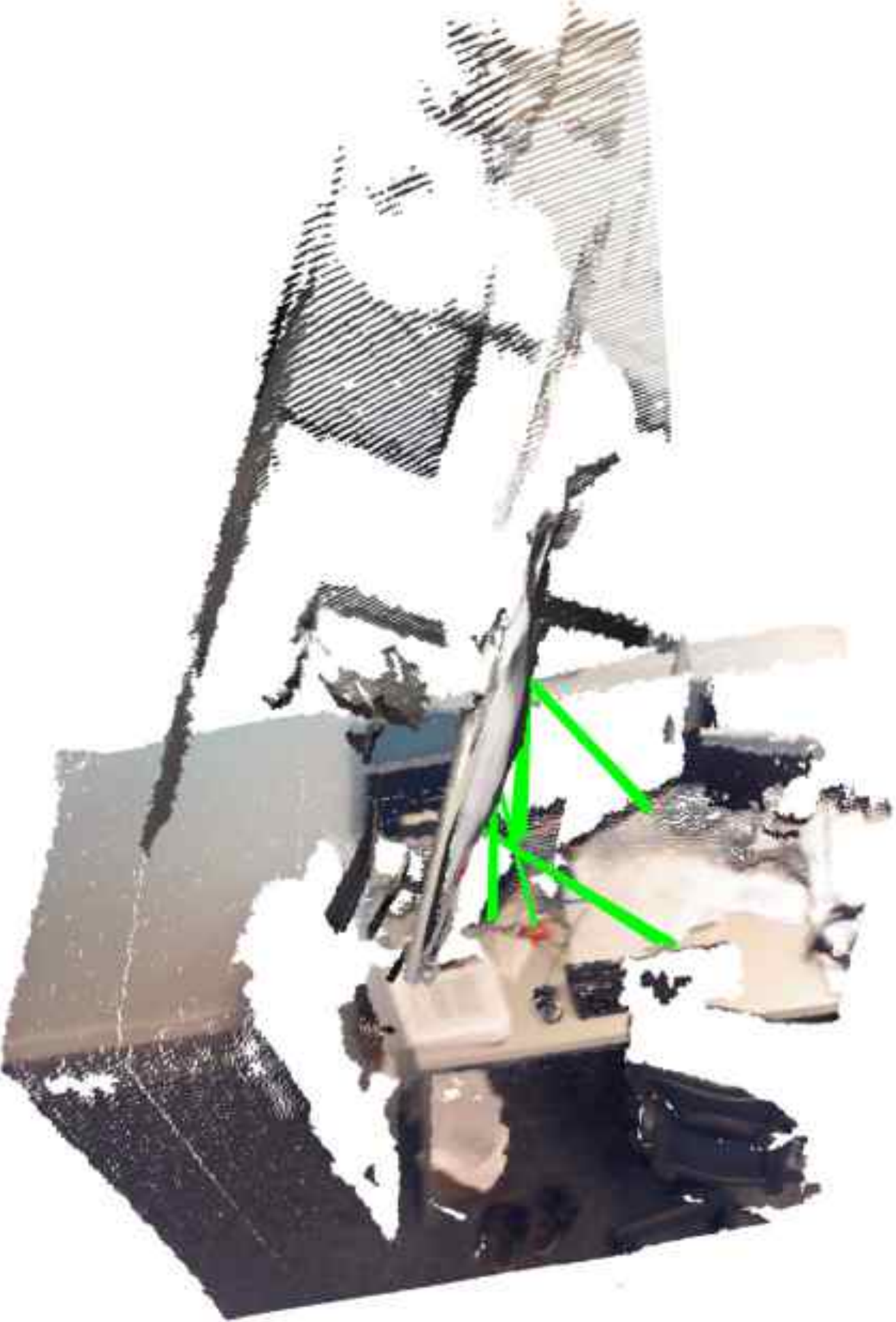}
\includegraphics[width=.4\linewidth]{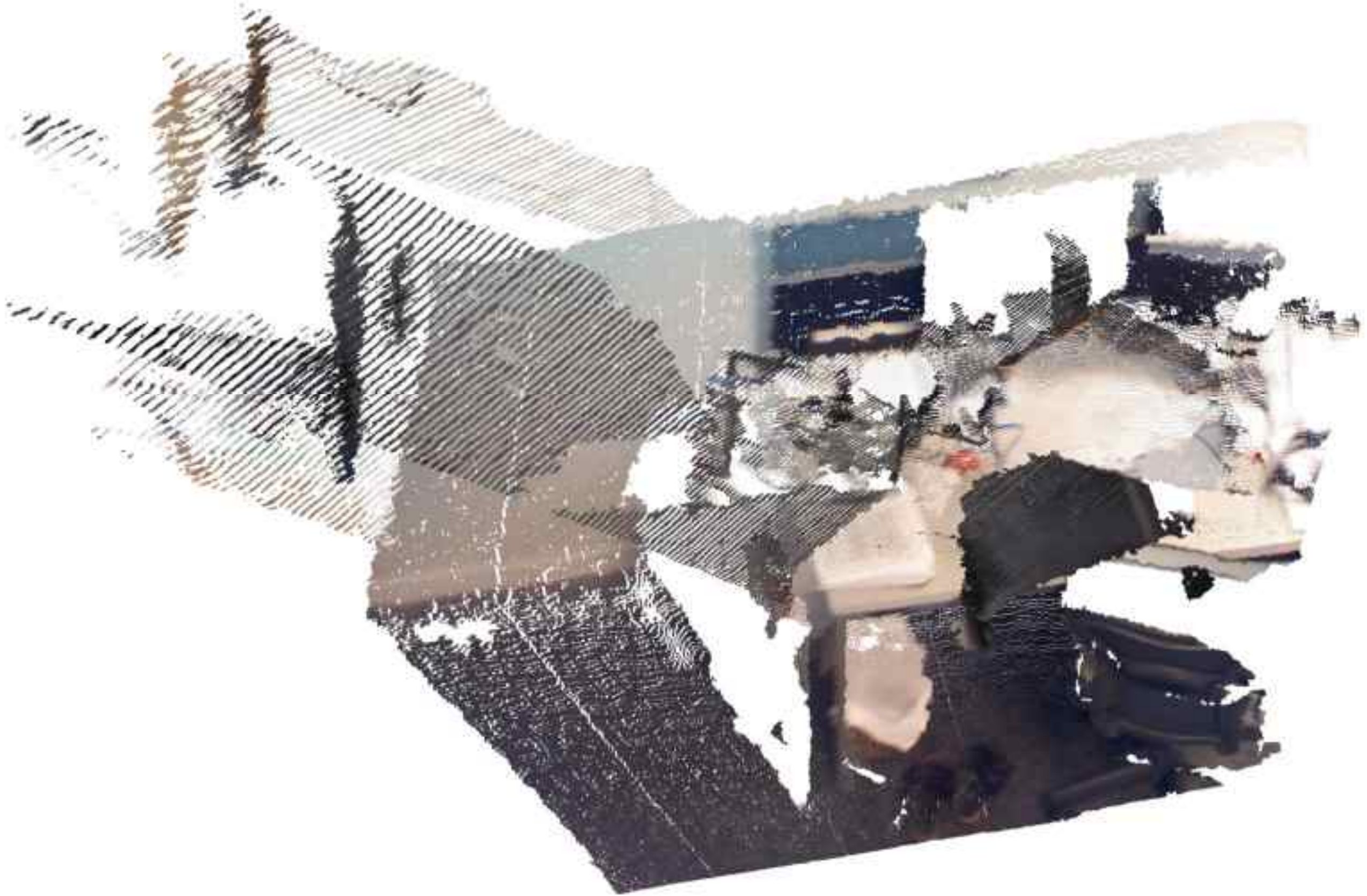}
\end{minipage}\,\,
&
\,\,
\begin{minipage}[t]{0.19\linewidth}
\centering
\includegraphics[width=.27\linewidth]{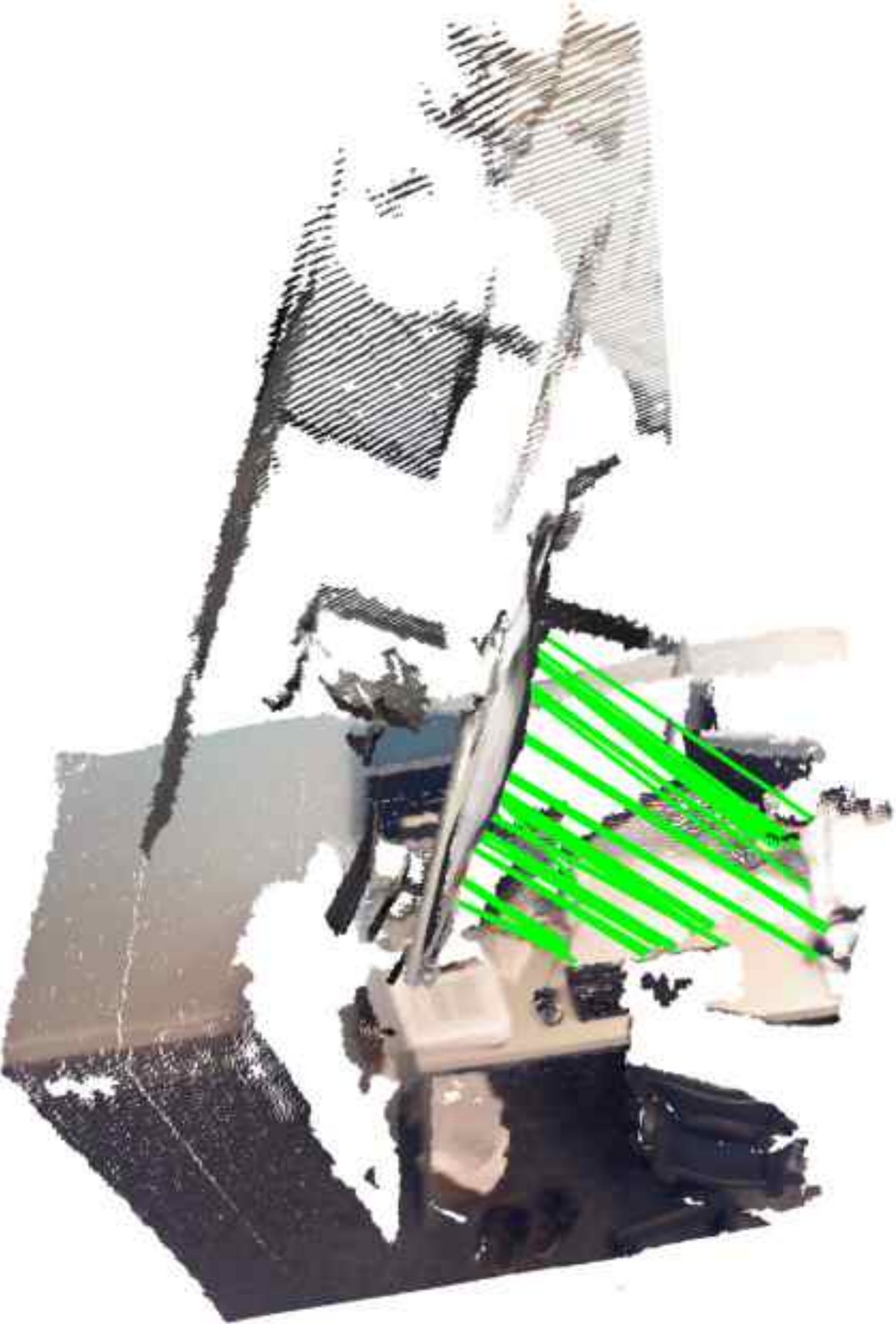}
\includegraphics[width=.4\linewidth]{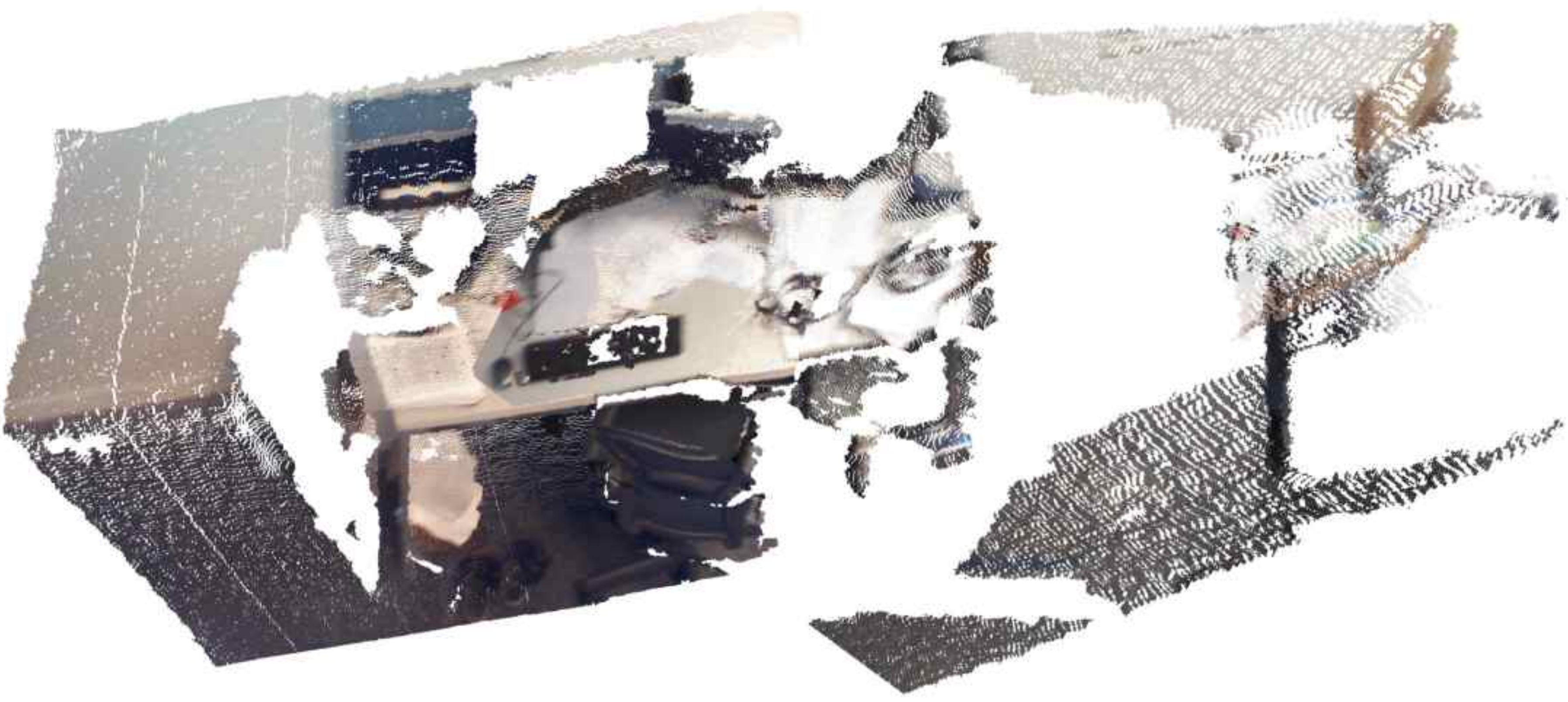}
\end{minipage}\,\,
&
\,\,
\begin{minipage}[t]{0.19\linewidth}
\centering
\includegraphics[width=.27\linewidth]{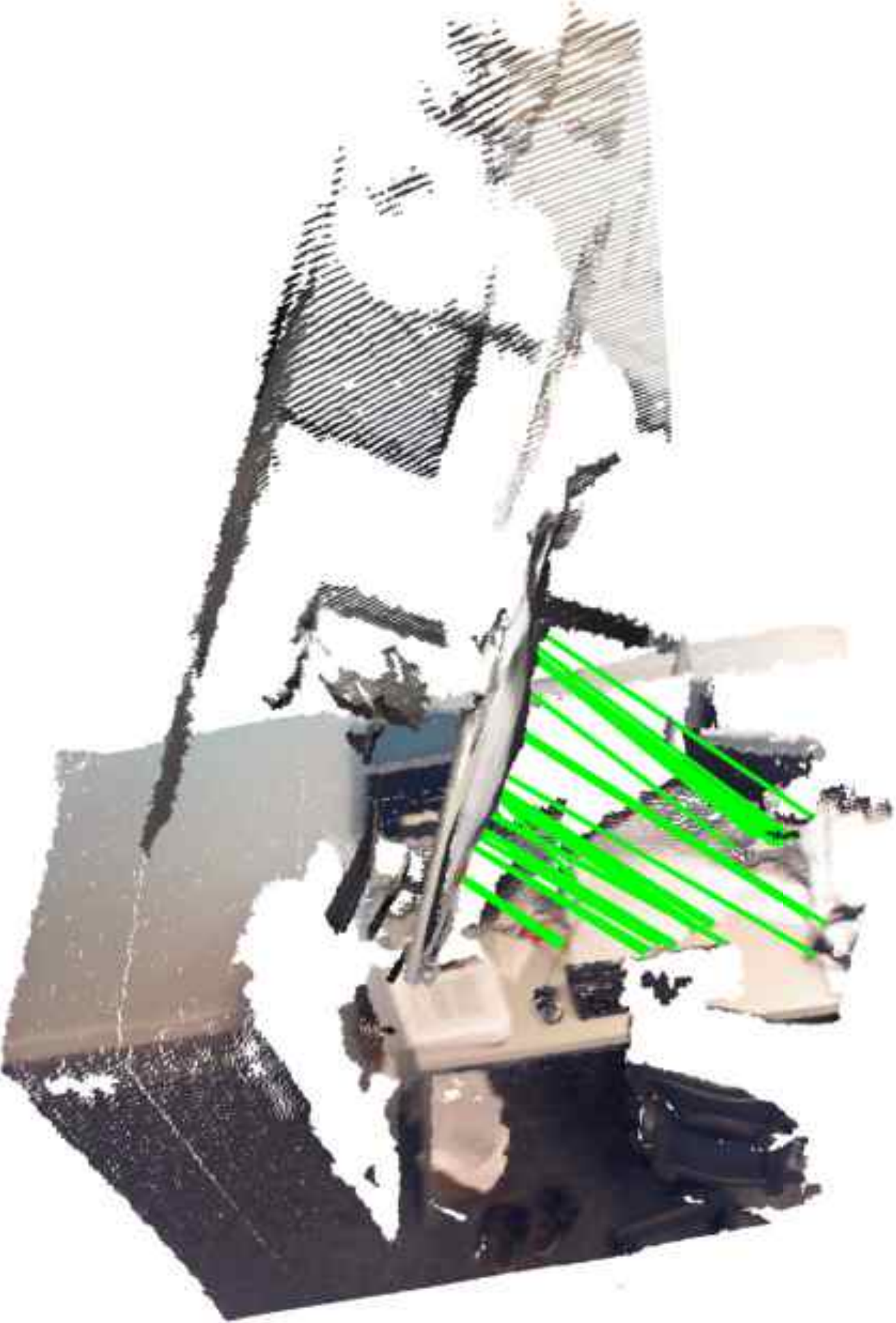}
\includegraphics[width=.4\linewidth]{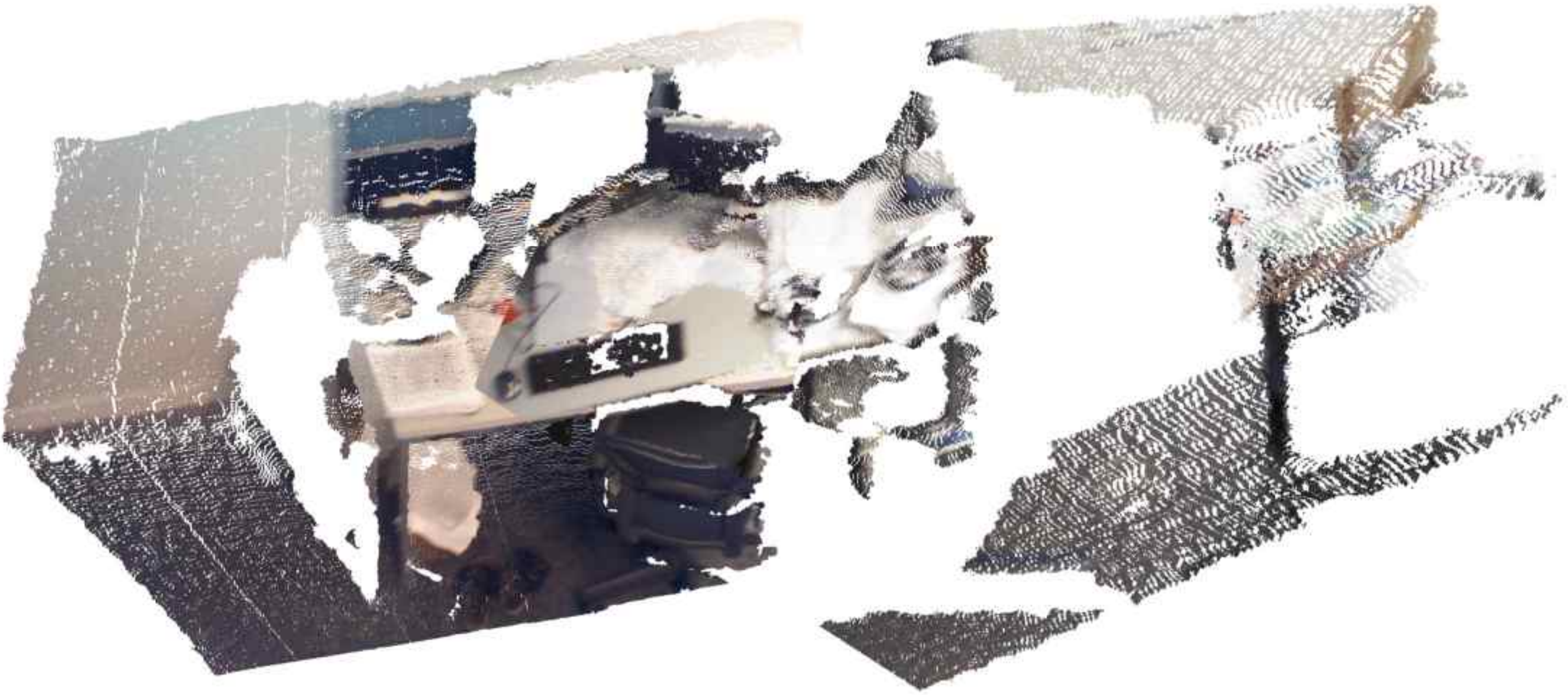}
\end{minipage}\,\,
&
\,\,
\begin{minipage}[t]{0.19\linewidth}
\centering
\includegraphics[width=.27\linewidth]{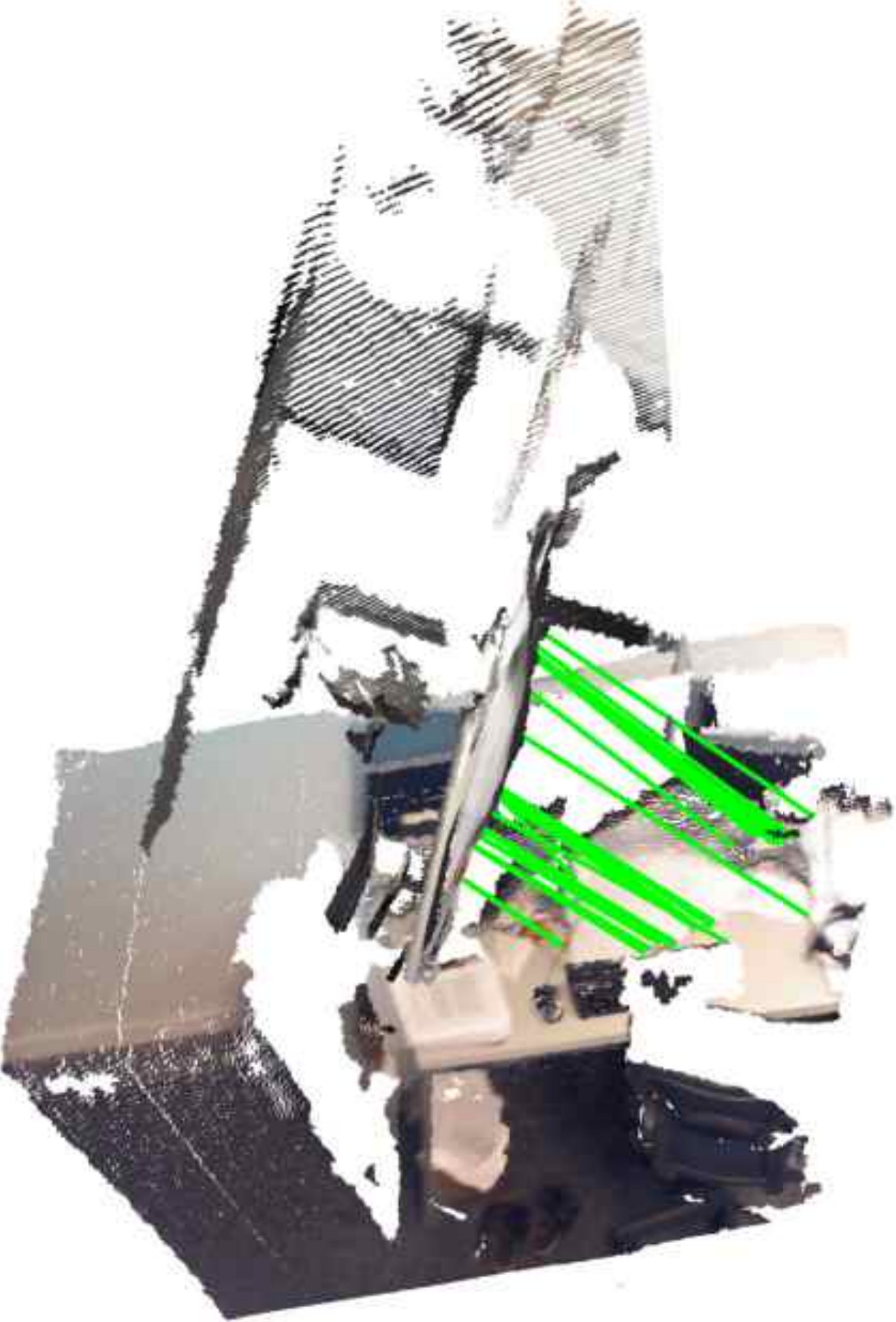}
\includegraphics[width=.4\linewidth]{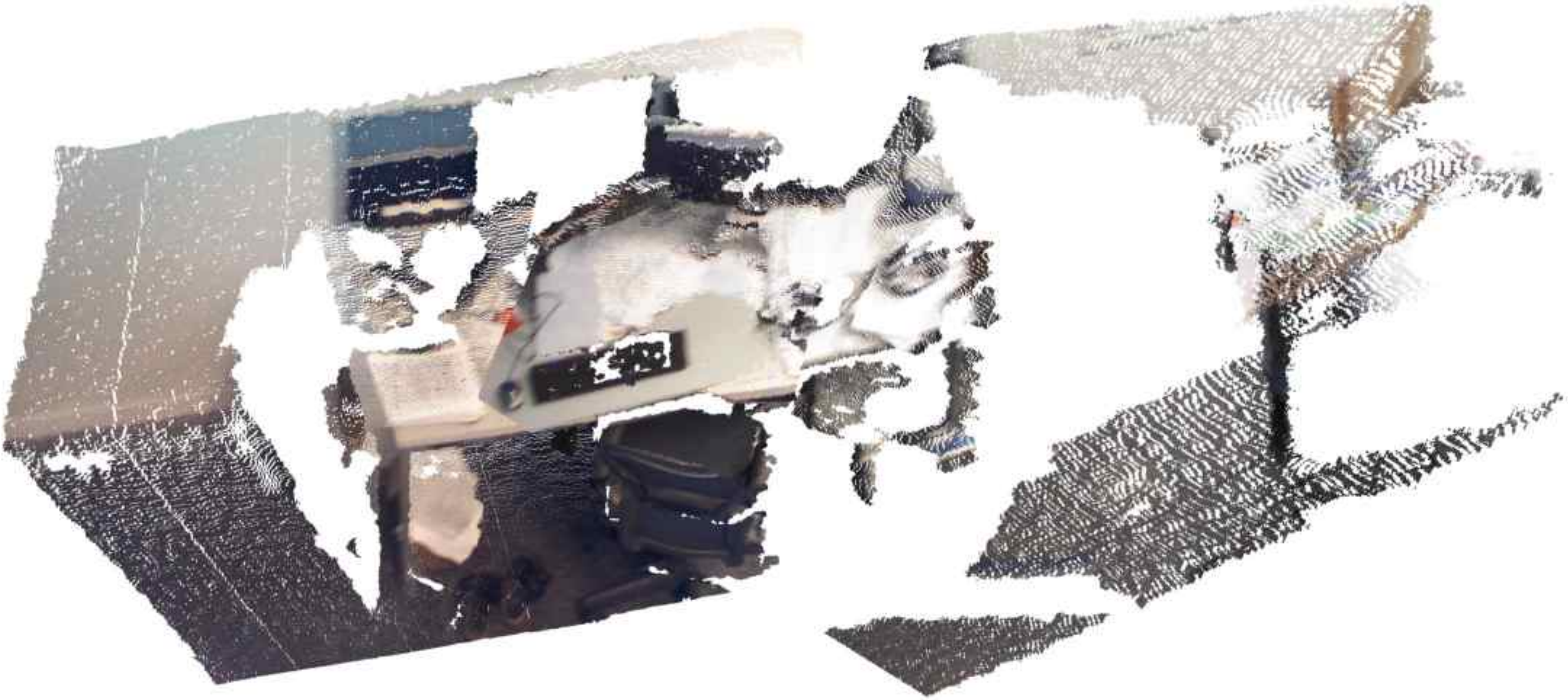}
\end{minipage}\,\,

\end{tabular}

\caption{Scan matching results on the Microsoft 7-scenes dataset~\cite{shotton2013scene}. The first column shows raw SURF correspondences, and the rest columns show: (i) the inliers found and (ii) the scene stitching results, using FLO-RANSAC, GNC-TLS, GORE+RANSAC and TriVoC. On top of each result, we show the stitching status (\textcolor[rgb]{1,0,0}{Fail} or \textcolor[rgb]{0,0.5,0}{Succeed}), the RMSE (Root Mean Square Error) and runtime. Note that when the stitching is failed, we no longer show the RMSE since it would become meaningless. Best results are shown in \textbf{bold} font.}
\label{scene-stitching}
\end{figure*}

\subsection{Standard Benchmarking}\label{sec-bench}

We evaluate TriVoC in benchmarking experiments with existing state-of-the-art robust solvers: FGR~\cite{zhou2016fast}, GNC-TLS/-GM~\cite{yang2020graduated}, ADAPT~\cite{tzoumas2019outlier}, RANSAC~\cite{fischler1981random}, FLO-RANSAC~\cite{lebeda2012fixing} (LO$^+$-RANSAC), GORE~\cite{bustos2017guaranteed} and GORE+RANSAC\footnote[2]{Using RANSAC to find the best consensus set after the guaranteed outlier removal of GORE.}. Two RANSAC solvers are set with 10000 maximum iterations and 0.99 confidence, and the local optimization is set with 10 iterations. The inlier threshold is constantly set to $\gamma=6\sigma$. We use the geodesic distance~\cite{hartley2013rotation} to denote rotation errors (in degrees): $\left|\arccos\left(\frac{trace({\boldsymbol{R}_{gt}}^{\top}\boldsymbol{R}^{\star})-1}{2}\right)\right|\cdot\frac{180}{\pi}^{\circ}$, and use L2-norm to denote translation errors (in meters): $
\left\|\boldsymbol{t}_{gt}-\boldsymbol{t}^{\star}\right\|\, m$.

Our setup is similar to~\cite{yang2020teaser}. We adopt the \textit{bunny} and \textit{armadillo} point clouds from Stanford 3D Repository~\cite{curless1996volumetric}. The point cloud is downsampled to $N=\{100,500,1000\}$ and resized to fit in a $[-0.5,0.5]^3m$ box as the initial point set $\mathcal{P}$. Then we transform $\mathcal{P}$ with a random transformation: $\boldsymbol{R}\in SO(3)$ and $\boldsymbol{t}\in\mathbb{R}^3$ ($||\boldsymbol{t}||\leq 3$) and also add random noise with $\sigma=0.01m$ to get the transformed point set $\mathcal{Y}$. To create outliers simulating cluttered scenes, we replace 20--99\% of the points in $\mathcal{Y}$ with random points in a 3D sphere of radius 1. All the results are based on 50 Monte Carlo runs.

From the boxplot results in Fig.~\ref{Benchmarking}, we can observe that: (i) our TriVoC is (one of) the most robust solver, tolerating over 95\% outliers (up to 99\% with $N=1000$), (ii) TriVoC has (one of) the highest estimation accuracy, indicating that it can find the largest consensus set for the final estimation, and (iii) like the non-minimal solvers (GNC, FGR, ADAPT), TriVoC has almost constant runtime w.r.t. varied outlier ratios, (but it is much more robust than them), and more importantly, TriVoC is always the fastest solver when the outlier ratio is extreme (over 95\%) excluding those already failed ones.

\subsection{Runtime Analysis of TriVoC}
We provide an explicit analysis on the runtime of the different parts in TriVoC. We divide TriVoC into 2 main parts: consistency matrix building (Algorithm~\ref{algo-consistency}) and consensus maximization (lines 2-21 in Algorithm~\ref{algo-main}), whose runtime with different $N$ is displayed in Fig~\ref{time-TriVoC}. We can see that: (i) the first part, having time complexity of $O(N^2)$, occupies a large portion of the time cost, so with smaller $N$ TriVoC would become significantly faster, and (ii) TriVoC has the merit that when the outlier ratio is no more than 98\%, its runtime almost stabilizes at a certain value (with relatively small fluctuations) and would not increase with the outlier ratio, which is a desirable characteristic for practical use.

\subsection{Correctness Guarantee of TriVoC}\label{sec-guarantee}

Since TriVoC is based on consensus maximization over 3-point sets, how could we know if we have obtained the full inlier set (or the maximum consensus set) after TriVoC is terminated? Now we further test the correctness guarantee of TriVoC (which means how confident we are to find all the inliers using TriVoC) compared against the traditional random-sampling paradigm of RANSAC in Fig.~\ref{guarantee-TriVoC}. We report the numbers of 3-point sets (including outlier ones) selected and computed with as well as the numbers of pure-inlier 3-point sets in RANSAC and TriVoC (with $N=1000$), respectively. It is apparent to observe that: (i)TriVoC can find and use at least over 50 different pure-inlier 3-point sets (at most over 2000) for consensus maximization even with extremely high outlier ratios (e.g. 99\%), whereas RANSAC can only find several pure-inlier sets when the outlier ratio is not high and may not find any pure-inlier set with extreme outliers (over 95\%), and (ii) the ratio of the number of pure-inlier sets found to the number of all the 3-point sets obtained in TriVoC is significantly higher than that in RANSAC. With such a sufficient number of pure-inlier sets, TriVoC can converge to the maximum consensus set very easily. (Note that the fewer the pure-inlier sets we obtain, the less likely it is to achieve the maximum consensus due to the possible influence of noise.) Thus, from this empirical evaluation, we see that TriVoC has a very promising guarantee of optimality (correctness of finding the full inlier set), much stronger than that of random-sampling (RANSAC).

\subsection{Real Application 1: Object Localization}

We test TriVoC in the realistic application of 3D object localization using RGB-D Scenes dataset~\cite{lai2011large} with different objects and RGB-D scenes. We extract the object from the scene with the labels provided, and transform the object with a random transformation ($||\boldsymbol{t}||\leq3$). To increase the outlier ratio, we impose noise $\sigma=0.1$ on the transformed object and use FPFH to build correspondences between the object and whole scene. We use GNC-TLS, FLO-RANSAC, GORE+RANSAC, and TriVoC to localize the object (estimating transformation). Results are shown in Fig.~\ref{obj-local}, where we find that in such high-outlier situations, GNC-TLS fails in all tests, FLO-RANSAC fails in most tests and runs too slowly, and GORE+RANSAC is slow in some cases, while TriVoC is accurate, robust and time-efficient all the time.

\subsection{Real Application 2: Scene Stitching}

We further evaluate TriVoC in the scene stitching application with Microsoft 7-scenes dataset~\cite{shotton2013scene}. We select 6 pairs of scans with low overlapping from the \textit{red kitchen} and \textit{office}. Since FPFH may yield too many outliers on RGB-D data, we use SURF~\cite{bay2006surf} to match 2D correspondences across the two RGB images and convert them into 3D ones using depth and intrinsic data. We also apply GNC-TLS, FLO-RANSAC, GORE+RANSAC and TriVoC for comparative evaluation. Results are show in Fig.~\ref{scene-stitching}. Though 2D keypoint matching is used, we see that the inliers are fairly sparse among the correspondences. We can observe that GNC-TLS breaks in all tests and FLO-RANSAC is too slow and occasionally fails, while GORE-RANSAC and our TriVoC both keep robust in all tests. Moreover, TriVoC is greatly faster than GORE+RANSAC, showing the best performance overall.

\section{Conclusion}

This paper presents a novel, deterministic and fast voting-based consensus maximization solver TriVoC for point cloud registration with high or even extreme outliers.  We introduce a new framework of triple-layered voting and correspondence sorting using the pairwise equal-length constraint to rapidly find the maximum consensus set with a strong guarantee of correctness. Multiple experiments validate that the proposed solver TriVoC remains highly robust and fast even when encountering extreme outliers (e.g. up to 99\%), and also applies well to realistic real applications including object localization and scene stitching, outperforming other state-of-the-art robust estimators. Besides, TriVoC has runtime hardly sensitive to the outlier ratio, showing great practicality and potential for real-world use.

{\small
\bibliographystyle{IEEEtran}
\bibliography{egbib}
}

\end{document}